\pdfoutput=1
\documentclass[3p,number]{elsarticle}%

\usepackage{tabularx}
\usepackage{array}
\usepackage{graphicx}
\graphicspath{{figures/}}
\usepackage{booktabs}
\usepackage{algorithm}
\usepackage{algpseudocode}
\usepackage{siunitx}
\usepackage{subcaption}
\usepackage{xcolor}
\usepackage{amsmath}
\usepackage{comment}

\usepackage{tikz}
\usepackage{tikz-3dplot}
\usetikzlibrary{shapes.geometric, arrows.meta, positioning, fit, calc, backgrounds, shadows, 3d, decorations.pathreplacing}

\usepackage{pgfplots}
\pgfplotsset{compat=1.18}
\pgfplotsset{
    meshplot/.style={
        width=\linewidth,
        height=0.62\linewidth,
        tick label style={font=\footnotesize},
        label style={font=\footnotesize},
        legend style={
            font=\footnotesize,
            draw=none,
            fill=none,
            at={(0.98,0.98)},
            anchor=north east,
            cells={anchor=west}
        },
        legend cell align={left},
        grid=major,
        grid style={dashed,gray!30},
        axis line style={black},
        tick style={black}
    }
}

\definecolor{meshcoarse}{HTML}{1F77B4}
\definecolor{meshmedium}{HTML}{FF7F0E}
\definecolor{meshfine}{HTML}{2CA02C}
\definecolor{meshextrafine}{HTML}{D62728}

\usetikzlibrary{patterns}

\definecolor{camcol}{HTML}{2D3748}
\definecolor{fggreen}{HTML}{276749}
\definecolor{fgdot}{HTML}{38A169}
\definecolor{disc}{HTML}{C53030}
\definecolor{ptA}{HTML}{2B6CB0}
\definecolor{ptB}{HTML}{38A169}
\definecolor{ptC}{HTML}{D69E2E}
\definecolor{ptD}{HTML}{9B2C2C}
\definecolor{maskfill}{HTML}{EDF2F7}
\definecolor{maskbdr}{HTML}{A0AEC0}
\definecolor{conefill}{HTML}{EBF8FF}
\definecolor{conebdr}{HTML}{90CDF4}
\definecolor{barkeep}{HTML}{2B6CB0}
\definecolor{barfail}{HTML}{CBD5E0}
\definecolor{threshcol}{HTML}{D69E2E}
\definecolor{secbg}{HTML}{F7FAFC}
\definecolor{keptbg}{HTML}{F0FFF4}
\definecolor{discbg}{HTML}{FFF5F5}
\definecolor{arrcol}{HTML}{4A5568}

\tikzset{
    basic/.style={draw, align=center, rounded corners, drop shadow, font=\small},
    io/.style={basic, trapezium, trapezium left angle=70, trapezium right angle=110, fill=blue!10, minimum height=0.8cm},
    process/.style={basic, rectangle, fill=white, minimum height=0.8cm, minimum width=2.5cm},
    decision/.style={basic, diamond, fill=green!10, aspect=2, inner sep=2pt, font=\footnotesize},
    arrow/.style={->, >=stealth, thick, color=black!70},
    group/.style={draw=black!30, dashed, inner sep=10pt, rounded corners, fill=gray!5}
}

\usepackage{makecell}
\usepackage{threeparttable}
\usepackage{multirow}

\usepackage{hyperref}
\hypersetup{colorlinks,
  linkcolor=blue,
  citecolor=blue,
  urlcolor=blue,
  pdftitle={Semi-automated reconstruction of indoor geometry from 360-degree video for CFD-based airflow analysis in classrooms},
  pdfauthor={Dhruv Gamdha, James Afful, Shambhavi Joshi, Ulrike Passe, Adarsh Krishnamurthy, Baskar Ganapathysubramanian}}
\newcommand{\cref}[3]{\hyperref[#2]{#1~\ref*{#2}{#3}}}
\newcommand{\colref}[2]{\hyperref[#2]{#1~\ref*{#2}}}
\newcommand{\algoref}[1]{\colref{Algorithm}{#1}}
\newcommand{\eqnref}[1]{\colref{Equation}{#1}}
\newcommand{\figref}[1]{\colref{Figure}{#1}}
\newcommand{\subfigref}[2]{\cref{Figure}{#1}{#2}}
\newcommand{\secref}[1]{\colref{Section}{#1}}
\newcommand{\tableref}[1]{\colref{Table}{#1}}
\newcommand{\coloredref}[2]{\hyperref[#2]{#1~\ref*{#2}}}
\newcommand{\coloredsubref}[3]{\hyperref[#2]{#1~\ref*{#2}{#3}}}

\newcommand{\classGridTopHeight}{0.155\textheight}
\newcommand{\classGridCFDHeight}{0.19\textheight}

\newcommand{\classGridTopPanel}[1]{%
\begin{minipage}[c][\classGridTopHeight][c]{\linewidth}
\centering
\includegraphics[
    width=\linewidth,
    height=\classGridTopHeight,
    keepaspectratio
]{#1}
\end{minipage}%
}

\newcommand{\classGridCFDPanel}[2][500bp 70bp 500bp 50bp]{%
\begin{minipage}[c][\classGridCFDHeight][c]{\linewidth}
\centering
\includegraphics[
    width=\linewidth,
    height=\classGridCFDHeight,
    keepaspectratio,
    trim=#1,
    clip
]{#2}
\end{minipage}%
}

\newcommand{\classGridDecayPanel}[4]{%
\begin{minipage}[c][\classGridCFDHeight][c]{\linewidth}
\centering
\begin{tikzpicture}
\begin{axis}[
    width=\linewidth,
    height=\classGridCFDHeight,
    xlabel={Time (s)},
    ylabel={$C/C_0$},
    xmin=0,
    xmax=#2,
    ymin=0,
    ymax=1.05,
    tick label style={font=\scriptsize},
    label style={font=\scriptsize},
    grid=major,
    grid style={dashed,gray!30},
    axis line style={black},
    tick style={black},
    legend style={
        font=\scriptsize,
        draw=none,
        fill=none,
        at={(0.97,0.97)},
        anchor=north east
    }
]

\addplot[thick, each nth point=4, filter discard warning=false, unbounded coords=discard] table[
    col sep=comma,
    header=false,
    x index=0,
    y index=1
] {#1};
\addlegendentry{#4}

\addplot[dashed] coordinates {(0,0.5) (#2,0.5)};
\addlegendentry{$C/C_0=0.5$}

\addplot[dotted, thick] coordinates {(#3,0) (#3,0.5)};
\node[anchor=south west, font=\scriptsize] at (axis cs:#3,0.52) {$t_{50}=#3$~s};

\end{axis}
\end{tikzpicture}
\end{minipage}%
}

\newcommand{\audiGridTopHeight}{0.18\textheight}
\newcommand{\audiGridCFDHeight}{0.19\textheight}

\newcommand{\audiGridTopPanel}[1]{%
\begin{minipage}[c][\audiGridTopHeight][c]{\linewidth}
\centering
\includegraphics[
    width=\linewidth,
    height=\audiGridTopHeight,
    keepaspectratio
]{#1}
\end{minipage}%
}

\newcommand{\audiGridCFDPanel}[2][160bp 80bp 160bp 70bp]{%
\begin{minipage}[c][\audiGridCFDHeight][c]{\linewidth}
\centering
\includegraphics[
    width=\linewidth,
    height=\audiGridCFDHeight,
    keepaspectratio,
    trim=#1,
    clip
]{#2}
\end{minipage}%
}

\newcommand{\classAVelPanel}[2][0 0 0 0]{%
\includegraphics[width=\linewidth, trim=#1, clip]{#2}%
}

\begin{document}

\begin{frontmatter}

\title{Semi-automated reconstruction of indoor geometry from 360-degree video for CFD-based airflow analysis in classrooms}

\author[me]{Dhruv Gamdha\fnref{fn1}}
\ead{dgamdha@iastate.edu}

\author[me]{James Afful\fnref{fn1}}
\ead{affulj@iastate.edu}

\author[me]{Shambhavi Joshi}
\ead{sjoshi@iastate.edu}

\author[arch]{Ulrike Passe}
\ead{upasse@iastate.edu}

\author[me]{Adarsh Krishnamurthy\texorpdfstring{\corref{cor1}}{}}
\ead{adarsh@iastate.edu}

\author[me]{Baskar Ganapathysubramanian\texorpdfstring{\corref{cor1}}{}}
\ead{baskarg@iastate.edu}

\cortext[cor1]{Corresponding authors}
\fntext[fn1]{These authors contributed equally to this work.}

\address[me]{Department of Mechanical Engineering, Iowa State University, Ames, IA 50011, USA}
\address[arch]{Department of Architecture, Iowa State University, Ames, IA 50011, USA}

\begin{abstract}
Computational Fluid Dynamics (CFD) is widely used to evaluate ventilation and contaminant transport in occupied buildings, but deployment at scale is limited by three bottlenecks: acquiring room geometry without costly scanning hardware or manual CAD modeling, decomposing the scene into individually manipulable furniture and occupant objects, and reconfiguring those objects for alternative layouts without re-capturing the room. We present a semi-automated workflow that converts a single 360-degree video of a room into individually editable, simulation-ready geometry assets. In the geometry pipeline, a dense point cloud is reconstructed using Neural Radiance Fields (NeRF), and 2D instance masks are extracted using text-prompted SAM~3 segmentation. These masks are then lifted to 3D using multi-view consensus and depth-band filtering. The resulting points are separated into object instances using an octree, and occlusion gaps are healed using a connectivity graph. Chair templates are fitted using Iterative Closest Point (ICP) alignment, and table geometry is generated procedurally. A browser-based editor supports quality assurance and rapid construction of alternative furniture, occupant, and layout configurations. This geometry is triangulated and a steady Reynolds-averaged solution is obtained using OpenFOAM. This solution is then used for transient passive-scalar transport. The OpenFOAM setup is verified using a mesh-sensitivity study and validated against an IEA Annex~20 benchmark. We apply the workflow to two university classrooms and a tiered lecture-hall auditorium. The capture-to-geometry pass takes two to five hours per room on a consumer workstation. In a controlled obstruction sequence within one classroom, the modeled half-clearance time varies non-monotonically as furniture is added, and a cross-room comparison indicates that clearance behavior cannot be reliably extrapolated between rooms, motivating per-room geometry acquisition. By making that acquisition low-cost, our workflow makes geometry-resolved comparative ventilation studies practical for public spaces such as classrooms.
\end{abstract}


\end{frontmatter}

\section{Introduction}
Indoor airflow governs thermal comfort, ventilation effectiveness, and contaminant transport in occupied buildings, and Computational Fluid Dynamics (CFD) simulation has become an established tool for evaluating these in classrooms, lecture halls, hospitals, and offices \citep{Chen2009VentilationReviewCFD, Mohamadi2022ReviewCFDCOVID}. Reliable CFD predictions, however, depend on accurate 3D geometry of the space being simulated. For mechanically ventilated rooms in particular, seemingly secondary geometric features such as the arrangement of furniture and occupants can materially restructure near-field jets, recirculation zones, and the ventilation effectiveness experienced in the breathing zone \citep{Zhuang2014FurnitureLayoutIAQ, Vita2023CFDAirborneRisk, Askari2022DisplacementVentCFD}. The COVID-19 pandemic intensified attention to these effects in shared occupied spaces, where high occupant densities and variable HVAC topologies create complex airflow environments \citep{Morawska2020HowCanAirborne, Bhagat2020DisplacementVentilation}. A major practical barrier to deploying CFD across many real spaces is therefore the acquisition, semantic decomposition, and preparation of room-specific geometry in a form that can be robustly meshed and readily reconfigured for alternative scenarios.

Three persistent challenges currently limit CFD-based indoor air quality (IAQ) analysis at scale. \emph{First}, acquiring accurate room geometry typically requires either manual CAD modeling from physical measurements or expensive scanning hardware such as terrestrial laser scanners, which typically cost \$50k--\$200k and require careful planning to manage occlusion. In our experience, manual modeling takes days per room even for an experienced engineer, making large-scale or repeated assessment impractical (cost and time figures here and below are author estimates rather than surveyed values). \emph{Second}, even when geometry is captured, the standard pipeline yields a monolithic surface representation: existing Scan-to-BIM (Building Information Modeling) and direct point-cloud analysis tools rarely decompose the scene into individually manipulable, semantically meaningful objects such as walls, individual chairs, tables, and occupants \citep{Abreu2023ProceduralPointCloudReview, Khoshelham2021ISPRSBenchmarkIndoorModelling, Skrzypczak2022ScanToBIMAccuracy}. Yet these objects are precisely the geometry that drives indoor airflow. \emph{Third}, evaluating ventilation under alternative occupancy or furniture configurations currently requires re-capturing or re-modeling the room each time, making parametric what-if studies costly and ad hoc. Together, these challenges make room-specific CFD studies difficult to reproduce and expensive to extend across multiple spaces or alternative geometric configurations, motivating workflows that reduce the geometry-preparation burden while preserving explicit control over the resulting simulation domain.

Recent advances in neural scene reconstruction and foundation vision models offer a promising path through these challenges. Neural radiance fields (NeRFs) \citep{Mildenhall2020NeRF} can produce dense scene representations from casual image capture, and a consumer 360-degree camera costing under \$500 can capture an entire room in minutes. Modular frameworks such as Nerfstudio \citep{Tancik2023Nerfstudio} have made NeRF training and point cloud export accessible to non-specialists. In parallel, foundation segmentation models such as the Segment Anything Model (SAM) \citep{Kirillov2023SAM} and its successors \citep{Ravi2024SAM2, Carion2025SAM3} enable zero-shot, concept-prompted instance segmentation, reducing the annotation burden on new indoor environments. A NeRF yields dense geometry without simulation-ready surfaces \citep{Remondino2023CriticalNeRF}, and 2D segmentation yields instance semantics without 3D structure. Bridging the gap to engineering simulation requires an integrated pipeline that combines reconstruction, semantic decomposition, geometric repair, and mesh generation as a unified workflow.

This paper presents an end-to-end semi-automated pipeline that addresses the three challenges above by converting a single 360-degree video of an indoor environment into individually editable, simulation-ready surface assets, and then using those assets to drive CFD-based airflow and passive-scalar transport simulations in OpenFOAM. The pipeline couples NeRF-based scene reconstruction \citep{Tancik2023Nerfstudio} and SAM~3 concept-prompted segmentation \citep{Carion2025SAM3} with automated geometric healing and STL generation, together with a lightweight browser-based human-in-the-loop scene editor that supports rapid construction of parametric what-if configurations from a single capture without re-scanning. Reconstructed geometry preserves the captured locations and footprints of supply and exhaust patches in the room coordinate system, while the CFD model applies idealized uniform inlet velocity boundary conditions at these patches. We demonstrate the workflow on a representative set of university instructional rooms (two seminar/lecture rooms with ceiling-supplied HVAC and a tiered lecture-hall auditorium); the workflow itself is application-agnostic across mechanically ventilated indoor spaces. The key contributions are:

\begin{enumerate}
    \item An end-to-end, semi-automated pipeline that converts a single consumer 360-degree video into individually editable, simulation-ready surface assets, integrating NeRF reconstruction, text-prompted semantic segmentation, multi-view consensus and depth-band filtering, octree-based instance extraction, graph-based healing, template and procedural Iterative Closest Point (ICP) alignment, and automated OpenFOAM case generation with geometrically resolved supply and exhaust patches.

    \item A browser-based human-in-the-loop scene editor that supports rapid construction of parametric what-if configurations, varying furniture layouts, occupant placement, and object orientation, from a single room capture without re-scanning.

    \item A multi-environment CFD demonstration spanning two university classrooms and a tiered lecture-hall auditorium, including a controlled obstruction sequence that resimulates four interior configurations from a single capture under fixed ventilation boundary conditions and reports the resulting monitor-based passive-scalar half-clearance times, together with an auditorium-scale case that exercises the workflow at substantially larger geometric and computational scale.
\end{enumerate}

The remainder of this paper is organized as follows. \secref{sec:related} reviews related work in indoor CFD, 3D reconstruction, and semantic segmentation. \secref{sec:methodology} details the pipeline methodology. \secref{sec:results} presents experimental results across the three environments, including multi-configuration CFD simulations. \secref{sec:discussion} discusses reconstruction failure modes, practical implications, and the limitations of the CFD analysis. \secref{sec:conclusion} concludes with future directions. \ref{app:verification} reports the mesh-sensitivity assessment, the room-scale documentation, and the IEA Annex~20 benchmark comparison supporting the CFD framework.

\section{Related Work}
\label{sec:related}

\subsection{CFD for Indoor Air Quality}
CFD simulations are widely used for evaluating ventilation effectiveness, thermal comfort, and contaminant transport in indoor environments \citep{Chen2009VentilationReviewCFD, fontanini2016ventilation, Liu2019ReviewCFDPV, Mohamadi2022ReviewCFDCOVID, Tan2023ClassroomRiskFramework}. The COVID-19 pandemic accelerated research in this area, with numerous studies applying CFD to classrooms \citep{Park2022VerticalAirflowClassroom}, hospitals, and offices to predict airborne pathogen dispersion and assess ventilation strategies \citep{Morawska2020HowCanAirborne, Bhagat2020DisplacementVentilation}. \citet{Vita2023CFDAirborneRisk} proposed a CFD-based framework for quantifying airborne infection risk in buildings that couples aerosol transport with occupant exposure models. A recurring finding across these studies is that the spatial distribution of furniture and occupants significantly influences local airflow patterns: \citet{Zhuang2014FurnitureLayoutIAQ} showed that furniture layout alone can be a first-order driver of indoor pollutant distributions even when the ventilation system itself is unchanged, and \citet{Askari2022DisplacementVentCFD} further demonstrated that room configuration substantially affects displacement ventilation performance, with CFD predictions varying significantly across layouts. Despite the growing capability of CFD solvers, many studies either use simplified ``box'' representations of furniture or manually construct CAD models from physical measurements, a process that is time-consuming, error-prone, and difficult to scale across multiple rooms or configurations. Closest to the premise of the present work, \citet{Hagbo2021GeometryAcquisitionPedestrianWind} showed for pedestrian-level wind that the choice of geometry-acquisition method itself measurably alters CFD predictions, reinforcing that geometry provenance is a first-order modeling decision.

\subsection{3D Reconstruction of Building Interiors}
Several approaches exist for capturing the 3D geometry of indoor environments. Terrestrial laser scanning produces accurate, dense point clouds but requires expensive equipment, careful planning of scan positions to minimize occlusion, and significant post-processing to register multiple scans and convert raw data into usable surfaces \citep{Khoshelham2021ISPRSBenchmarkIndoorModelling, Skrzypczak2022ScanToBIMAccuracy, Wong2025ImageScanToBIM}. Multi-view stereo (MVS) photogrammetry offers a lower-cost alternative using consumer cameras, but struggles with the textureless surfaces (white walls, uniform floors) and reflective materials common in indoor environments \citep{Remondino2023CriticalNeRF}.

Neural radiance fields (NeRFs) \citep{Mildenhall2020NeRF} have emerged as a promising alternative that can produce dense scene representations from casual image capture. NeRFs synthesize novel views and export dense point clouds even where traditional stereo matching fails, by training a neural network to encode volumetric density and appearance. Critical analyses, however, have shown that NeRF reconstruction quality is scene-dependent and can degrade under reflective surfaces, weak texture, and limited viewpoints \citep{Remondino2023CriticalNeRF}. Work in the built-environment domain has begun integrating NeRF with geometric priors (such as planar and structural constraints) to improve the structural plausibility of reconstructed building elements \citep{Cui2024NeRFusionBuildingConstraints}, and surfel-based variants target online photorealistic reconstruction of indoor scenes \citep{Gao2023SurfelNeRF}. A NeRF representation is not itself a simulation-ready artifact: CFD requires watertight boundary representations and robust meshing pipelines, necessitating additional processing steps to bridge the gap between neural reconstruction and engineering analysis.

\subsection{Towards Simulation-Ready Geometry}
The translation from raw point clouds to simulation-ready surfaces has been addressed from multiple directions. Scan-to-BIM pipelines aim to reconstruct parametric building information models from laser scans or photogrammetry, but reviews emphasize that occlusion, missing data, and topology defects repeatedly force manual intervention \citep{Abreu2023ProceduralPointCloudReview}, and benchmarking efforts confirm that reconstruction quality varies substantially across environments \citep{Khoshelham2021ISPRSBenchmarkIndoorModelling, Skrzypczak2022ScanToBIMAccuracy}. Existing Scan-to-BIM tools also focus primarily on architectural elements (walls, floors, ceilings) rather than movable furniture and assets, which is precisely the geometry most relevant to indoor airflow studies.

An alternative direction bypasses the need for watertight surfaces entirely. \citet{Wang2023PhotogrammetryCFD} integrated photogrammetry-derived point clouds directly with analysis formulations, \citet{Balu2023IMGA_PointClouds} developed immersogeometric methods for flow analysis around point-cloud geometries, \citet{Ishikawa2021CartesianGridAsBuiltCFD} proposed generating CFD grids directly from laser-scanned point clouds without explicit CAD reconstruction, and \citet{Shah2022GPUCollisionPointCloud} demonstrated GPU-accelerated collision analysis of objects within point cloud environments. A closely related family of unfitted finite-element methods, including the Shifted Boundary Method (SBM), enforces boundary conditions on a surrogate boundary near the true geometry rather than on a body-fitted mesh, and has recently been extended to incomplete adaptive octree meshes for incompressible and thermal-flow simulations \citep{Yang2025OctreeSBM, Yang2025SBMThermal}, enabling efficient solution on complex geometries without body-fitted meshing. The reconstructed geometry produced by a pipeline like ours could in principle be consumed directly by such an unfitted solver; here we instead target compatibility with mainstream body-fitted meshing (\texttt{snappyHexMesh}) so that the reconstructed assets are usable within the OpenFOAM toolchain widely adopted for indoor CFD.

Methods for indoor reconstruction from point clouds have addressed multi-room scenes with curved walls \citep{Yang2019IndoorMultiRoomCurvedWalls}, multi-level buildings with inter-floor connections \citep{Lim2021MultiLevelIndoor}, and occlusion handling using structural priors \citep{Liang2025GestaltIndoorOcclusion}. While these approaches reduce the geometry burden, they share a limitation: none of them provides a semantic decomposition of the scene into individually manipulable objects, which is the prerequisite for parametric what-if studies of alternative occupancy or furniture configurations.

\subsection{Semantic Segmentation for Indoor Scenes}
The Segment Anything Model (SAM) \citep{Kirillov2023SAM} and its successors (SAM~2 \citep{Ravi2024SAM2} for video segmentation and tracking, and SAM~3 \citep{Carion2025SAM3} for concept-prompted segmentation) enable zero-shot segmentation without task-specific training. SAM~3 in particular introduces \emph{promptable concept segmentation}, where a short noun phrase (e.g., ``chair'' or ``table'') serves as the prompt and the model detects, segments, and tracks all matching instances across images and video. This reduces the annotation burden for new indoor environments. Our pipeline uses SAM~3's concept prompting for semantic segmentation of furniture categories in 2D video frames, then lifts the masks to 3D through multi-view consensus voting and depth-band filtering. Beyond segmentation accuracy in isolation, \citet{Patil2025ScanToBIMSegAccuracy} showed that downstream reconstruction accuracy does not degrade monotonically with segmentation errors: the critical requirement for CFD-oriented pipelines is end-to-end performance, with segmentation, geometric repair, and meshing assessed together against the final simulation objectives. Domain-adaptation strategies that bridge synthetic BIM-derived training data and real scans \citep{Hu2024DawNetBIMtoScan} are complementary to this end-to-end view, providing improved 2D/3D segmentation quality where annotation cost remains a bottleneck.

Taken together, prior work supplies the individual ingredients, dense neural reconstruction, promptable segmentation, and point-cloud-tolerant CFD formulations, but not an integrated route from a casual room capture to an editable, simulation-ready scene. Assembling and evaluating that route end-to-end is the subject of the methodology that follows.

\section{Methodology}
\label{sec:methodology}

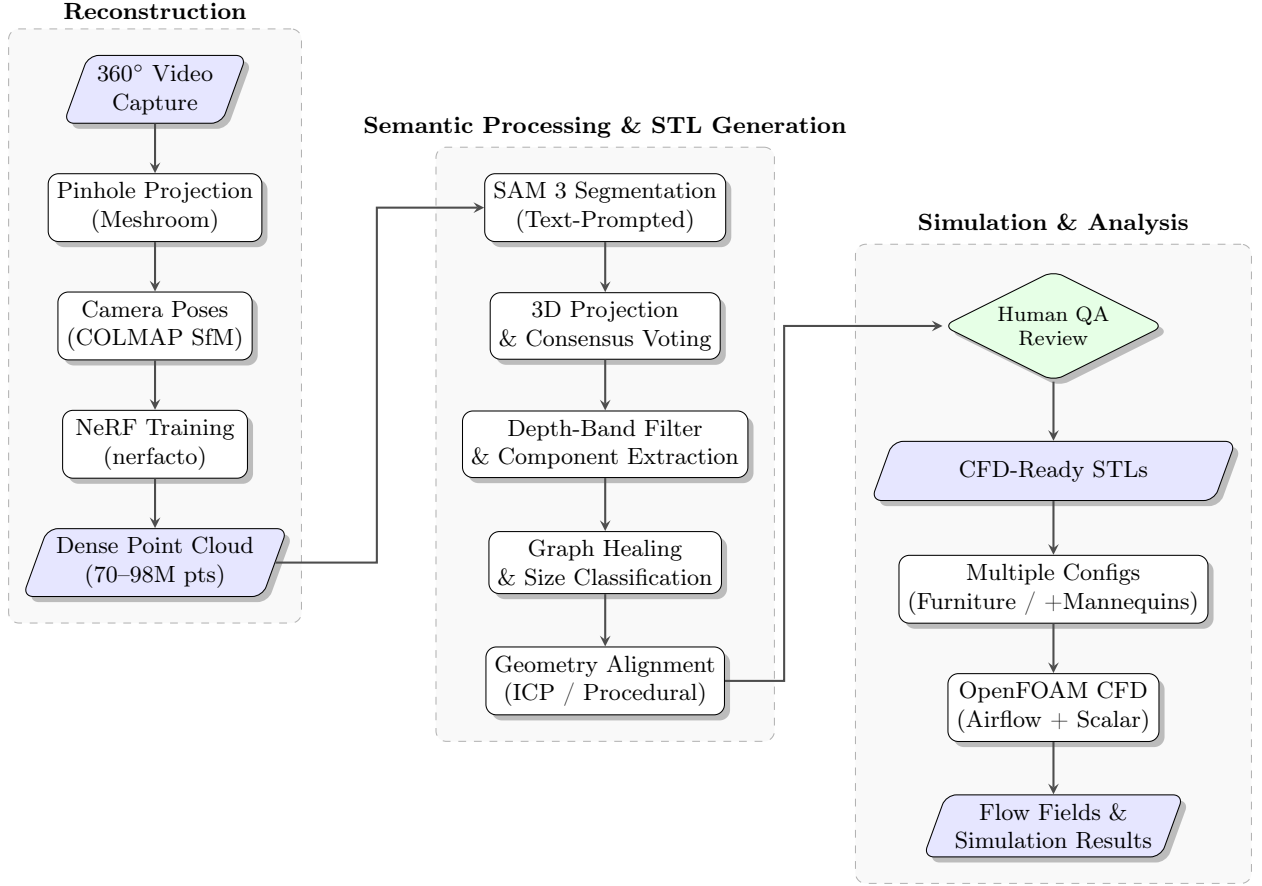
\begin{figure*}[t!]
\centering
\resizebox{\textwidth}{!}{%
\begin{tikzpicture}[node distance=1.6cm, every node/.style={font=\small}]

\node (input) [io] {360$^\circ$ Video \\ Capture};
\node (prep) [process, below of=input] {Pinhole Projection \\ (Meshroom)};
\node (colmap) [process, below of=prep] {Camera Poses \\ (COLMAP SfM)};
\node (nerf) [process, below of=colmap] {NeRF Training \\ (nerfacto)};
\node (pc) [io, below of=nerf] {Dense Point Cloud \\ (70--98M pts)};

\node (sam) [process, right=3cm of prep] {SAM~3 Segmentation \\ (Text-Prompted)};
\node (proj) [process, below of=sam] {3D Projection \\ \& Consensus Voting};
\node (filter) [process, below of=proj] {Depth-Band Filter \\ \& Component Extraction};
\node (heal) [process, below of=filter] {Graph Healing \\ \& Size Classification};
\node (icp) [process, below of=heal] {Geometry Alignment \\ (ICP / Procedural)};

\node (human) [decision, right=3cm of proj] {Human QA \\ Review};
\node (stl) [io, below=0.8cm of human] {CFD-Ready STLs};
\node (configs) [process, below of=stl] {Multiple Configs \\ (Furniture / +Mannequins)};
\node (cfd) [process, below of=configs] {OpenFOAM CFD \\ (Airflow + Scalar)};
\node (output) [io, below of=cfd] {Flow Fields \& \\ Simulation Results};

\draw [arrow] (input) -- (prep);
\draw [arrow] (prep) -- (colmap);
\draw [arrow] (colmap) -- (nerf);
\draw [arrow] (nerf) -- (pc);

\draw [arrow] (pc) -- ++(3,0) |- (sam);
\draw [arrow] (sam) -- (proj);
\draw [arrow] (proj) -- (filter);
\draw [arrow] (filter) -- (heal);
\draw [arrow] (heal) -- (icp);

\draw [arrow] (icp.east) -- ++(0.8,0) |- (human.west);
\draw [arrow] (human) -- (stl);
\draw [arrow] (stl) -- (configs);
\draw [arrow] (configs) -- (cfd);
\draw [arrow] (cfd) -- (output);

\begin{scope}[on background layer]
    \node [group, fit=(input) (prep) (colmap) (nerf) (pc), label=above:\textbf{Reconstruction}] {};
    \node [group, fit=(sam) (proj) (filter) (heal) (icp), label=above:\textbf{Semantic Processing \& STL Generation}] {};
    \node [group, fit=(human) (stl) (configs) (cfd) (output), label=above:\textbf{Simulation \& Analysis}] {};
\end{scope}

\end{tikzpicture}%
}
\caption{End-to-end workflow from 360-degree video capture to CFD simulation. The figure groups the five stages of \secref{sec:methodology} into three phases: (1)~reconstruction (camera pose estimation and NeRF-based scene capture), (2)~semantic processing (concept-prompted segmentation, multi-view filtering, and STL generation, with human-in-the-loop scene editing through a browser-based GUI), and (3)~simulation (CFD analysis across multiple room configurations).}
\label{fig:pipeline}
\end{figure*}

The pipeline transforms raw 360-degree video footage into semantically labeled, simulation-ready 3D assets and consumes them in OpenFOAM for steady airflow and transient passive-scalar transport. The workflow consists of five stages: (1)~data acquisition and preprocessing, (2)~neural scene reconstruction, (3)~semantic extraction and instance segmentation, (4)~geometric alignment, healing, and human-in-the-loop scene editing, and (5)~CFD integration and simulation. \figref{fig:pipeline} presents the complete workflow. Configurable parameters that appear throughout the methodology are summarized in \tableref{tab:params} at the end of the section.

\subsection{Data Acquisition and Preprocessing}
The reconstruction phase converts raw omnidirectional video into a dense, colored 3D point cloud with known camera poses. It consists of three stages: 360-degree video capture, equirectangular-to-pinhole projection, and Structure-from-Motion (SfM) camera pose estimation.

\subsubsection{360-Degree Video Capture}
Data collection is performed using an Insta360 X4 camera recording at 8K resolution ($7680 \times 3840$) and 30~FPS. The camera produces equirectangular frames covering the full $360^\circ \times 180^\circ$ field of view, so that most room surfaces are observed from multiple angles. We move the camera in a systematic grid pattern, traversing the width of the room back and forth while incrementally sliding along the length, to improve volumetric coverage. This scanning strategy provides sufficient baseline between viewpoints for accurate depth triangulation and dense overlap between consecutive frames.

We affix ArUco fiducial markers \citep{GarridoJurado2014ArUco} to the walls to mitigate the tracking failures common in indoor environments with textureless surfaces (e.g., white walls and ceilings). Each of the two classrooms uses 16 markers (4 per wall on each of the four walls). The auditorium, with its larger wall area, uses 18 markers (8 on each of the two long side walls and 2 on the back wall). These markers serve as high-contrast anchors for the SfM pipeline, providing reliable feature correspondences in otherwise featureless regions.

\subsubsection{Equirectangular-to-Pinhole Projection}
Standard computer vision algorithms, including feature detection, matching, and neural radiance field training, are designed for perspective (pinhole) images and perform poorly on equirectangular projections due to the severe distortion at the poles. Each equirectangular frame is therefore decomposed into a set of overlapping perspective views.

We extract eight overlapping $1200 \times 1200$-pixel pinhole projections from each equirectangular frame using the Meshroom framework \citep{Meshroom2021}. The eight views are distributed to cover the full sphere with sufficient overlap between adjacent views. This decomposition preserves the geometric fidelity of straight lines and produces images compatible with standard feature detectors (e.g., SIFT \citep{Lowe2004SIFT}) used in the SfM pipeline. The total number of pinhole images produced per environment is reported in \tableref{tab:capture_stats}.

\begin{table}[t!]
\centering
\caption{Video capture and frame processing statistics. Pinhole images = equirectangular frames $\times$ 8 views; COLMAP poses are the subset of pinhole images successfully registered.}
\label{tab:capture_stats}
\begin{tabularx}{\textwidth}{@{} l *{5}{>{\raggedleft\arraybackslash}X} @{}}
\toprule
\textbf{Environment} & \textbf{Duration (s)} & \textbf{Sampling} & \textbf{Equirect.\ frames} & \textbf{Pinhole images} & \textbf{COLMAP poses} \\
\midrule
Mixed-Furniture Classroom & 61  & 1 per 22 & 84  & 672     & 664   \\
Chair-Dominant Classroom  & 53  & 1 per 20 & 80  & 640     & 620   \\
Auditorium                & 280 & 1 per 42 & 201 & 1{,}608 & 1{,}472 \\
\bottomrule
\end{tabularx}
\end{table}

\subsubsection{Structure-from-Motion via COLMAP}
The resulting set of pinhole images is processed using COLMAP \citep{Schonberger2016COLMAP} to estimate camera intrinsics and extrinsic poses. Sequential matching constraints exploit the temporal continuity of the video feed: each frame is matched primarily against its temporal neighbors, reducing the cost of the matching phase relative to exhaustive matching while maintaining robust pose estimates.

COLMAP performs incremental bundle adjustment to jointly optimize the camera parameters $\{K_i, T_{cw,i}\}$ and a sparse 3D point cloud, where $K_i$ denotes the intrinsic matrix and $T_{cw,i} \in SE(3)$ the world-to-camera transformation for view $i$. The number of successfully registered poses per environment is reported in \tableref{tab:capture_stats}. The sparse reconstruction and calibrated camera poses serve as input for the neural scene reconstruction stage, with camera parameters exported in the format required by the Nerfstudio framework, including the coordinate-system transformation between the COLMAP world frame and the Nerfstudio scene frame.

\subsection{Neural Scene Reconstruction}
Given the calibrated camera poses from COLMAP and the corresponding pinhole images, a neural radiance field is trained to obtain a dense photorealistic 3D representation of the scene. This representation is then exported as an explicit point cloud for downstream semantic processing.

\subsubsection{NeRF Training}
We use the Nerfstudio framework \citep{Tancik2023Nerfstudio} to train a \texttt{nerfacto} model under its default configuration; this method is optimized for real-world scene capture. Unlike traditional mesh-based photogrammetry (e.g., Meshroom's Multi-View Stereo \citep{Meshroom2021}), which reconstructs explicit surfaces and struggles with textureless or reflective regions, the NeRF model represents the scene as a continuous volumetric radiance field $F_\theta: (\mathbf{x}, \mathbf{d}) \rightarrow (\mathbf{c}, \sigma)$, mapping a 3D position $\mathbf{x}$ and viewing direction $\mathbf{d}$ to a color $\mathbf{c}$ and density $\sigma$. The model is trained by minimizing the photometric reconstruction loss between rendered and observed pixel colors across all training views.

\subsubsection{Dense Point Cloud Extraction}
\begin{figure*}[t!]
\centering
\begin{subfigure}[b]{0.32\linewidth}
    
    \centering
    \includegraphics[width=\linewidth]{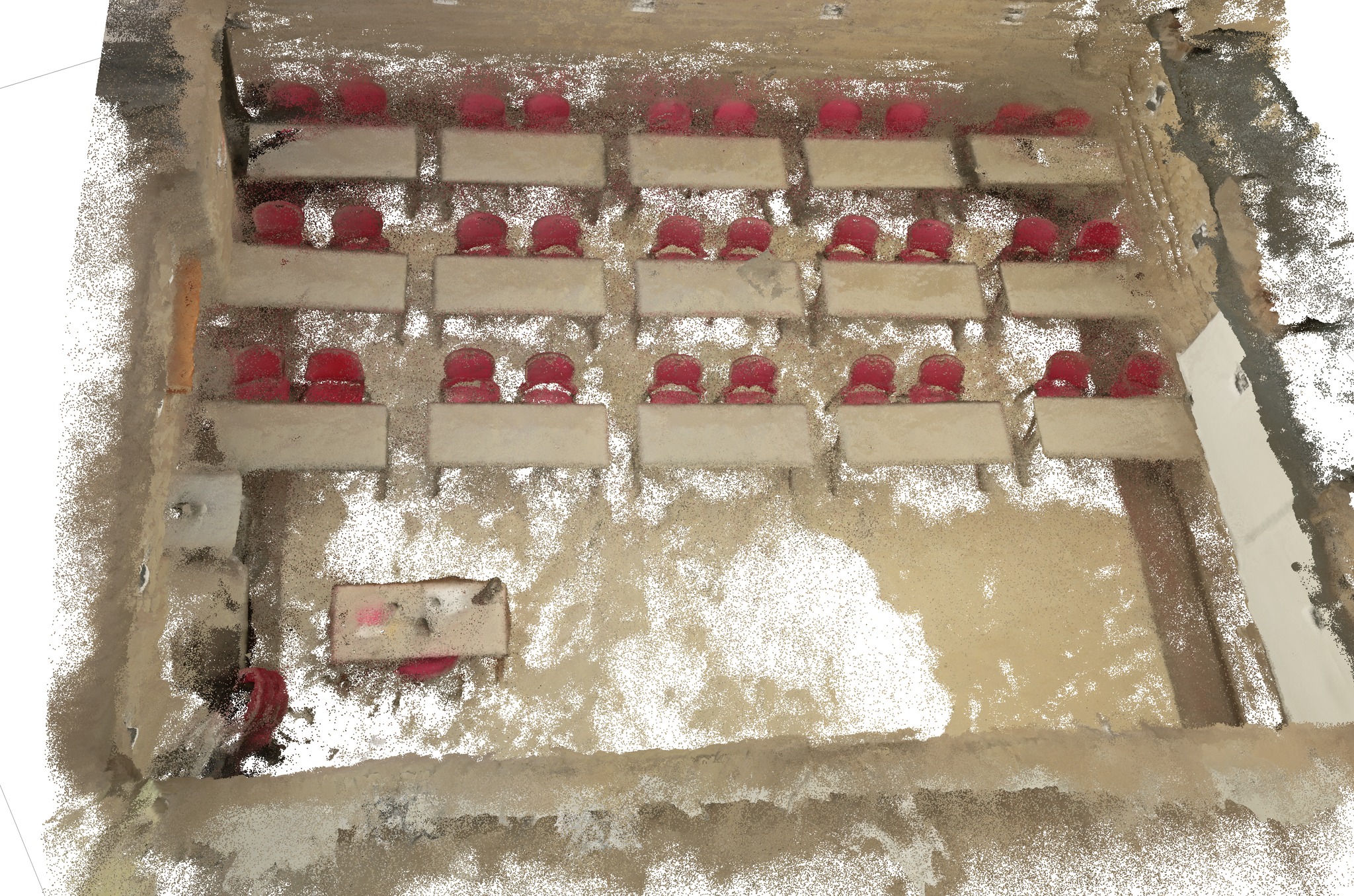}
    \caption{Top-down interior view}
\end{subfigure}\hfill
\begin{subfigure}[b]{0.32\linewidth}
    
    \centering
    \includegraphics[width=\linewidth]{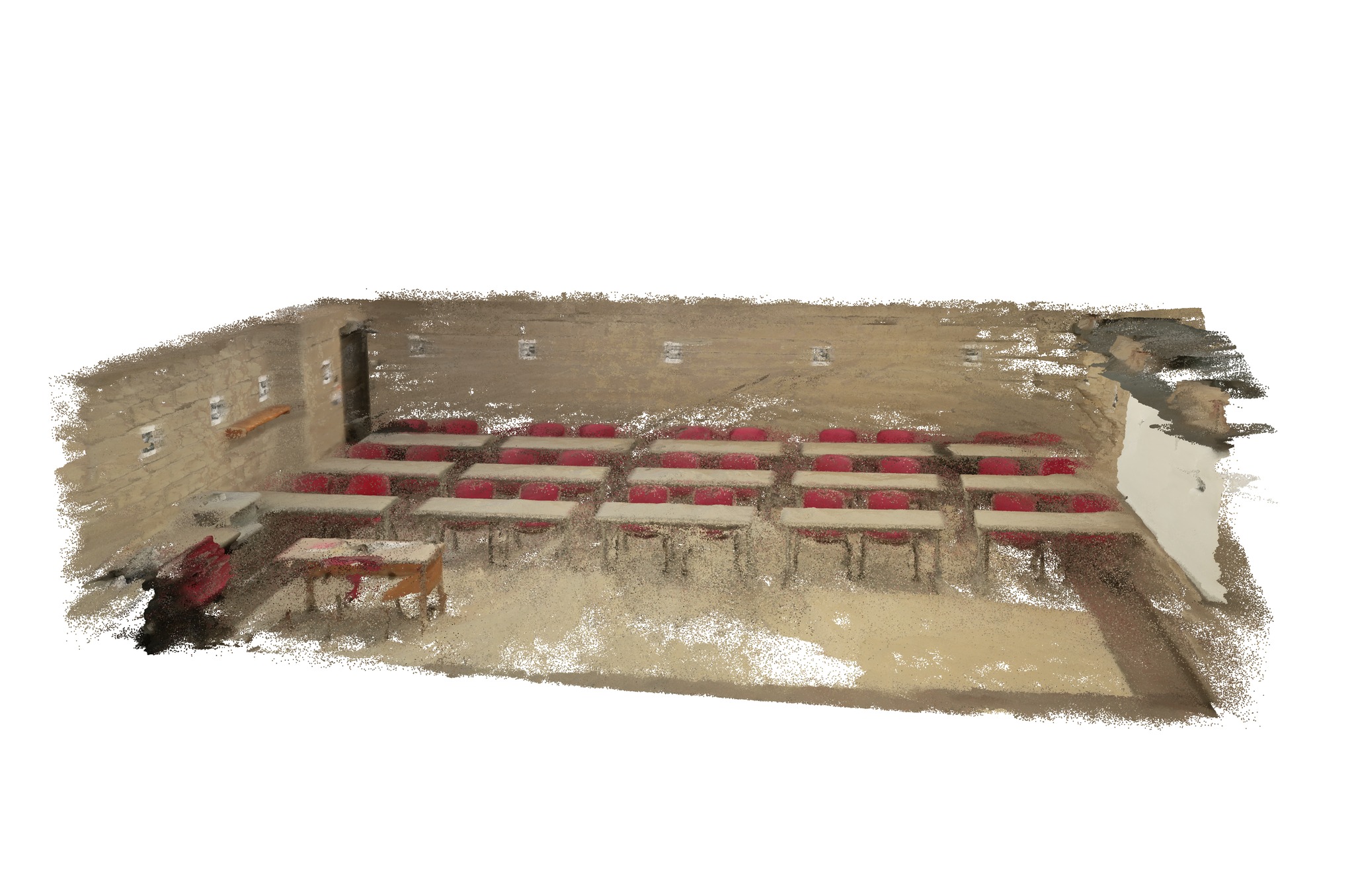}
    \caption{Perspective interior view}
\end{subfigure}\hfill
\begin{subfigure}[b]{0.32\linewidth}
    
    \centering
    \includegraphics[width=\linewidth]{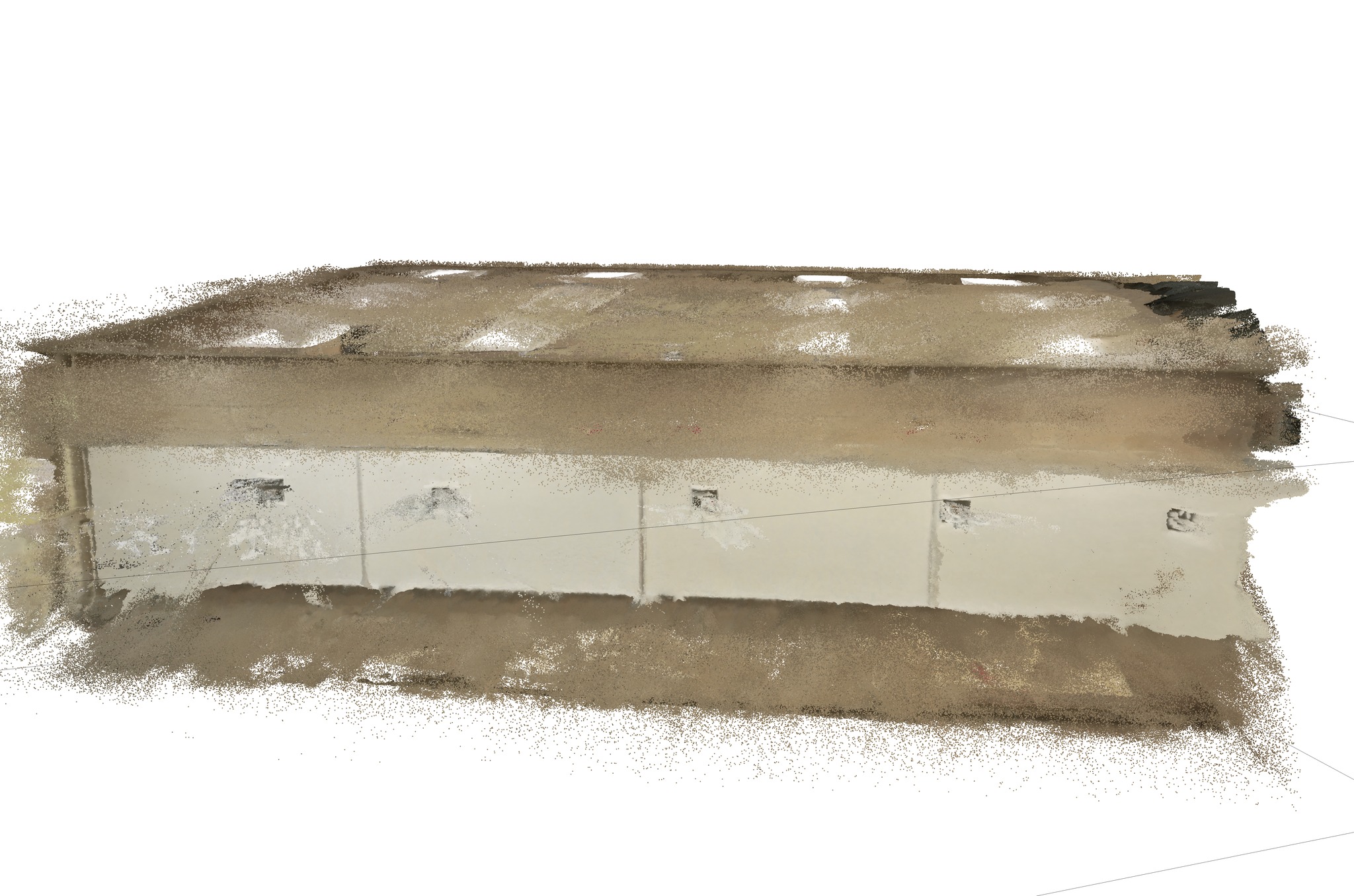}
    \caption{Exterior view}
\end{subfigure}
\caption{Raw dense point cloud extracted from the trained \texttt{nerfacto} model. (a,\,b)~Interior views showing the reconstructed furniture layout with visible noise and floating artifacts. (c)~Exterior view revealing the overall room envelope with scattered outlier points before semantic processing.}
\label{fig:raw_nerf}
\end{figure*}

Following training, a dense point cloud is extracted using Nerfstudio's \texttt{ns-export pointcloud} command; after cropping to the room volume, the exported clouds contain 70--98 million points across the three environments. No outlier removal is applied at this stage. This deliberately produces a noisy point cloud containing floating artifacts and density blobs in free space (\figref{fig:raw_nerf}). Noise removal is delegated entirely to the semantic post-processing pipeline (\secref{sec:semantic}), which uses multi-view geometric consistency to distinguish true geometry from artifacts. Decoupling reconstruction from filtering in this way favors geometric recall, with precision recovered downstream where semantic context is available.

\begin{figure}[t!]
\centering
\begin{subfigure}[b]{0.4\linewidth}
    
    \centering
    \includegraphics[width=\linewidth]{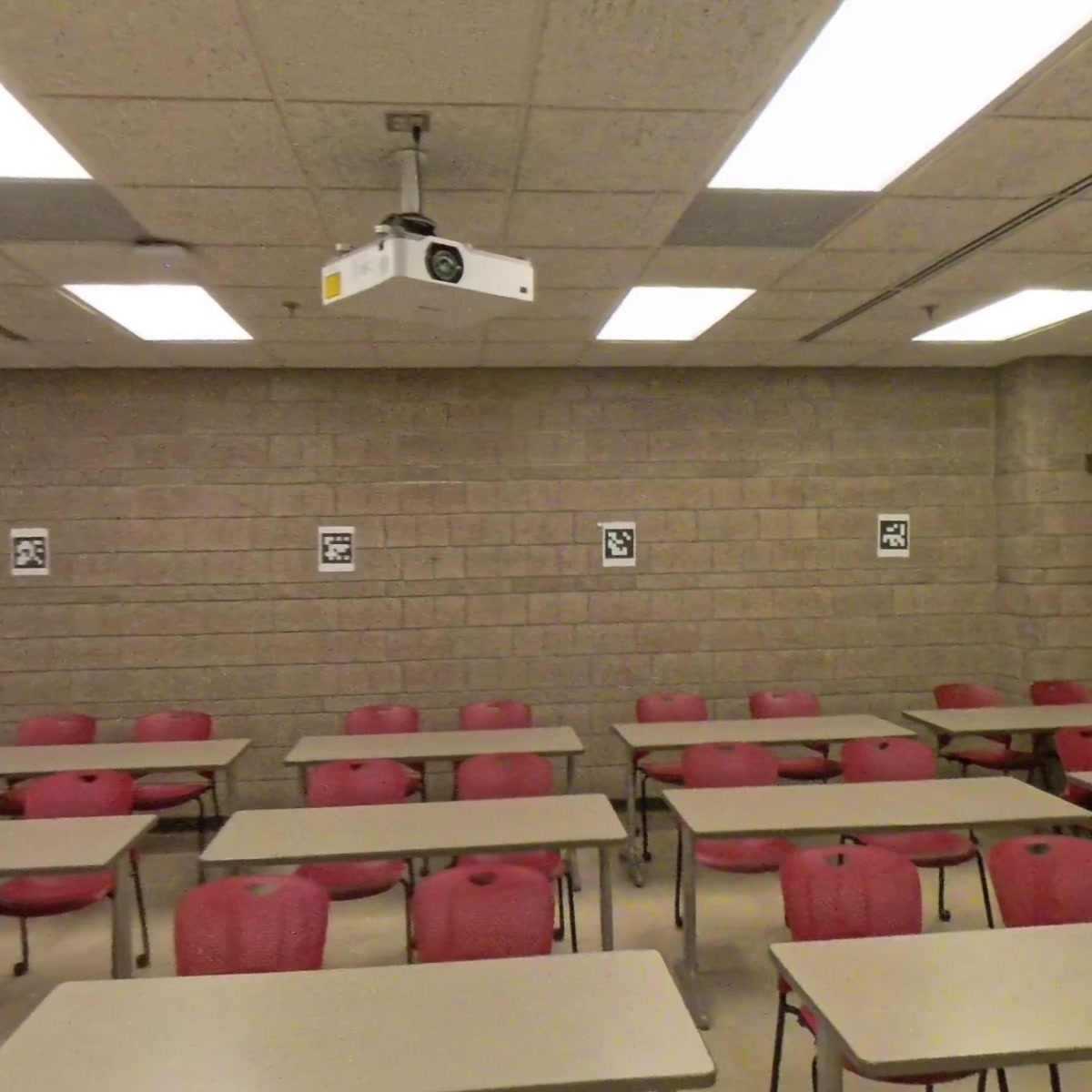}
    \caption{Original photograph}
\end{subfigure}
\begin{subfigure}[b]{0.4\linewidth}
    
    \centering
    \includegraphics[width=\linewidth]{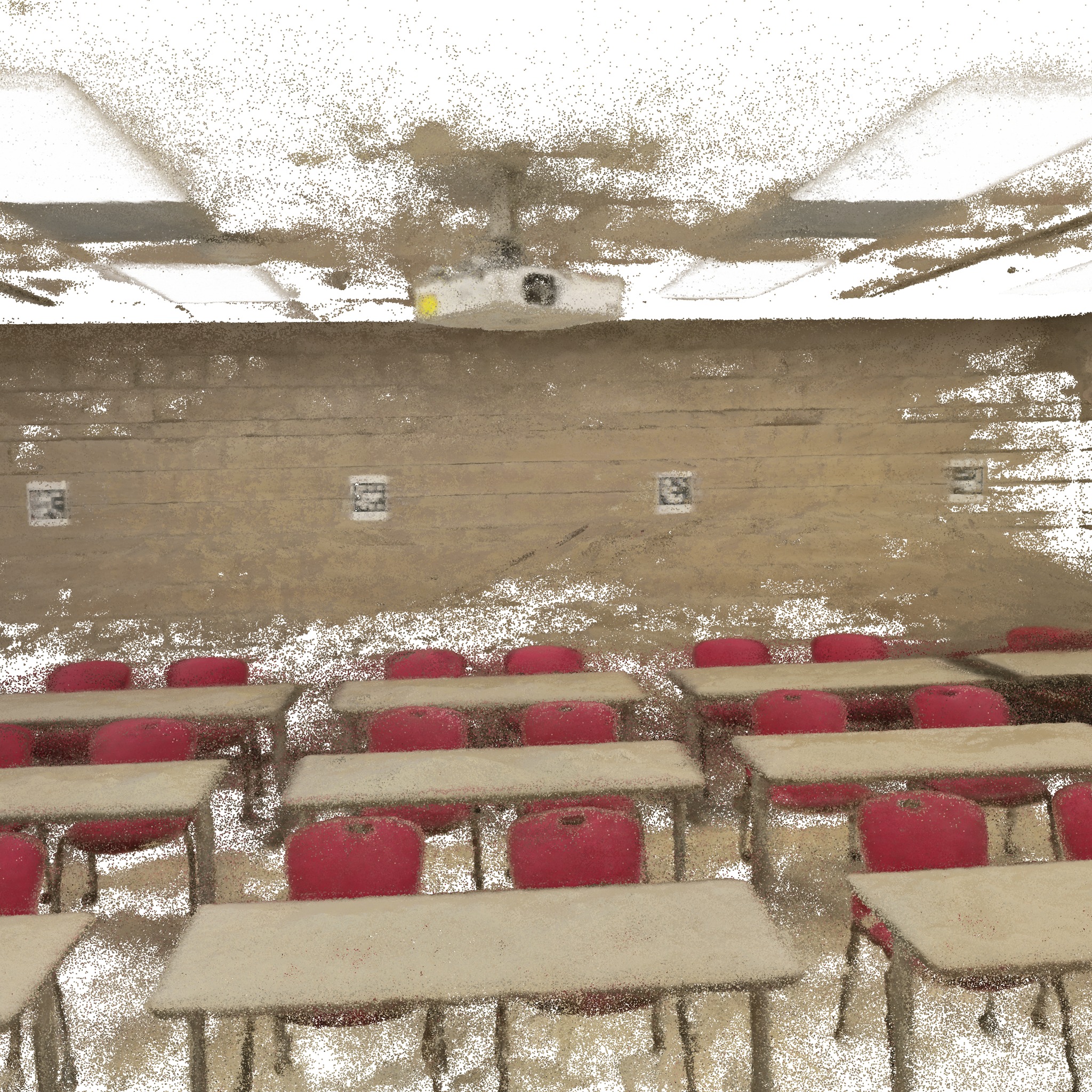}
    \caption{NeRF point cloud render}
\end{subfigure}
\caption{Visual fidelity comparison. (a)~Original photograph captured from the 360-degree camera. (b)~NeRF point cloud rendered from the same camera viewpoint, illustrating the geometric and color fidelity of the reconstruction.}
\label{fig:visual_comparison}
\end{figure}

\subsubsection{Visual Fidelity Assessment}
We render comparison images from matched viewpoints to assess the reconstruction qualitatively (\figref{fig:visual_comparison}): for each reference camera pose, the original captured photograph is shown alongside the corresponding rendered view of the colorized NeRF point cloud. This comparison enables visual inspection of the geometric fidelity of furniture shapes, the color accuracy of textures and lighting, and the completeness of thin structures such as chair legs. Quantitative novel-view-synthesis metrics, peak signal-to-noise ratio (PSNR), structural similarity (SSIM), and learned perceptual image patch similarity (LPIPS), for held-out views are reported in \tableref{tab:nerf_metrics} as part of the multi-environment evaluation.

\subsection{Semantic Extraction and Instance Segmentation}
\label{sec:semantic}
The semantic-extraction stage transforms the raw, unorganized NeRF point cloud into individual object instances (e.g., \emph{Chair~1}, \emph{Chair~2}). It combines text-prompted 2D segmentation, 3D-to-2D projection of the reconstructed point cloud, multi-view consensus voting, depth-band filtering, and octree-based connected-component analysis.

\subsubsection{SAM~3 Model Configuration}
\begin{figure*}[t!]
\centering
\begin{subfigure}[b]{0.32\linewidth}
    \centering
    \includegraphics[width=\linewidth]{sam3_1_orig.jpg}
    \caption{Original classroom image}
\end{subfigure}\hfill
\begin{subfigure}[b]{0.32\linewidth}
    \centering
    \includegraphics[width=\linewidth]{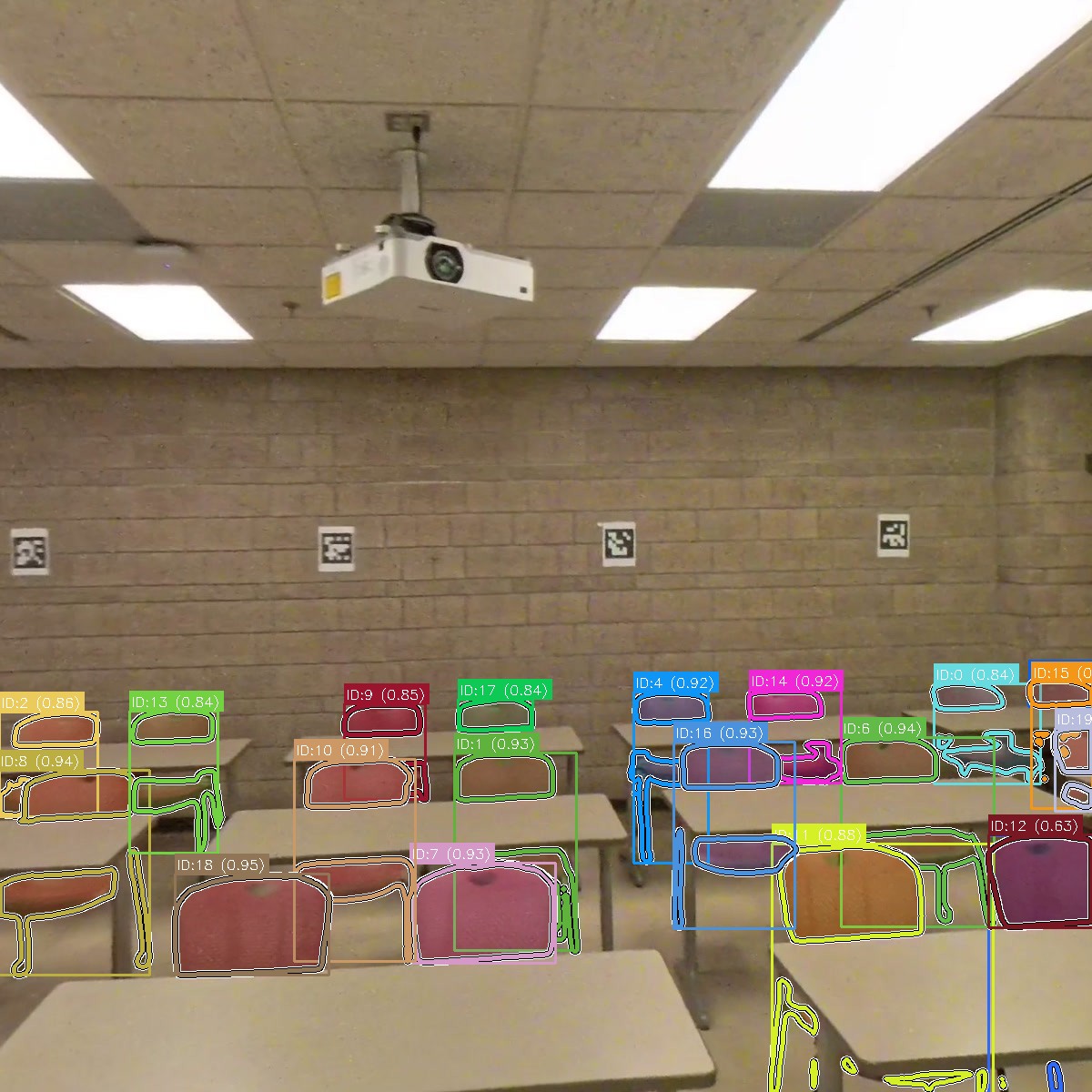}
    \caption{SAM~3 instance segmentation overlay}
\end{subfigure}\hfill
\begin{subfigure}[b]{0.32\linewidth}
    \centering
    \includegraphics[width=0.48\linewidth]{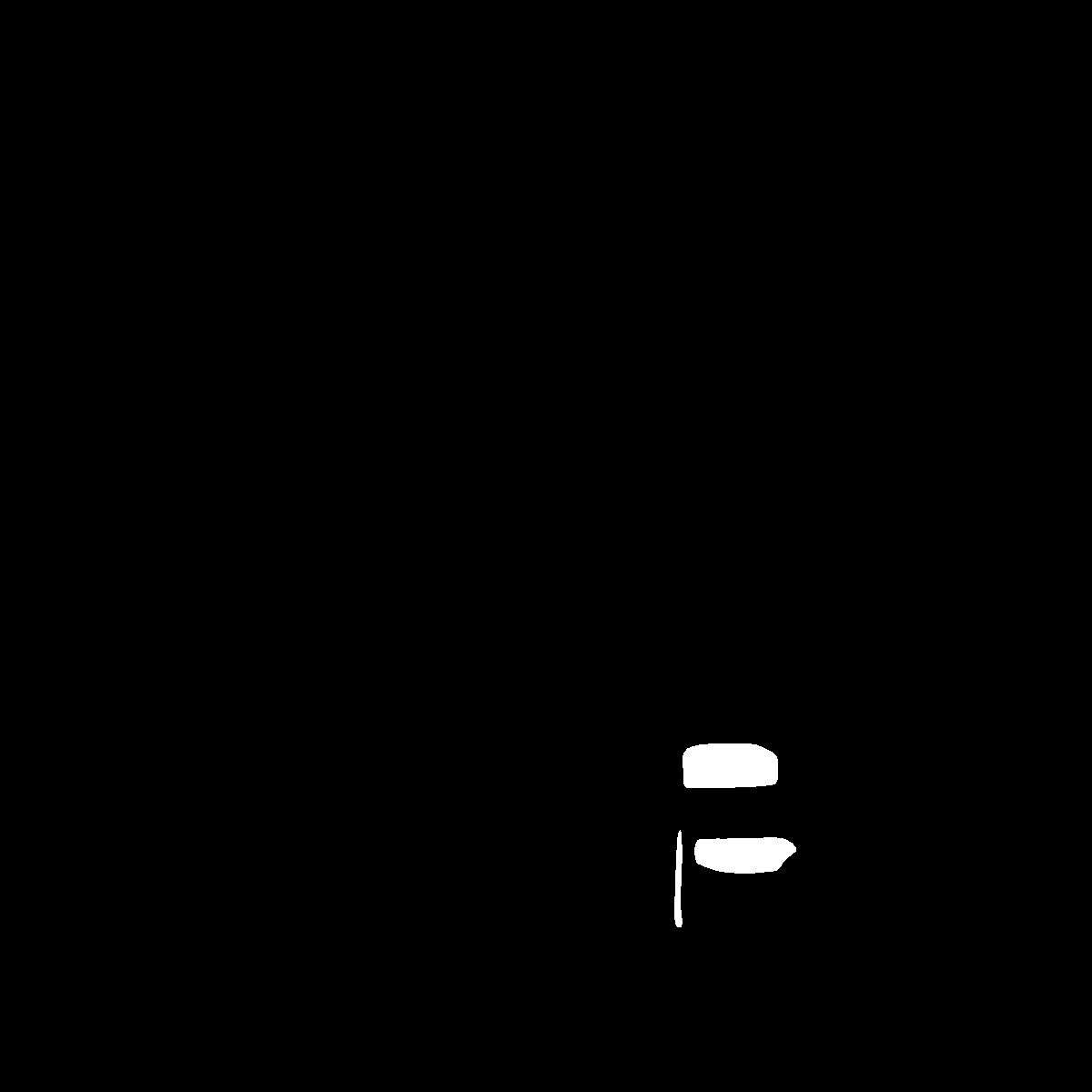}\hfill
    \includegraphics[width=0.48\linewidth]{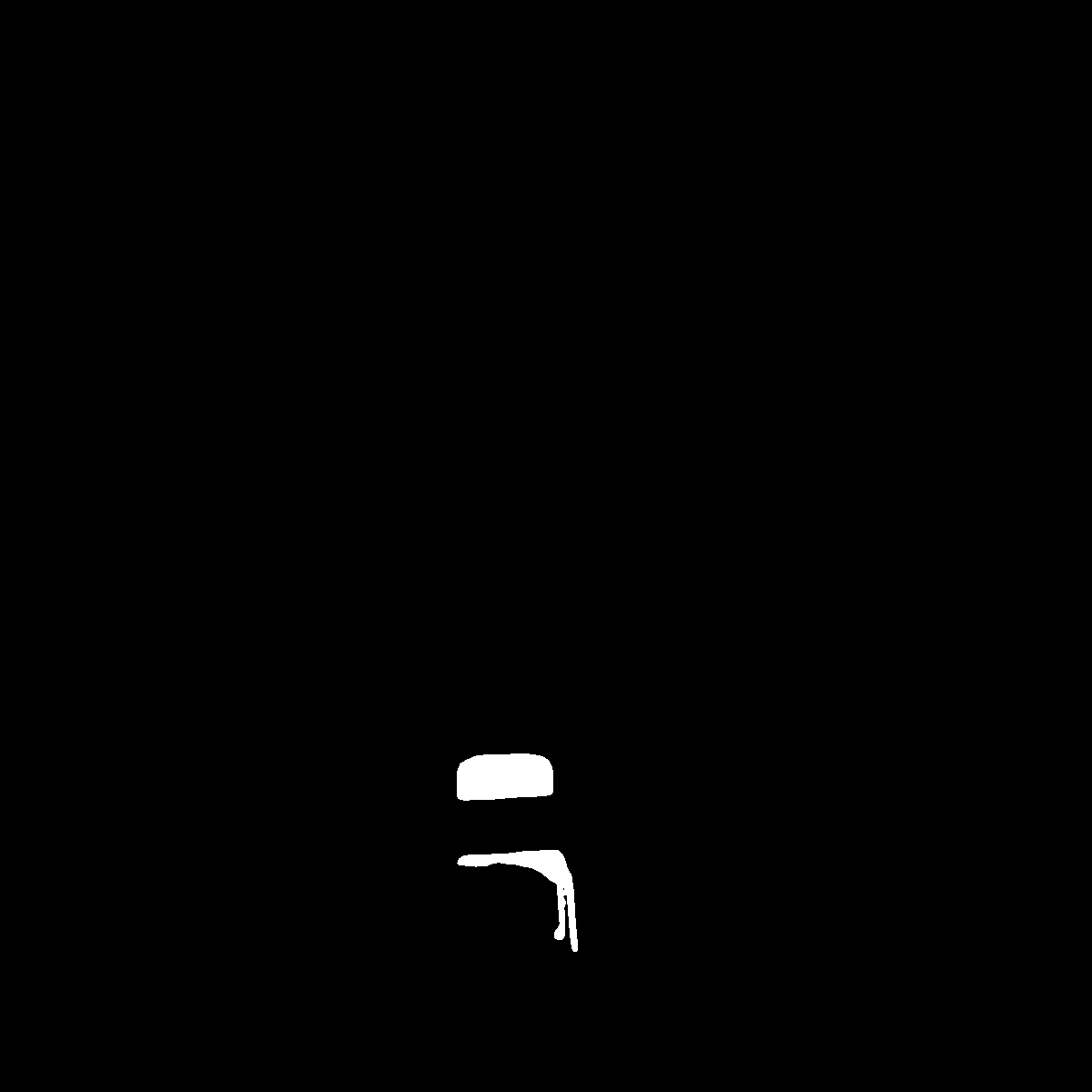}\\[2pt]
    \includegraphics[width=0.48\linewidth]{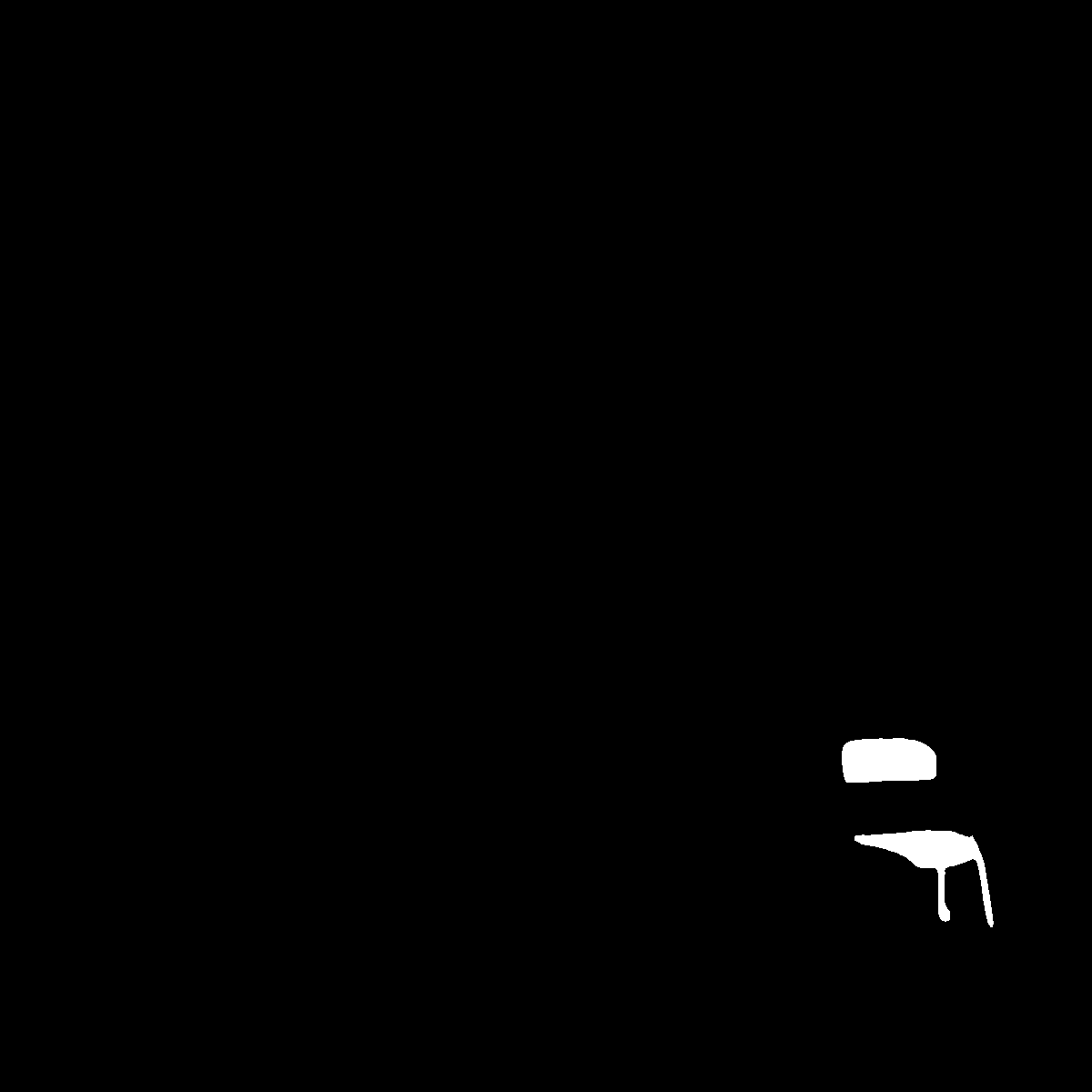}\hfill
    \includegraphics[width=0.48\linewidth]{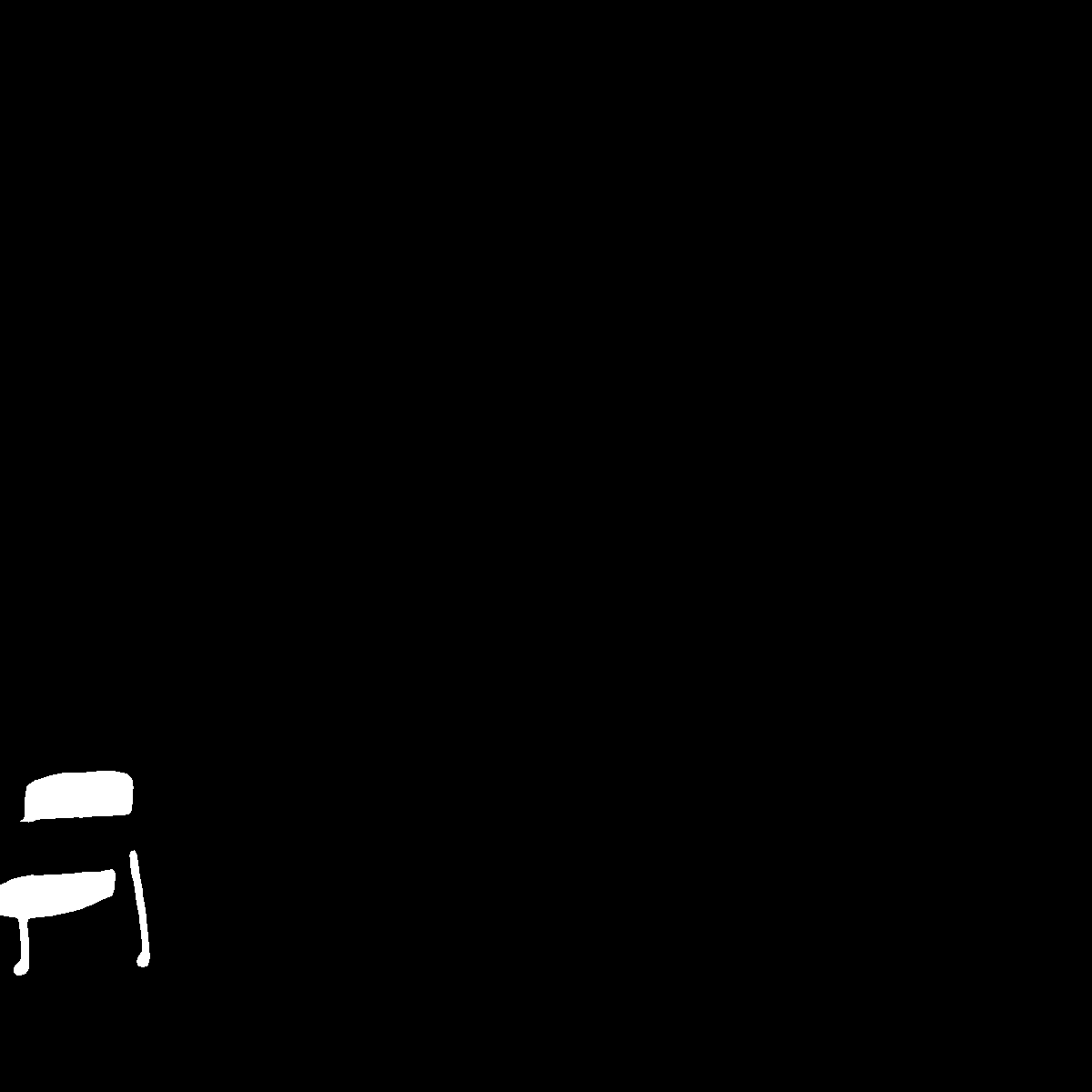}
    \caption{Individual binary masks}
\end{subfigure}
\caption{SAM~3 text-prompted segmentation. (a)~Pinhole-projected training image from the 360\textdegree{} video. (b)~SAM~3 segmentation output with per-instance colored masks overlaid on the image, prompted with the text query \emph{chair}. (c)~Four individual binary masks extracted for separate chair instances; each isolates a single object for the subsequent 3D projection step.}
\label{fig:sam3_segmentation}
\end{figure*}

The segmentation engine is the Segment Anything Model~3 (SAM~3) \citep{Carion2025SAM3}, run under its default configuration. We use SAM~3's native \emph{text-prompted} interface, which avoids the over-segmentation and irrelevant regions that box prompts and fully automatic modes often produce in cluttered indoor scenes. The model is queried with category names (e.g., \emph{chair}, \emph{big table}) and outputs binary masks for regions where the predicted confidence score exceeds $\tau_{conf}$. These 2D masks (\figref{fig:sam3_segmentation}) serve as input to the 3D projection step.

\subsubsection{3D-to-2D Projection}
To label the 3D point cloud, we project global 3D points onto the 2D image planes of the reference cameras and check each projection against the SAM~3 binary masks. This direction (3D~$\rightarrow$~2D) avoids the divergence problems associated with ray-casting from individual pixels into 3D space.

\begin{figure*}[t!]
\centering
\includegraphics[width=\linewidth,trim=0 100 0 100,clip]{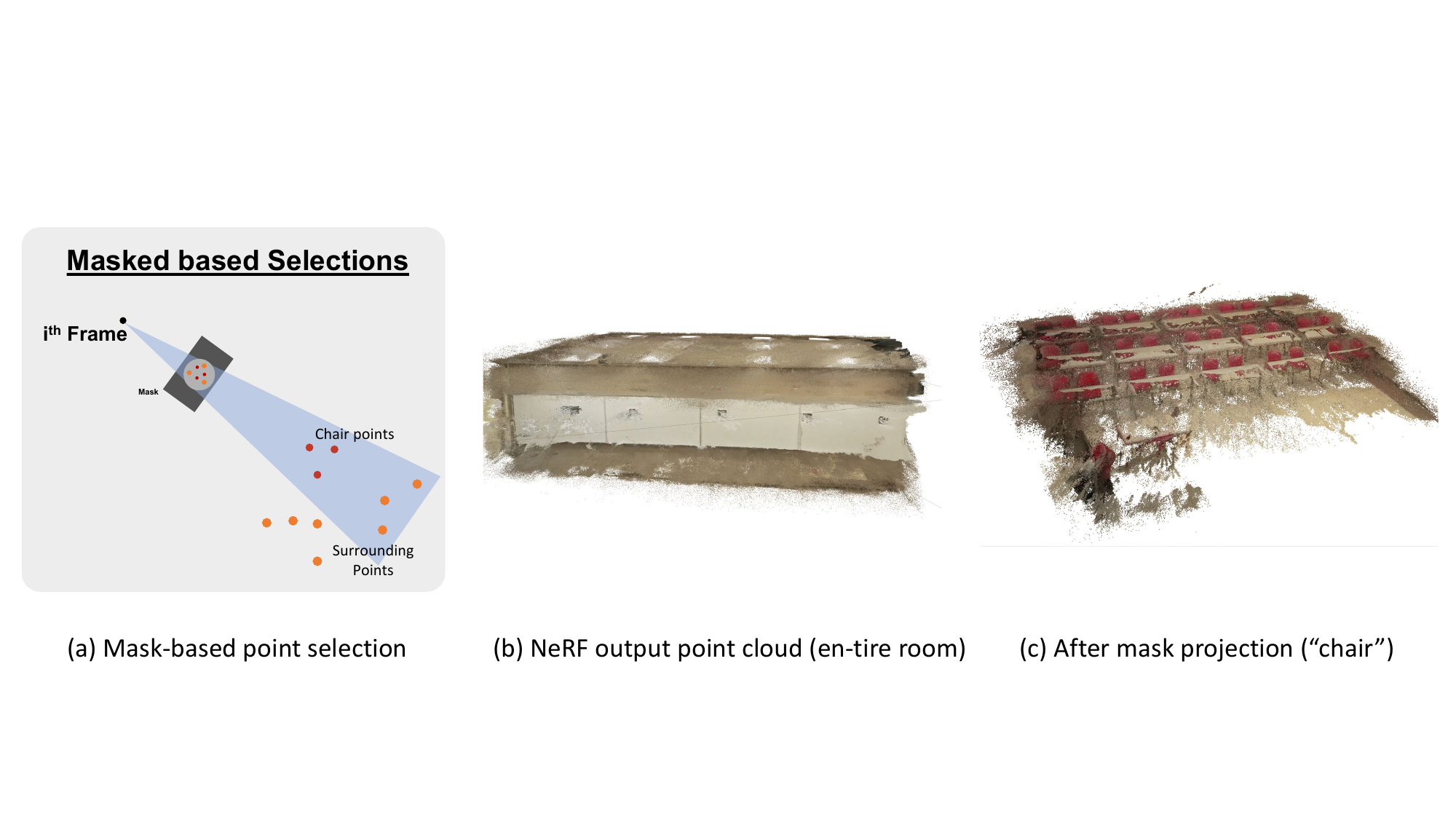}
\caption{Category point cloud extraction via mask-based projection. (a)~For each camera frame, 3D points are projected onto the image plane and checked against the SAM~3 binary mask; points falling inside the mask (chair points) are retained while surrounding background points are discarded. (b)~Full NeRF-reconstructed point cloud of the classroom. (c)~Resulting raw category point cloud after projecting across all frames and retaining only points labeled \emph{chair}, with significant noise from background leakage and projection errors that motivates the multi-view consensus and depth-band filtering steps.}
\label{fig:category_cloud}
\end{figure*}

\begin{figure}[t!]
\centering
\scalebox{0.88}{
\begin{tikzpicture}[node distance=1.2cm, auto]
    \node (point) [io] {3D Point $P_w$};
    \node (proj) [process, below of=point] {Project to Image $(u,v)$};
    \node (maskcheck) [decision, below of=proj, yshift=-0.5cm] {Inside Mask?};
    \node (discard1) [process, left=2.5cm of maskcheck, fill=red!10] {Discard};
    \node (vote) [process, below of=maskcheck, yshift=-0.25cm] {Vote $v_{ij} = 1$};
    \node (repeat) [io, below of=vote, text width=3.5cm, align=center] {Repeat for all $N$ frames};
    \node (consensus) [decision, below of=repeat, yshift=-0.7cm] {$c_j = \sum v_{ij} \geq \tau_{mv}$?};
    \node (discard2) [process, left=2.5cm of consensus, fill=red!10] {Discard (Noise)};
    \node (depthcheck) [decision, below of=consensus, yshift=-1.2cm] {Depth Band Check? \\ $z \in [d_{med} \pm \sigma_{depth} \cdot \mathrm{MAD}]$};
    \node (discard3) [process, left=2.5cm of depthcheck, fill=red!10] {Discard (Occlusion)};
    \node (keep) [process, below of=depthcheck, fill=green!20, yshift=-0.8cm] {Add to Category PC};

    \draw [arrow] (point) -- (proj);
    \draw [arrow] (proj) -- (maskcheck);
    \draw [arrow] (maskcheck) -- node[above] {No} (discard1);
    \draw [arrow] (maskcheck) -- node[right] {Yes} (vote);
    \draw [arrow] (vote) -- (repeat);
    \draw [arrow] (repeat) -- (consensus);
    \draw [arrow] (consensus) -- node[above] {No} (discard2);
    \draw [arrow] (consensus) -- node[right] {Yes} (depthcheck);
    \draw [arrow] (depthcheck) -- node[above] {No} (discard3);
    \draw [arrow] (depthcheck) -- node[right] {Yes} (keep);
    \node [right=0.5cm of consensus, text width=3.5cm, font=\footnotesize, align=left] {Multi-view consensus retains only points consistently selected across $\geq \tau_{mv}$ viewpoints.};
\end{tikzpicture}
}
\caption{Projection-based filtering logic. For each frame, points falling within the semantic mask receive a vote. Votes are accumulated across all $N$ frames, and only points meeting the multi-view consensus threshold $\tau_{mv}$ proceed to depth-band filtering, which removes occluded background geometry.}
\label{fig:projection_logic}
\end{figure}
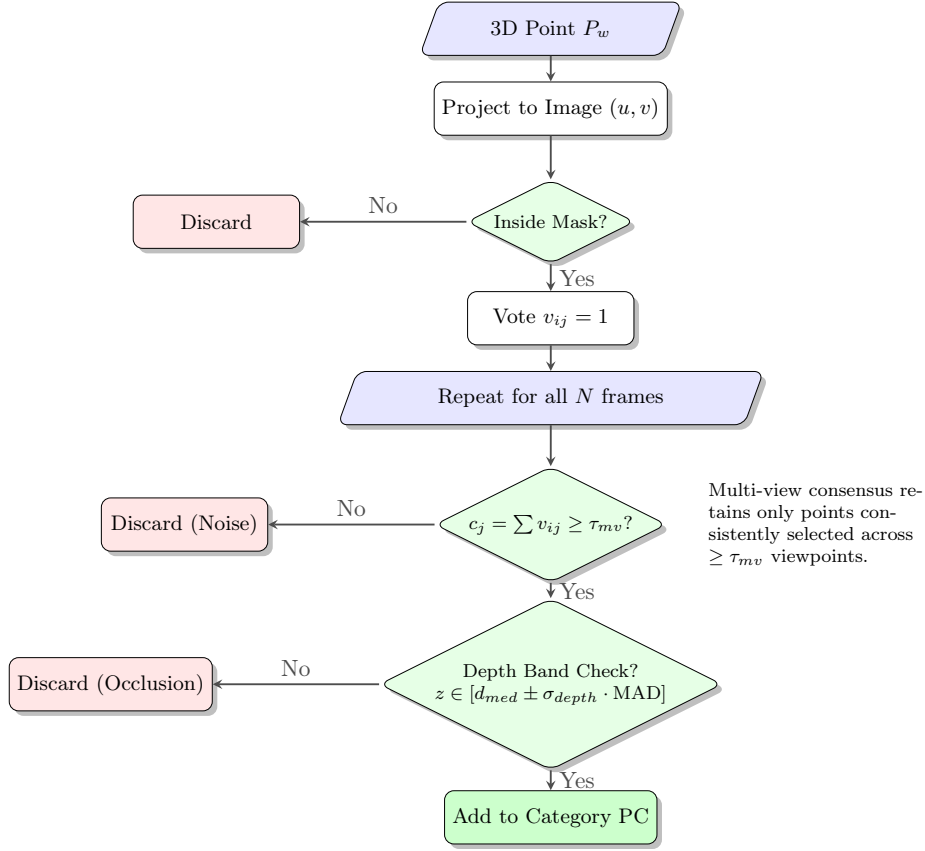

For each camera view $i$, raw NeRF points $P_w$ in the COLMAP world frame are transformed to the camera frame through the world-to-camera transformation $T_{cw,i}$, then projected onto the image plane through the intrinsic matrix $K_i$:
\begin{equation}
    \lambda \begin{bmatrix} u \\ v \\ 1 \end{bmatrix}
    = K_i \, T_{cw,i} \begin{bmatrix} X_w \\ Y_w \\ Z_w \\ 1 \end{bmatrix},
    \qquad
    K_i = \begin{bmatrix} f_x & 0 & c_x \\ 0 & f_y & c_y \\ 0 & 0 & 1 \end{bmatrix}.
\end{equation}
Here $f_x, f_y$ are the focal lengths and $c_x, c_y$ are the principal-point offsets. A point $P_w$ is accepted into the category cloud for view $i$ if and only if its projected pixel falls inside a valid mask: $\mathrm{Mask}_i(u,v) = 1$ (\figref{fig:category_cloud}). The full filtering chain (mask check, multi-view consensus, depth-band filtering) is summarized in \figref{fig:projection_logic}.

\clearpage

\subsubsection{Multi-View Consensus}
\label{sec:multiview_consensus}
\begin{figure*}[t!]
\centering
\includegraphics[width=0.85\linewidth, trim=40bp 86bp 40bp 86bp, clip]{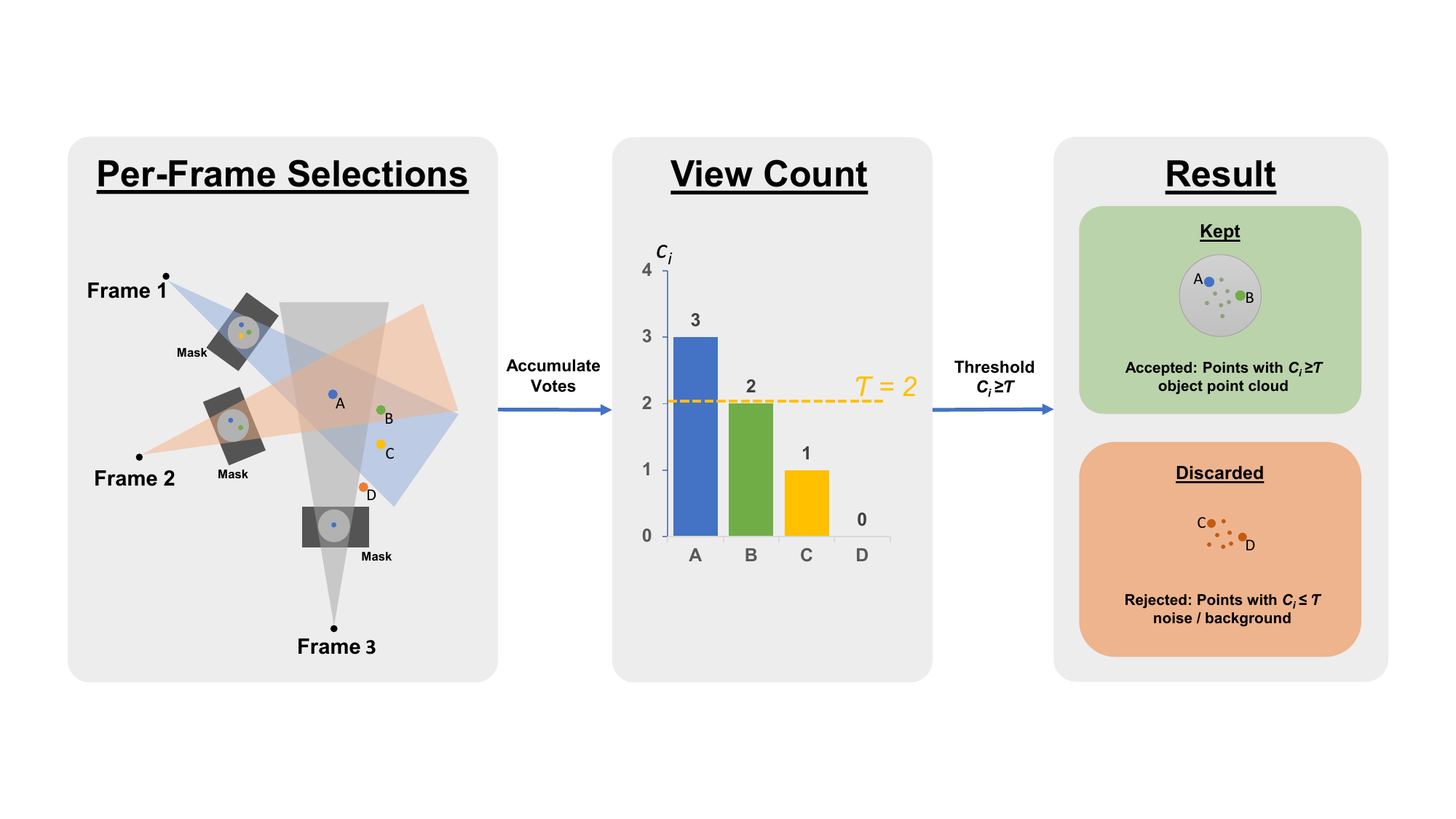}
\caption{Multi-view consensus for point selection. (Left)~3D points ($A, B, C, D$) are projected into multiple camera frames; each frame's semantic mask determines per-frame votes. (Center)~Accumulated view counts for each point, with the consensus threshold $\tau_{mv}$ (labeled $\tau$ in the figure) indicated. (Right)~Points meeting the threshold are retained as the object point cloud; points below the threshold are discarded as noise or background.}
\label{fig:multiview_selection}
\end{figure*}

A single-frame mask projection is inherently noisy: a 3D point on a background wall may project onto a foreground chair mask in one view through coincidental occlusion alignment. To enforce geometric consistency across viewpoints, we use a multi-view consensus voting mechanism (\figref{fig:multiview_selection}).

For each 3D point $P_j$ and each reference camera frame $i \in \{1, \ldots, N\}$, the projection and mask-lookup procedure produces a binary vote $v_{ij} \in \{0, 1\}$, where $v_{ij} = 1$ indicates that $P_j$ projected onto a valid mask pixel in frame $i$. Per-point view counts are accumulated as
\begin{equation}
    c_j = \sum_{i=1}^{N} v_{ij},
\end{equation}
and a point is admitted to the multi-view consensus set $\mathcal{C}$ if its accumulated count meets or exceeds a configurable threshold $\tau_{mv}$ (\algoref{alg:consensus}):
\begin{equation}
    \mathcal{C} = \{ P_j \mid c_j \geq \tau_{mv} \}.
\end{equation}
The depth-band filter described in the next subsubsection is applied after this consensus stage, to the consensus points only: each frame re-projects $\mathcal{C}$, and a point enters the final filtered category cloud $\mathcal{P}^{*}$ if its depth falls within the local depth band in at least one frame (the per-frame union in \algoref{alg:consensus}). The union is deliberately permissive at this stage: the consensus test has already removed spurious single-view projections, and the depth band is applied to reject points lying far behind the visible surface, so requiring agreement in every frame would discard correctly segmented points that are occluded in most views.

The threshold $\tau_{mv}$ controls a recall--precision trade-off: a low threshold retains more points (including potential noise from spurious single-view projections), while a high threshold is more conservative and produces cleaner object clouds at the risk of discarding sparsely observed regions. For the mixed-furniture classroom we use $\tau_{mv} = 10$ for chairs and $\tau_{mv} = 5$ for tables (\tableref{tab:params}): chairs are visible in many frames and support a high consensus requirement, whereas tables appear in fewer views and use a lower threshold. The threshold is additionally adjusted per environment with the number of registered views available. Each retained point must be selected across at least $\tau_{mv}$ distinct viewpoints. This suppresses transient projection artifacts: background points that coincidentally align with a foreground mask in one or two views are unlikely to do so consistently across many viewpoints. It also implicitly enforces multi-view geometric consistency, since points that survive the threshold tend to lie on actual object surfaces visible from multiple angles.

\begin{algorithm}[t!]
\caption{Multi-view consensus point selection}
\label{alg:consensus}
\begin{algorithmic}[1]
\Require Point cloud $\mathcal{P}=\{P_j\}$; calibrated frames $\{(T_{cw,i},K_i,M_i)\}_{i=1}^{N}$ with binary masks $M_i$; consensus threshold $\tau_{mv}$; patch size $S_{patch}$; band width $\sigma_{depth}$
\Ensure Filtered category point cloud $\mathcal{P}^{*}$
\State $c_j \gets 0$ for all $j$ \Comment{per-point view counts}
\For{$i = 1,\ldots,N$}
    \State $(u,v,z,\mathcal{I}) \gets \Call{Project}{\mathcal{P},T_{cw,i},K_i}$ \Comment{visible projections}
    \ForAll{$j \in \mathcal{I}$ with $M_i[u_j,v_j]=1$}
        \State $c_j \gets c_j + 1$
    \EndFor
\EndFor
\State $\mathcal{C} \gets \{\, j : c_j \ge \tau_{mv} \,\}$ \Comment{multi-view consensus}
\State $\mathcal{P}^{*} \gets \emptyset$
\For{$i = 1,\ldots,N$}
    \State project $\mathcal{C}$ into frame $i$; group pixels into $S_{patch}\times S_{patch}$ patches
    \ForAll{patches $q$}
        \State $d_{med}\gets \operatorname{median}(z_q)$;\quad $\mathrm{MAD}\gets \operatorname{median}(|z_q-d_{med}|)$
        \State keep $P_j$ with $z_j\in[\,d_{med}-\sigma_{depth}\,\mathrm{MAD},\ d_{med}+\sigma_{depth}\,\mathrm{MAD}\,]$
    \EndFor
    \State $\mathcal{P}^{*} \gets \mathcal{P}^{*}\cup\{\text{points kept in frame } i\}$
\EndFor
\State \Return $\mathcal{P}^{*}$
\end{algorithmic}
\end{algorithm}

\subsubsection{Depth-Band Filtering}
\begin{figure*}[t!]
\centering
\includegraphics[width=0.94\linewidth,trim=0 30 0 30,clip]{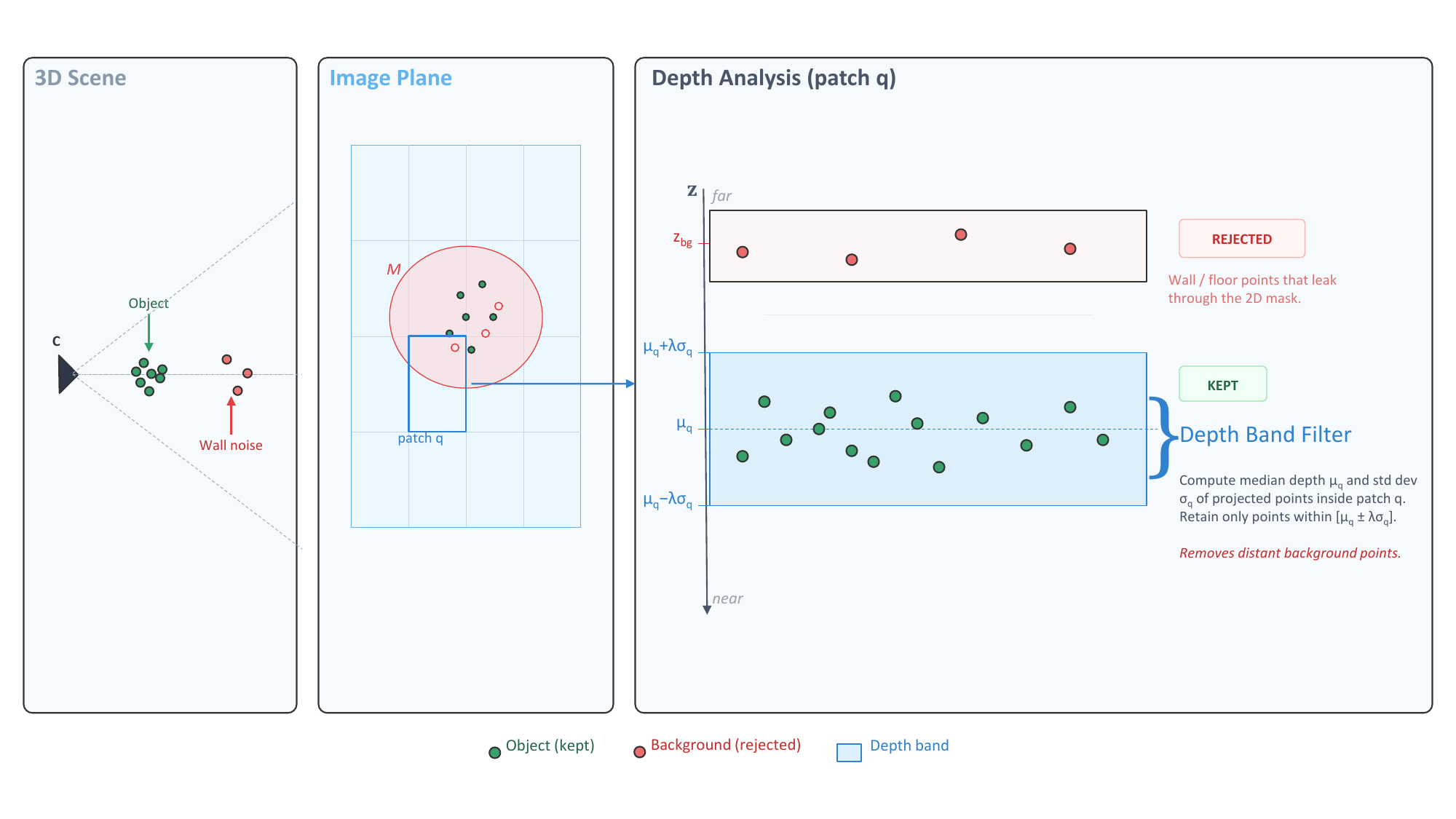}
\caption{Depth-band filtering. (Left)~3D scene with object points and background wall noise viewed from camera $C$. (Center)~Image plane showing the semantic mask $M$ and a selected patch $q$. (Right)~Depth analysis for patch $q$: the median depth $\mu_q$ and spread $\sigma_q$ define a depth band $[\mu_q - \lambda\sigma_q,\ \mu_q + \lambda\sigma_q]$. Points within the band (green) are retained; distant background points at $z_{\mathrm{bg}}$ (red) that leak through the 2D mask are rejected. The patch statistics $\mu_q$, $\sigma_q$, and multiplier $\lambda$ shown in the figure correspond to $d_{med}$, $\mathrm{MAD}$, and $\sigma_{depth}$ in the text.}
\label{fig:depth_band}
\end{figure*}

We apply a patch-based depth-band filter to resolve occlusions, for example a point on a back wall projecting onto a foreground chair mask (\figref{fig:depth_band}). The image is divided into patches of size $S_{patch} \times S_{patch}$. For each patch, we compute the median depth $d_{med}$ and the median absolute deviation (MAD), and define a valid depth band parametrized by $\sigma_{depth}$:
\begin{equation}
    \mathrm{Band} = \left[\, d_{med} - \sigma_{depth} \cdot \mathrm{MAD},\ \ d_{med} + \sigma_{depth} \cdot \mathrm{MAD} \,\right].
\end{equation}
Points falling outside this band are discarded as background clutter.

\subsubsection{Instance Extraction via Octree-Based Connected Components}
\begin{figure}[t!]
\centering
\begin{subfigure}[b]{0.24\linewidth}
    
    \centering
    \includegraphics[width=\linewidth]{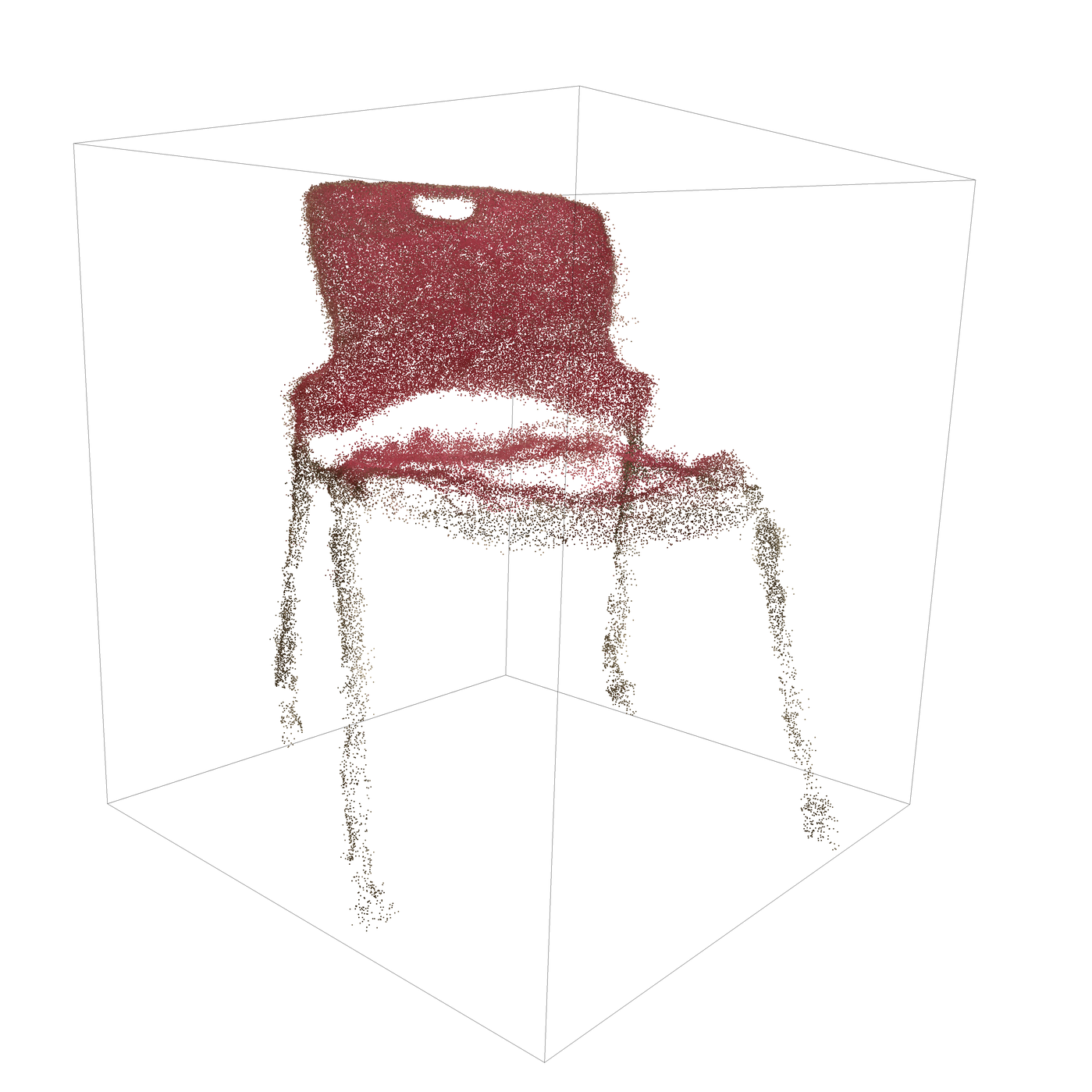}
    \caption{Chair instance 1}
\end{subfigure}\hfill
\begin{subfigure}[b]{0.24\linewidth}
    
    \centering
    \includegraphics[width=\linewidth]{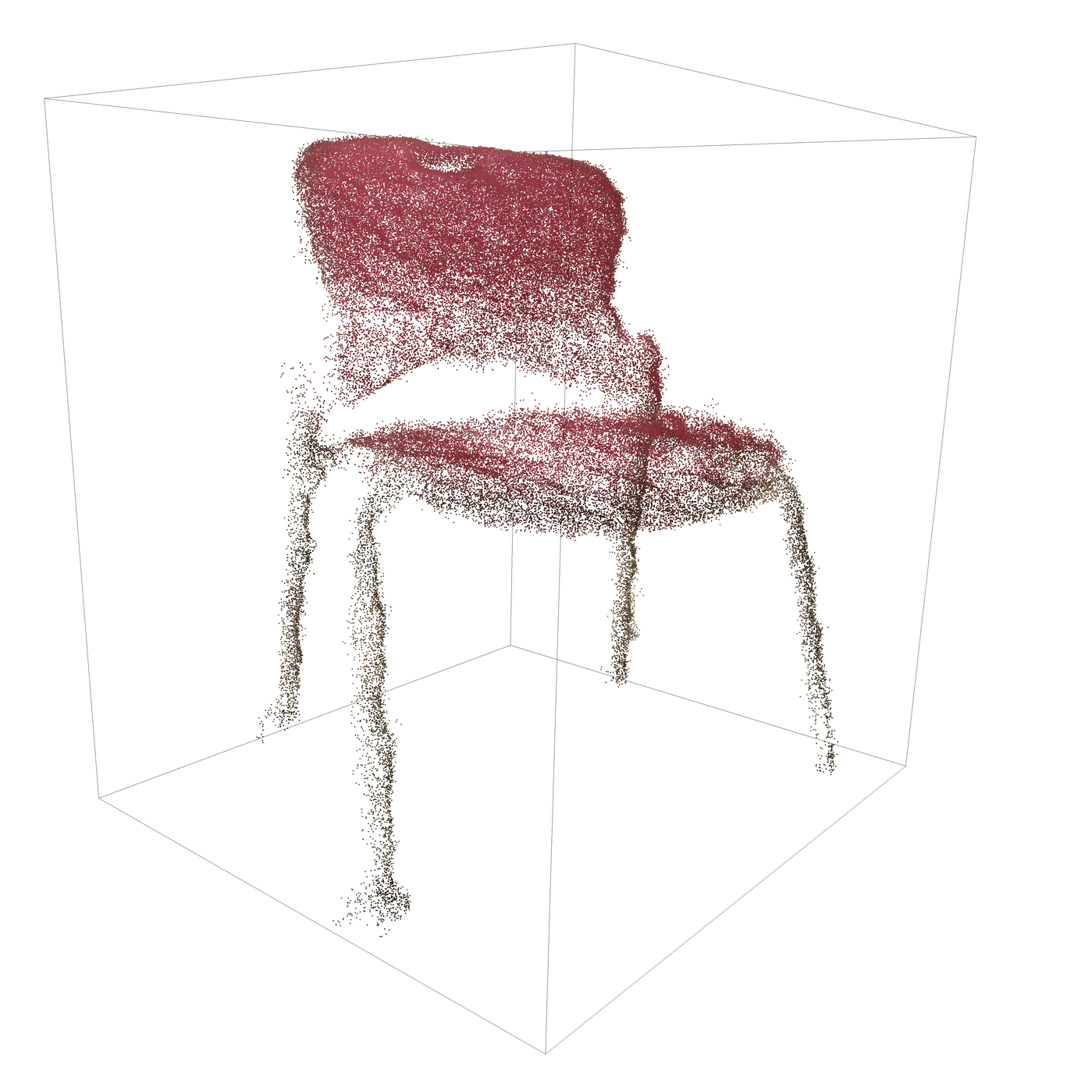}
    \caption{Chair instance 2}
\end{subfigure}\hfill
\begin{subfigure}[b]{0.24\linewidth}
    
    \centering
    \includegraphics[width=\linewidth]{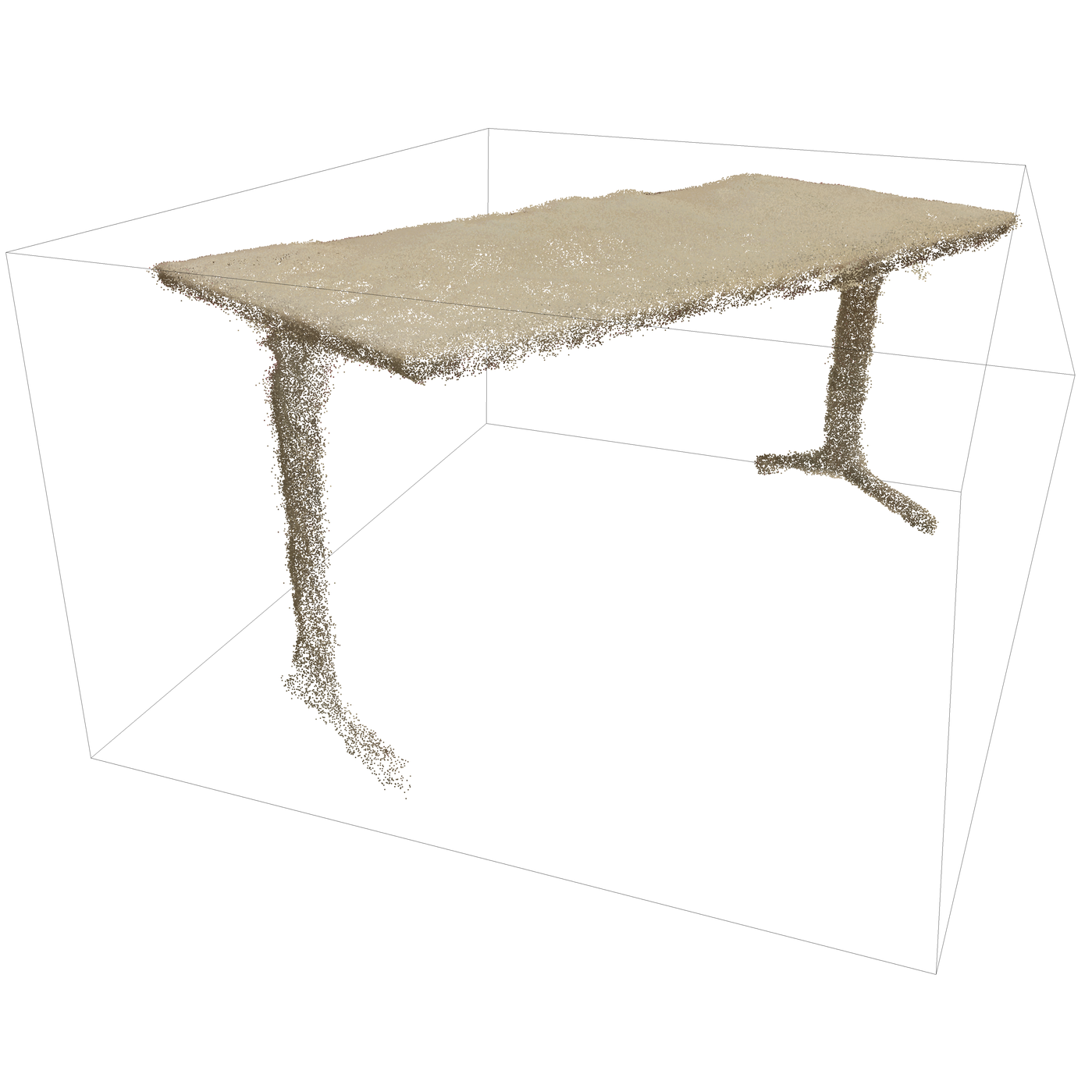}
    \caption{Table instance}
\end{subfigure}\hfill
\begin{subfigure}[b]{0.24\linewidth}
    
    \centering
    \includegraphics[width=\linewidth]{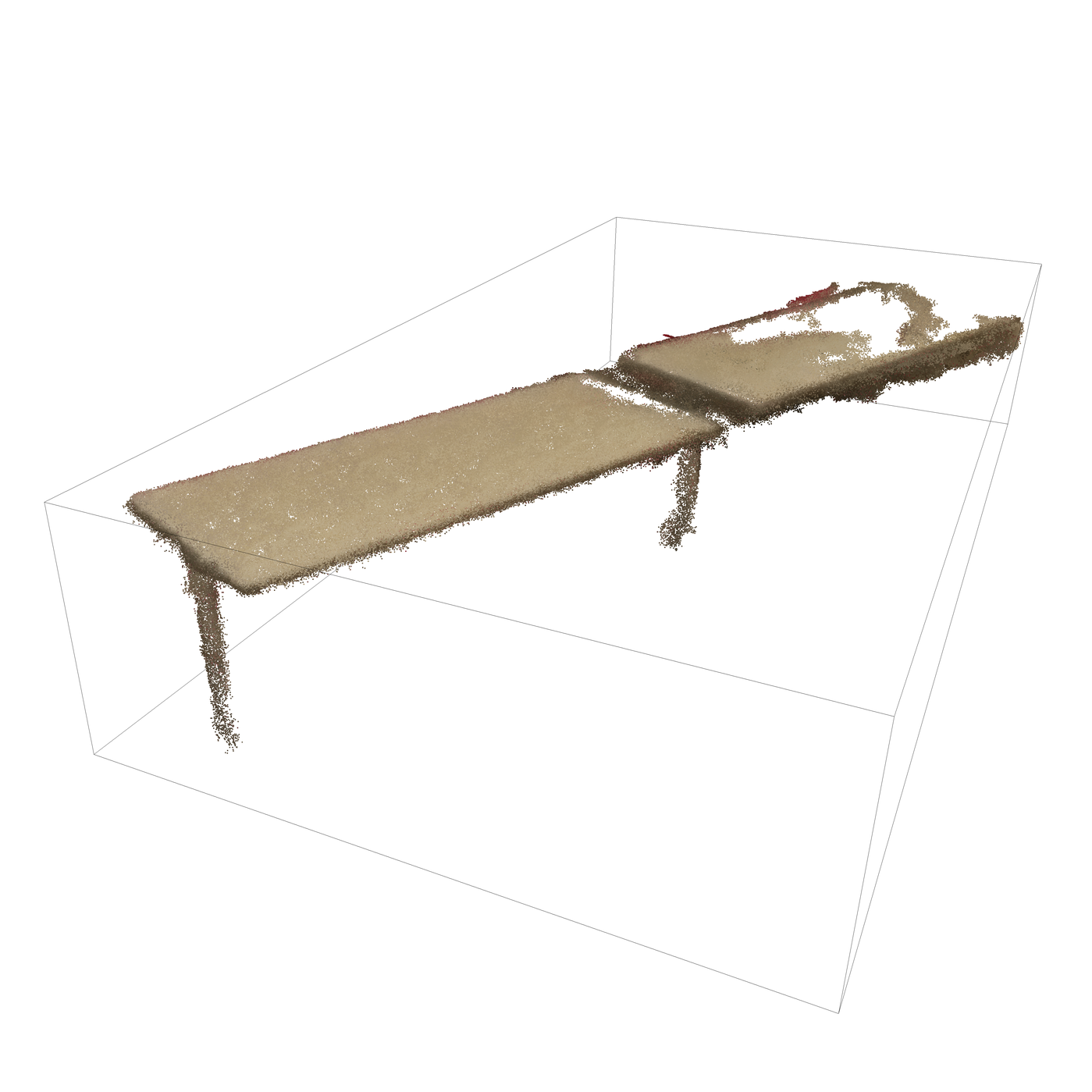}
    \caption{Merged instance}
\end{subfigure}
\caption{Individual object instances extracted via octree-based connected components from the category cloud. Quality of extracted instances depends on the underlying point cloud: (a,\,b)~well-separated chairs yield clean single-object instances, (c)~an isolated table is correctly extracted, and (d)~two adjacent tables with insufficient spatial gap are merged into a single connected component, with a nearby chair fragment also included.}
\label{fig:instances}
\end{figure}

The output of the previous steps is a category point cloud in which all chairs (or all tables) are merged into a single set of points spread across the room; individual instances must still be separated. We use an octree-based union--find algorithm (\algoref{alg:octree}) to perform this segmentation. The space is voxelized at a resolution defined by level $L_{octree}$, occupied voxels form the nodes of a graph, and 26-connectivity (face, edge, and corner adjacency) defines the edges. Distinct connected components in this graph correspond to spatially separated objects (\figref{fig:instances}). Components with fewer than $N_{min}$ points are discarded as noise. At $L_{octree}=11$, the voxel size is approximately $1/2000$ of the category cloud's largest bounding-box extent, and the healing gap threshold $\delta_{gap}$ (\tableref{tab:params}) exceeds the corresponding voxel diagonal by a factor of four or more, so the subsequent gap bridging operates well above the voxelization resolution. The result is a set of individual instance point clouds, ready for geometric healing. For the Mixed-Furniture Classroom, this step produced 35 chair clusters and 16 table clusters prior to size-based pruning, yielding 32 chairs and 16 tables after pruning and human quality-assurance (QA) review (see \tableref{tab:recon_counts}).

\begin{table}[t!]
\centering
\caption{Ground-truth and final reconstruction furniture counts per environment. Ground truth was verified manually; final reconstruction counts are after the full pipeline including human QA correction.}
\label{tab:recon_counts}
\setlength{\tabcolsep}{6pt}
\begin{tabular}{lrrrr}
\toprule
 & \multicolumn{2}{c}{Ground truth} & \multicolumn{2}{c}{Final reconstruction} \\
\cmidrule(lr){2-3} \cmidrule(lr){4-5}
\textbf{Environment} & \textbf{Chairs} & \textbf{Tables} & \textbf{Chairs} & \textbf{Tables} \\
\midrule
Mixed-Furniture Classroom & 32  & 17 & 32  & 16 \\
Chair-Dominant Classroom  & 47  & 3  & 45  & 2  \\
Auditorium                & 242 & 2  & 273 & 2  \\
\bottomrule
\end{tabular}
\end{table}

\begin{algorithm}[t!]
\caption{Octree-based instance extraction}
\label{alg:octree}
\begin{algorithmic}[1]
\Require Category cloud $\mathcal{P}^{*}$; octree level $L_{octree}$; minimum size $N_{min}$
\Ensure Instance point clouds $\{\mathcal{O}_k\}$
\State $s \gets \dfrac{\max\big(\text{bounding-box extent of }\mathcal{P}^{*}\big)}{2^{L_{octree}}}$ \Comment{voxel size}
\State $\mathbf{v}_j \gets \lfloor (P_j-\mathbf{p}_{\min})/s \rfloor$ for all $j$ \Comment{voxel index}
\State initialise union--find with each point in its own set
\ForAll{occupied voxels $\mathbf{v}$}
    \State union all points sharing voxel $\mathbf{v}$
    \ForAll{occupied 26-neighbours $\mathbf{v}'$ of $\mathbf{v}$}
        \State $\Call{Union}{\mathbf{v},\mathbf{v}'}$ \Comment{merge voxel representatives}
    \EndFor
\EndFor
\State label connected components; discard those with $<N_{min}$ points
\State \Return one instance cloud $\mathcal{O}_k$ per surviving component
\end{algorithmic}
\end{algorithm}

\subsection{Geometric Alignment and Healing}
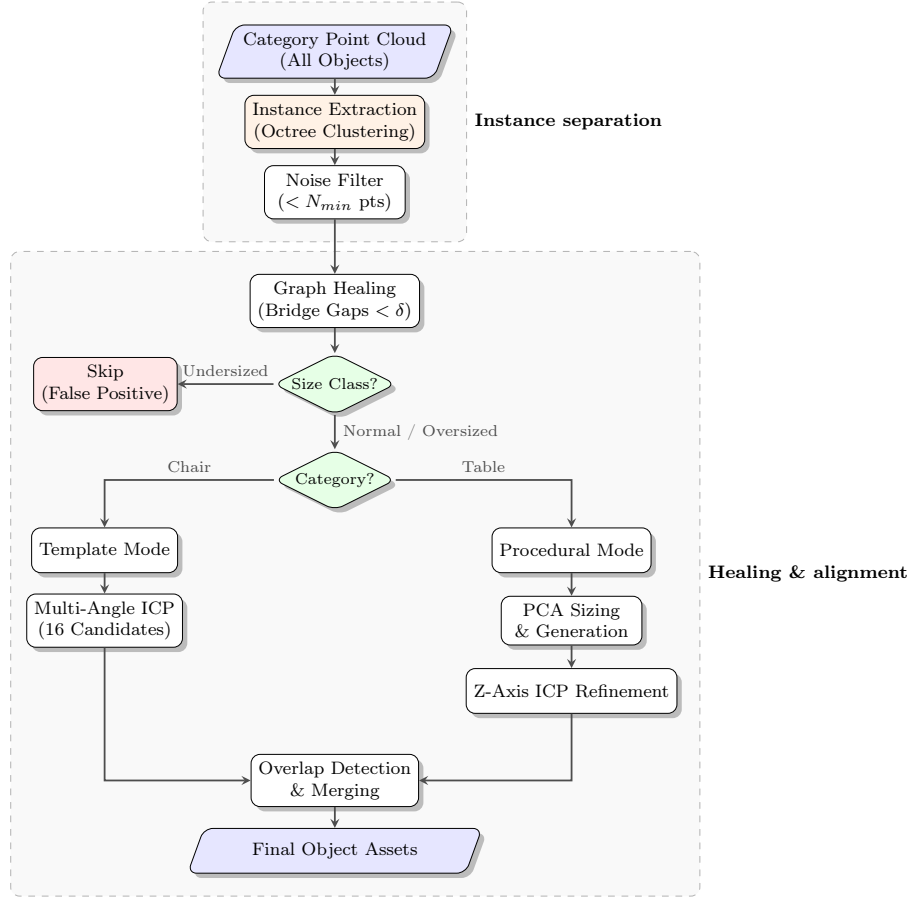
\begin{figure}[t!]
\centering
\resizebox{0.74\linewidth}{!}{
\begin{tikzpicture}[node distance=1.1cm, auto,
    sprocess/.style={basic, rectangle, fill=white, minimum height=0.7cm, minimum width=2.2cm, font=\footnotesize},
    sio/.style={basic, trapezium, trapezium left angle=70, trapezium right angle=110, fill=blue!10, minimum height=0.7cm, font=\footnotesize},
    sdecision/.style={basic, diamond, fill=green!10, aspect=2, inner sep=1pt, font=\scriptsize}]
    \node (start) [sio] {Category Point Cloud \\ (All Objects)};
    \node (cluster) [sprocess, below of=start, fill=orange!10] {Instance Extraction \\ (Octree Clustering)};
    \node (filter) [sprocess, below of=cluster] {Noise Filter \\ ($< N_{min}$ pts)};
    \node (heal) [sprocess, below of=filter, yshift=-0.6cm] {Graph Healing \\ (Bridge Gaps $<\delta$)};
    \node (size) [sdecision, below of=heal, yshift=-0.2cm] {Size Class?};
    \node (skip) [sprocess, left=1.5cm of size, fill=red!10] {Skip \\ (False Positive)};
    \node (decide) [sdecision, below of=size, yshift=-0.4cm] {Category?};
    \node (chair) [sprocess, left=1.5cm of decide, yshift=-1.1cm] {Template Mode};
    \node (icp_chair) [sprocess, below of=chair] {Multi-Angle ICP \\ (16 Candidates)};
    \node (table) [sprocess, right=1.5cm of decide, yshift=-1.1cm] {Procedural Mode};
    \node (pca) [sprocess, below of=table] {PCA Sizing \\ \& Generation};
    \node (icp_table) [sprocess, below of=pca] {Z-Axis ICP Refinement};
    \node (overlap) [sprocess, below of=decide, yshift=-3.6cm] {Overlap Detection \\ \& Merging};
    \node (final) [sio, below of=overlap] {Final Object Assets};

    \draw [arrow] (start) -- (cluster);
    \draw [arrow] (cluster) -- (filter);
    \draw [arrow] (filter) -- (heal);
    \draw [arrow] (heal) -- (size);
    \draw [arrow] (size) -- node[above, font=\scriptsize] {Undersized} (skip);
    \draw [arrow] (size) -- node[right, font=\scriptsize] {Normal / Oversized} (decide);
    \draw [arrow] (decide) -| node[above, near start, font=\scriptsize] {Chair} (chair);
    \draw [arrow] (chair) -- (icp_chair);
    \draw [arrow] (icp_chair) |- (overlap);
    \draw [arrow] (decide) -| node[above, near start, font=\scriptsize] {Table} (table);
    \draw [arrow] (table) -- (pca);
    \draw [arrow] (pca) -- (icp_table);
    \draw [arrow] (icp_table) |- (overlap);
    \draw [arrow] (overlap) -- (final);

    \begin{scope}[on background layer]
        \node [group, fit=(start) (cluster) (filter), label=right:{\footnotesize\textbf{Instance separation}}] {};
        \node [group, fit=(heal) (skip) (overlap) (final) (table) (icp_table), label=right:{\footnotesize\textbf{Healing \& alignment}}] {};
    \end{scope}
\end{tikzpicture}
}
\caption{Instance-processing pipeline: instance separation, noise filtering, geometric healing, size classification, and category-specific alignment producing the final CFD-ready assets.}
\label{fig:alignment}
\end{figure}

The extracted instance point clouds are often fragmented due to occlusion (e.g., legs separated from seats) or NeRF density artifacts. A graph-based healing algorithm followed by CAD-template or procedural alignment converts these instances into watertight surface assets (\figref{fig:alignment}). A human-in-the-loop scene editor (\secref{sec:human_qa}) provides quality assurance and parametric scene augmentation before the geometry is consumed by the CFD solver.

\subsubsection{Graph-Based Point Cloud Healing}
\begin{figure}[t!]
\centering
\begin{subfigure}[b]{0.4\linewidth}
    \centering
    \includegraphics[width=\linewidth]{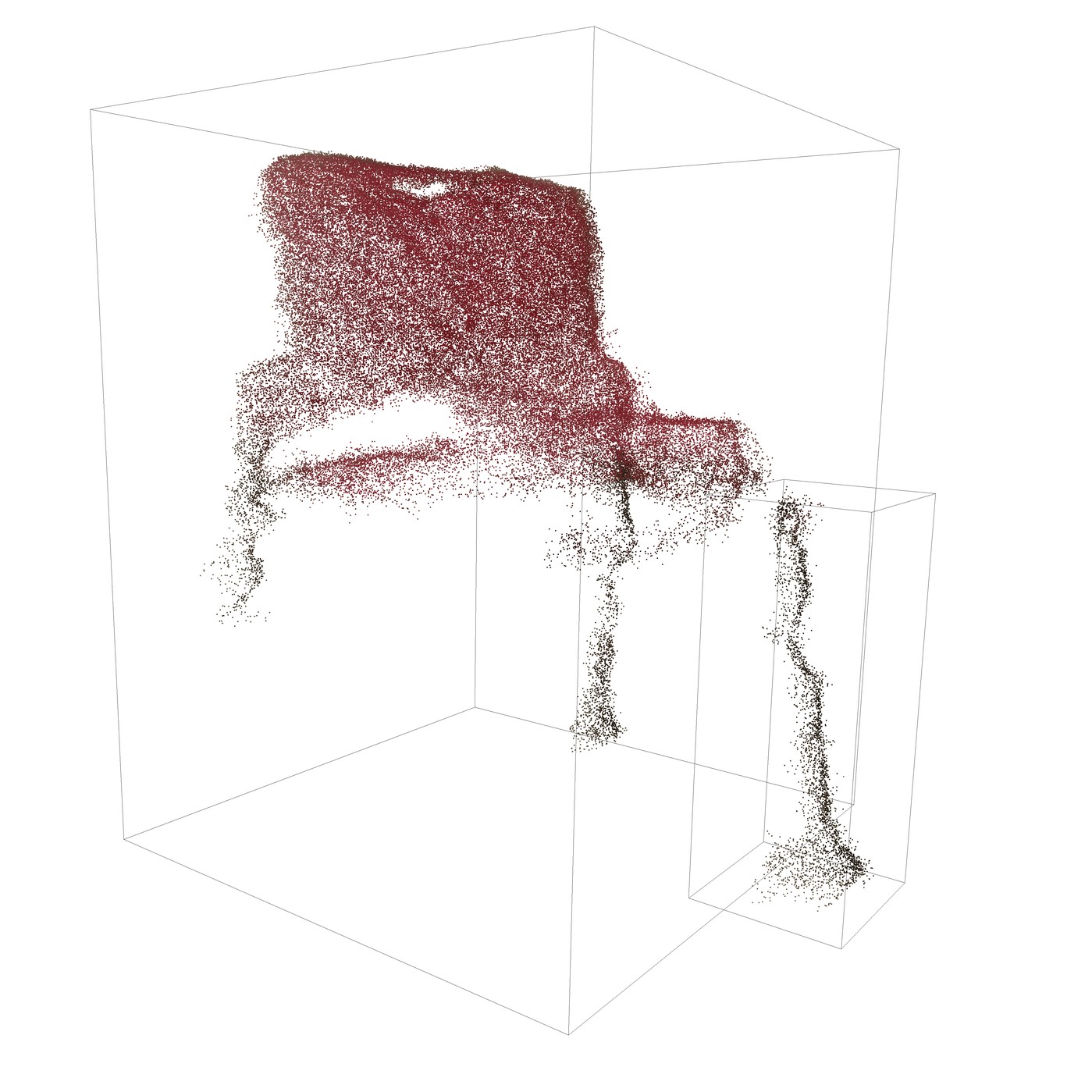}
    \caption{Before healing}
\end{subfigure}
\begin{subfigure}[b]{0.4\linewidth}
    \centering
    \includegraphics[width=\linewidth]{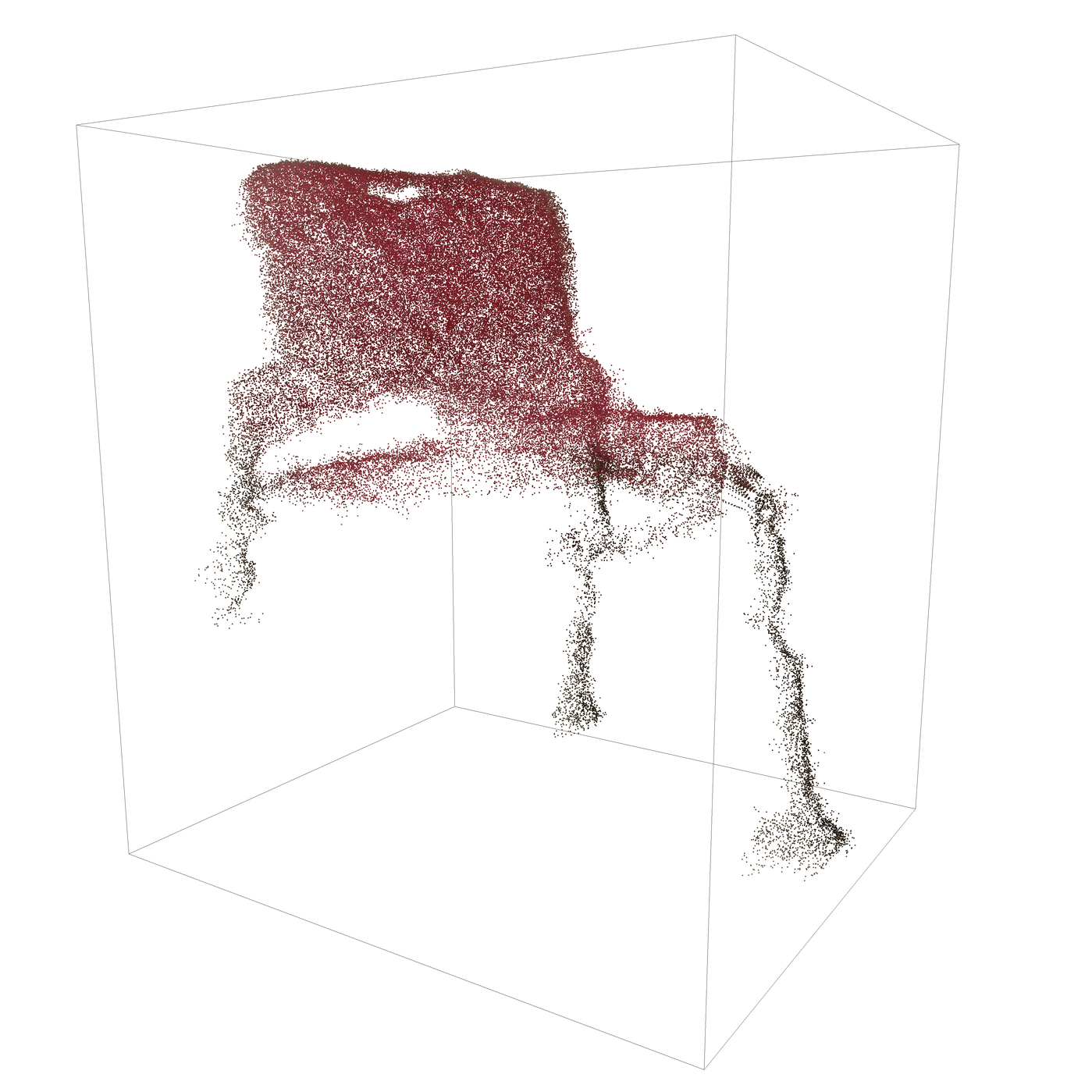}
    \caption{After healing}
\end{subfigure}
\caption{Effect of graph-based healing on a fragmented chair instance. (a)~A front leg is disconnected from the main body due to occlusion gaps in the NeRF reconstruction, appearing as a separate connected component within its own bounding box. (b)~After bridging gaps smaller than $\delta_{gap}$, the leg is merged back into a single unified point cloud.}
\label{fig:healing_result}
\end{figure}

We repair fragmented objects by constructing a connectivity graph $G=(V, E)$ (\algoref{alg:healing}) in which the nodes $V$ are the disconnected components of a single instance cluster and an edge $(u, v)$ is created if the minimum nearest-neighbour distance between the points of $u$ and the points of $v$ is less than a gap threshold $\delta_{gap}$. Distance evaluation uses a sliced Kd-tree to avoid global distance calculations. Detected gaps are filled by linear interpolation of point position and color, producing a unified per-instance point cloud (\figref{fig:healing_result}).

\begin{algorithm}[t!]
\caption{Graph-based fragment healing}
\label{alg:healing}
\begin{algorithmic}[1]
\Require Instance fragments $\{F_m\}$; gap threshold $\delta_{gap}$; fill spacing $r$
\Ensure Healed (merged) instance clouds
\State graph $G \gets (V,E)$ with one node per fragment, $E \gets \emptyset$
\ForAll{fragment pairs $(F_a,F_b)$ co-occurring in a spatial slab}
    \State $d_{ab} \gets \min_{p\in F_a}\operatorname{dist}(p,F_b)$ \Comment{sliced KD-tree}
    \If{$d_{ab} \le \delta_{gap}$}
        \State $E \gets E \cup \{(a,b)\}$
        \State insert points along each bridging pair by linear interpolation of position and colour at spacing $r$
    \EndIf
\EndFor
\State \Return one merged cloud per connected component of $G$
\end{algorithmic}
\end{algorithm}

\subsubsection{Size Classification and Post-Processing}
\label{sec:size_classification}
After healing, each instance cluster is classified by comparing its axis-aligned bounding-box dimensions against expected size ranges for the given object category. This classification determines how the cluster is processed:
\begin{itemize}
    \item \emph{Undersized clusters} (bounding box significantly smaller than the expected object dimensions) are classified as false positives, typically residual noise fragments that survived earlier filtering, and are discarded.
    \item \emph{Normal-sized clusters} (bounding box within the expected range for a single object) receive a single STL placement via the template or procedural pipeline described below.
    \item \emph{Oversized clusters} (bounding box exceeding the expected range, typically $1.5\times$ or more of the single-object dimensions along one or more axes) indicate that multiple objects were merged into a single connected component due to spatial proximity (e.g., adjacent chairs touching). These clusters receive multiple STL placements: the bounding box is subdivided along its longest axis into segments matching the expected single-object width, and an independent STL is placed at each subdivision centroid.
\end{itemize}

An overlap-detection pass then checks for intersecting bounding boxes between placed STL models. When two placements overlap beyond a configurable intersection-over-union threshold, the lower-fitness placement is removed to prevent degenerate CFD geometry. The final output is a set of per-object STL files together with a combined \texttt{furniture.stl} asset containing all placed objects.

\subsubsection{Model Integration via ICP}
The final geometric step integrates simulation-ready surfaces using the point-to-plane Iterative Closest Point (ICP) algorithm \citep{Besl1992ICP, Chen1992ICPRange} under two strategies:

\paragraph{Template matching (chairs)}
For standardized objects, a CAD template is used. A multi-start ICP is initialized at the centroid of the target cloud and searches over 16 candidate rotations about the vertical axis in two phases: a coarse phase evaluates eight candidates at $45^\circ$ increments spanning the full $360^\circ$, and a fine phase evaluates eight additional candidates at $5^\circ$ increments within $\pm20^\circ$ of the best coarse angle. Each candidate is refined by point-to-plane ICP for up to $K_{icp}$ iterations, and the alignment with the highest fitness score is selected. Fitness is Open3D's inlier fraction: the proportion of template sample points with a target correspondence within the ICP distance threshold (range 0--1, higher is better). The correspondence distance is set adaptively from the target cloud density (the configured threshold, enlarged to at least three times the estimated point spacing), and surface normals are estimated within a radius of at least five times the point spacing. The iteration budgets differ by an order of magnitude ($K_{icp}=500$ for chairs versus $50$ for tables, \tableref{tab:params}) because template matching must correct an arbitrary initial orientation through the multi-start search, whereas the procedural table mesh starts from a PCA-aligned initialization and refines only the vertical-axis rotation. Placements with fitness below a threshold $\tau_{fit}$ are flagged as low-confidence; the threshold is tuned to the target geometry. Chairs use a permissive value ($\tau_{fit}=0.05$) because the fixed CAD template only approximates the captured chairs and partial overlap is expected, whereas the procedurally generated tables are dimensioned from the observed point cloud itself, so a substantially higher fitness ($\tau_{fit}=0.3$) is required before an alignment is accepted.

\begin{figure*}[t!]
\centering
\begin{subfigure}[b]{0.24\linewidth}
    \centering
    \includegraphics[width=\linewidth]{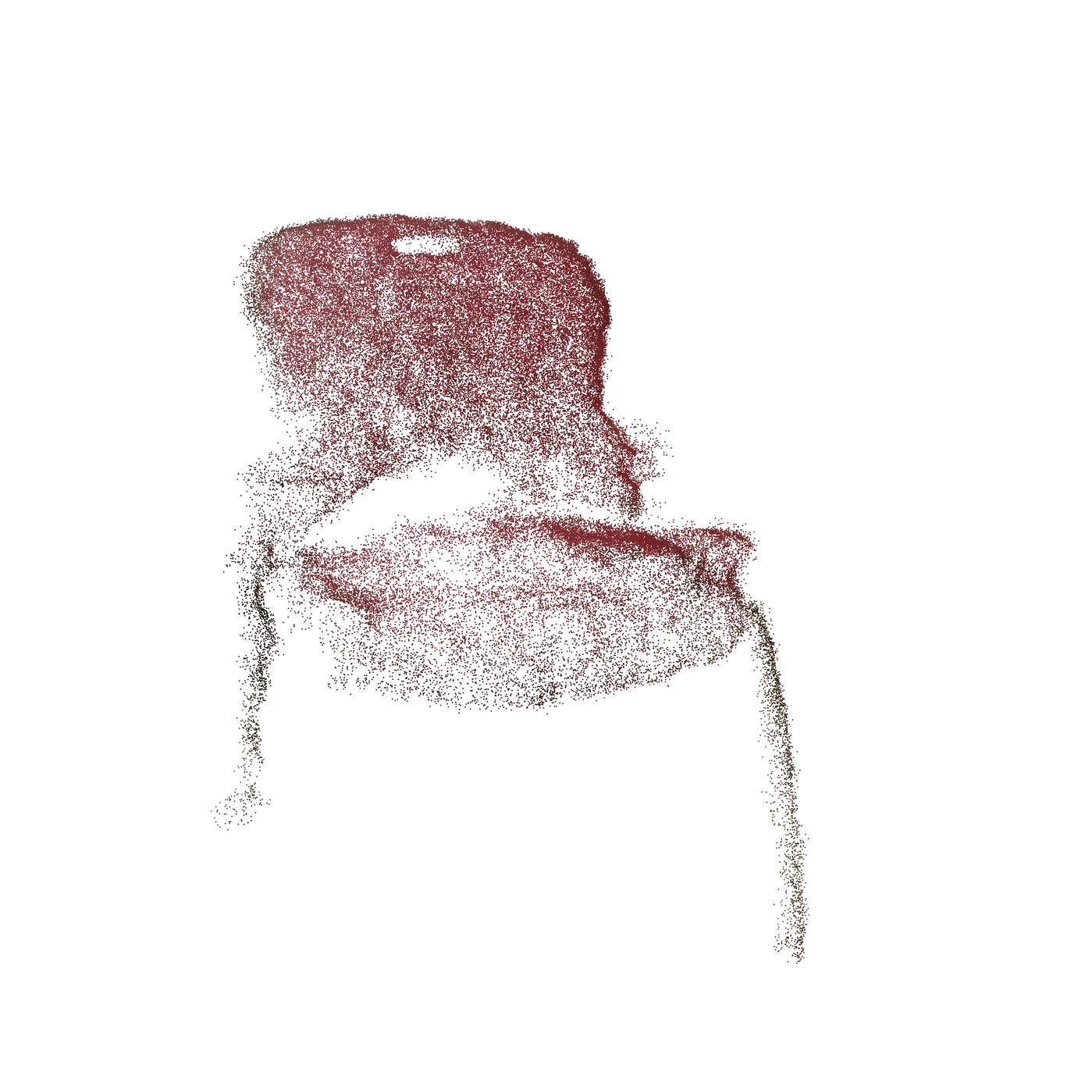}\\[2pt]
    \includegraphics[width=\linewidth]{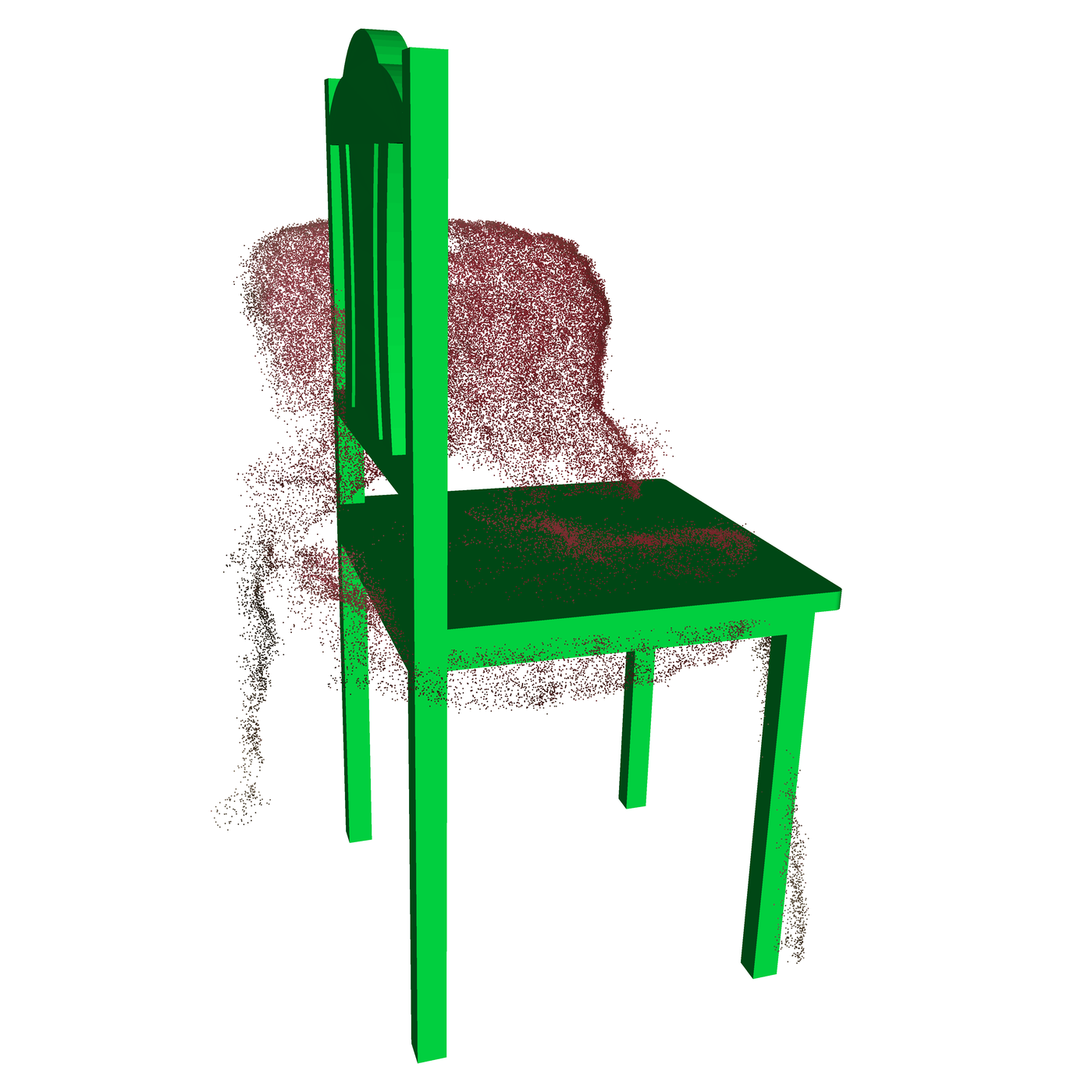}
    \caption{Chair (misaligned)}
\end{subfigure}\hfill
\begin{subfigure}[b]{0.24\linewidth}
    \centering
    \includegraphics[width=\linewidth]{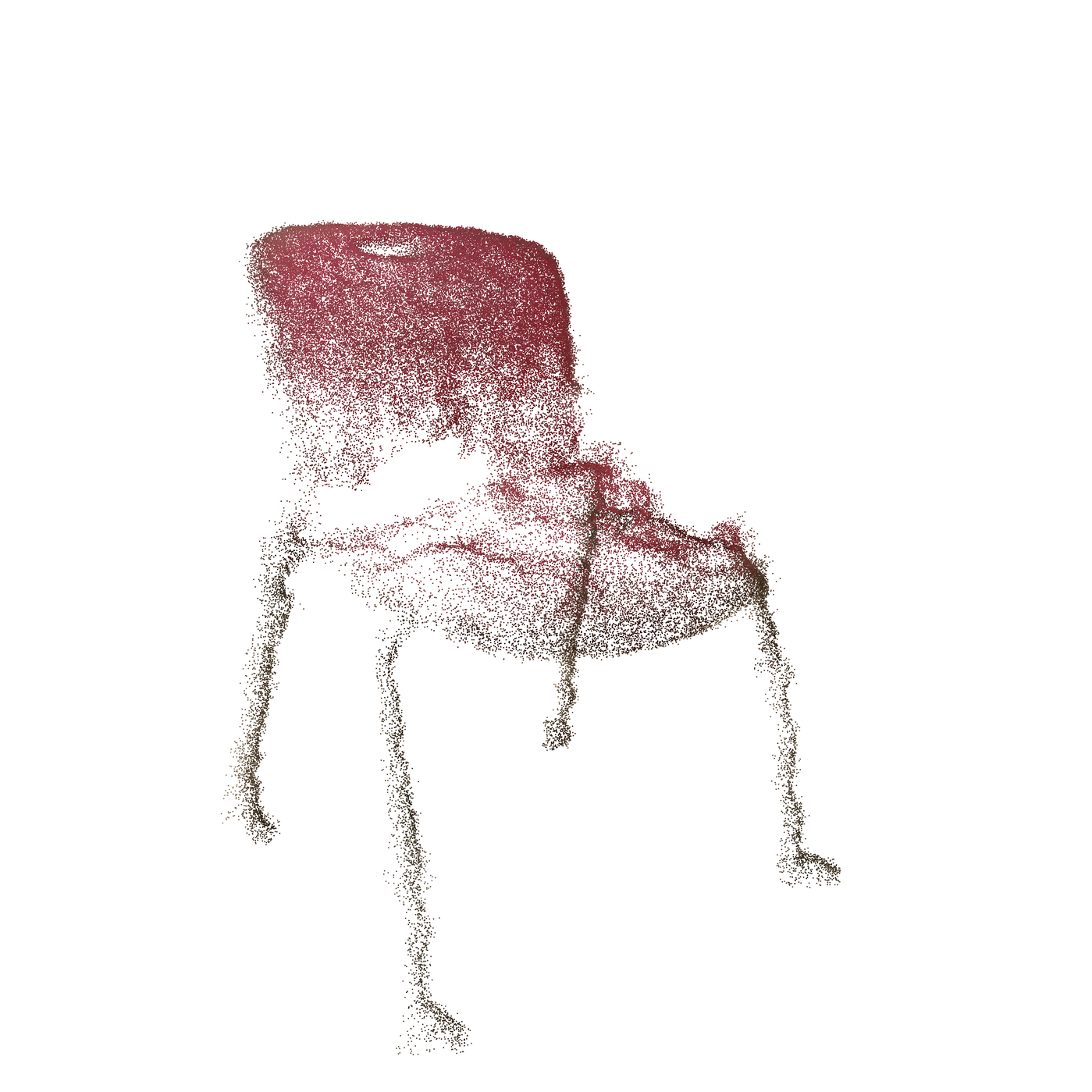}\\[2pt]
    \includegraphics[width=\linewidth]{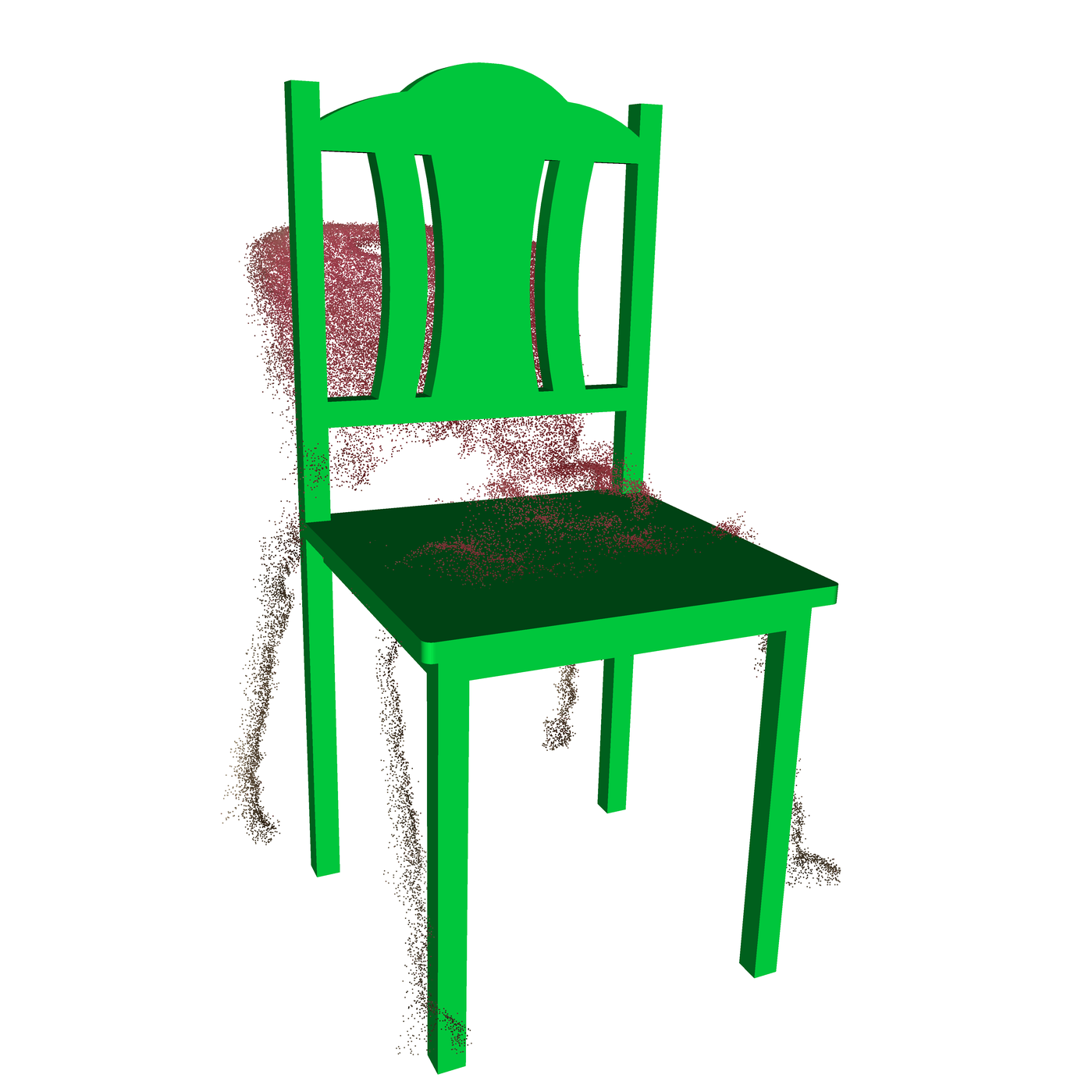}
    \caption{Chair (aligned)}
\end{subfigure}\hfill
\begin{subfigure}[b]{0.24\linewidth}
    \centering
    \includegraphics[width=\linewidth]{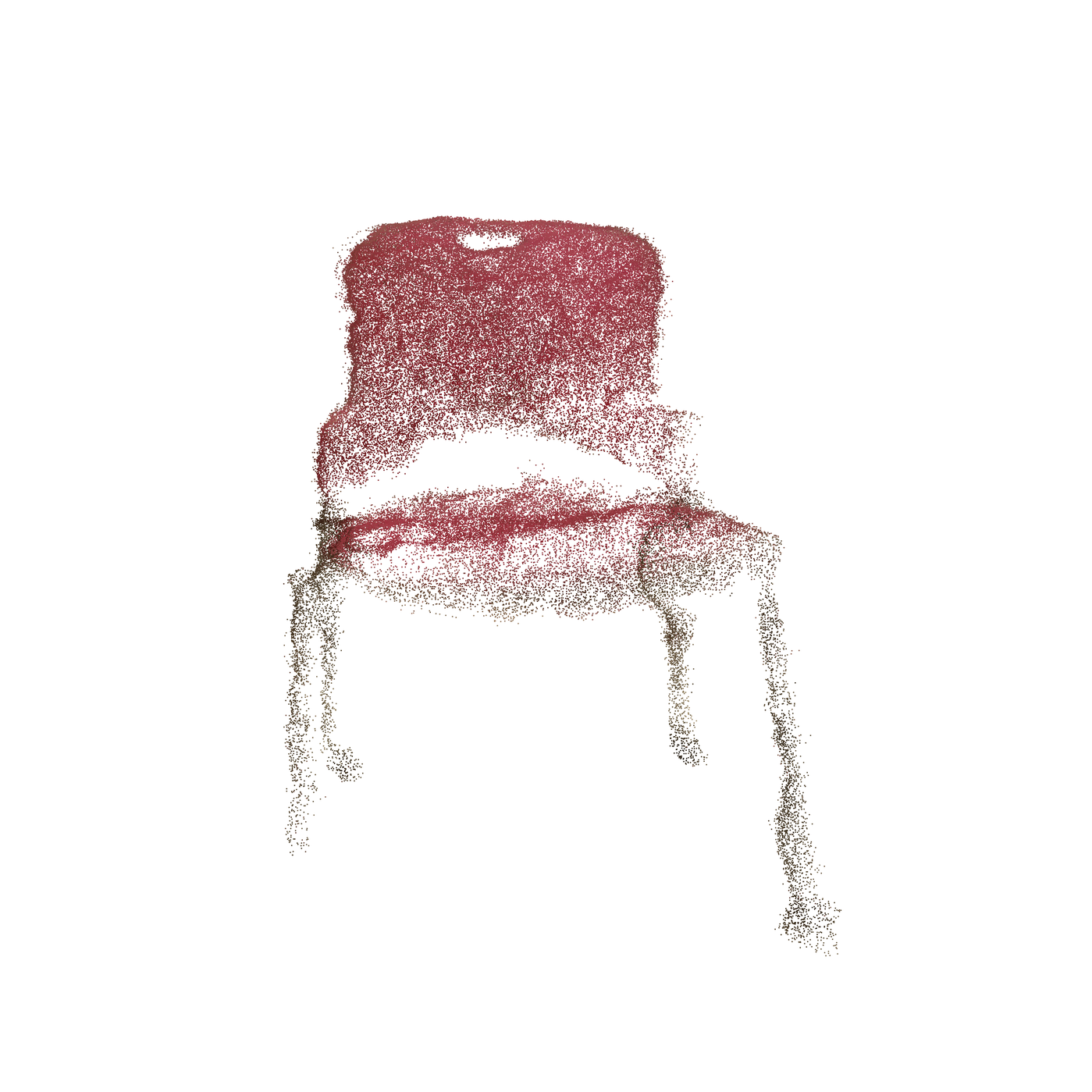}\\[2pt]
    \includegraphics[width=\linewidth]{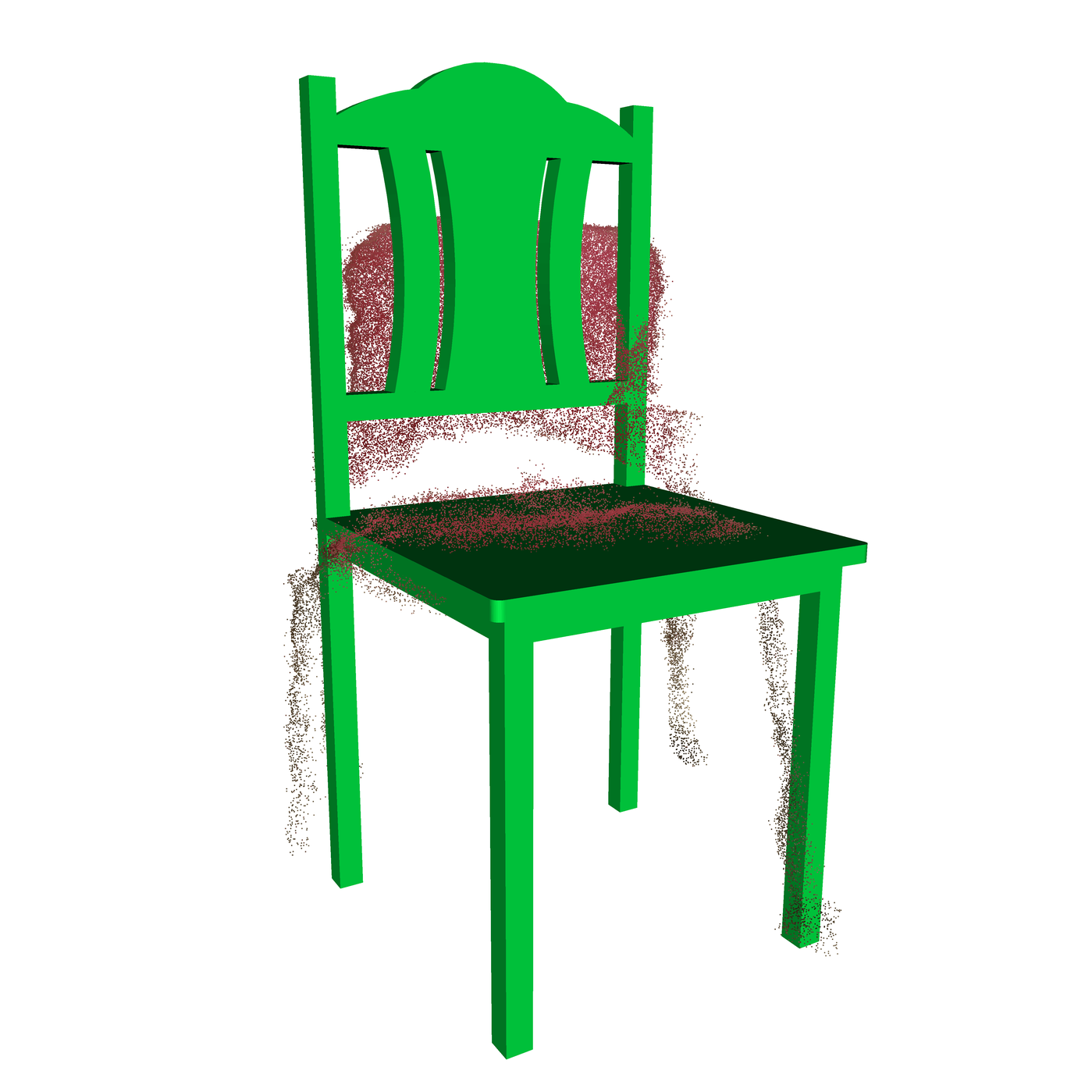}
    \caption{Chair (aligned)}
\end{subfigure}\hfill
\begin{subfigure}[b]{0.24\linewidth}
    \centering
    \includegraphics[width=\linewidth]{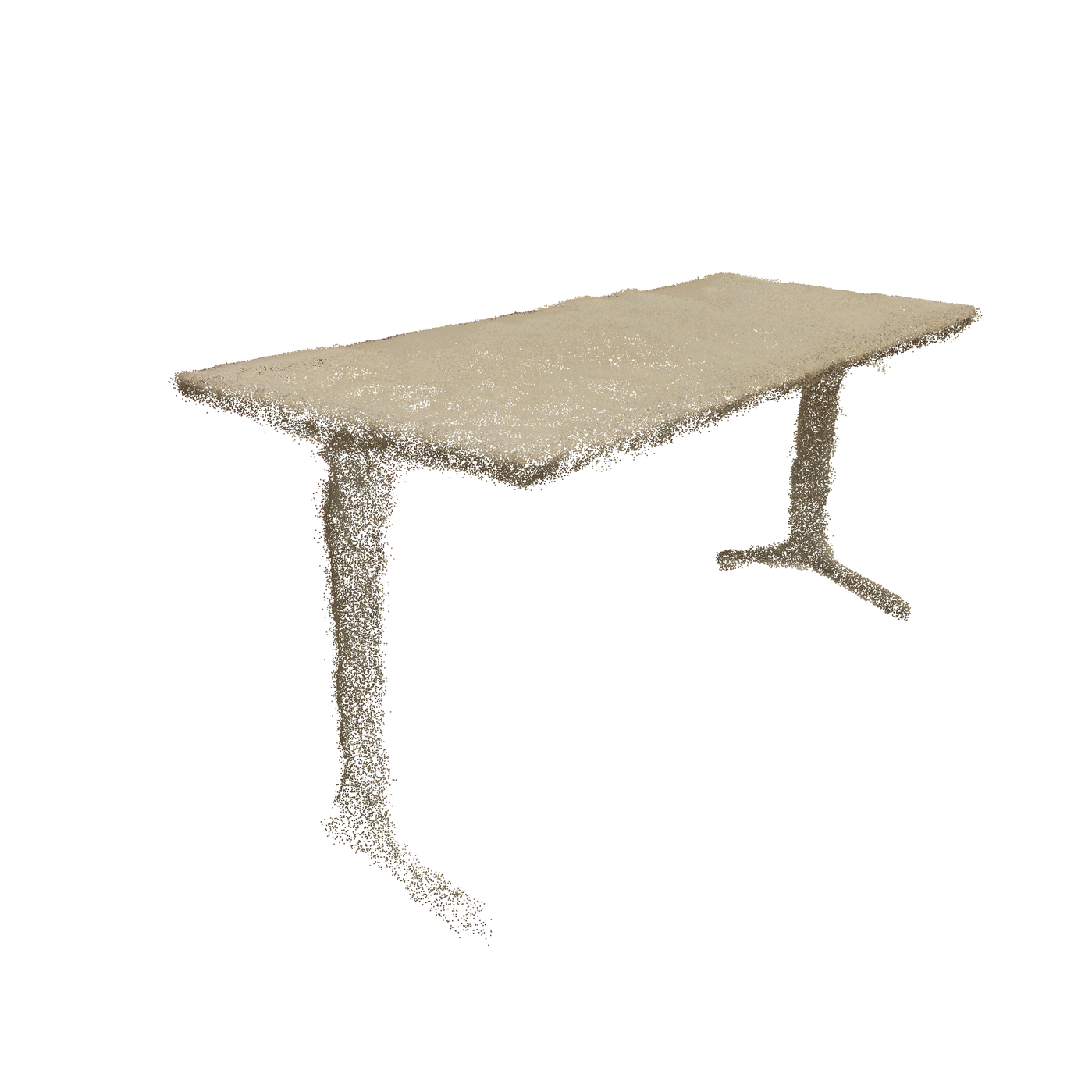}\\[2pt]
    \includegraphics[width=\linewidth]{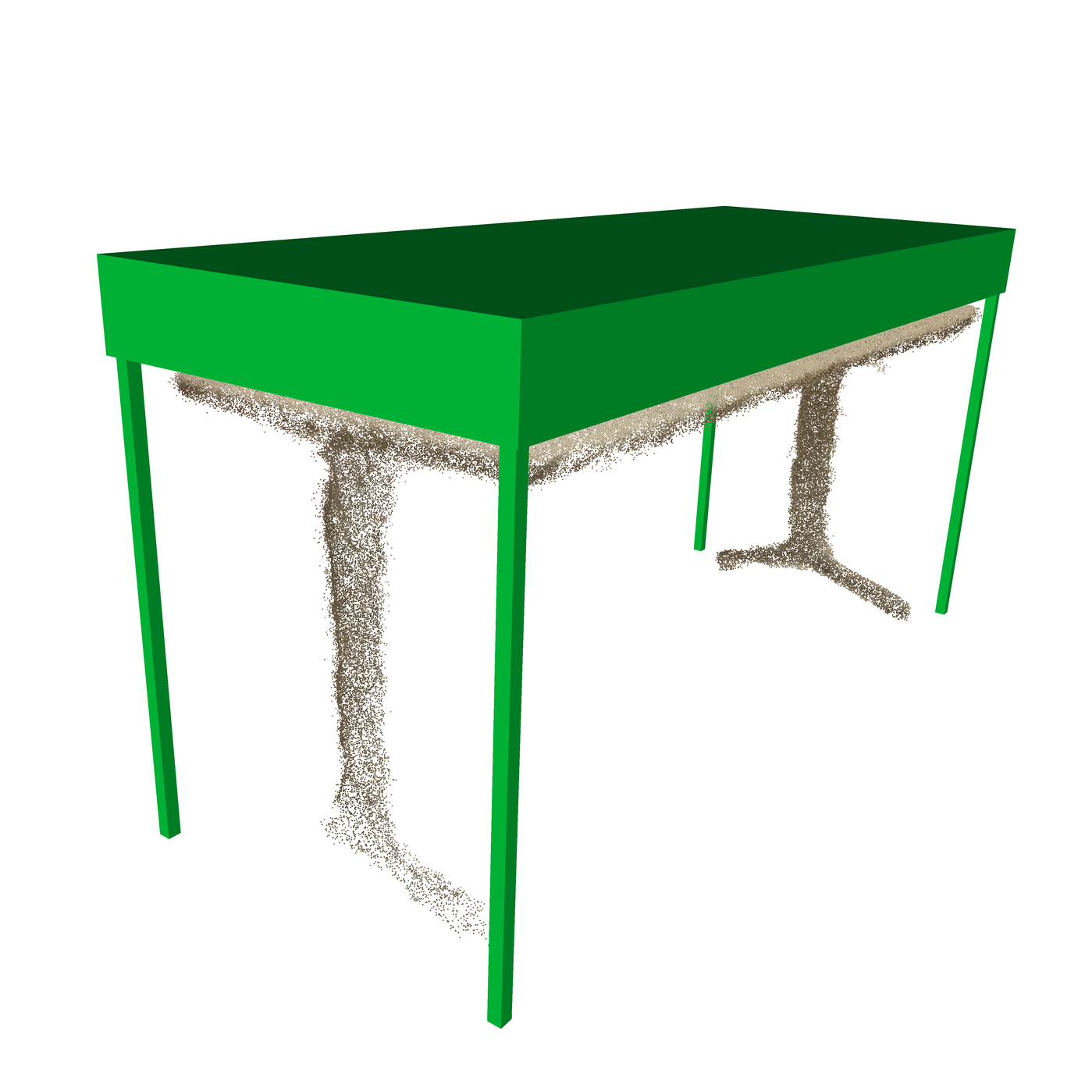}
    \caption{Table (aligned)}
\end{subfigure}
\caption{Final geometric alignment results. Each column shows the healed point cloud (top) and the corresponding STL template overlay in green (bottom). (a)~Failure case where multi-start ICP converges to a local minimum, placing the chair template approximately $90^{\circ}$ from the correct orientation. (b,\,c)~Successful chair alignments where the template closely matches the point cloud geometry. (d)~Procedurally generated table mesh fitted via PCA-based orientation and $z$-axis ICP refinement, in close agreement with the scanned point cloud.}
\label{fig:final_alignment}
\end{figure*}

\paragraph{Procedural generation (tables)}
For variable-dimension objects, geometry is generated procedurally. Principal Component Analysis (PCA) \citep{Jolliffe2016PCA} on the projected point cloud extracts the principal axes ($v_1, v_2$) corresponding to the table's length and width. A parametric mesh is generated from these dimensions and refined via single-pass ICP restricted to $z$-axis rotation (\figref{fig:final_alignment}). The procedural model represents a table as a box slab (tabletop) with four box legs, with all dimensions derived directly from the point cloud extents. This avoids the need for a template library for geometrically simple objects and should extend to other regular shapes such as shelves or ventilation ducts.

\subsubsection{Human-in-the-Loop Scene Editing}
\label{sec:human_qa}
\begin{figure*}[t!]
\centering
\begin{subfigure}[b]{0.52\linewidth}
    \centering
    \includegraphics[width=\linewidth]{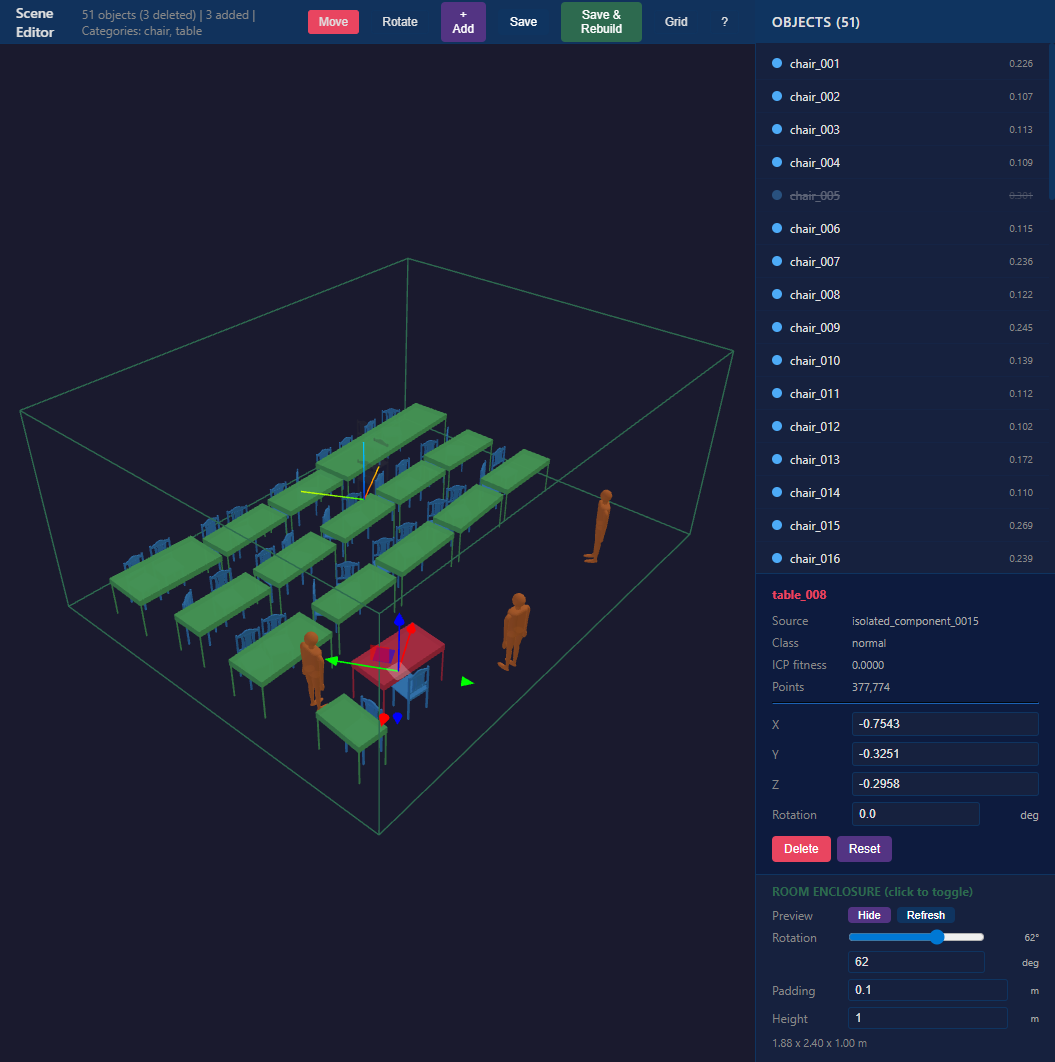}
    \caption{Browser-based scene editor with interactive 3D viewport.}
\end{subfigure}\\[6pt]
\begin{subfigure}[b]{0.43\linewidth}
    \centering
    \includegraphics[width=\linewidth]{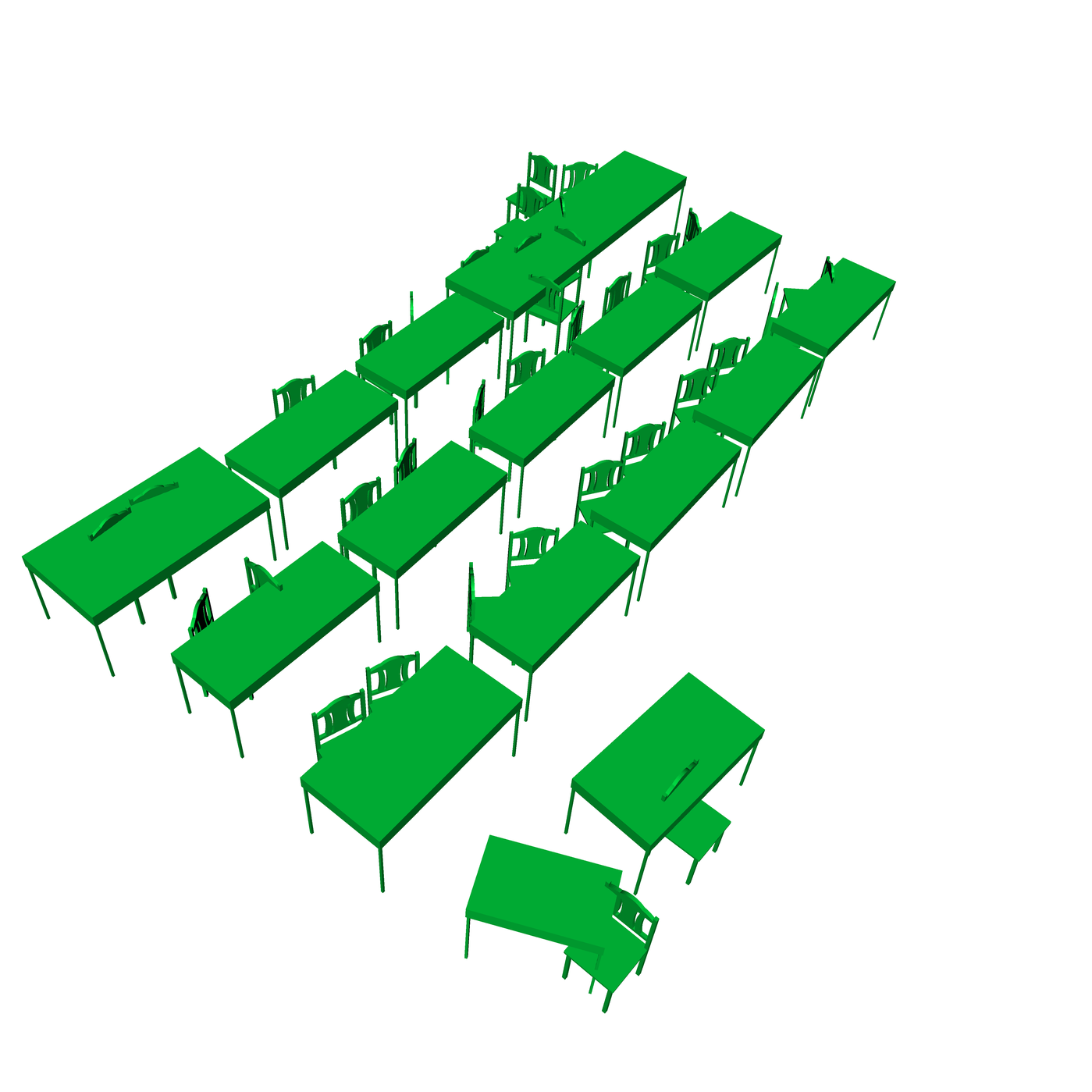}
    \caption{Automated pipeline output (furniture only, no occupants).}
\end{subfigure}
\begin{subfigure}[b]{0.43\linewidth}
    \centering
    \includegraphics[width=\linewidth]{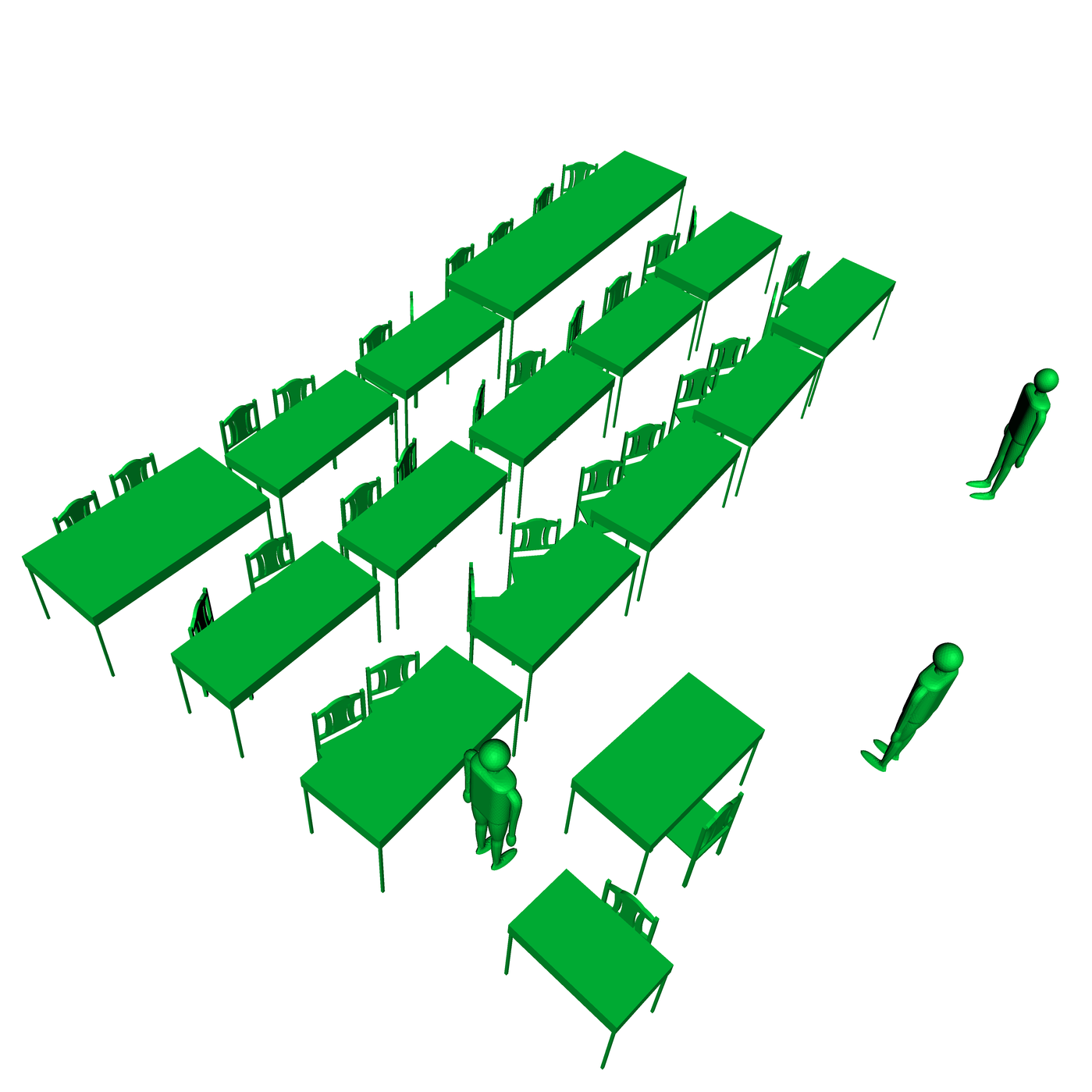}
    \caption{Edited scene with mannequins added and misalignments corrected.}
\end{subfigure}
\caption{Human-in-the-loop scene editing. (a)~The browser-based GUI displays the reconstructed classroom with automatically placed furniture STLs, allowing the operator to inspect and modify placements; the editor also supports batch placement of seated mannequins on detected chairs through an adjustable occupancy slider. (b)~Raw automated pipeline output with furniture only. (c)~Scene after human editing: standing mannequins have been added and misaligned furniture corrected, producing an alternative simulation configuration without re-scanning the physical room.}
\label{fig:gui}
\end{figure*}

The automated pipeline produces usable STL placements for most detected instances; a human-in-the-loop editing stage is nonetheless incorporated before the geometry is consumed by the CFD solver. This stage serves two purposes: (i)~quality assurance of automated placements, and (ii)~scene augmentation to construct alternative simulation configurations (e.g., adding occupant mannequins or rearranging furniture). Both tasks are performed through a browser-based graphical user interface that renders the placed STL models overlaid on the original NeRF point cloud in an interactive 3D viewport (\figref{fig:gui}).

\paragraph{Quality assurance}
The operator inspects the automatically placed furniture and corrects three categories of error:
\begin{enumerate}
    \item \emph{Gross misplacements:} ICP alignments that converged to an incorrect local minimum, producing a visually misaligned or inverted chair or table. These are corrected by selecting the object and applying a rotation or translation through the GUI controls.
    \item \emph{Missing objects:} Furniture that was present in the physical room but lost during segmentation (e.g., heavily occluded items). The operator adds a new furniture item from the asset library at the approximate location.
    \item \emph{Phantom objects:} False-positive placements arising from noise clusters that passed the size filter. These are selected and removed.
\end{enumerate}

\paragraph{Scene augmentation}
Beyond error correction, the GUI enables construction of alternative room configurations for parametric CFD studies without re-scanning the physical environment:
\begin{itemize}
    \item \emph{Occupant placement.} Seated mannequin STLs can be placed on chairs individually or in batch using an occupancy-percentage slider (e.g., 50\% randomly fills half of the detected chairs). Standing mannequins can also be added at arbitrary positions to represent occupants in aisles or near doorways.
    \item \emph{Furniture addition and removal.} Any furniture item (chair, table, or other asset) can be added from the template library or removed from the scene, enabling the study of alternative furniture layouts.
    \item \emph{Object manipulation.} Every placed object can be selected and interactively rotated or translated along any axis, allowing fine-grained adjustments to position and orientation.
\end{itemize}

This editing stage is deliberately lightweight and does not involve manual CAD modeling: the operator works exclusively with pre-built parametric assets, preserving the core automation benefit of the pipeline. The ability to rearrange furniture and add mannequins without re-capturing the room enables the multi-configuration CFD studies presented in \secref{sec:results}, where the same reconstructed classroom is simulated under multiple distinct geometric configurations.

\subsection{CFD Integration and Simulation Setup}
\label{sec:cfd_method}
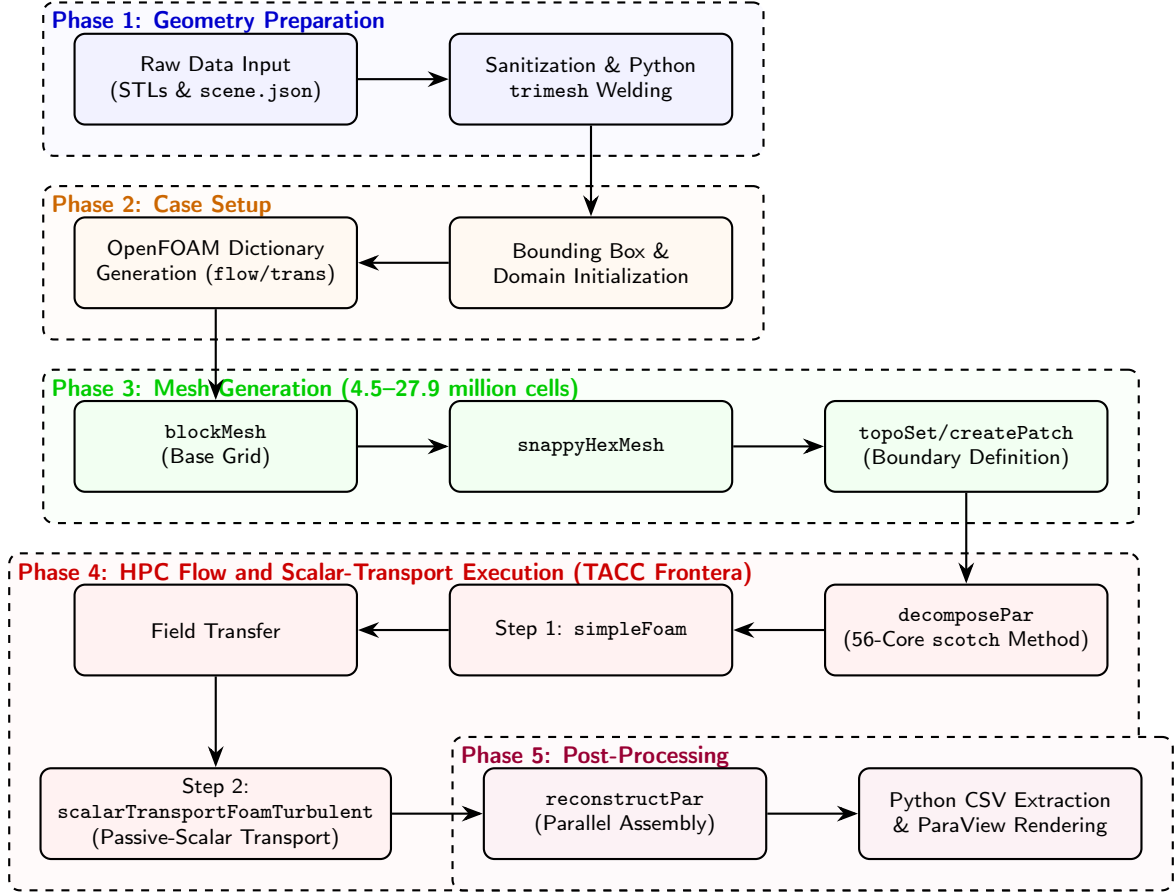
\begin{figure*}[t!]
    \centering
    \begin{tikzpicture}[
        node distance = 1.2cm and 1.2cm,
        block/.style = {rectangle, draw, thick, fill=white, text width=3.5cm, text centered, rounded corners, minimum height=1.2cm, font=\footnotesize\sffamily},
        phase_box/.style = {rectangle, draw, dashed, thick, inner sep=0.4cm, rounded corners},
        arrow/.style = {thick, -{Stealth[scale=1.2]}},
        phase_title/.style = {font=\bfseries\sffamily\small, anchor=north west}
    ]

    \node [block, fill=blue!5] (n1) {Raw Data Input \\ (STLs \& \texttt{scene.json})};
    \node [block, fill=blue!5, right=of n1] (n2) {Sanitization \& Python \\ \texttt{trimesh} Welding};
    \draw [arrow] (n1) -- (n2);

    \node [block, fill=orange!5, below=of n2] (n3) {Bounding Box \& \\ Domain Initialization};
    \node [block, fill=orange!5, left=of n3] (n4) {OpenFOAM Dictionary \\ Generation (\texttt{flow/trans})};
    \draw [arrow] (n2) -- (n3);
    \draw [arrow] (n3) -- (n4);

    \node [block, fill=green!5, below=of n4] (n5) {\texttt{blockMesh} \\ (Base Grid)};
    \node [block, fill=green!5, right=of n5] (n6) {\texttt{snappyHexMesh}};
    \node [block, fill=green!5, right=of n6] (n7) {\texttt{topoSet/createPatch} \\ (Boundary Definition)};
    \draw [arrow] (n4) -- (n5);
    \draw [arrow] (n5) -- (n6);
    \draw [arrow] (n6) -- (n7);

    \node [block, fill=red!5, below=of n7] (n8) {\texttt{decomposePar} \\ (56-Core \texttt{scotch} Method)};
    \node [block, fill=red!5, left=of n8] (n9) {Step 1: \texttt{simpleFoam}};
    \node [block, fill=red!5, left=of n9] (n10) {Field Transfer};
    \node [block, fill=red!5, below=of n10, text width=4.4cm] (n11) {Step 2: \texttt{scalarTransportFoamTurbulent} \\ (Passive-Scalar Transport)};
    \draw [arrow] (n7) -- (n8);
    \draw [arrow] (n8) -- (n9);
    \draw [arrow] (n9) -- (n10);
    \draw [arrow] (n10) -- (n11);

    \node [block, fill=purple!5, right=of n11] (n12) {\texttt{reconstructPar} \\ (Parallel Assembly)};
    \node [block, fill=purple!5, right=of n12] (n13) {Python CSV Extraction \\ \& ParaView Rendering};
    \draw [arrow] (n11) -- (n12);
    \draw [arrow] (n12) -- (n13);

    \begin{scope}[on background layer]
        \node [phase_box, fill=blue!2, fit=(n1) (n2)] (box1) {};
        \node [phase_title, text=blue!80!black] at (box1.north west) {Phase 1: Geometry Preparation};
        \node [phase_box, fill=orange!2, fit=(n3) (n4)] (box2) {};
        \node [phase_title, text=orange!80!black] at (box2.north west) {Phase 2: Case Setup};
        \node [phase_box, fill=green!2, fit=(n5) (n6) (n7)] (box3) {};
        \node [phase_title, text=green!80!black] at (box3.north west) {Phase 3: Mesh Generation (4.5--27.9 million cells)};
        \node [phase_box, fill=red!2, fit=(n8) (n9) (n10) (n11)] (box4) {};
        \node [phase_title, text=red!80!black] at (box4.north west) {Phase 4: HPC Flow and Scalar-Transport Execution (TACC Frontera)};
        \node [phase_box, fill=purple!2, fit=(n12) (n13)] (box5) {};
        \node [phase_title, text=purple!80!black] at (box5.north west) {Phase 5: Post-Processing};
    \end{scope}

    \end{tikzpicture}
    \caption{OpenFOAM CFD execution workflow, from reconstructed STL geometry to steady RANS airflow and transient turbulent-diffusivity passive-scalar flushing simulations, executed on the Frontera system at the Texas Advanced Computing Center (TACC).}
    \label{fig:cfd_pipeline}
\end{figure*}

The CFD stage uses the reconstructed STL geometries to generate OpenFOAM \citep{Weller1998OpenFOAM} cases for steady isothermal incompressible airflow followed by transient passive-scalar transport. The CFD domain consists of a closed room enclosure and reconstructed interior objects, including chairs, tables, and mannequin geometries where applicable. These geometries are exported as STL files from the reconstruction pipeline and placed in the OpenFOAM case directory under \texttt{constant/triSurface/}. The airflow field is first solved with \texttt{simpleFoam}. The converged steady velocity field and turbulent kinematic viscosity field are then held fixed and used for the passive-scalar transport solve; the governing equation and effective-diffusivity model are given in \eqnref{eq:scalar_transport_turbulent}. All simulations are isothermal; the modeling assumptions and their consequences are discussed in \secref{sec:cfd_limitations}. Solver fidelity and production-mesh sensitivity are assessed in \ref{app:verification}, which reports a mesh-sensitivity assessment and an IEA Annex~20 benchmark comparison. The complete CFD execution workflow is illustrated in \figref{fig:cfd_pipeline}.

\subsubsection{Domain and Mesh}
The computational domain is discretized with a hybrid Cartesian hex-dominant strategy in OpenFOAM. A structured base mesh is generated with \texttt{blockMesh} and refined and conformed to the reconstructed STL boundaries using \texttt{snappyHexMesh} \citep{OpenCFD2024UserGuide}. To capture localized flow phenomena without globally over-refining the domain, three targeted refinement strategies are applied. First, structural feature edges are extracted with \texttt{surfaceFeatureExtract} to preserve geometric curvature of the room and interior objects. Second, surface-distance refinement is applied to furniture and mannequin surfaces: cells within $0.02$~m of these surfaces are refined to Level~4, yielding an effective local cell resolution of $5$~mm. Third, a volumetric refinement region (\texttt{searchableBox}) directly beneath the ceiling supply diffusers is refined to Level~3 ($1$~cm cells), extending down to the occupied zone ($z = 0.5$~m). Wall functions are used throughout, so the mesh is not intended to be wall-resolved at the boundary-layer scale.

For the production runs, a fine background cell size of \(0.08\)~m before local refinement is used. Mesh selection is informed by a four-level convergence assessment performed for the fully obstructed mixed-furniture classroom configuration (\secref{sec:results}), which contains the room shell, reconstructed chairs, reconstructed tables, and mannequin obstructions. The convergence study uses the turbulent scalar-transport formulation with \(\mathit{Sc}_t=0.7\) and compares coarse, medium, fine, and extra-fine meshes containing approximately \(2.15\), \(5.99\), \(15.91\), and \(55.37\) million cells, respectively. Because the extra-fine mesh produced valid samples over a shorter portion of the requested vertical sampling line, the four-mesh convergence comparison is performed over the common valid interval from approximately \(z=0.05\)~m to \(z=2.00\)~m above the floor. The full convergence study is reported in \ref{app:verification}.

\subsubsection{Boundary Conditions}
Inlet and outlet patches are identified directly on the ceiling background grid through topological face sets generated with \texttt{topoSet} and converted to physical OpenFOAM patches with \texttt{createPatch}. Supply diffusers and exhaust vents are represented as rectangular ceiling patches in the reconstructed room coordinate system, with locations and footprints taken from the captured ceiling layout in each room.

A fixed uniform downward velocity of $U_z = -0.5$~m/s is prescribed at the inlet diffuser patches. Outlet patches use zero-gradient velocity and a fixed modified kinematic pressure of $p = 0$ (\eqnref{eq:momentum}) to permit exhaust flow. All solid boundaries (room enclosure, furniture, and mannequin surfaces) are treated as no-slip walls with standard $k$--$\varepsilon$ wall functions for the turbulence quantities. We simulate removal of an initially well-mixed passive contaminant by initializing the scalar field uniformly as $C = 1$ throughout the domain, with $C = 0$ prescribed at the inlet patches (representing contaminant-free supply air); outlet, wall, furniture, and mannequin patches use zero-gradient scalar conditions. Identical ventilation boundary conditions are applied across the multi-configuration sequences described in \secref{sec:results} so that differences in scalar decay can be associated with the change in interior geometry, within the assumptions and numerical resolution of the present CFD framework, rather than with changes in HVAC forcing. Patch definitions and boundary-condition schemes are summarized in \tableref{tab:patches} and \tableref{tab:bcs}.

\begin{table}[t!]
\centering
\caption{Boundary patch definitions.}
\begin{tabular}{p{2.4cm} p{2.6cm} p{8cm}}
\toprule
Patch group & Type & Description \\
\midrule
Inlet patches & Velocity inlet & Four ceiling supply diffusers (classroom cases); six ceiling supply diffusers (auditorium case) \\
Outlet patches & Pressure outlet & Two ceiling exhaust vents at opposing corners \\
Walls & No-slip wall & Room enclosure (ceiling, floor, vertical walls) \\
Furniture & No-slip wall & Internal complex geometries (chairs, tables, mannequins) \\
\bottomrule
\end{tabular}
\label{tab:patches}
\end{table}

\begin{table}[t!]
\centering
\caption{Boundary condition summary.}
\begin{tabular}{lll}
\toprule
Field & Patch group & Boundary condition \\
\midrule
$p$ & Inlets & \texttt{zeroGradient} \\
$p$ & Outlets & \texttt{fixedValue uniform 0} \\
$p$ & Walls / furniture & \texttt{zeroGradient} \\
\midrule
$\mathbf{U}$ & Inlets & \texttt{fixedValue uniform (0 0 -0.5)} \\
$\mathbf{U}$ & Outlets & \texttt{zeroGradient} \\
$\mathbf{U}$ & Walls / furniture & \texttt{noSlip} \\
\midrule
$k$ & Inlets & \texttt{fixedValue uniform 0.015} \\
$k$ & Outlets & \texttt{zeroGradient} \\
$k$ & Walls / furniture & \texttt{kqRWallFunction uniform 0.015} \\
\midrule
$\varepsilon$ & Inlets & \texttt{fixedValue uniform 0.004} \\
$\varepsilon$ & Outlets & \texttt{zeroGradient} \\
$\varepsilon$ & Walls / furniture & \texttt{epsilonWallFunction uniform 0.004} \\
\midrule
$C$ & Internal field & \texttt{uniform 1} (initial condition) \\
$C$ & Inlets & \texttt{fixedValue uniform 0} \\
$C$ & Outlets / walls / furniture & \texttt{zeroGradient} \\
\bottomrule
\end{tabular}
\label{tab:bcs}
\end{table}

\subsubsection{Governing Equations and Solver Configuration}
\label{sec:gov_eqs}
The CFD workflow consists of two sequential solves on the same geometric domain: a steady-state airflow solution using a Reynolds-averaged turbulence closure, followed by a transient passive-scalar transport solve driven by the previously computed velocity field. Air is modeled as an incompressible Newtonian fluid with constant properties throughout.

\paragraph{Steady RANS airflow.}
The time-averaged velocity field $\mathbf{U}(\mathbf{x})$ and pressure $p(\mathbf{x})$ satisfy the steady incompressible Reynolds-averaged Navier--Stokes (RANS) equations:
\begin{equation}
\nabla \cdot \mathbf{U} = 0,
\label{eq:continuity}
\end{equation}
\begin{equation}
\nabla \cdot (\mathbf{U}\mathbf{U})
= -\nabla p
+ \nabla \cdot \left[\, \nu_{\mathrm{eff}} \left(\nabla \mathbf{U} + (\nabla \mathbf{U})^{T}\right)\,\right],
\label{eq:momentum}
\end{equation}
where $p$ denotes the modified kinematic pressure solved by the OpenFOAM incompressible solvers, $p = p_{\mathrm{static}} / \rho + \tfrac{2}{3} k$, which absorbs the isotropic part of the Reynolds stress, and $\nu_{\mathrm{eff}} = \nu + \nu_t$ is the effective kinematic viscosity, with molecular viscosity $\nu = 1.5 \times 10^{-5}~\si{m^2/s}$. The eddy viscosity $\nu_t$ is modeled with the standard $k$--$\varepsilon$ closure \citep{Launder1974KEpsilon},
\begin{equation}
\nu_t = C_\mu \frac{k^2}{\varepsilon},
\label{eq:nut_keps}
\end{equation}
with steady transport equations for $k$ and $\varepsilon$:
\begin{equation}
\nabla \cdot (\mathbf{U}k)
= \nabla \cdot \left[\left(\nu + \frac{\nu_t}{\sigma_k}\right)\nabla k\right]
+ P_k - \varepsilon,
\label{eq:k_eq}
\end{equation}
\begin{equation}
\nabla \cdot (\mathbf{U}\varepsilon)
= \nabla \cdot \left[\left(\nu + \frac{\nu_t}{\sigma_\varepsilon}\right)\nabla \varepsilon\right]
+ C_{\varepsilon 1}\frac{\varepsilon}{k} P_k
- C_{\varepsilon 2}\frac{\varepsilon^2}{k}.
\label{eq:eps_eq}
\end{equation}
Here $P_k = \nu_t \left(\nabla \mathbf{U} + (\nabla \mathbf{U})^{T}\right) : \nabla \mathbf{U}$ is the production of turbulent kinetic energy from mean shear. The closure coefficients take their standard values, $C_\mu = 0.09$, $C_{\varepsilon 1} = 1.44$, $C_{\varepsilon 2} = 1.92$, $\sigma_k = 1.0$, and $\sigma_\varepsilon = 1.3$ \citep{Launder1974KEpsilon}. Wall-bounded turbulence is treated with standard wall functions for $k$ and $\varepsilon$ consistent with \tableref{tab:bcs}. Inlet turbulence quantities are prescribed as $k = 0.015~\si{m^2/s^2}$ and $\varepsilon = 0.004~\si{m^2/s^3}$. These values correspond to an inlet turbulence intensity of $I = \sqrt{2k/3}/U \approx 20\%$ at the prescribed inlet magnitude $U = 0.5$~m/s, with an implied turbulent length scale $\ell = C_\mu^{3/4} k^{3/2} / \varepsilon \approx 0.075$~m using $C_\mu = 0.09$. These turbulence boundary values are held fixed across the comparative cases. The steady RANS system (\eqnref{eq:continuity}--\eqnref{eq:eps_eq}) is solved in \texttt{simpleFoam} with the Semi-Implicit Method for Pressure-Linked Equations (SIMPLE) \citep{Patankar1972SIMPLE}, run in its consistent SIMPLEC variant (\texttt{consistent yes} in \tableref{tab:numerics}), under which the pressure under-relaxation factor is unity. Residuals and integral quantities of interest are monitored during the solve. Numerical schemes and linear-solver settings are listed in \tableref{tab:numerics}. The steady airflow solves were advanced to \texttt{simpleFoam} pseudo-time/iteration 1000 for the final production cases before the velocity and turbulent-viscosity fields were frozen for scalar transport, to keep the workflow uniform across cases. This fixed iteration budget, rather than a residual threshold, is the stopping rule used for the production cases. Post-processing of the outlet patches did not indicate substantial persistent reverse flow, so the zero-gradient outlet velocity treatment was retained for the comparative simulations.

\paragraph{Transient passive-scalar transport.}
After convergence of the steady airflow field, the velocity field is frozen and used to advect a dimensionless passive scalar \(C\). The scalar is initialized as \(C=1\) throughout the room, and clean supply air is imposed through \(C=0\) at the inlet patches. The transient scalar equation is
\begin{equation}
    \frac{\partial C}{\partial t}
    + \nabla \cdot (\mathbf{U} C)
    =
    \nabla \cdot
    \left[
        \left(
        D + \frac{\nu_t}{\mathit{Sc}_t}
        \right)
        \nabla C
    \right],
    \label{eq:scalar_transport_turbulent}
\end{equation}
where \(D=1.5\times10^{-5}\,\mathrm{m^2\,s^{-1}}\) is the molecular scalar diffusivity, taken equal to the molecular kinematic viscosity so that the molecular Schmidt number is unity, and negligible against the turbulent diffusivity \(\nu_t/\mathit{Sc}_t\) throughout the domain, \(\nu_t\) is the turbulent kinematic viscosity obtained from the steady RANS solution, and \(\mathit{Sc}_t=0.7\) is the turbulent Schmidt number. In the OpenFOAM case files, the scalar field is stored as \texttt{T}; throughout the manuscript it is denoted \(C\) to avoid confusion with temperature. The scalar is dimensionless and is used only as a passive contaminant-clearance marker. The flow and scalar-transport model settings are summarized in \tableref{tab:models}. The turbulent scalar diffusivity is modeled with the standard gradient-diffusion closure, \(D_t=\nu_t/\mathit{Sc}_t\), using \(\mathit{Sc}_t=0.7\), a widely used default for RANS pollutant-dispersion and room-airflow simulations. Reported values vary substantially, so the choice is itself a modeling uncertainty~\citep{Gualtieri2017TurbulentSchmidt, Li2018ModifiedSchmidt}.

\begin{table}[t!]
\centering
\caption{Flow and passive-scalar transport models used in the OpenFOAM simulations.}
\label{tab:models}
\begin{tabular}{ll}
\toprule
Item & Setting \\
\midrule
Flow model & Incompressible steady RANS, SIMPLE algorithm \\
Turbulence closure & Standard \(k\)--\(\varepsilon\) with wall functions \\
Air kinematic viscosity, \(\nu\) & \(1.5\times10^{-5}\,\mathrm{m^2\,s^{-1}}\) \\
Passive-scalar field & Dimensionless scalar \(C\) stored as \texttt{T} in OpenFOAM \\
Molecular scalar diffusivity, \(D\) & \(1.5\times10^{-5}\,\mathrm{m^2\,s^{-1}}\) \\
Turbulent Schmidt number, \(\mathit{Sc}_t\) & \(0.7\) \\
Effective scalar diffusivity & \(D+\nu_t/\mathit{Sc}_t\) \\
Scalar solver & Custom \texttt{scalarTransportFoamTurbulent} \\
\bottomrule
\end{tabular}
\end{table}

\begin{table}[t!]
\centering
\caption{Numerical schemes and solver settings.}
\begin{tabular}{ll}
\toprule
Item & Setting \\
\midrule
Time discretization & Euler \\
Gradients & Gauss linear \\
Scalar convection & Gauss upwind \\
Laplacian & Gauss linear corrected \\
\midrule
$p$ solver & PCG + DIC; tol $10^{-6}$; relTol $0.01$ \\
$\mathbf{U}, k, \varepsilon$ solver & PBiCGStab + DILU; tol $10^{-7}$; relTol $0.1$ \\
$C$ solver & PBiCGStab + DILU; tol $10^{-8}$; relTol $0$ \\
SIMPLE & \texttt{nNonOrthogonalCorrectors 0}, \texttt{consistent yes} \\
Relaxation factors & $\mathbf{U}, k, \varepsilon: 0.7$;\quad $C: 1.0$ \\
\bottomrule
\end{tabular}
\label{tab:numerics}
\end{table}

\subsubsection{Clearance Metrics}
We compare contaminant removal across configurations through a half-clearance time $t_{50}$, computed from the normalized scalar concentration history at each monitor location. Here $C(t)$ is the scalar concentration recorded at a fixed monitor point in the occupied zone (a single monitor per classroom and three monitors in the auditorium), not a spatially averaged quantity. The half-clearance time is the elapsed time required for the monitored normalized concentration $C^*(t) = C(t) / C_0$ to fall to half its initial value:
\begin{equation}
C^*(t_{50}) = 0.5,
\end{equation}
where $C_0$ is the concentration at the start of the analyzed decay window. At a given monitor location, smaller \(t_{50}\) indicates more rapid decay of the modeled passive scalar at that location. When the concentration crosses $C^* = 0.5$ between two recorded time levels, $t_{50}$ is obtained by linear interpolation:
\begin{equation}
t_{50}
= t_i
+ \frac{0.5 - C^*(t_i)}{C^*(t_{i+1}) - C^*(t_i)} (t_{i+1} - t_i),
\end{equation}
with $C^*(t_i) > 0.5$ and $C^*(t_{i+1}) \leq 0.5$. All reported half-clearance times are measured relative to the start of the corresponding decay analysis window; because the scalar field is initialized uniformly at $C=1$, the reference value $C_0$ equals this initial concentration at every monitor. Later-stage clearance metrics are defined analogously: $t_{80}$ and $t_{90}$ are the first recorded elapsed times at which $C^{*}$ falls below $0.2$ and $0.1$, respectively, and are reported at the sampling resolution of the recorded history without the linear interpolation applied to $t_{50}$; the area under the curve, $\mathrm{AUC}=\int C^{*}(t)\,dt$, is integrated over the simulated decay window (\tableref{tab:clearance_metrics}). The same extraction procedure is applied to each metric across all configurations.

\subsubsection{Physical Scale Recovery}
\label{sec:scale_recovery}
Because NeRF reconstruction is metrically scale-ambiguous, each reconstructed room is scaled to physical dimensions through an independently measured room-height reference. A single uniform scale factor is computed from the ratio of measured to reconstructed room height and applied to the complete geometry before meshing. The resulting room dimensions and the height references used are documented in \ref{app:room_scale_patch_checks}, and the uncertainty associated with this single-reference scaling is discussed in \secref{sec:cfd_limitations}.

\begin{table}[t!]
\centering
\caption{Configurable pipeline parameters and the values used for the mixed-furniture classroom demonstration of \secref{sec:results}. Chair and table values are listed separately where category-specific tuning is applied; a small subset of thresholds is additionally adjusted per environment (\secref{sec:discussion_generalization}). Length-valued parameters are specified in the scale-normalized reconstruction frame, in which the room height is approximately one unit; physical lengths follow from the uniform metric scaling of \secref{sec:scale_recovery}.}
\label{tab:params}
\begin{tabular}{p{6.2cm} l c c}
\toprule
Parameter & Symbol & Chairs & Tables \\
\midrule
\multicolumn{4}{l}{\emph{Semantic segmentation and projection (\secref{sec:semantic})}} \\
SAM~3 segmentation confidence threshold & $\tau_{conf}$ & 0.5 & 0.5 \\
Depth patch size (pixels) & $S_{patch}$ & 32 & 48 \\
Depth-band tolerance (MAD multiplier) & $\sigma_{depth}$ & 0.75 & 1.25 \\
Multi-view consensus threshold (votes) & $\tau_{mv}$ & 10 & 5 \\
\midrule
\multicolumn{4}{l}{\emph{Instance extraction and healing (\secref{sec:size_classification})}} \\
Octree resolution level & $L_{octree}$ & 11 & 11 \\
Minimum component size (points) & $N_{min}$ & 2048 & 1024 \\
Healing gap threshold (scene units) & $\delta_{gap}$ & 0.008 & 0.01 \\
Healing filler-point spacing (scene units) & $r$ & 0.001 & 0.002 \\
\midrule
\multicolumn{4}{l}{\emph{ICP alignment (\secref{sec:size_classification})}} \\
ICP maximum iterations & $K_{icp}$ & 500 & 50 \\
ICP fitness threshold & $\tau_{fit}$ & 0.05 & 0.3 \\
\bottomrule
\end{tabular}
\end{table}

\section{Results}
\label{sec:results}
We evaluate the pipeline on three indoor environments: a mixed-furniture classroom, a chair-dominant classroom, and an auditorium with tiered seating. Room dimensions are listed in \tableref{tab:room_scale_checks}. The mixed-furniture classroom serves as the primary study environment, in which a progressive what-if analysis is conducted by incrementally adding geometric complexity to the CFD domain (room shell, +chairs, +tables, +mannequins). The chair-dominant classroom and the auditorium demonstrate generalizability of the workflow to different room types and scales. For each environment we present the reconstruction output and the corresponding CFD simulation results together, reflecting the end-to-end nature of the workflow.

\subsection{Implementation Details}
The pipeline is implemented in Python, using the Nerfstudio library for reconstruction, Open3D \citep{Zhou2018Open3D} for geometric processing, and \texttt{trimesh} \citep{DawsonHaggerty2019Trimesh} for STL manipulation. All experiments were conducted on a workstation equipped with an NVIDIA RTX~4070 GPU, an Intel Core i9-14900KF CPU (3.20~GHz), and 64~GB of RAM. The NeRF model was trained for 30{,}000 steps (the default \texttt{nerfacto} configuration) for all three environments, requiring approximately 11, 13, and 31 minutes of wall-clock time for the mixed-furniture classroom, chair-dominant classroom, and auditorium, respectively. Wall-clock times for the remaining stages, estimated from output-file timestamps, are approximately 1.5~h for frame extraction and structure-from-motion (mixed-furniture classroom); 5--20 minutes for SAM~3 segmentation in the classrooms and roughly 2~h in the auditorium; 5--20 minutes for projection and filtering; under 10 minutes for healing; and 2--30 minutes for STL placement per configuration. A complete capture-to-STL pass therefore takes on the order of two to five hours per room on this workstation. CFD simulations were executed on one TACC Frontera normal node \citep{Stanzione2020Frontera}, with domain decomposition via \texttt{scotch} \citep{Pellegrini1996Scotch} and field visualization in ParaView \citep{Ahrens2005ParaView}.

\tableref{tab:capture_stats} summarizes the video capture and frame processing statistics for each environment. Frame sampling rate was selected per environment to balance reconstruction quality with processing time, with sparser sampling for the longer auditorium video. \tableref{tab:recon_counts} reports the ground-truth furniture counts (manually verified) alongside the final reconstruction counts after the full pipeline including human QA. \tableref{tab:nerf_metrics} reports NeRF reconstruction quality on held-out test views.

\begin{table}[t!]
\centering
\caption{NeRF reconstruction quality metrics (mean $\pm$ std over held-out test views), computed with Nerfstudio's \texttt{ns-eval} on the default \texttt{nerfacto} evaluation split of each capture. Higher PSNR and SSIM indicate better fidelity; lower LPIPS indicates better perceptual similarity.}
\label{tab:nerf_metrics}
\setlength{\tabcolsep}{7pt}
\begin{tabular}{lccc}
\toprule
\textbf{Environment} & \textbf{PSNR (dB)} $\uparrow$ & \textbf{SSIM} $\uparrow$ & \textbf{LPIPS} $\downarrow$ \\
\midrule
Mixed-Furniture Classroom & $24.51 \pm 2.53$ & $0.819 \pm 0.066$ & $0.370 \pm 0.129$ \\
Chair-Dominant Classroom  & $23.09 \pm 1.68$ & $0.750 \pm 0.063$ & $0.418 \pm 0.125$ \\
Auditorium                & $24.55 \pm 3.13$ & $0.836 \pm 0.066$ & $0.233 \pm 0.080$ \\
\bottomrule
\end{tabular}
\end{table}

The auditorium attains the highest perceptual fidelity (lowest LPIPS) despite its larger scale, plausibly due to the more textured surfaces (seat upholstery patterning, ceiling acoustic panels) that aid view synthesis relative to the textureless walls dominant in the classrooms. The reconstruction count for the auditorium (273 chairs vs.\ 242 ground truth) overestimates by 31 chairs; we discuss the source of this overcount in \secref{sec:discussion_failures}.

\subsection{Mixed-Furniture Classroom: What-If Scenarios}
\label{sec:results_mixed}
The Mixed-Furniture Classroom contains 32~chairs and 17~tables (ground truth) in approximately 2:1 proportion and serves as the primary environment for a controlled what-if study. Starting from an empty room, four configurations are simulated with increasing interior obstruction:
\begin{enumerate}
    \item[(i)] \emph{Empty}: room shell with ventilation patches only.
    \item[(ii)] \emph{+Chairs}: empty room plus 32 reconstructed chair STLs.
    \item[(iii)] \emph{+Chairs+Tables}: empty room plus 32 chairs and 16 tables (16 of the 17 physical tables were recovered; \tableref{tab:recon_counts}).
    \item[(iv)] \emph{+Chairs+Tables+Mannequins}: full furniture configuration plus 25 mannequins (22 seated and 3 standing; 69\% of the 32 chairs occupied).
\end{enumerate}
The room was scanned once, and the pipeline produced individually movable STL assets that enabled all four configurations through the human-in-the-loop scene editor (\secref{sec:human_qa}) without re-scanning. Identical ventilation boundary conditions were applied across all four cases (\secref{sec:cfd_method}) so that differences in scalar decay can be associated with the change in interior geometry.

\begin{figure*}[t!]
\centering
\footnotesize
\setlength{\tabcolsep}{4pt}
\renewcommand{\arraystretch}{1.3}
\newcommand{\reconcellpad}{\rule[-6pt]{0pt}{6pt}}
\begin{tabular}{@{}|>{\centering\arraybackslash}m{0.15\textwidth}|>{\centering\arraybackslash}m{0.245\textwidth}|>{\centering\arraybackslash}m{0.245\textwidth}|>{\centering\arraybackslash}m{0.245\textwidth}|@{}}
\hline
&
\textbf{(a) STL on point cloud}
&
\textbf{(b) STL only}
&
\textbf{(c) With room enclosure}
\\
\hline
\makecell[c]{\textbf{(i)}\\[2pt]\textbf{Empty}}
&
\makecell[c]{\scriptsize Not applicable}
&
\makecell[c]{\scriptsize Not applicable}
&
\includegraphics[width=\linewidth]{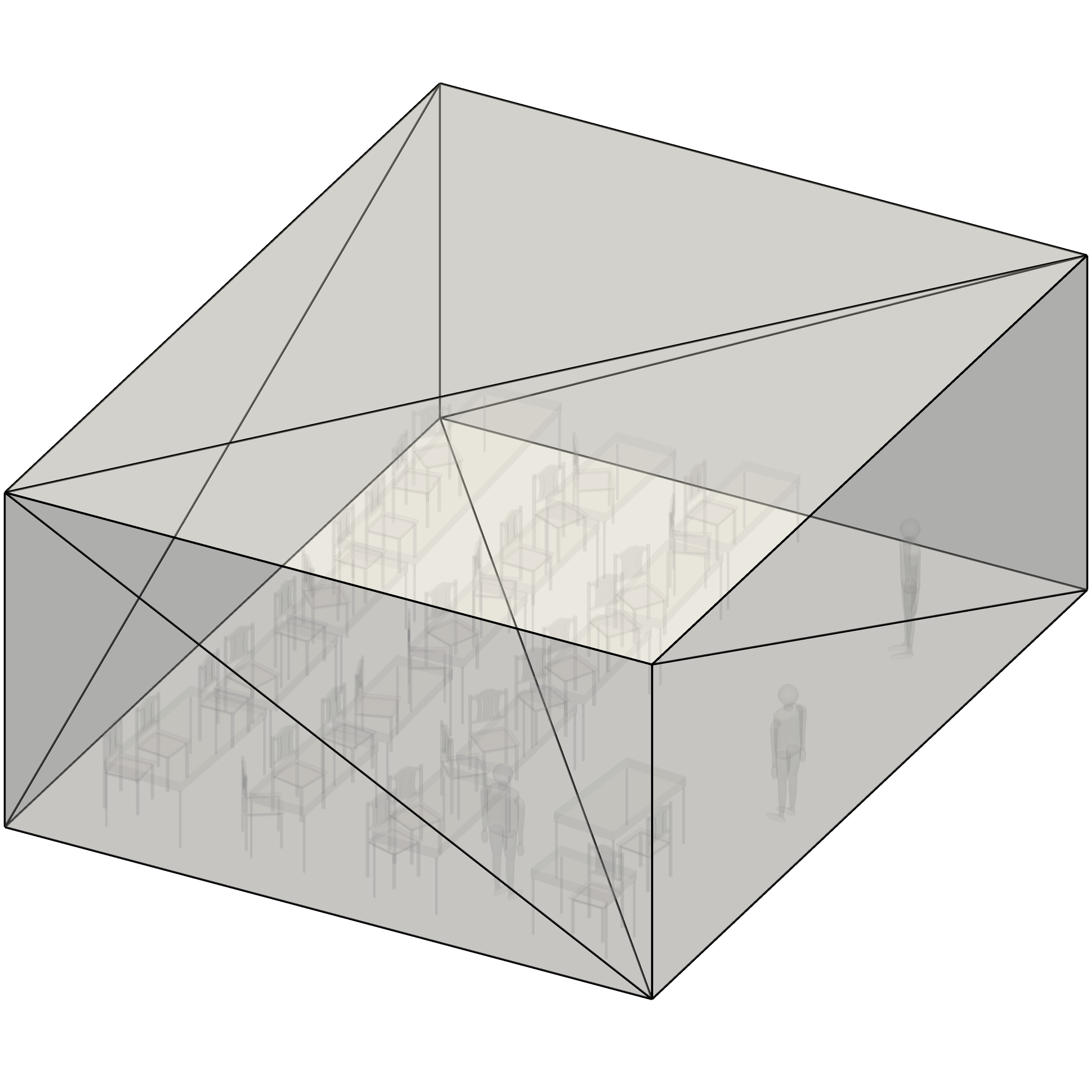}\reconcellpad
\\
\hline
\makecell[c]{\textbf{(ii)}\\[2pt]\textbf{+Chairs}}
&
\includegraphics[width=\linewidth]{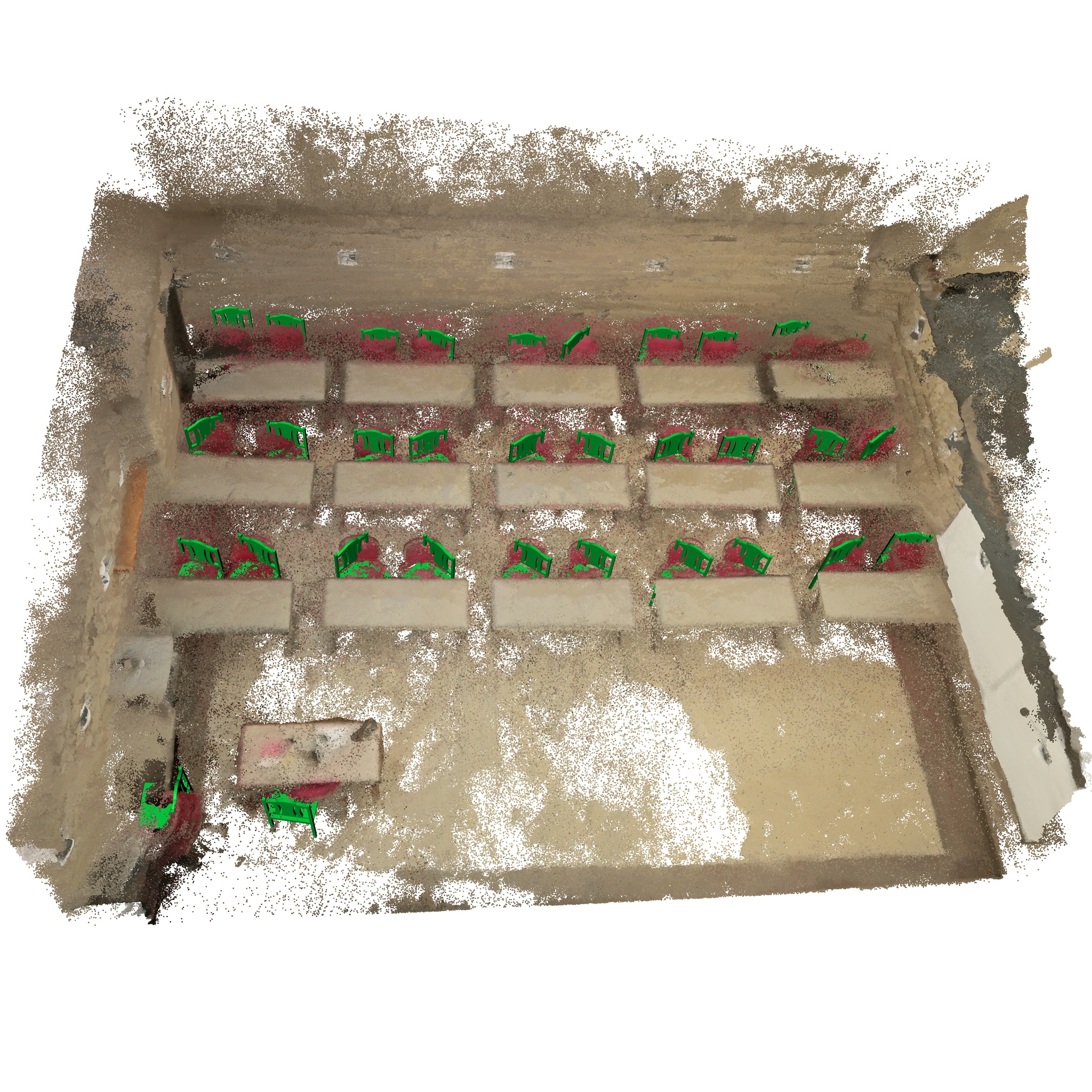}
&
\includegraphics[width=\linewidth]{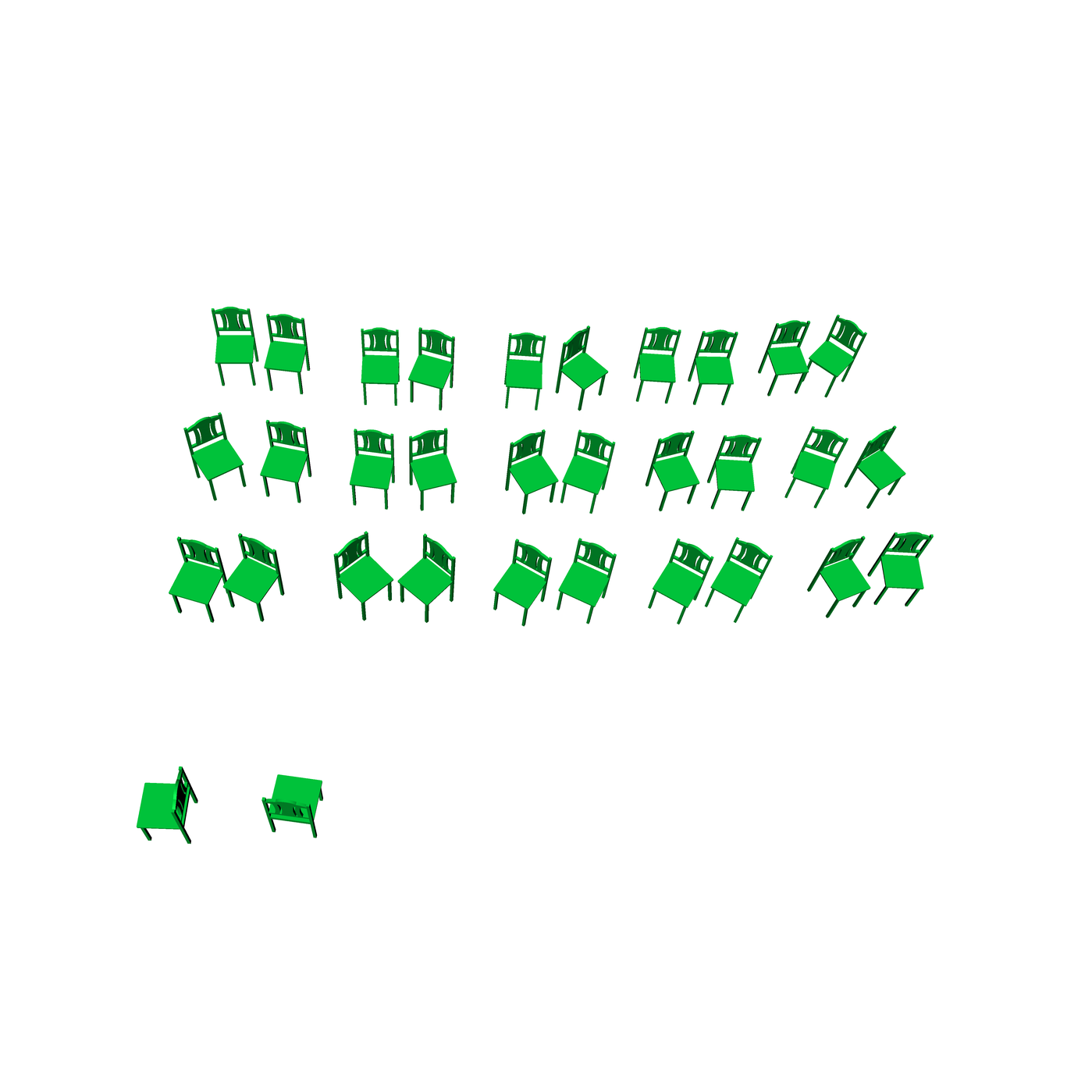}
&
\includegraphics[width=\linewidth]{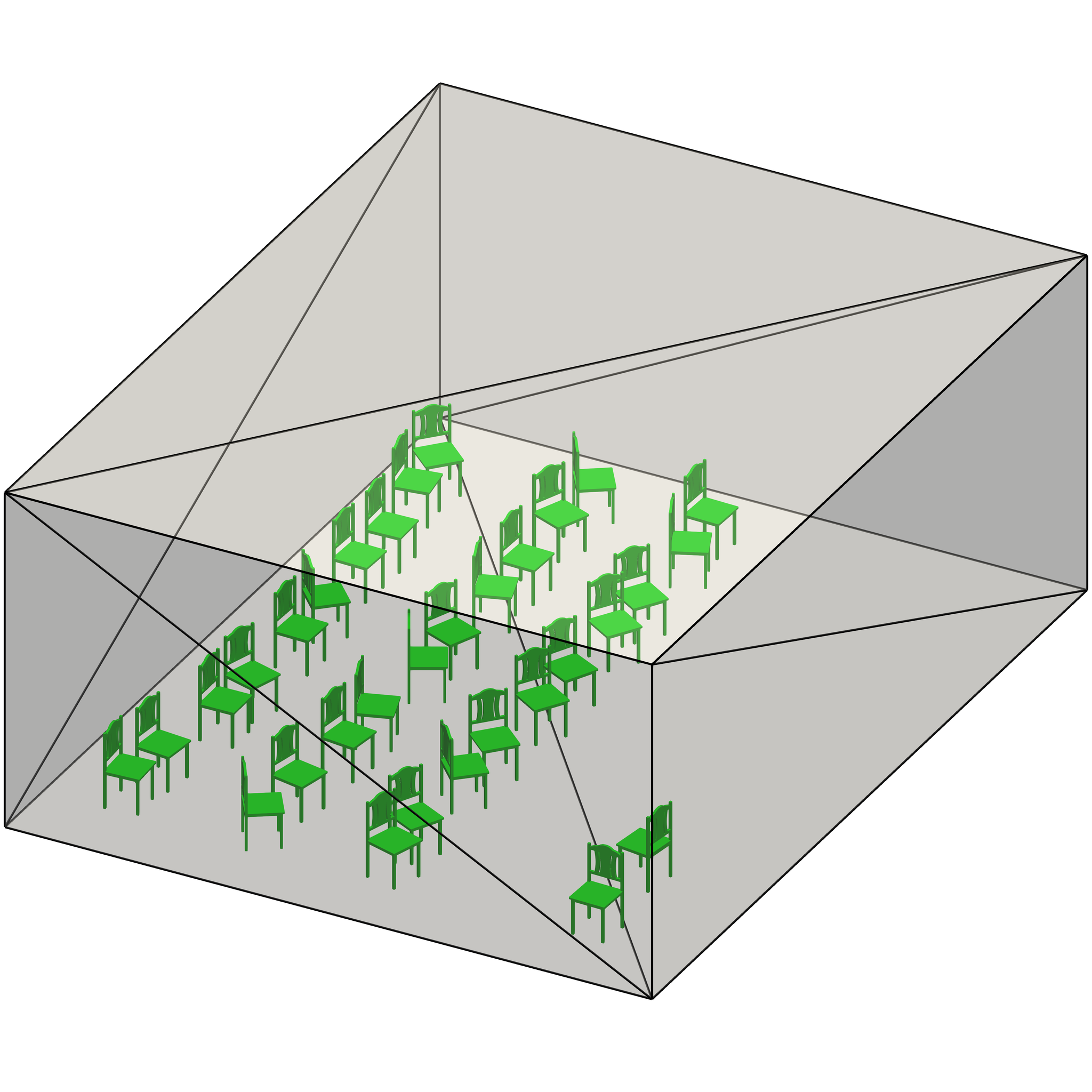}\reconcellpad
\\
\hline
\makecell[c]{\textbf{(iii)}\\[2pt]\textbf{+Chairs}\\\textbf{+Tables}}
&
\includegraphics[width=\linewidth]{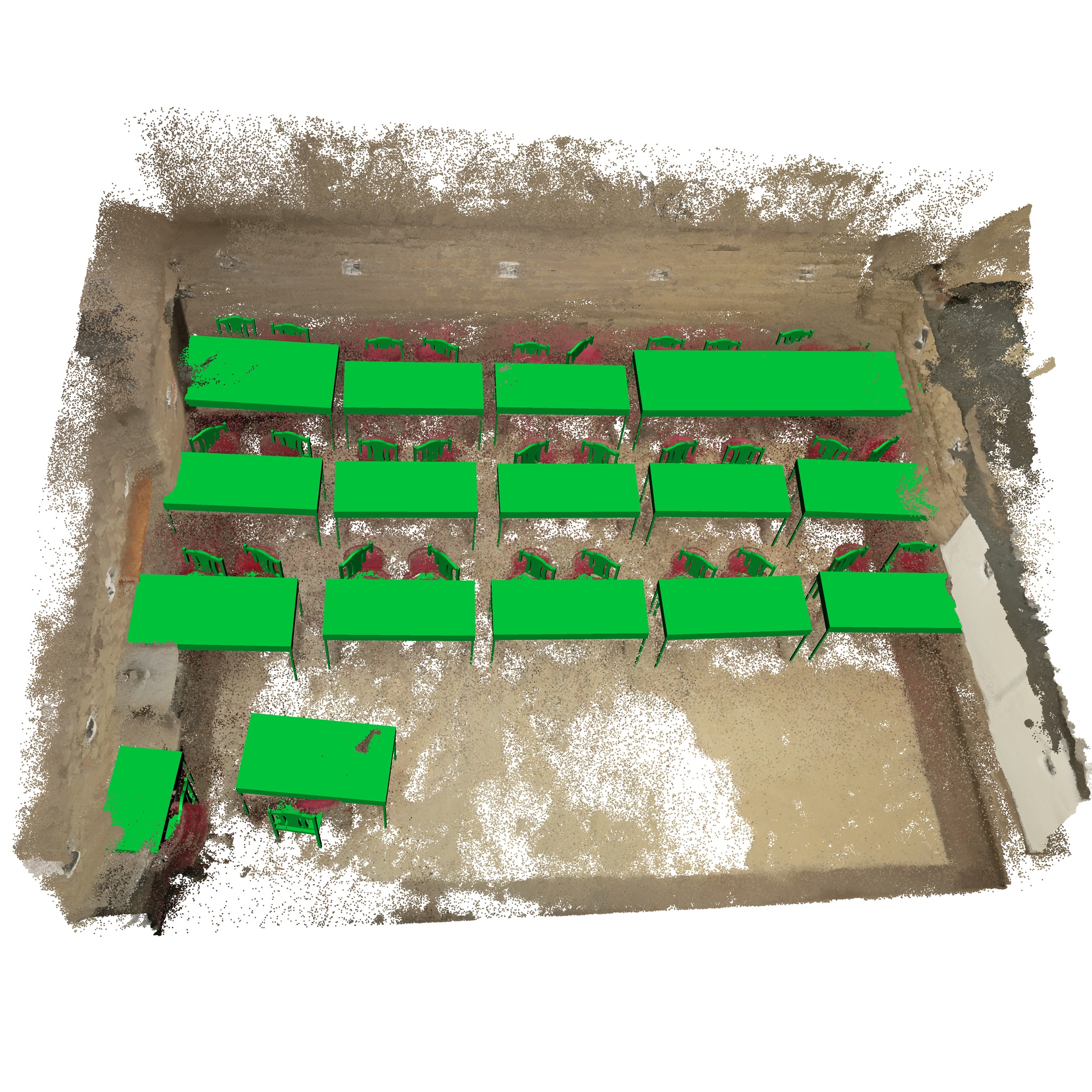}
&
\includegraphics[width=\linewidth]{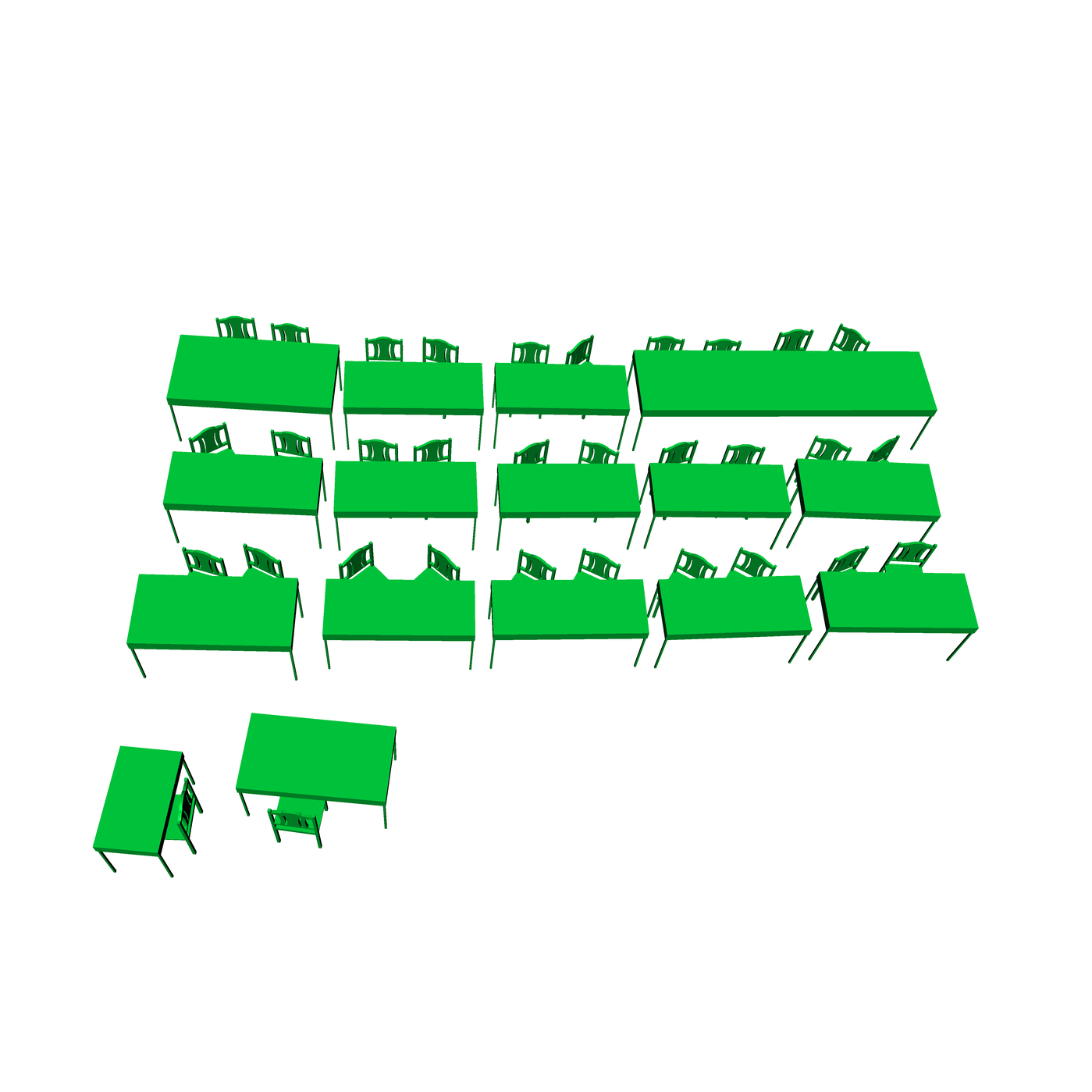}
&
\includegraphics[width=\linewidth]{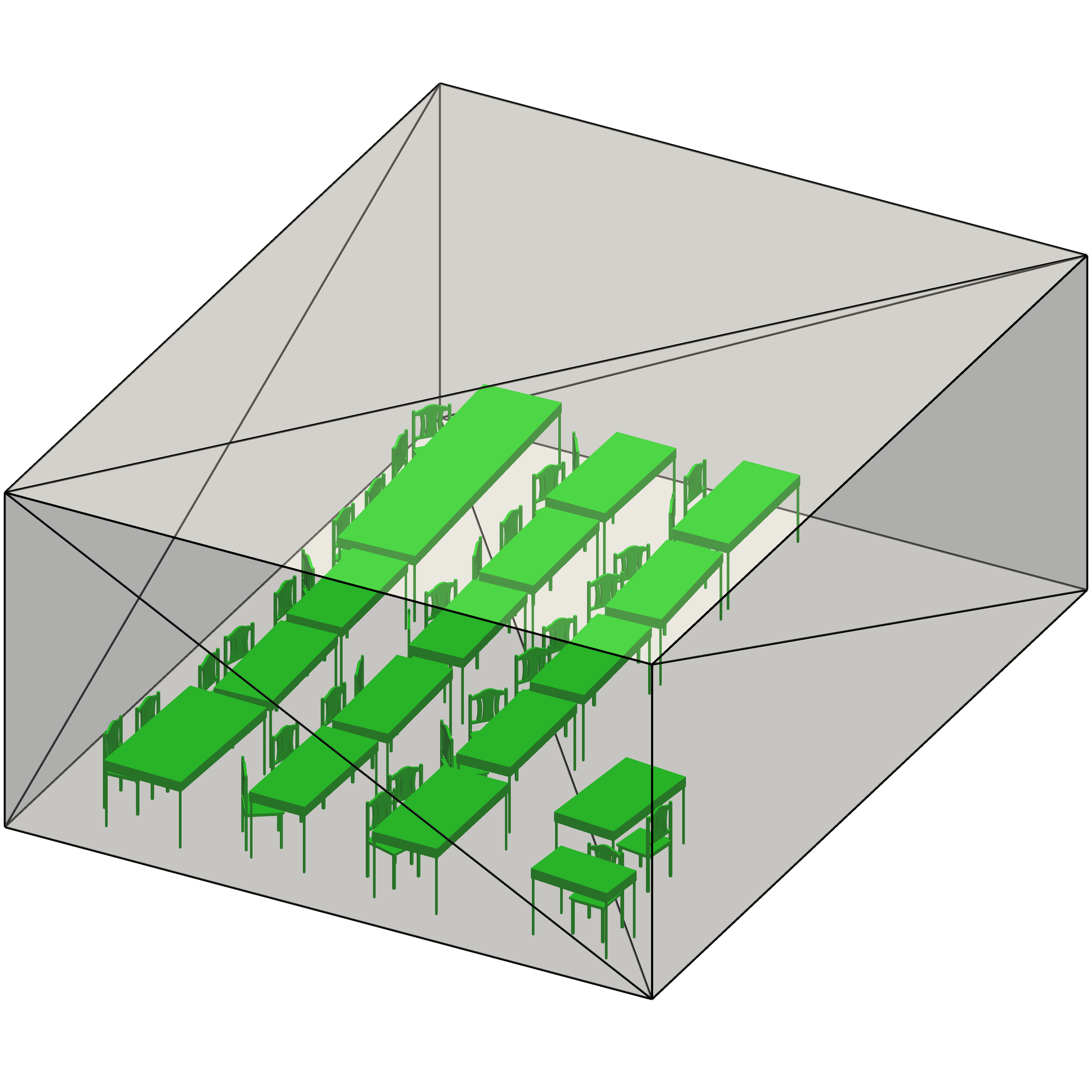}\reconcellpad
\\
\hline
{\scriptsize\makecell[c]{\textbf{(iv)}\\[2pt]\textbf{+Chairs}\\\textbf{+Tables}\\\textbf{+Mannequins}}}
&
\includegraphics[width=\linewidth]{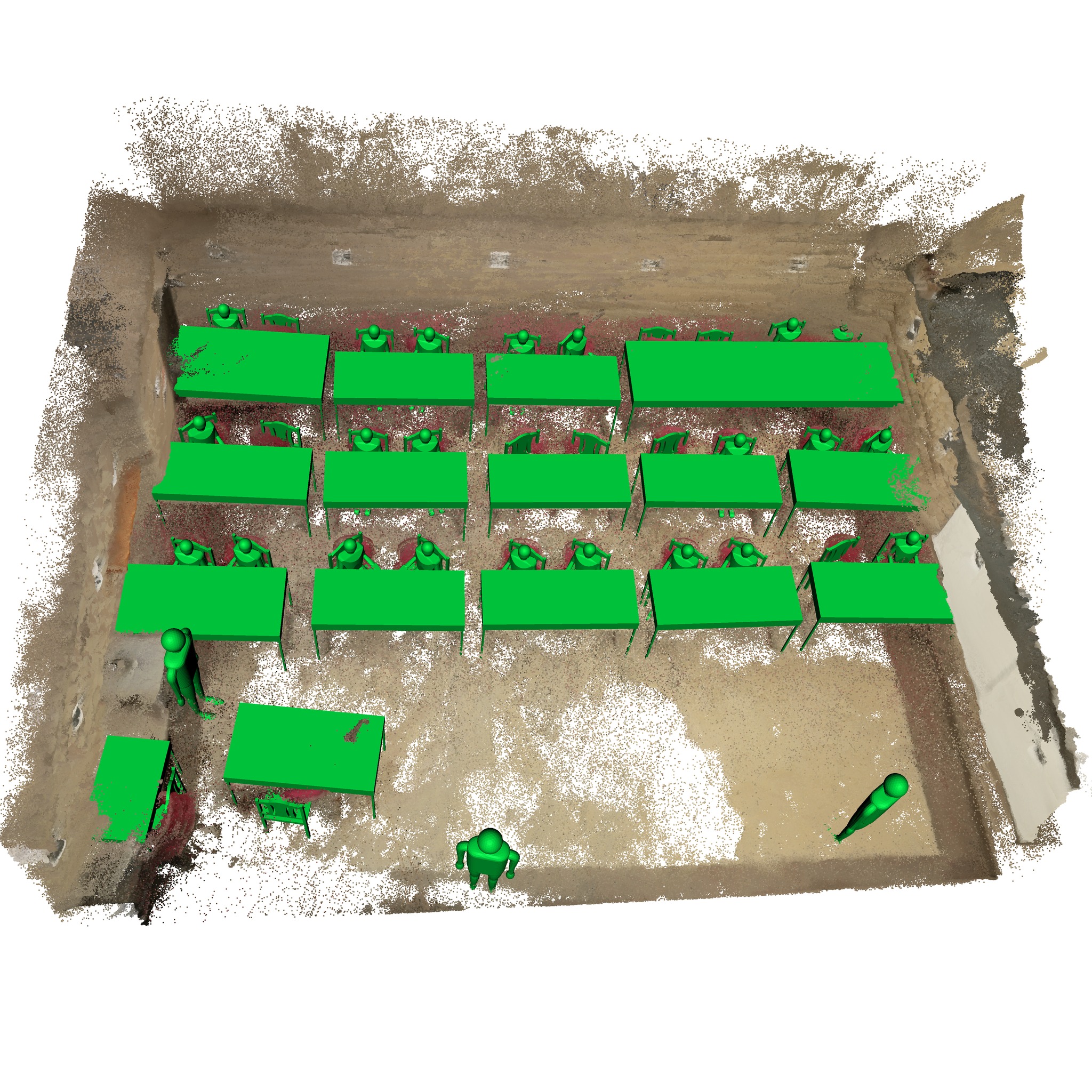}
&
\includegraphics[width=\linewidth]{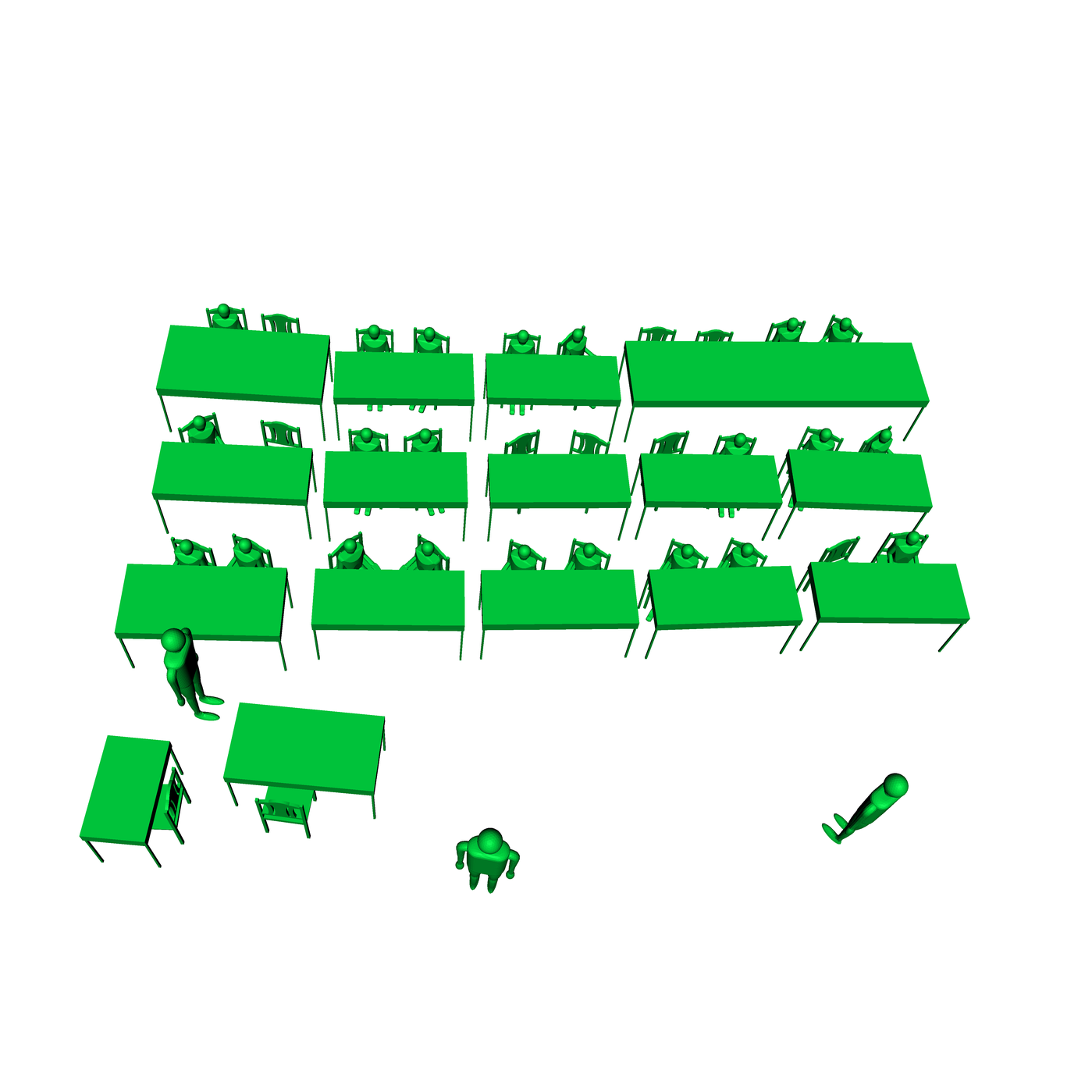}
&
\includegraphics[width=\linewidth]{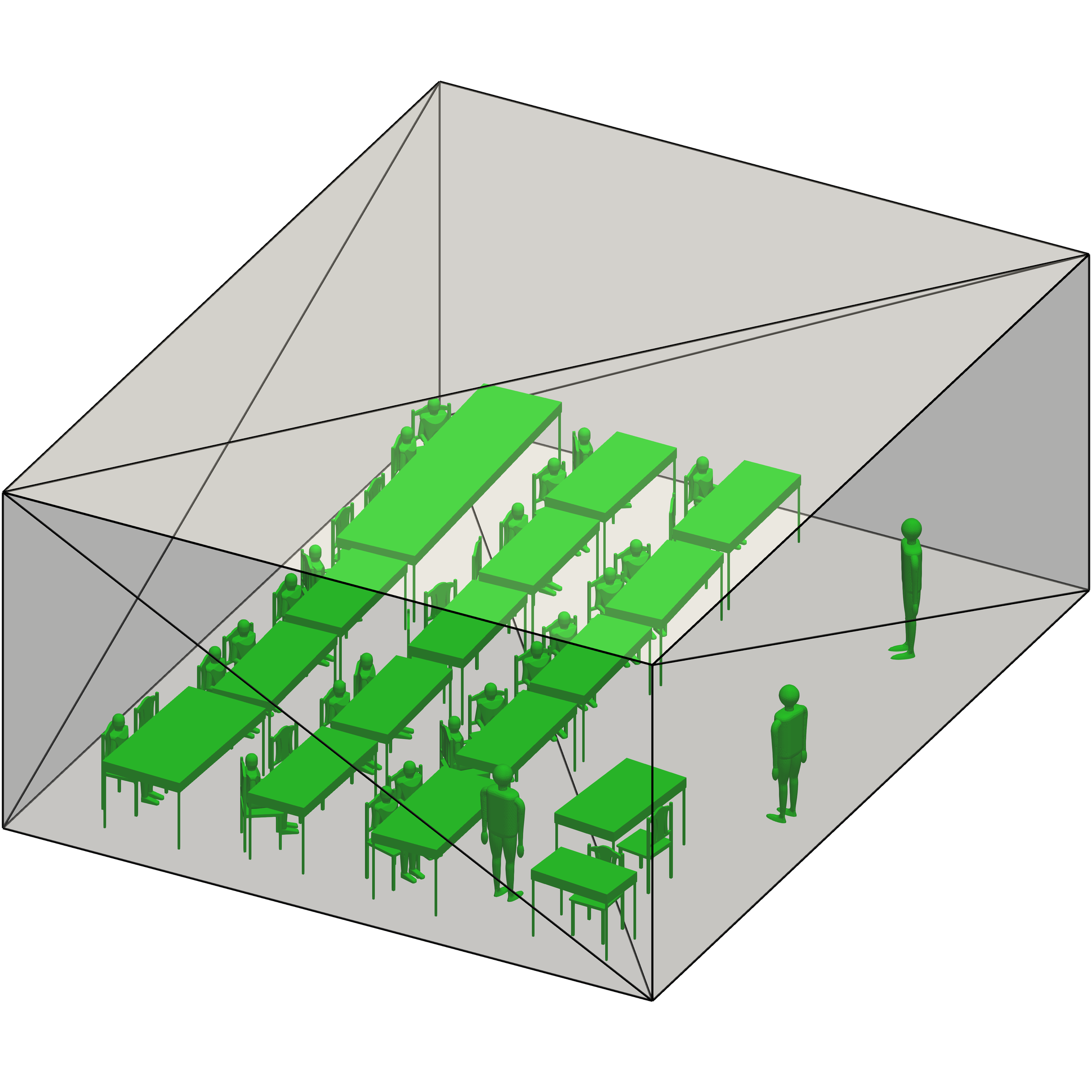}\reconcellpad
\\
\hline
\end{tabular}

\caption{What-if reconstruction sequence for the mixed-furniture classroom. Rows show four progressively modified configurations: (i) the empty room enclosure, (ii) the room with 32 chairs, (iii) the room with 32 chairs and 16 tables, and (iv) the full furniture configuration with 25 mannequins (22 seated, 3 standing). Columns show, from left to right, the reconstructed STL assets overlaid on the NeRF point cloud, the STL assets with the point cloud removed, and the final simulation-ready geometry including the room enclosure. For the empty-room case, only the enclosure geometry is shown because no furniture or occupant assets are present.}
\label{fig:classA_recon}
\end{figure*}

\begin{figure*}[t!]
\centering
\footnotesize
\setlength{\tabcolsep}{4pt}
\renewcommand{\arraystretch}{1.05}

\begin{minipage}[c]{0.90\textwidth}
\centering
\begin{tabular}{@{}cc@{}}
\begin{subfigure}[b]{0.39\textwidth}
    \centering
    \classAVelPanel[596bp 72bp 596bp 20bp]{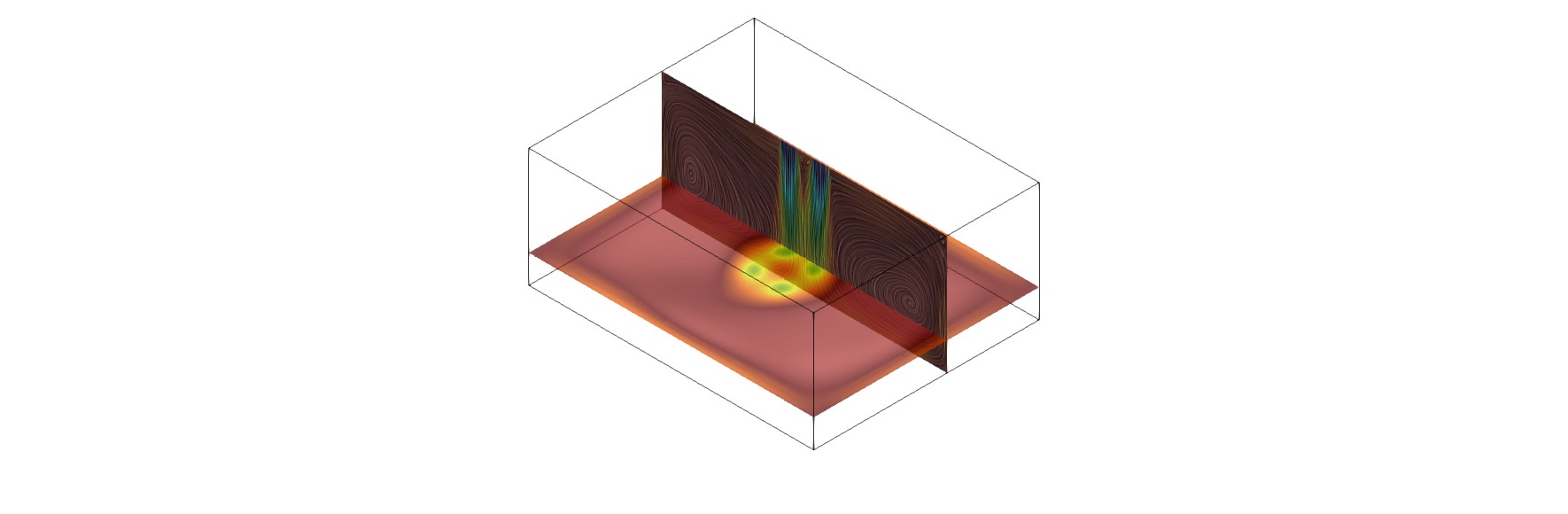}
    \caption{Empty}
\end{subfigure}
&
\begin{subfigure}[b]{0.39\textwidth}
    \centering
    \classAVelPanel[608bp 48bp 608bp 48bp]{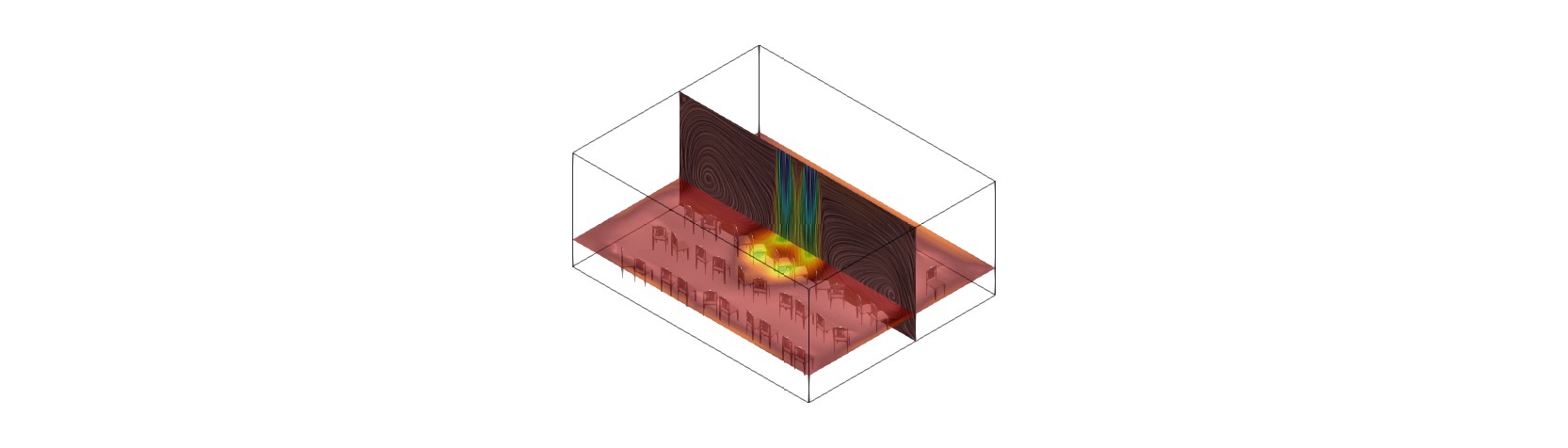}
    \caption{+Chairs}
\end{subfigure}
\\[8pt]
\begin{subfigure}[b]{0.39\textwidth}
    \centering
    \classAVelPanel[670bp 45bp 670bp 45bp]{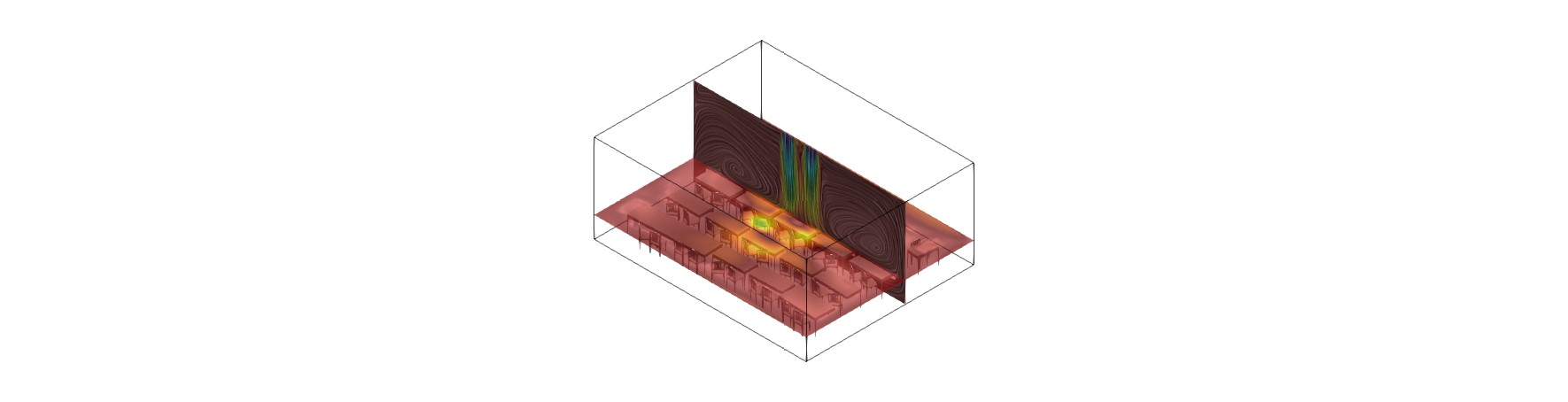}
    \caption{+Chairs +Tables}
\end{subfigure}
&
\begin{subfigure}[b]{0.39\textwidth}
    \centering
    \classAVelPanel[640bp 52bp 640bp 52bp]{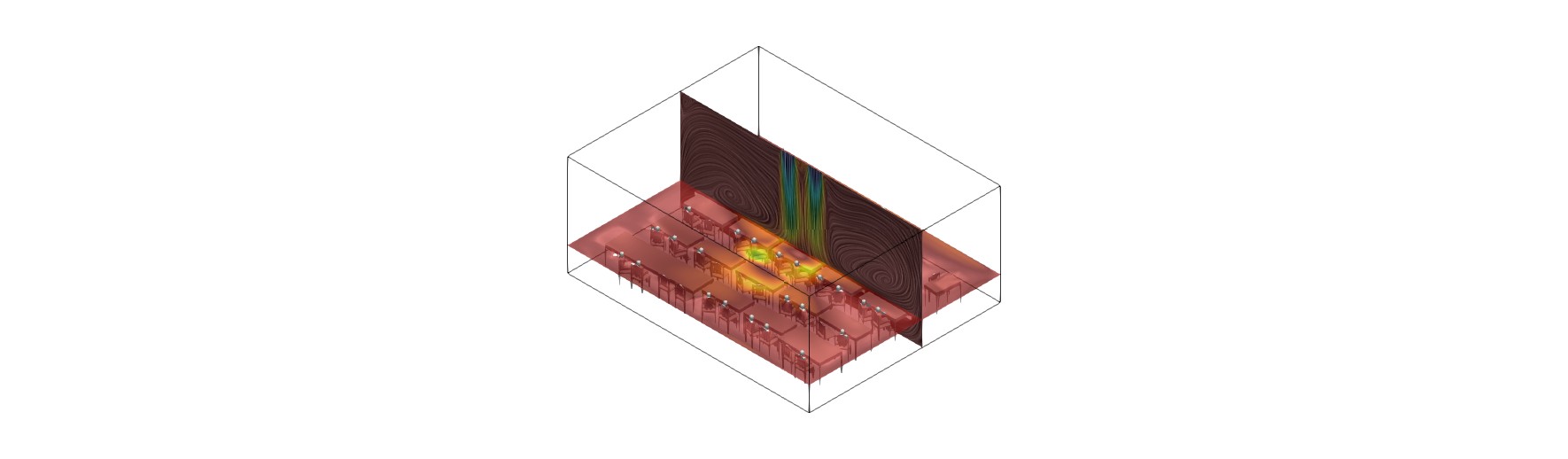}
    \caption{+Chairs +Tables +Mannequins}
\end{subfigure}
\end{tabular}
\end{minipage}
\hfill
\begin{minipage}[c]{0.075\textwidth}
\centering
\includegraphics[
    width=\linewidth,
    height=0.2\textheight,
    keepaspectratio,
    trim=1215bp 340bp 520bp 315bp,
    clip
]{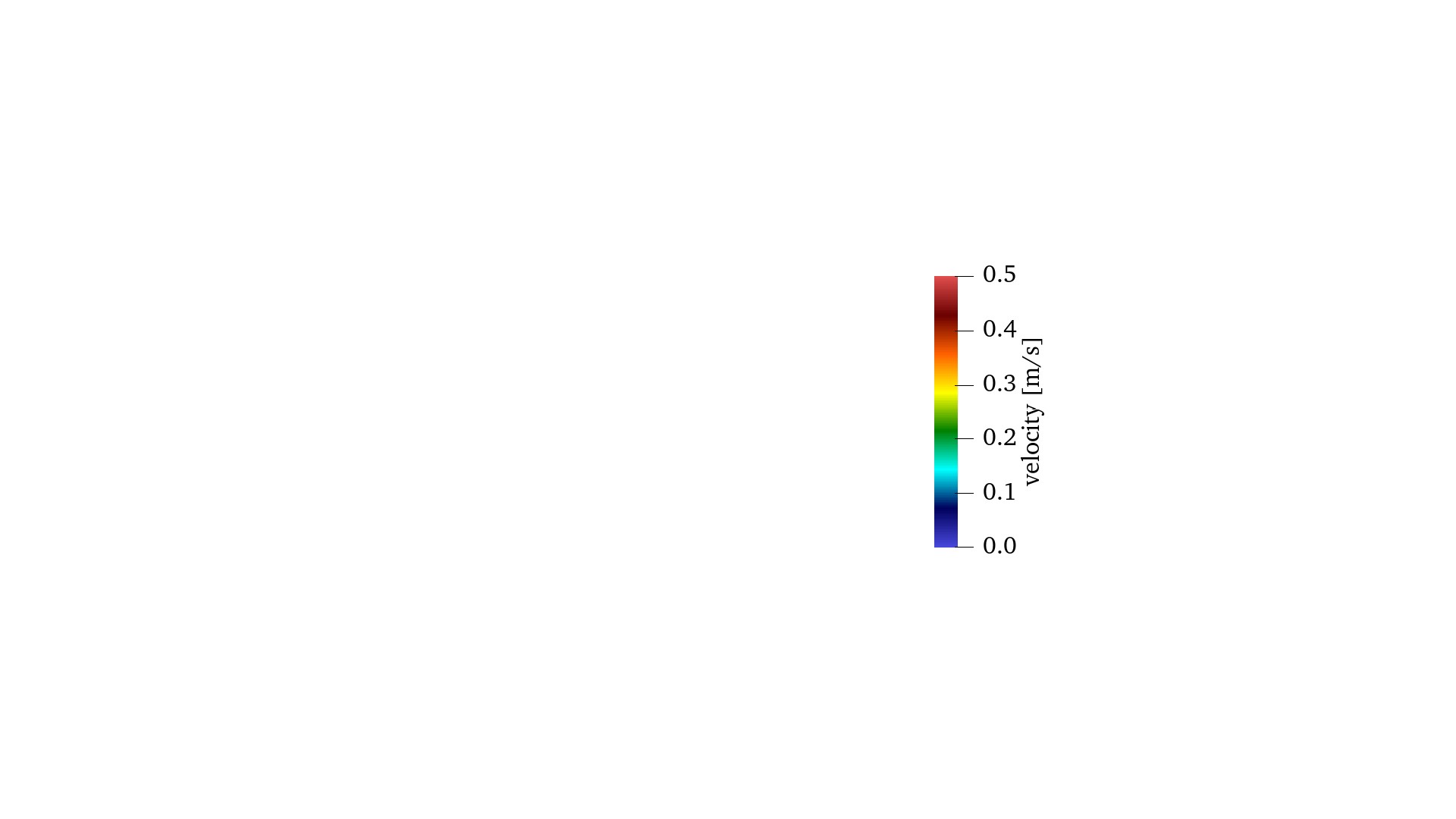}
\end{minipage}

\caption{Steady-state velocity-magnitude fields for the four
Mixed-Furniture Classroom configurations shown in three-dimensional views. Panels (a)--(d) correspond to configurations (i)--(iv): the empty room, +chairs, +chairs+tables, and +chairs+tables+mannequins. The shared colour bar indicates velocity magnitude in m/s. The panels show the steady velocity fields obtained for each configuration.}
\label{fig:classA_velocity}
\end{figure*}

\begin{figure*}[t!]
\centering
\footnotesize
\setlength{\tabcolsep}{4pt}
\renewcommand{\arraystretch}{1.05}

\begin{minipage}[c]{0.90\textwidth}
\centering
\begin{tabular}{@{}cc@{}}
\begin{subfigure}[b]{0.46\textwidth}
    \centering
    \classAVelPanel[668bp 199bp 668bp 199bp]{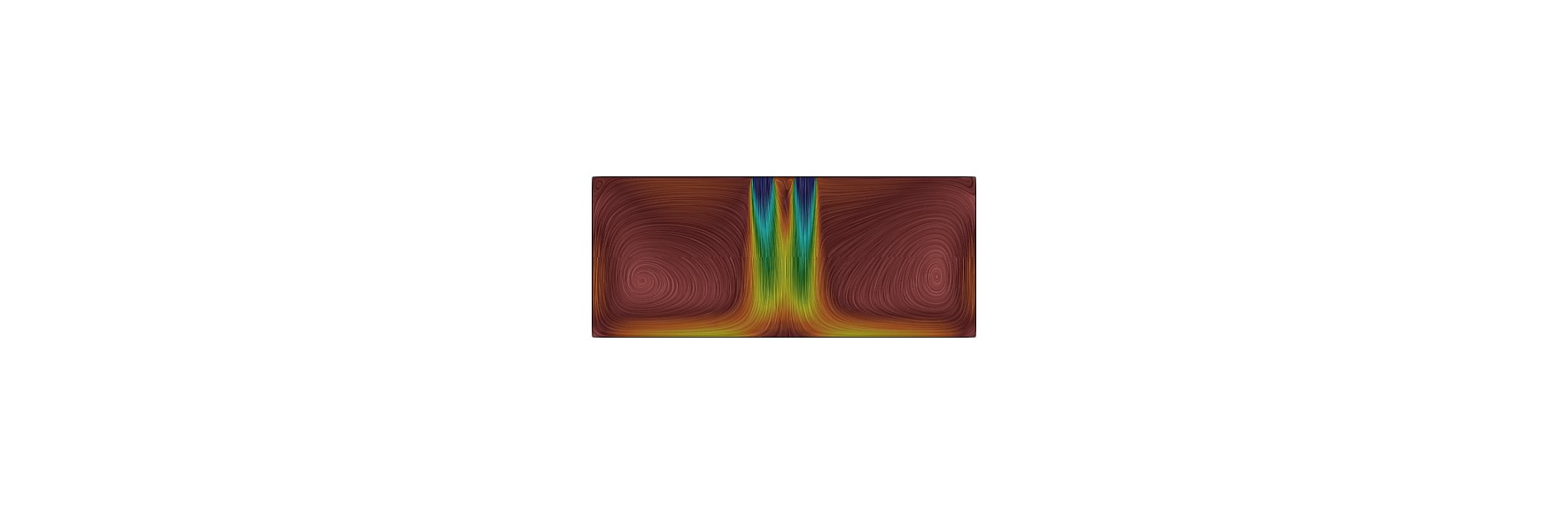}
    \caption{Empty}
\end{subfigure}
&
\begin{subfigure}[b]{0.46\textwidth}
    \centering
    \classAVelPanel[613bp 146bp 613bp 146bp]{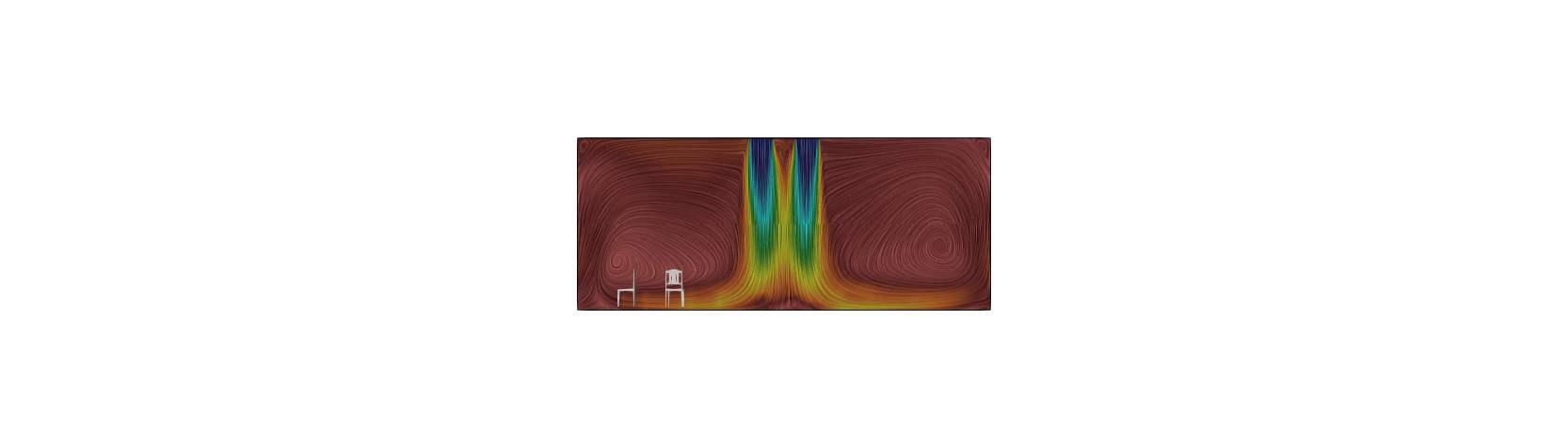}
    \caption{+Chairs}
\end{subfigure}
\\[8pt]
\begin{subfigure}[b]{0.46\textwidth}
    \centering
    \classAVelPanel[715bp 156bp 715bp 156bp]{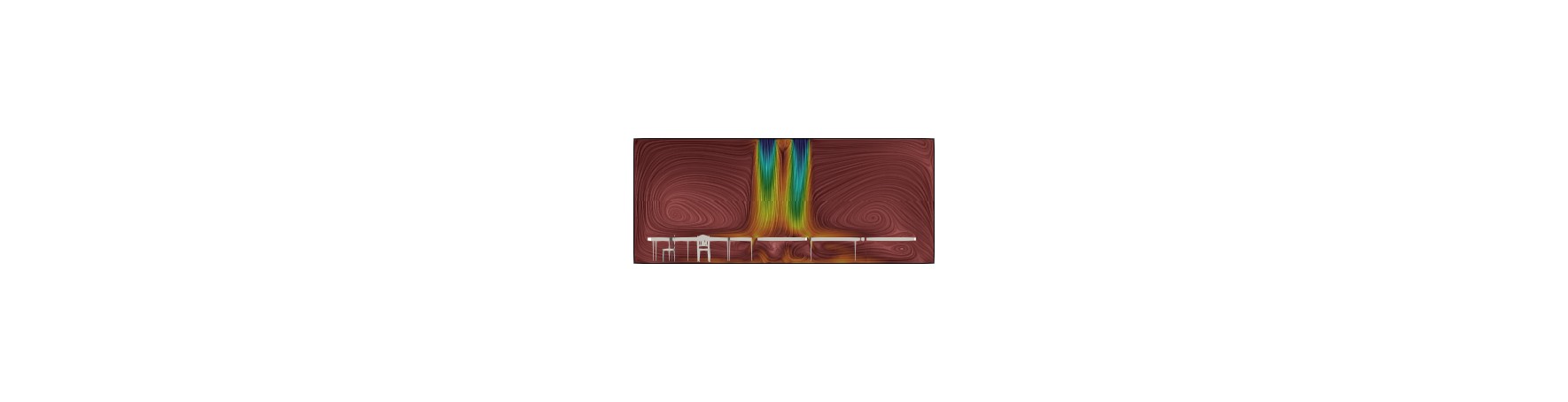}
    \caption{+Chairs +Tables}
\end{subfigure}
&
\begin{subfigure}[b]{0.46\textwidth}
    \centering
    \classAVelPanel[655bp 220bp 651bp 104bp]{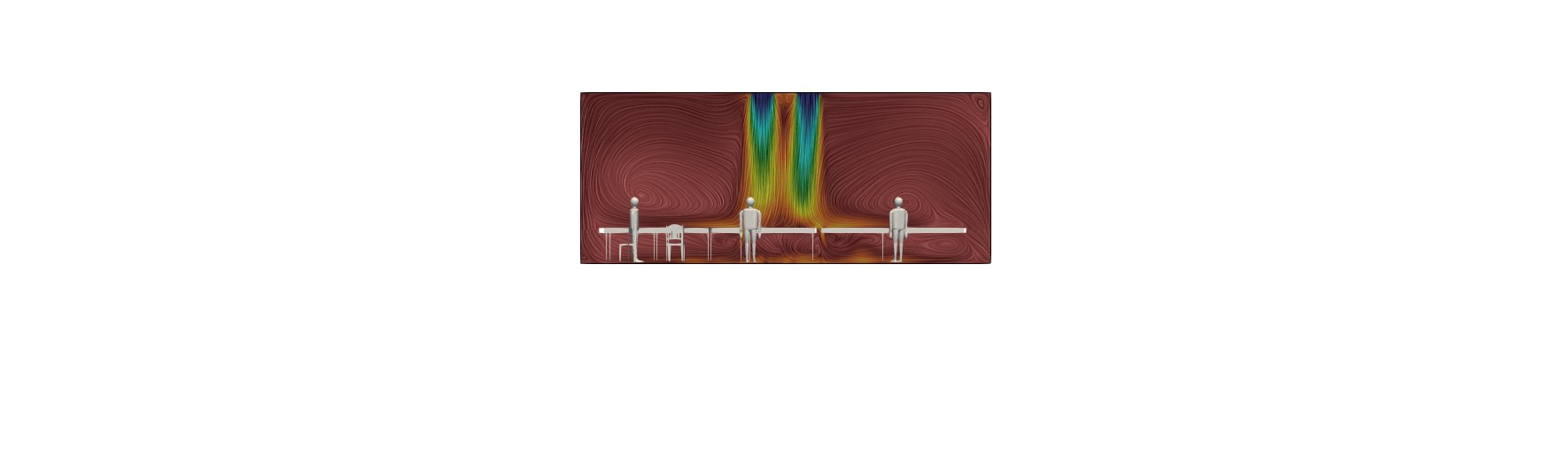}
    \caption{+Chairs +Tables +Mannequins}
\end{subfigure}
\end{tabular}
\end{minipage}
\hfill
\begin{minipage}[c]{0.075\textwidth}
\centering
\includegraphics[
    width=\linewidth,
    height=0.2\textheight,
    keepaspectratio,
    trim=1215bp 340bp 520bp 315bp,
    clip
]{figures/v2-colormap_velocity-3.jpg}
\end{minipage}

\caption{Steady-state vertical mid-plane velocity-magnitude slices for the four Mixed-Furniture Classroom configurations. Panels (a)--(d) correspond to configurations (i)--(iv): the empty room, +chairs, +chairs+tables, and +chairs+tables+mannequins. The shared colour bar indicates velocity magnitude in m/s. The panels show the mid-plane velocity fields obtained for each configuration.}
\label{fig:classA_velocity_test2}
\end{figure*}

\begin{figure*}[t!]
\centering
\footnotesize
\setlength{\tabcolsep}{2pt}
\renewcommand{\arraystretch}{1.15}

\newcommand{\scalarpanel}[1]{%
    \includegraphics[
        width=\linewidth,
        trim=480bp 55bp 480bp 35bp,
        clip
    ]{#1}%
}

\begin{tabular}{@{}|>{\centering\arraybackslash}m{0.12\textwidth}|>{\centering\arraybackslash}m{0.26\textwidth}|>{\centering\arraybackslash}m{0.26\textwidth}|>{\centering\arraybackslash}m{0.26\textwidth}|@{}}
\hline
&
\textbf{$t=5$~s}
&
\textbf{$t=60$~s}
&
\textbf{$t=300$~s}
\\
\hline
\makecell[c]{\textbf{(i)}\\[2pt]\textbf{Empty}}
&
\scalarpanel{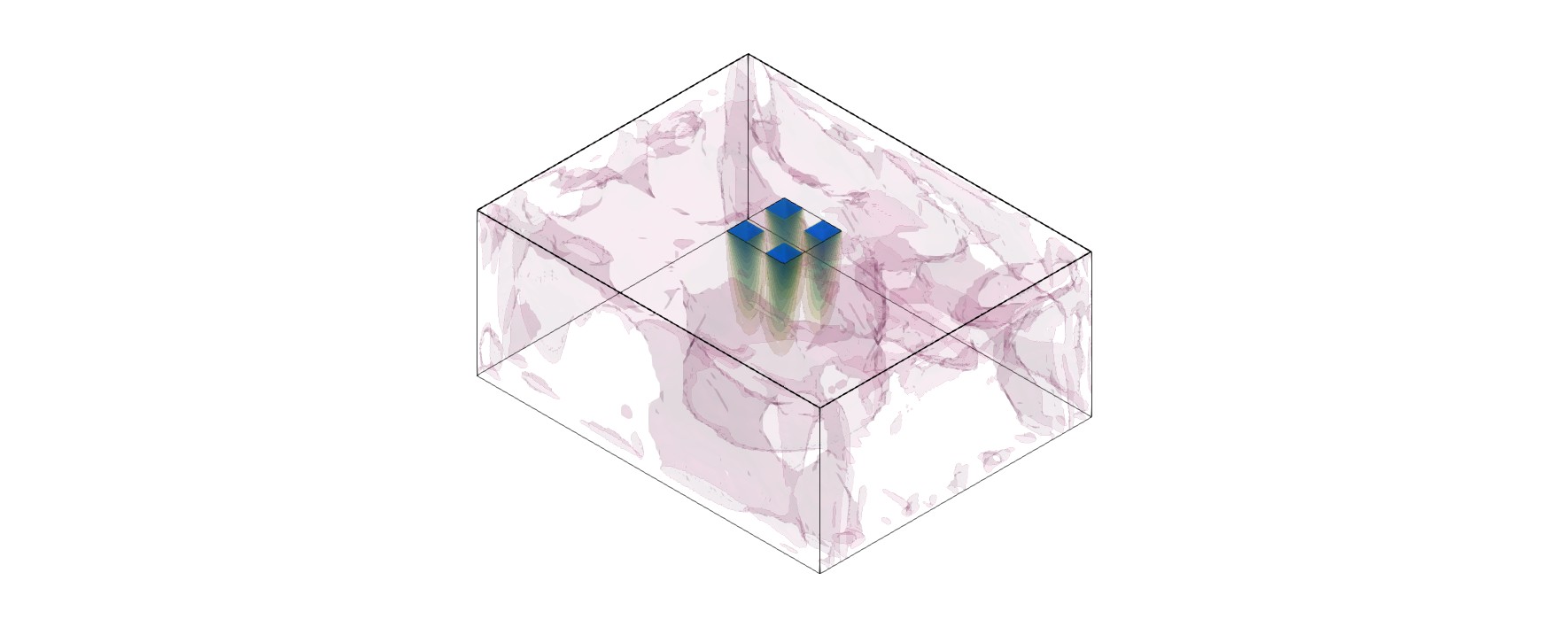}
&
\scalarpanel{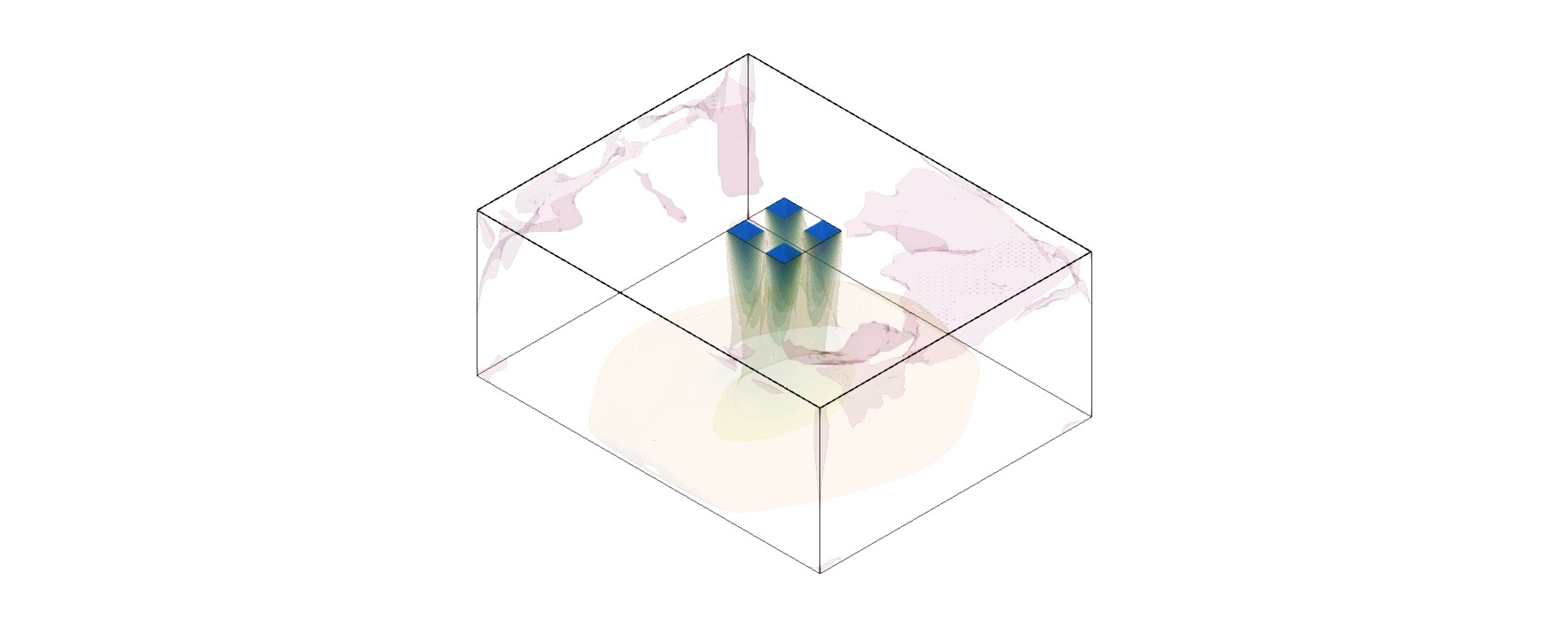}
&
\scalarpanel{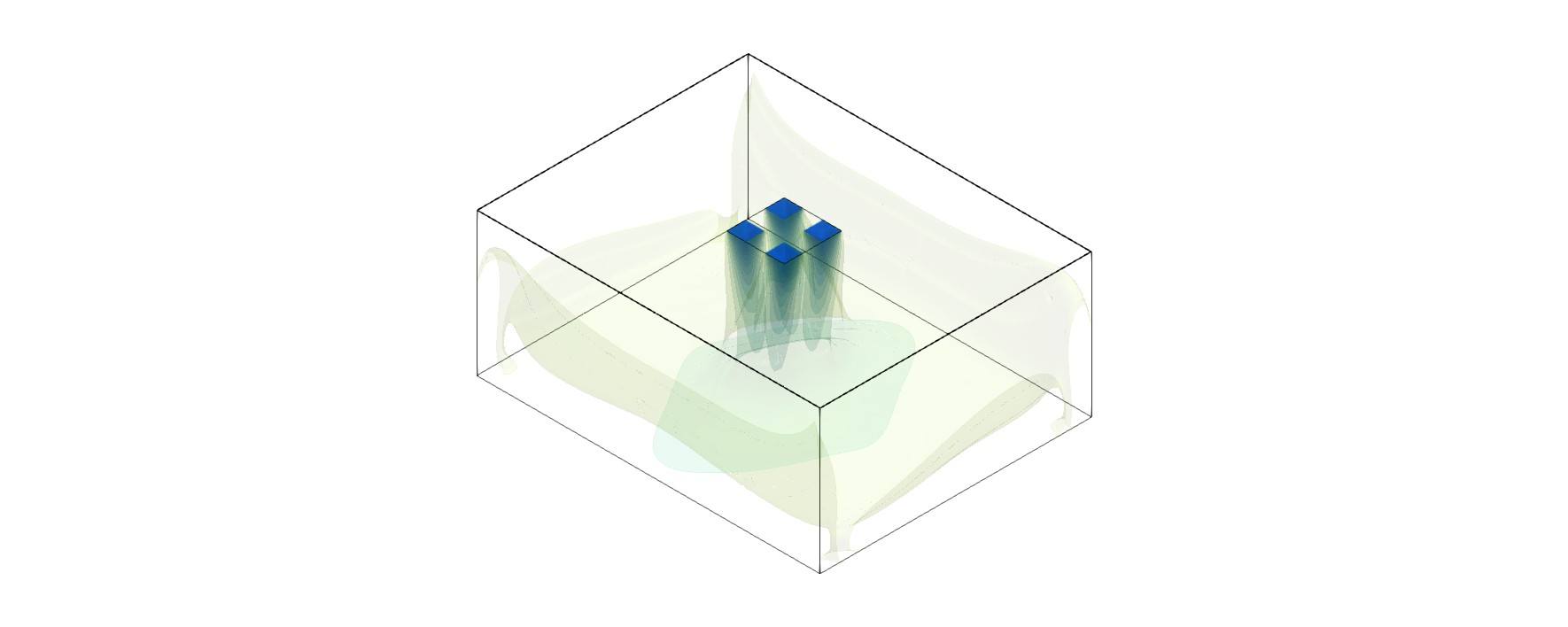}
\\
\hline
\makecell[c]{\textbf{(ii)}\\[2pt]\textbf{+Chairs}}
&
\scalarpanel{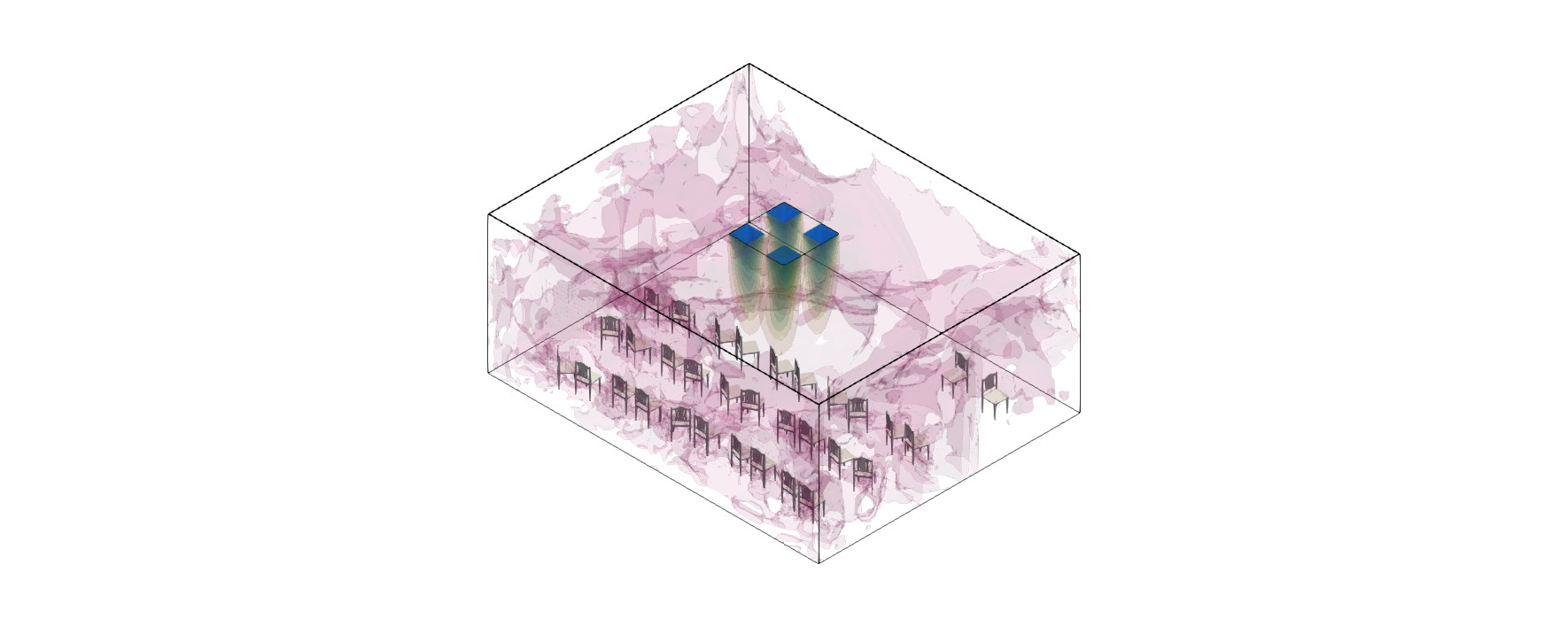}
&
\scalarpanel{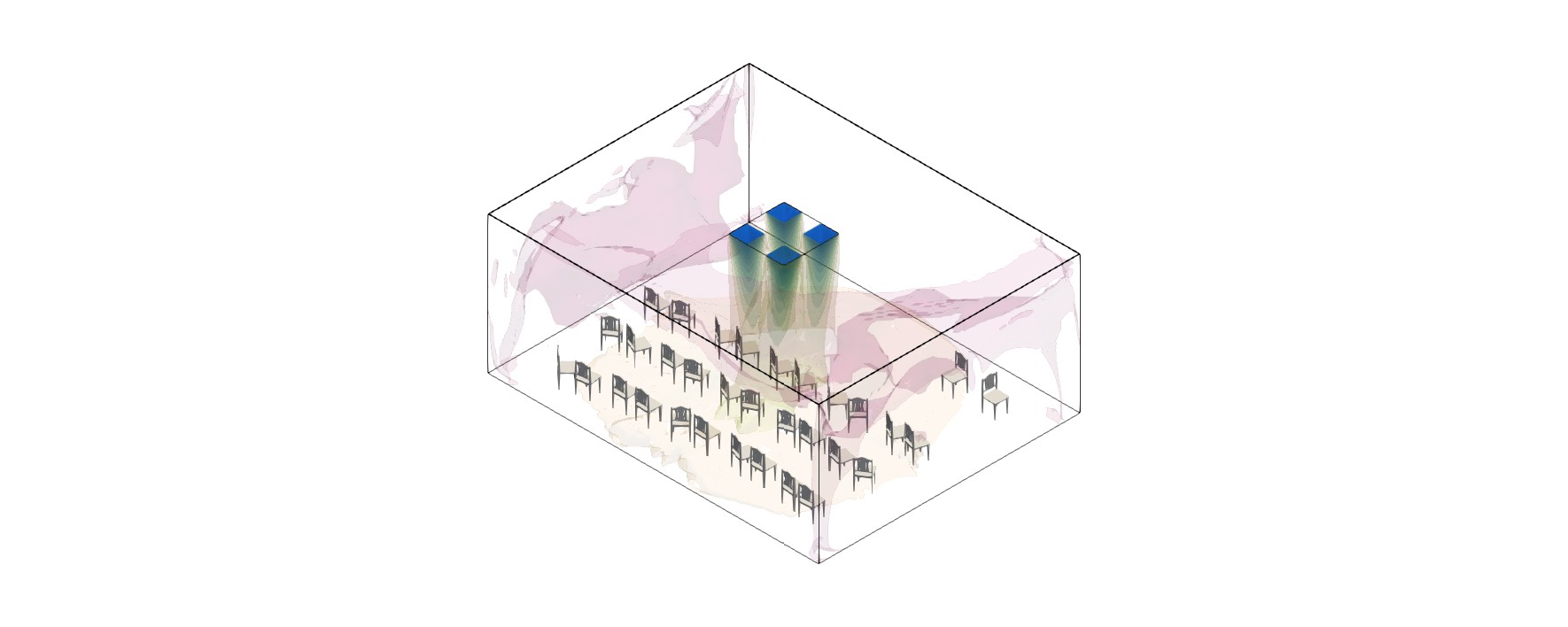}
&
\scalarpanel{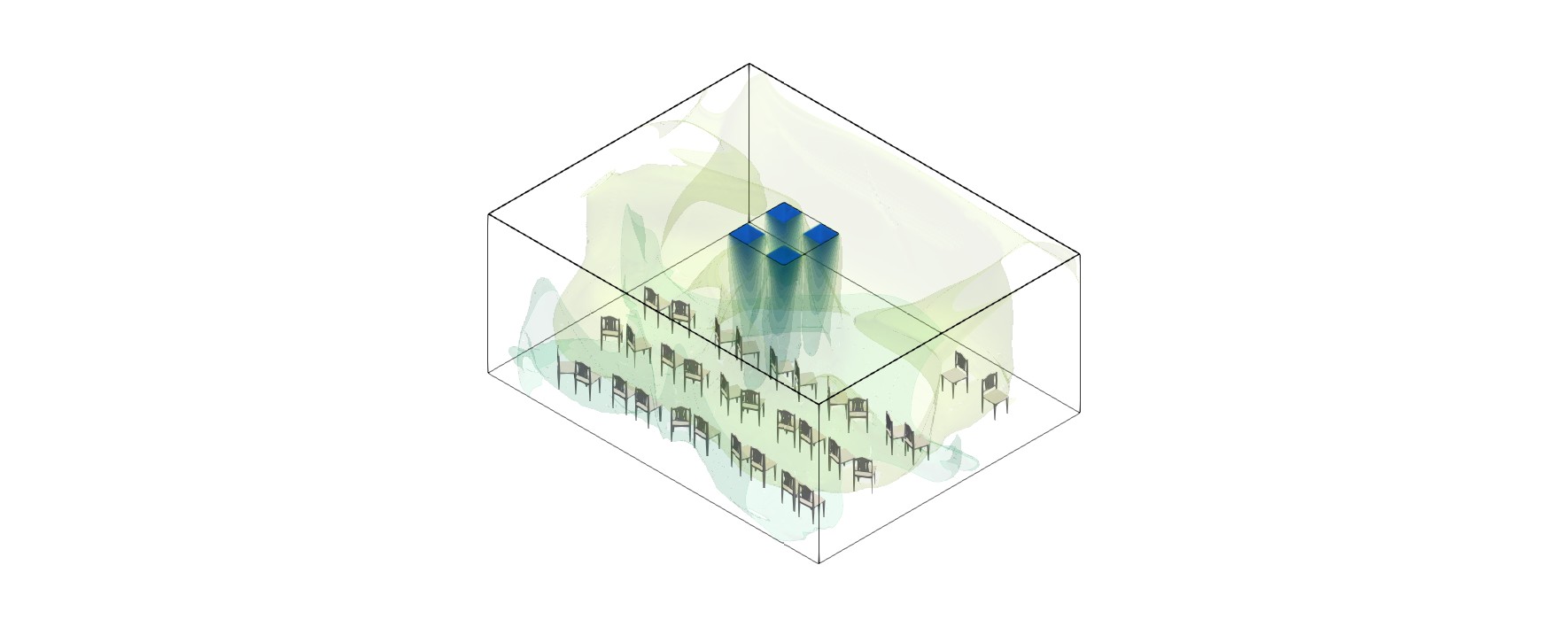}
\\
\hline
\makecell[c]{\textbf{(iii)}\\[2pt]\textbf{+Chairs}\\\textbf{+Tables}}
&
\scalarpanel{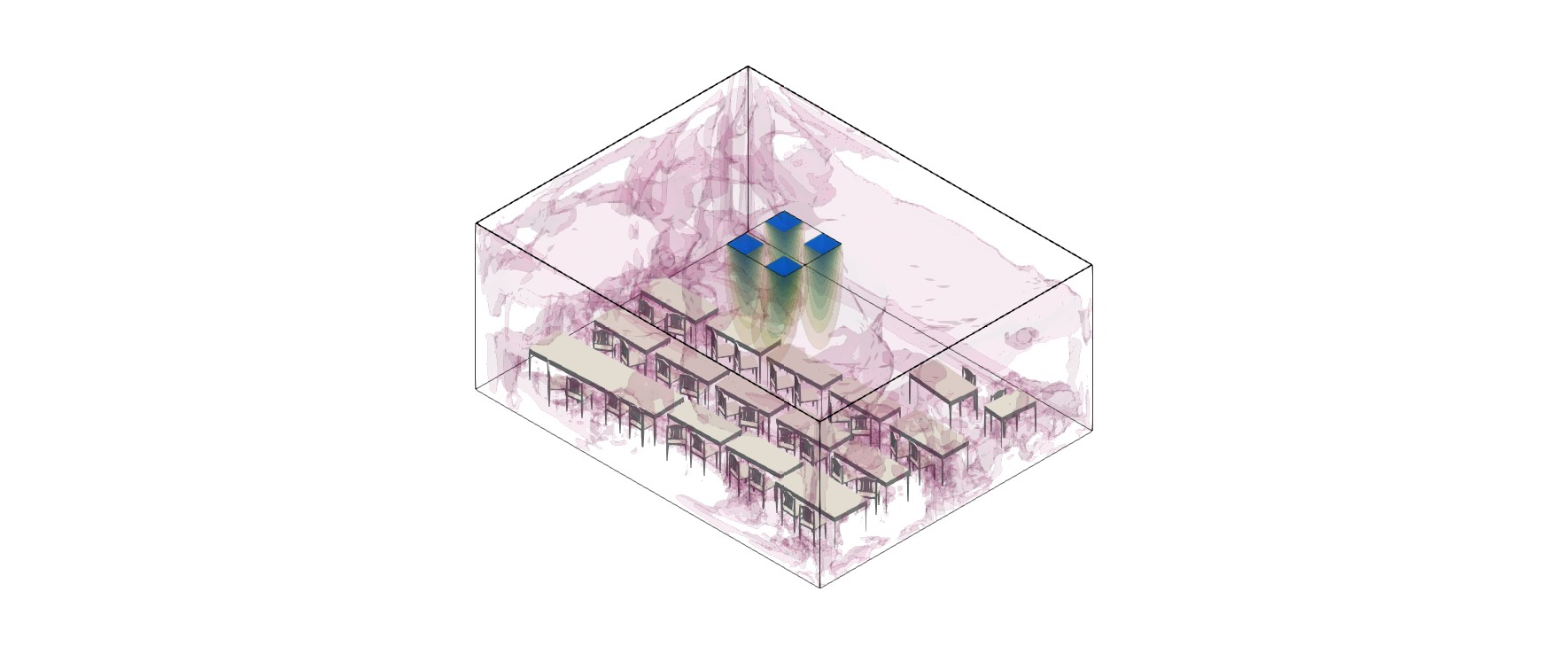}
&
\scalarpanel{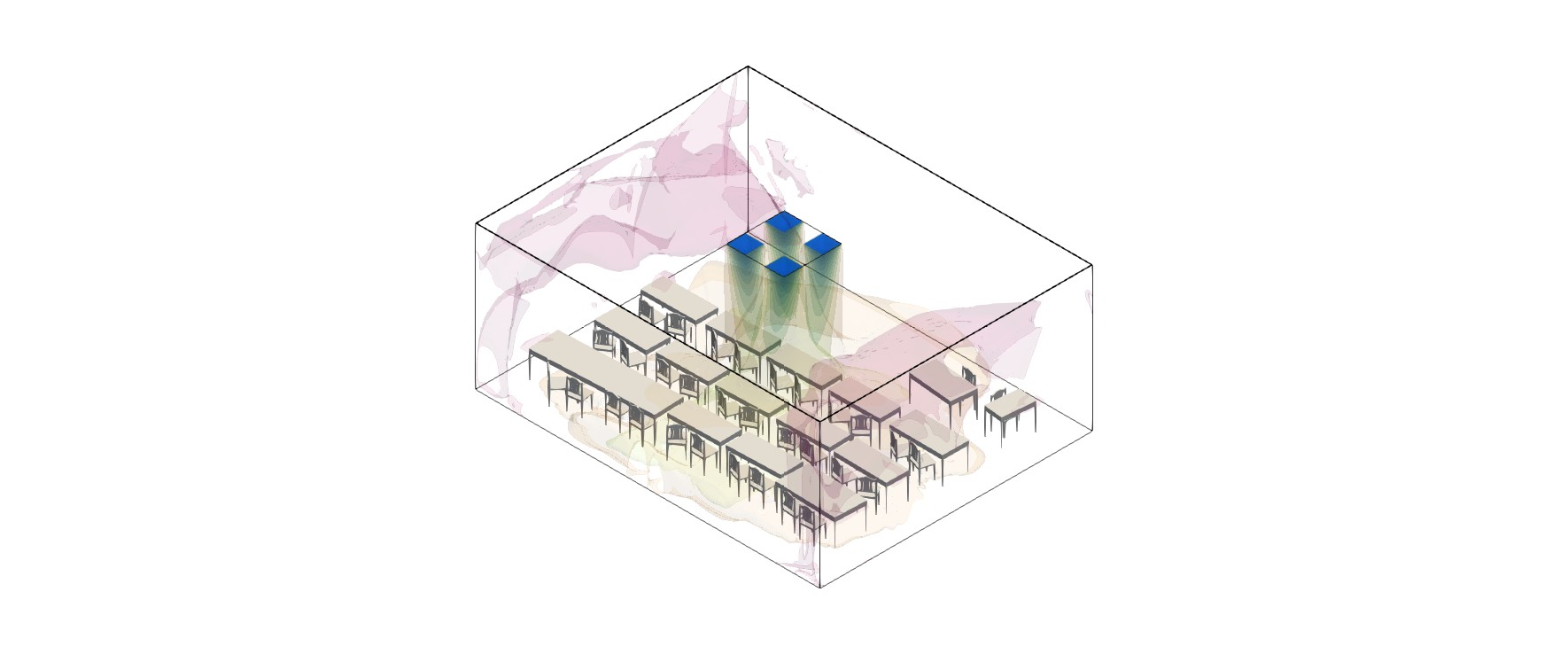}
&
\scalarpanel{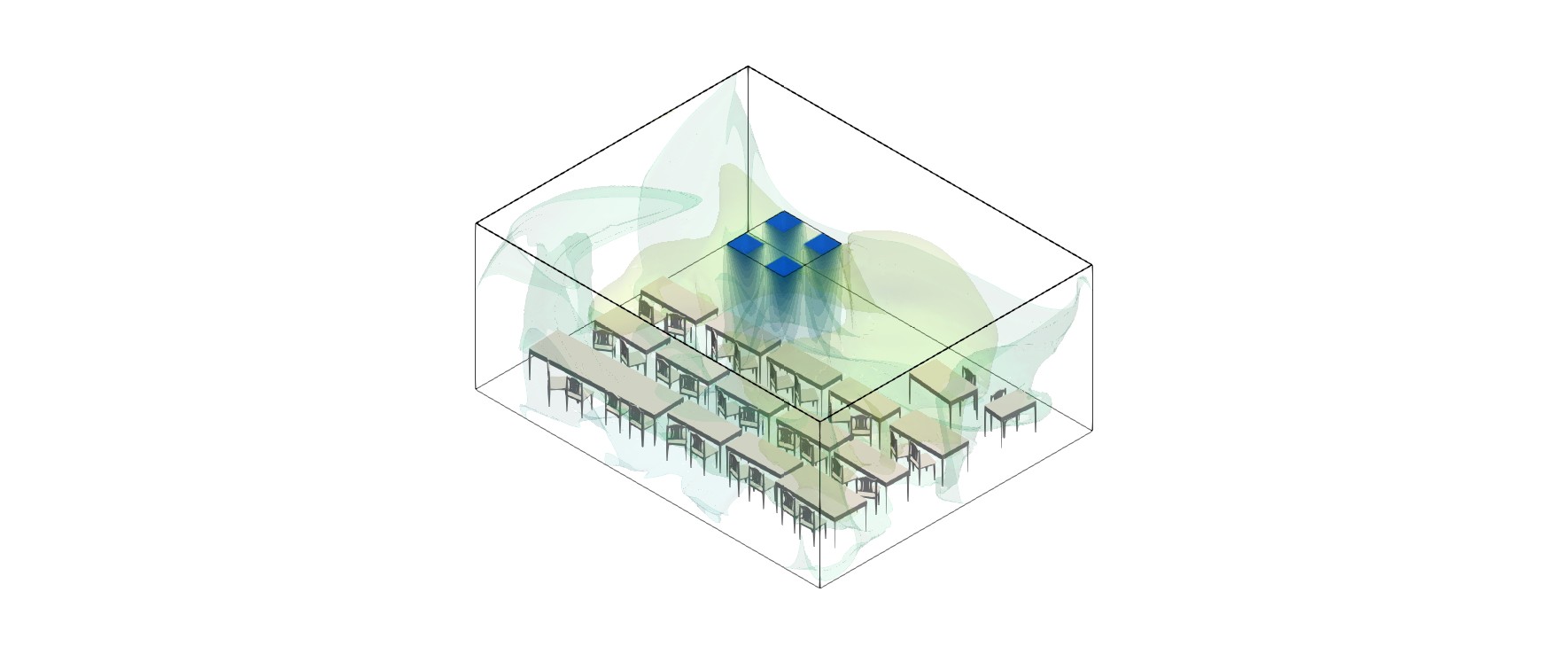}
\\
\hline
{\scriptsize\makecell[c]{\textbf{(iv)}\\[2pt]\textbf{+Chairs}\\\textbf{+Tables}\\\textbf{+Mannequins}}}
&
\scalarpanel{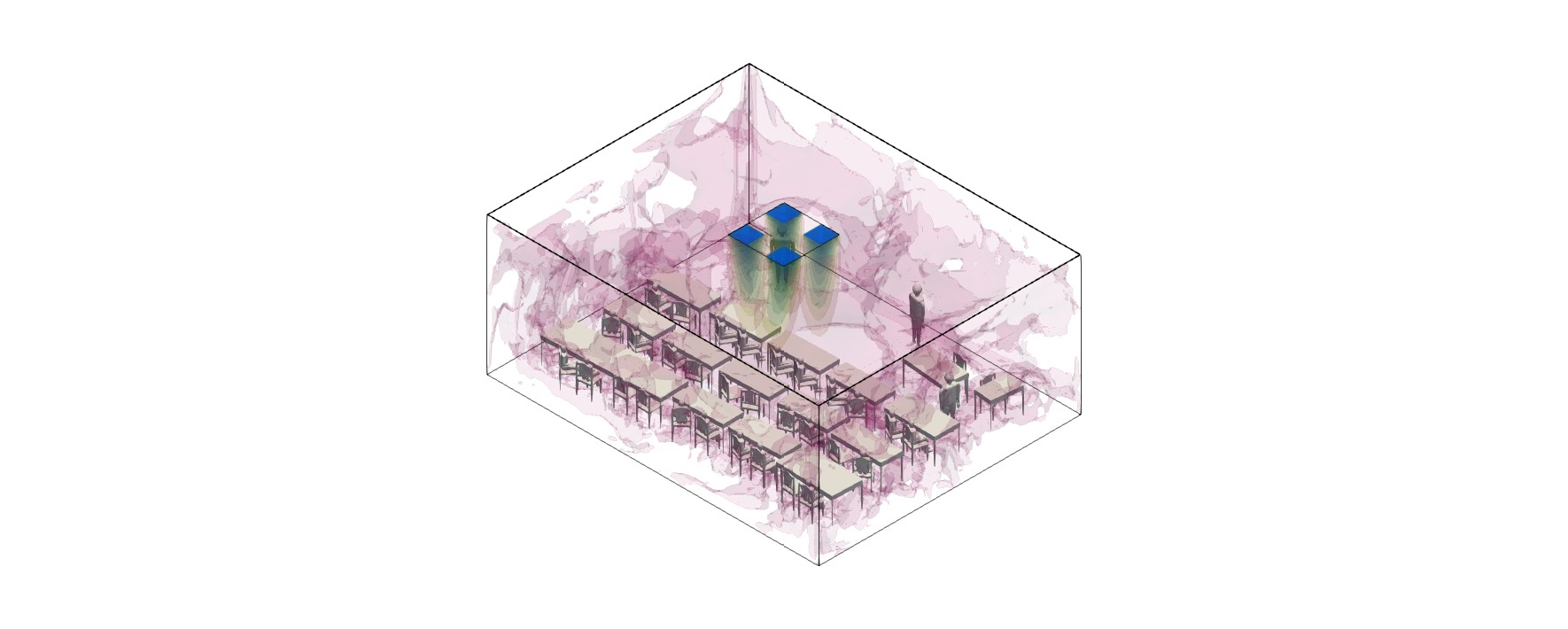}
&
\scalarpanel{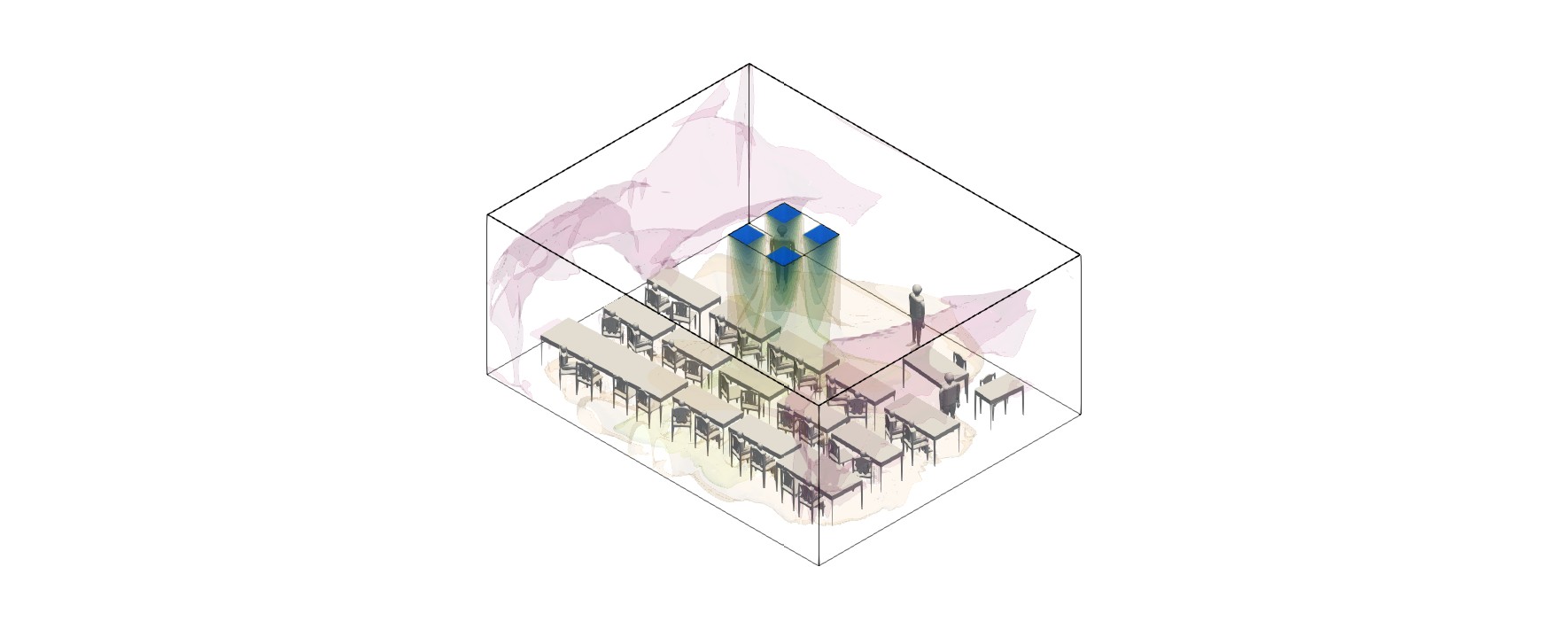}
&
\scalarpanel{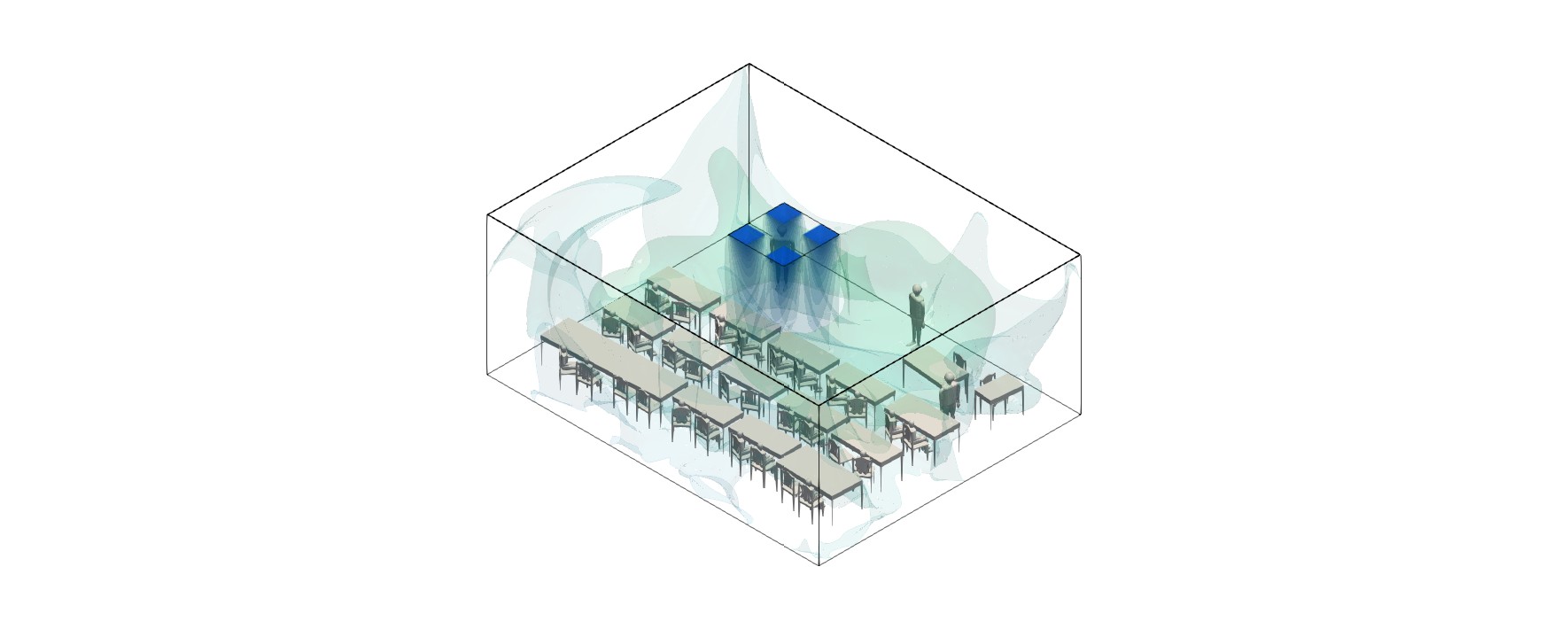}
\\
\hline
\end{tabular}

\vspace{3pt}

\includegraphics[
    width=0.5\textwidth,
    keepaspectratio,
    trim=500bp 300bp 500bp 650bp,
    clip
]{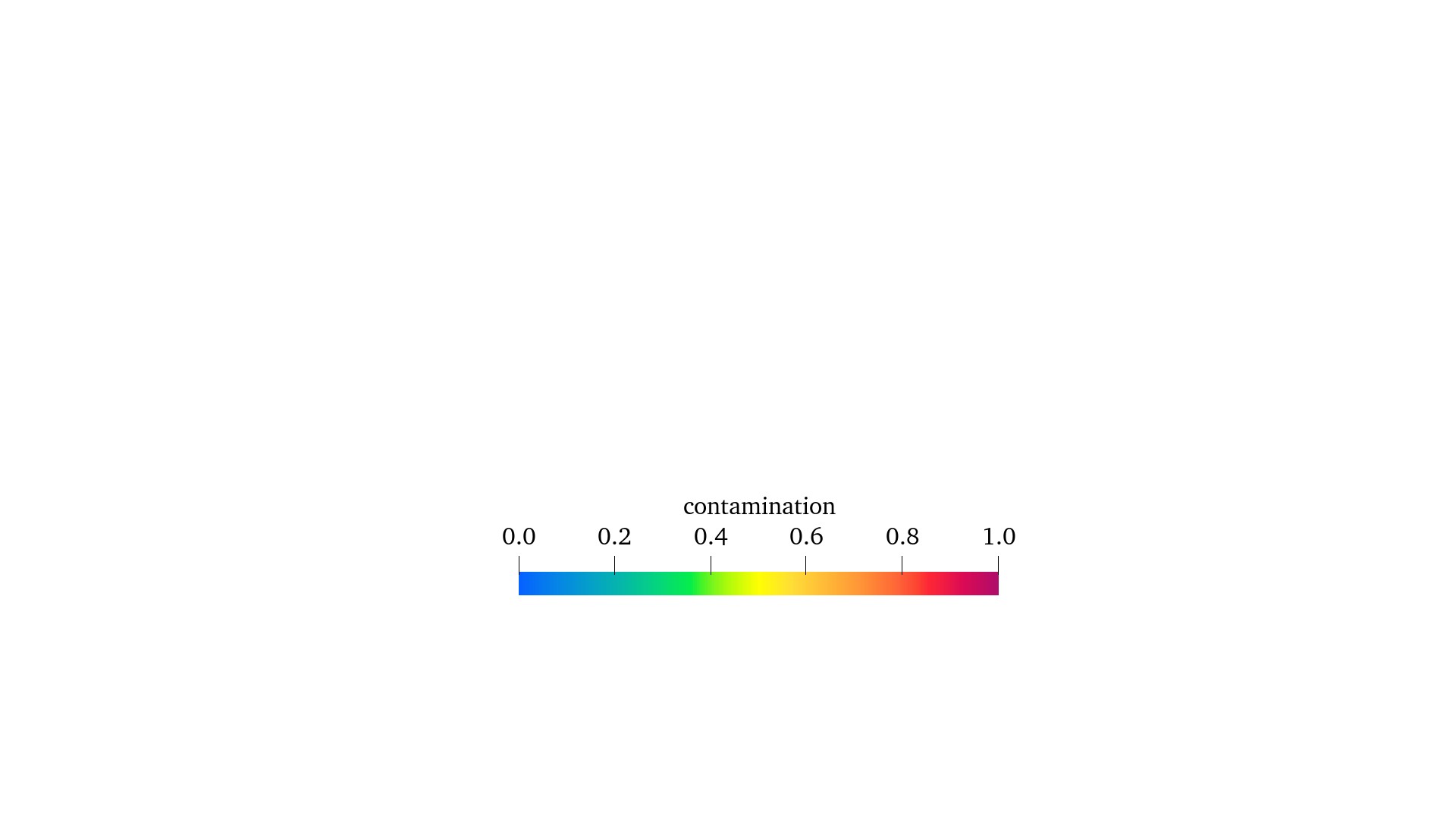}

\caption{Passive-scalar concentration evolution for the Mixed-Furniture Classroom what-if sequence. Rows correspond to the four configurations shown in \figref{fig:classA_recon}: (i)~room shell only, (ii)~room shell with 32 chairs, (iii)~room shell with 32 chairs and 16 tables, and (iv)~room shell with 32 chairs, 16 tables, and 25 mannequins (22 seated, 3 standing). Columns show matched scalar-field snapshots at $t=5$~s, $60$~s, and $300$~s after the start of the scalar-decay window. All panels are rendered from the same viewpoint and use the same normalized concentration scale, $C/C_0 \in [0,1]$, shown by the shared color bar.}
\label{fig:classA_scalar}
\end{figure*}

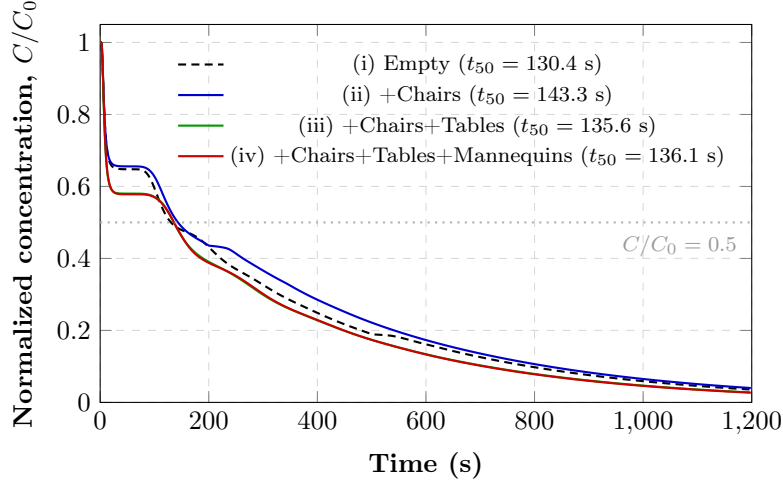
\begin{figure}[t!]
\centering
\begin{tikzpicture}
\begin{axis}[
    width=0.62\linewidth, height=0.40\linewidth,
    xlabel={\textbf{Time (s)}}, ylabel={\textbf{Normalized concentration, $C/C_0$}},
    xmin=0, xmax=1200, ymin=0, ymax=1.05,
    grid=major, grid style={dashed, gray!30},
    legend style={font=\footnotesize, at={(0.98,0.95)}, anchor=north east, draw=none, fill=none}
]
\addplot[color=black, thick, densely dashed, each nth point=10, filter discard warning=false, unbounded coords=discard]
    table[col sep=comma, header=false, x index=0, y index=1]
    {figures/Class_A_v1_Empty.csv};
\addlegendentry{(i) Empty ($t_{50}=130.4$ s)}

\addplot[color=blue!80!black, thick, each nth point=5, filter discard warning=false, unbounded coords=discard]
    table[col sep=comma, header=false, x index=0, y index=1]
    {figures/Class_A_v2_Chairs.csv};
\addlegendentry{(ii) +Chairs ($t_{50}=143.3$ s)}

\addplot[color=green!60!black, thick, each nth point=4, filter discard warning=false, unbounded coords=discard]
    table[col sep=comma, header=false, x index=0, y index=1]
    {figures/Class_A_v3_Tables.csv};
\addlegendentry{(iii) +Chairs+Tables ($t_{50}=135.6$ s)}

\addplot[color=red!80!black, thick, each nth point=3, filter discard warning=false, unbounded coords=discard]
    table[col sep=comma, header=false, x index=0, y index=1]
    {figures/Class_A_v4_Heavy.csv};
\addlegendentry{(iv) +Chairs+Tables+Mannequins ($t_{50}=136.1$ s)}

\draw[gray!60, dotted, thick] (axis cs:0,0.5) -- (axis cs:1200,0.5);
\node[gray!80, anchor=north east, font=\footnotesize] at (axis cs:1190,0.49) {$C/C_0 = 0.5$};
\end{axis}
\end{tikzpicture}
\caption{Mixed-furniture classroom what-if sequence: normalized passive-scalar concentration histories for the four configurations. The horizontal reference line marks \(C/C_0=0.5\), corresponding to the \(t_{50}\) half-clearance threshold. The curves show non-monotonic variation of the monitor-based clearance across the four configurations; additional metrics are reported in \tableref{tab:clearance_metrics}.}
\label{fig:classA_decay}
\end{figure}

\figref{fig:classA_recon} presents the reconstructed assets for all four configurations together; \figref{fig:classA_velocity} and \figref{fig:classA_velocity_test2} compare the steady velocity field across configurations in three-dimensional views and vertical mid-plane slice views, respectively; \figref{fig:classA_scalar} compares passive-scalar transport snapshots; and \figref{fig:classA_decay} compares the normalized concentration decay curves with computed half-clearance times.

The mixed-furniture classroom decay curves (\figref{fig:classA_decay}) show that the monitor-based scalar clearance does not vary monotonically with the amount of interior furniture. The empty room-shell baseline clears fastest at $t_{50}=130.4$~s. Adding chairs slows the clearance ($t_{50}=143.3$~s, $+9.9\%$ relative to the empty room). Adding tables in addition to the chairs returns the value to $t_{50}=135.6$~s, and the fully obstructed configuration with mannequins yields $t_{50}=136.1$~s, a further $0.5$~s that is too small, relative to the numerical evidence presented here, to be treated as a resolved effect. Because the ventilation forcing is held fixed across all four cases, these differences track the change in interior geometry; the associated flow field is examined in \secref{sec:discussion_classA}.

\subsection{Chair-Dominant Classroom}
\label{sec:results_chairdominant}
The Chair-Dominant Classroom contains 47~chairs and 3~tables (ground truth), with the pipeline producing 45~chairs and 2~tables in the final reconstruction (\tableref{tab:recon_counts}). It is a different mechanically ventilated classroom with a chair-heavy layout and limited table count, and its successful processing suggests that the reconstruction-to-CFD workflow transfers beyond the primary study environment. A single configuration with full furniture and mannequins is simulated. \figref{fig:chair_dominant_combined} presents the reconstruction together with the steady velocity field, scalar transport snapshots, and decay curve.

\newcommand{\classGridVelColorbar}{%
    \includegraphics[
        height=\dimexpr\classGridCFDHeight*5/10\relax,
        keepaspectratio,
        trim=1215bp 340bp 520bp 315bp,
        clip
    ]{figures/v2-colormap_velocity-3.jpg}%
}

\newcommand{\classGridContamColorbar}{%
    \includegraphics[
        height=\dimexpr\classGridCFDHeight*5/10\relax,
        keepaspectratio,
        trim=1045bp 315bp 700bp 320bp,
        clip
    ]{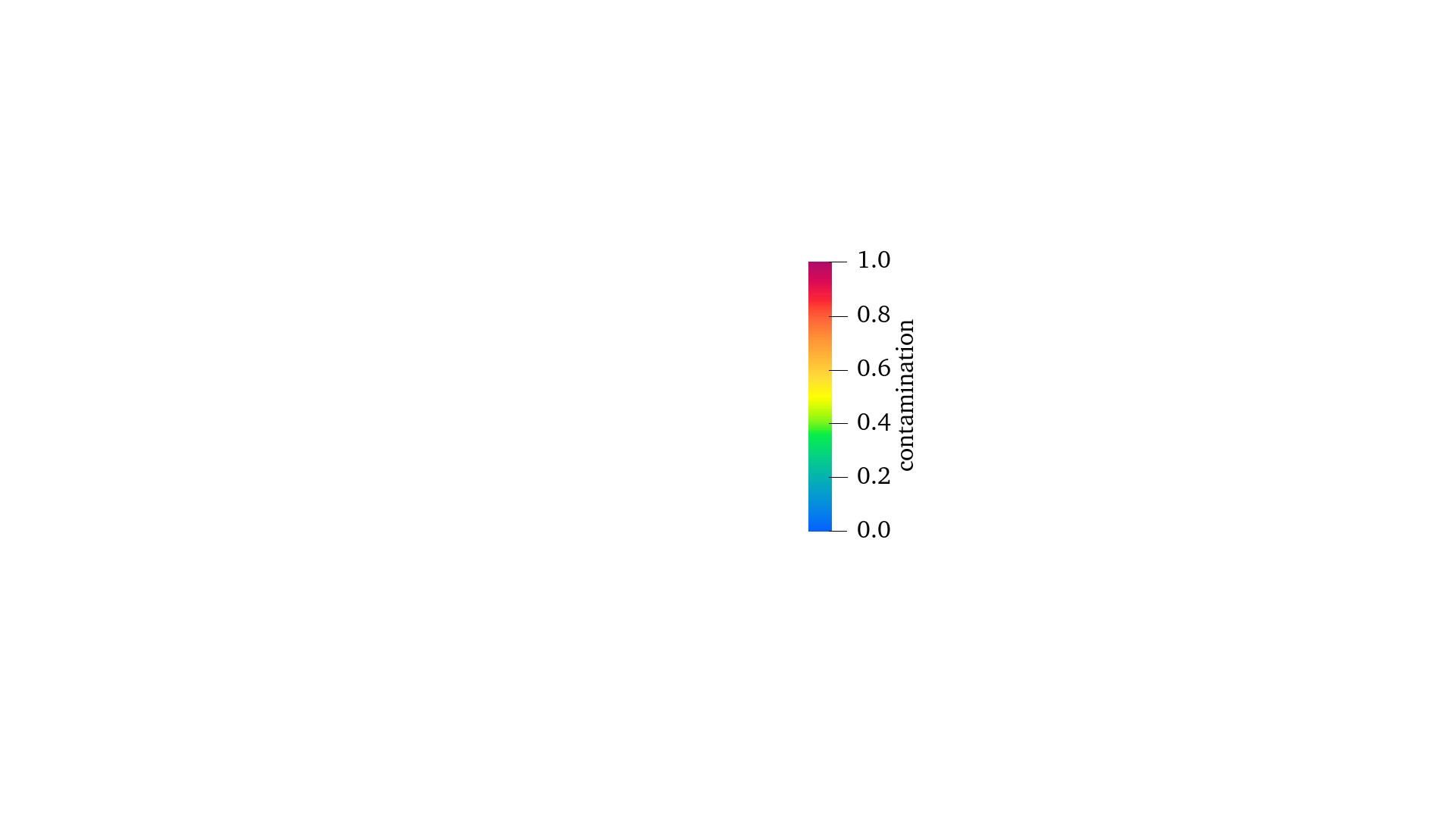}%
}

\newcommand{\classGridCFDPanelWithBar}[3]{%
    \begin{minipage}[c]{0.78\linewidth}
        \centering
        \classGridCFDPanel[#1]{#2}
    \end{minipage}%
    \hspace{1pt}%
    \begin{minipage}[c]{0.19\linewidth}
        \centering
        #3
    \end{minipage}%
}

\begin{figure*}[t!]
\centering
\footnotesize
\begin{tabular}{@{}ccc@{}}
\begin{subfigure}[b]{0.3\linewidth}
    \centering
    \classGridTopPanel{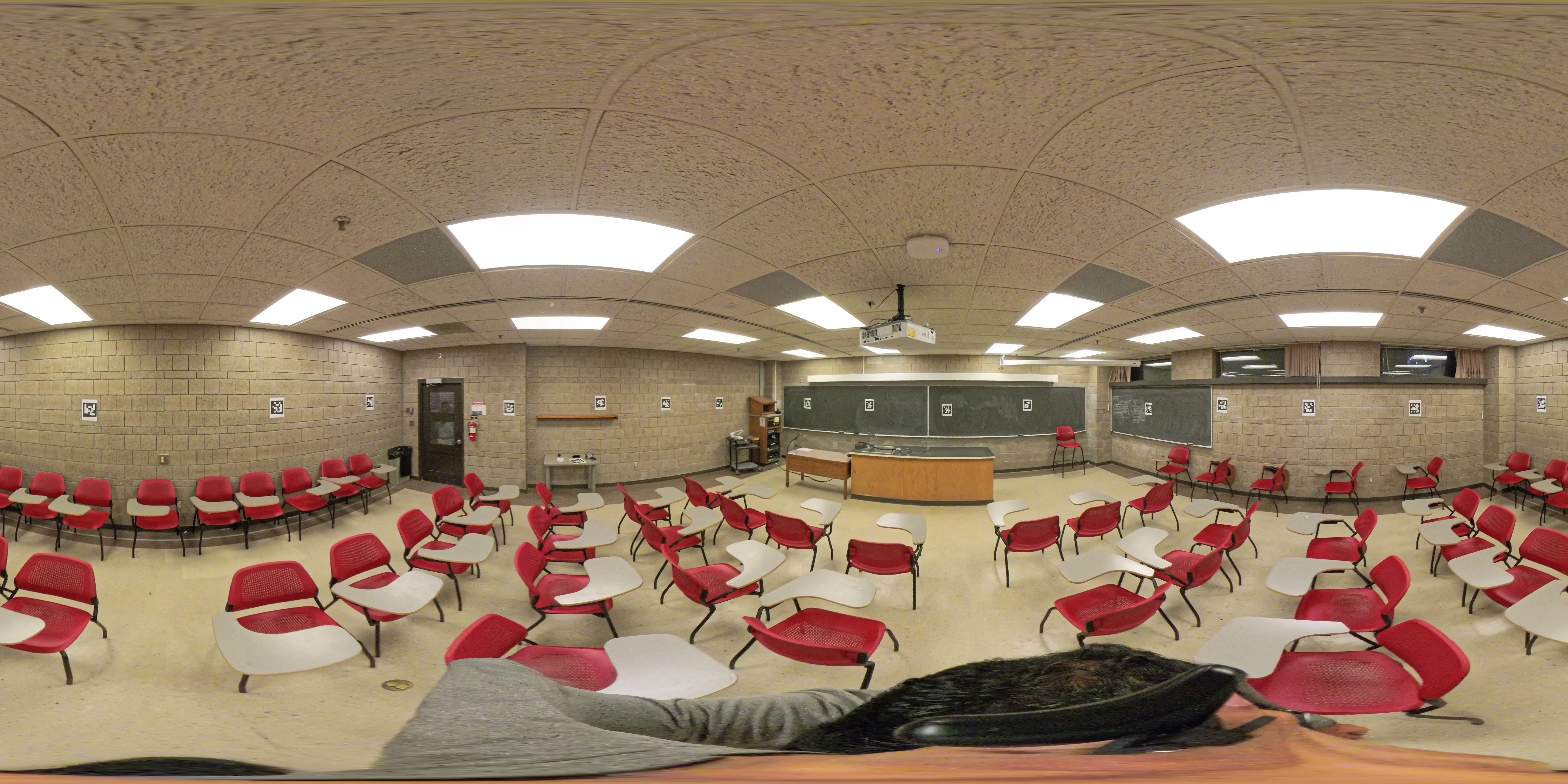}
    \caption{Input 360\textdegree{} frame}
\end{subfigure}
&
\begin{subfigure}[b]{0.3\linewidth}
    \centering
    \classGridTopPanel{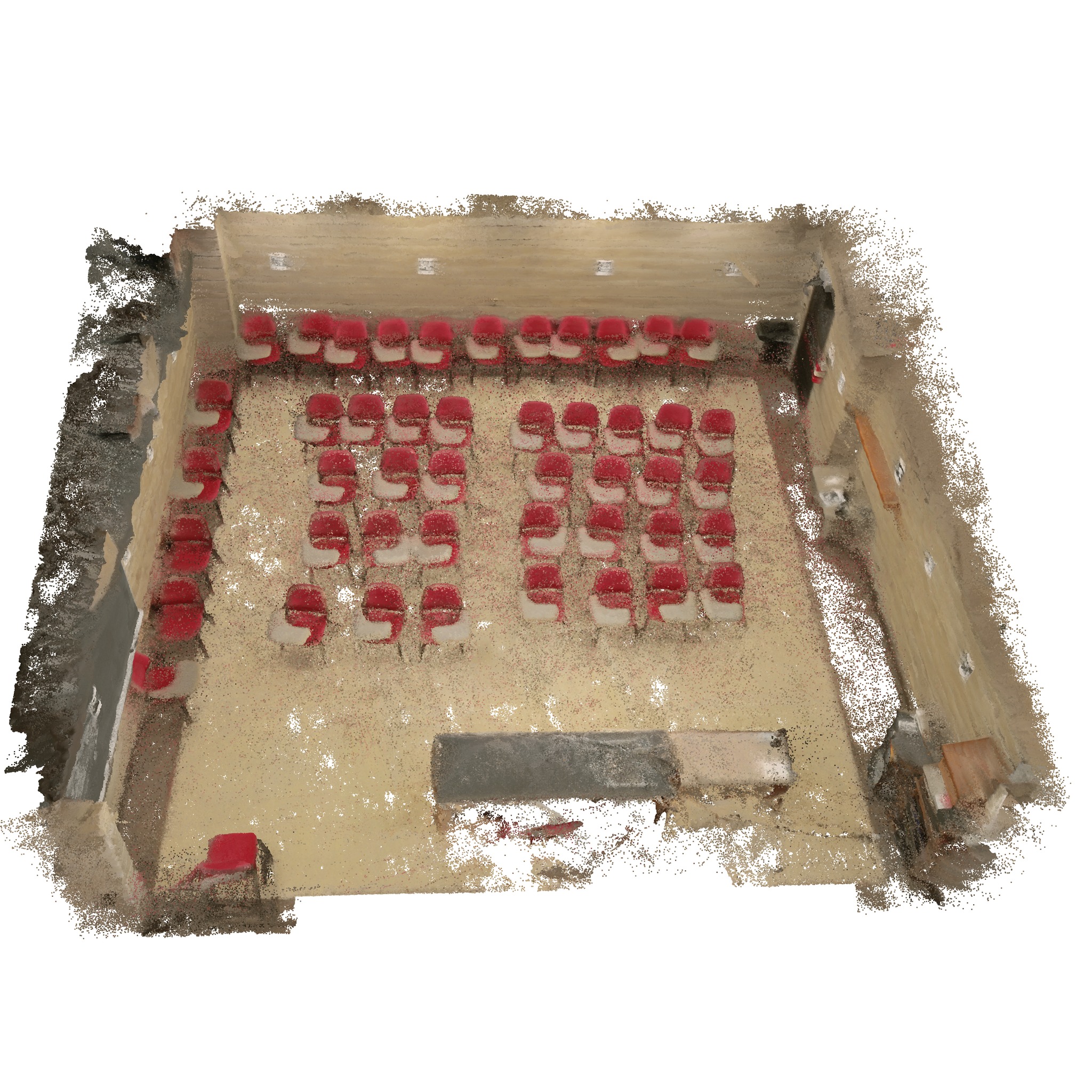}
    \caption{NeRF point cloud}
\end{subfigure}
&
\begin{subfigure}[b]{0.3\linewidth}
    \centering
    \classGridTopPanel{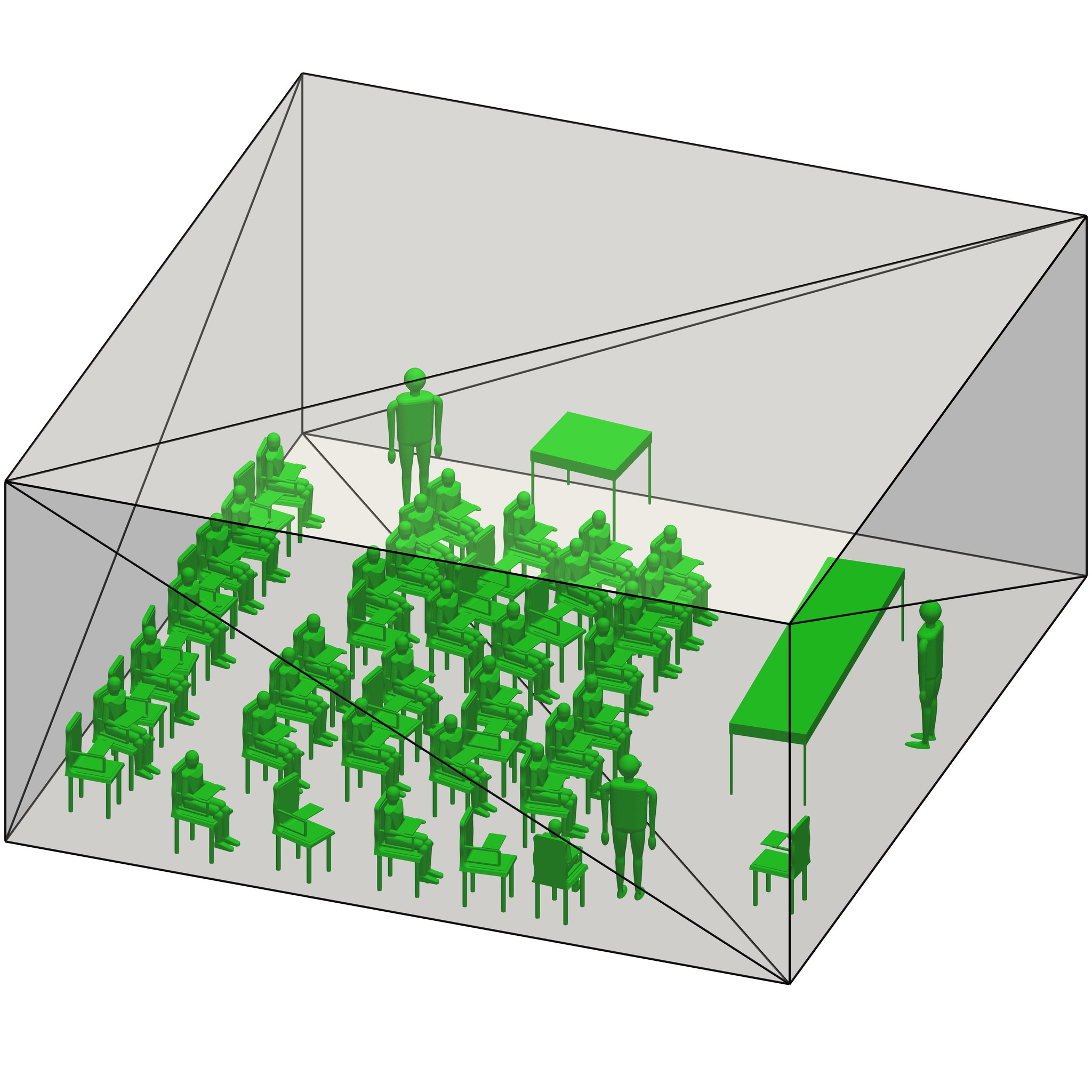}
    \caption{Simulation geometry}
\end{subfigure}
\end{tabular}
\begin{tabular}{@{}cc@{}}
\begin{subfigure}[b]{0.45\linewidth}
    \centering
    \classGridCFDPanelWithBar
        {530bp 51bp 530bp 50bp}
        {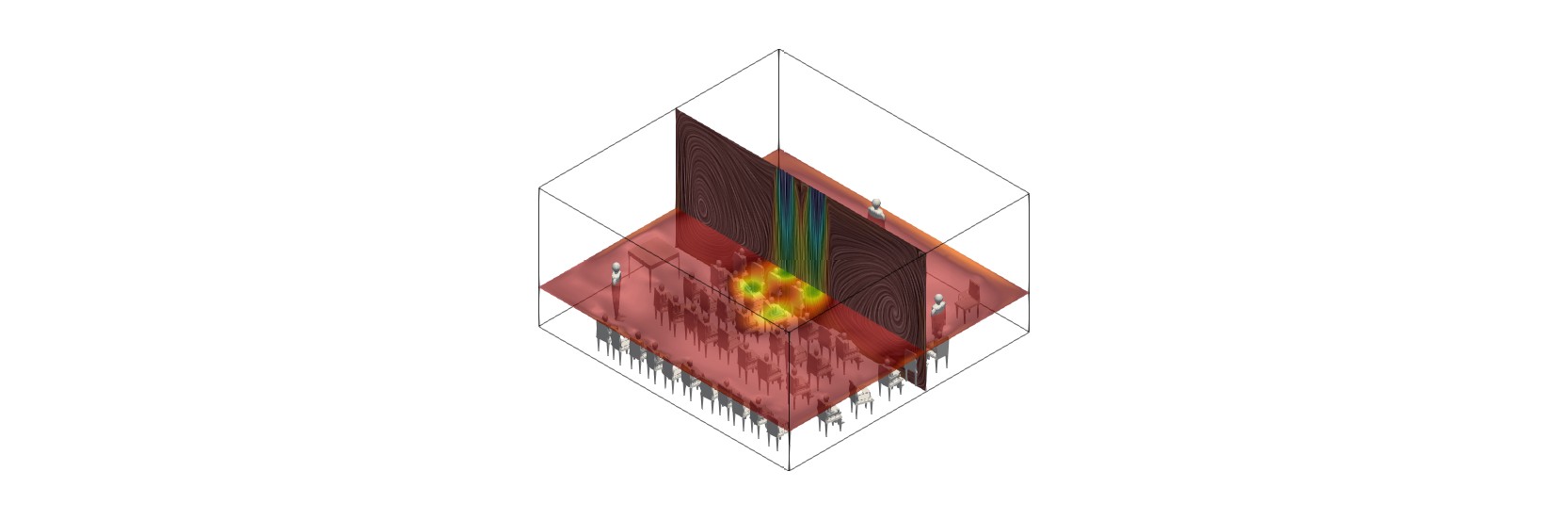}
        {\classGridVelColorbar}
    \caption{Velocity magnitude, view 1}
\end{subfigure}
&
\begin{subfigure}[b]{0.45\linewidth}
    \centering
    \classGridCFDPanelWithBar
        {530bp 51bp 530bp 50bp}
        {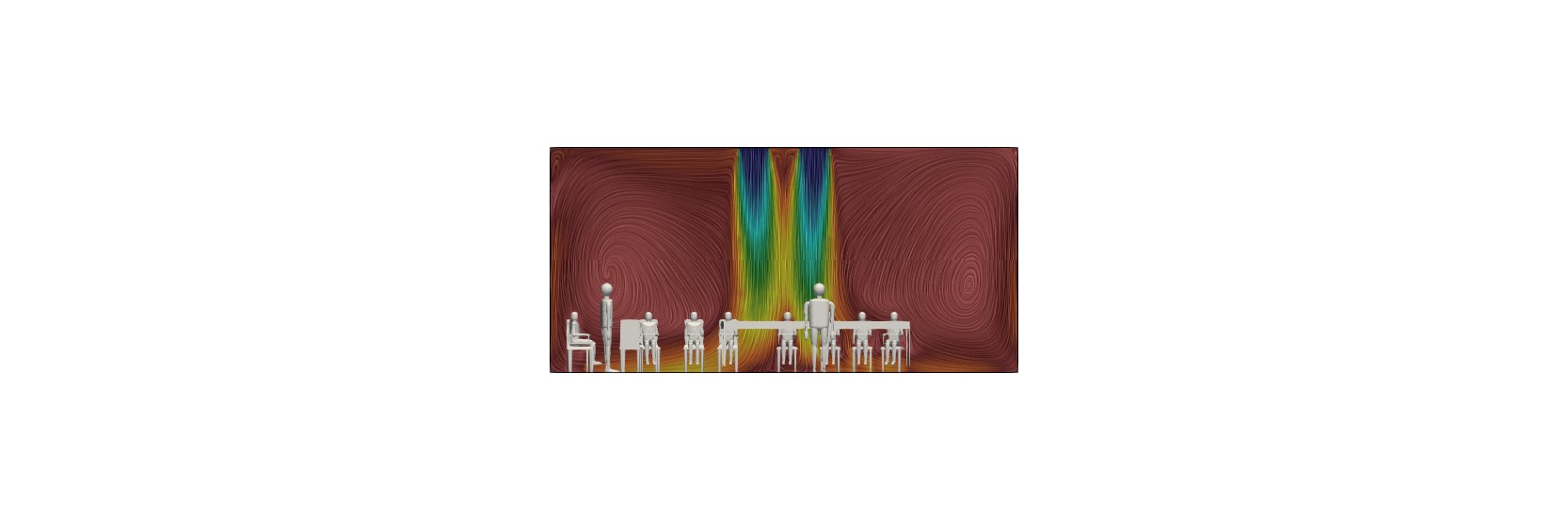}
        {\classGridVelColorbar}
    \caption{Velocity magnitude, view 2}
\end{subfigure}
\\[8pt]
\begin{subfigure}[b]{0.45\linewidth}
    \centering
    \classGridCFDPanelWithBar
        {530bp 51bp 530bp 50bp}
        {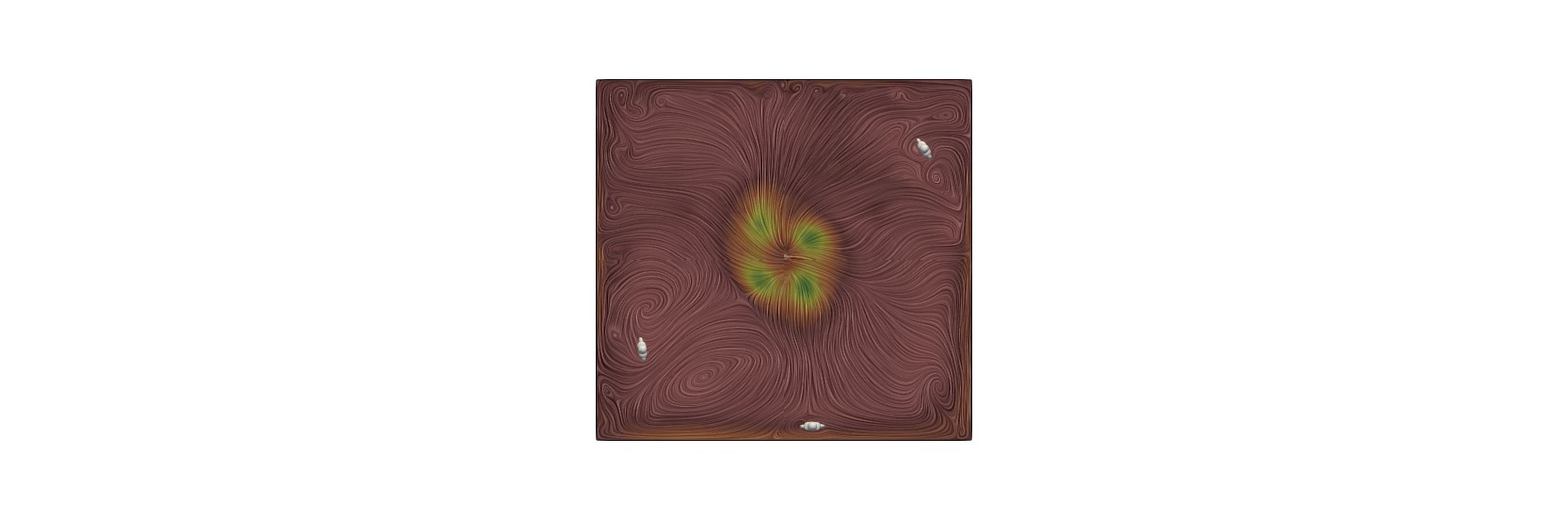}
        {\classGridVelColorbar}
    \caption{Velocity magnitude, view 3}
\end{subfigure}
&
\begin{subfigure}[b]{0.45\linewidth}
    \centering
    \classGridCFDPanelWithBar
        {530bp 60bp 530bp 50bp}
        {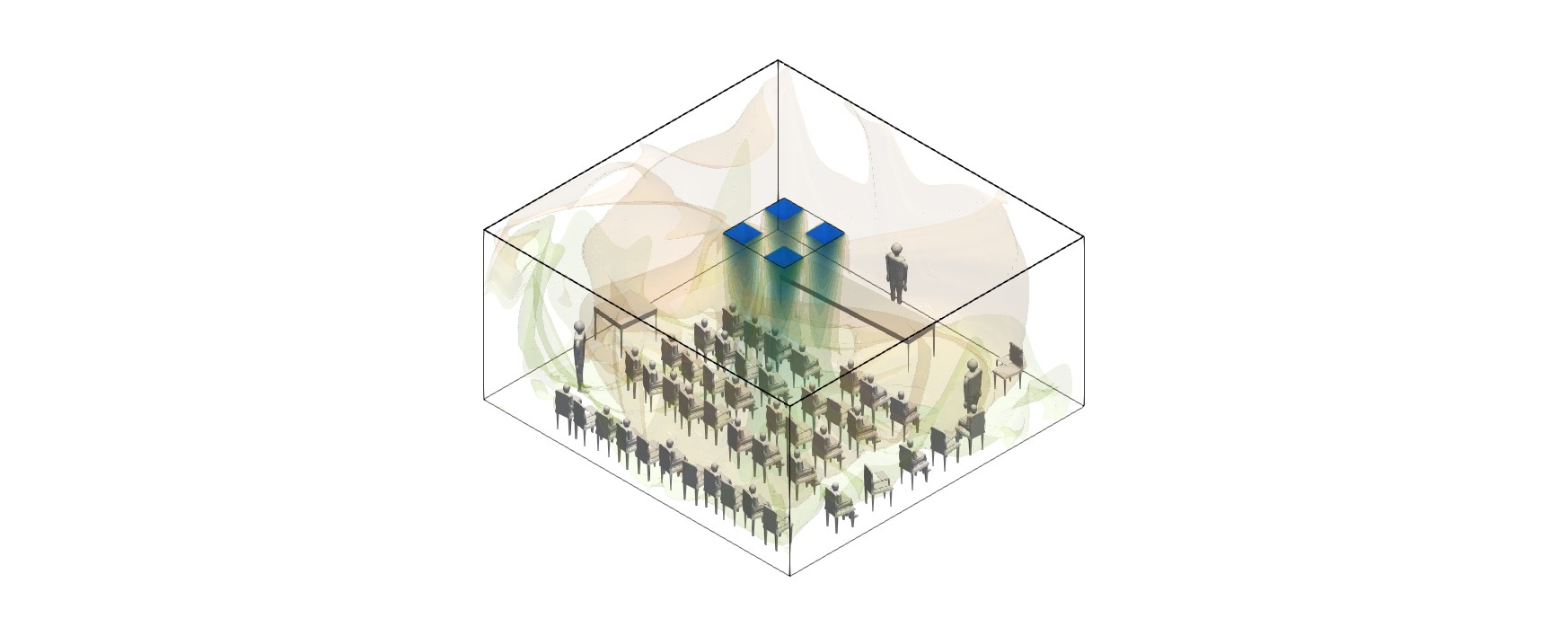}
        {\classGridContamColorbar}
    \caption{Passive-scalar concentration}
\end{subfigure}
\end{tabular}
\begin{subfigure}[b]{0.9\linewidth}
    \centering
    \classGridDecayPanel{figures/Class_B_Complex.csv}{1200}{123.1}{Chair-Dominant}
    \caption{Normalized scalar decay}
\end{subfigure}
\caption{Chair-dominant classroom reconstruction and CFD-analysis summary. The top row shows (a)~an input 360\textdegree{} frame, (b)~the reconstructed NeRF point cloud, and (c)~the simulation-ready room geometry. Panels (d)--(f) show three views of the steady velocity-magnitude field, all sharing the velocity-magnitude scale (m/s) indicated beside each panel, and panel~(g) shows a representative passive-scalar concentration field with its corresponding normalized concentration scale. Panel~(h) shows the normalized scalar concentration decay history.}
\label{fig:chair_dominant_combined}
\end{figure*}

The half-clearance time for this configuration is $t_{50}=123.1$~s (\tableref{tab:clearance_metrics}). The chair-dominant classroom has a different room envelope, furniture distribution, and reconstructed geometry, so its $t_{50}$ is reported as a standalone value rather than as a continuation of the mixed-furniture sequence; a cautious cross-room reading is offered in \secref{sec:discussion_generalization}. The case is reported here as evidence that the reconstruction-to-CFD workflow can be deployed on a new classroom with no manual modeling and produce a complete CFD case with comparable mesh density (\tableref{tab:geometry_mesh_summary}).

\subsection{Auditorium}
\label{sec:results_auditorium}
We apply the workflow to a tiered lecture-hall auditorium containing 242~chairs and 2~tables (ground truth) to test scalability to larger and more complex venues. The auditorium presents three challenges that classrooms do not: a substantially larger spatial extent (mesh size 27.9M cells, see \tableref{tab:geometry_mesh_summary}), tiered seating geometry, and more complex lighting conditions. The pipeline produces 273~chairs and 2~tables in the final reconstruction; the over-count of 31 chairs is examined in \secref{sec:discussion_failures}. \figref{fig:auditorium_combined} presents the reconstruction together with the velocity field, scalar transport snapshots, and row-resolved decay curves.

\newcommand{\audiGridVelColorbar}{%
    \includegraphics[
        height=\dimexpr\audiGridCFDHeight*5/10\relax,
        keepaspectratio,
        trim=1215bp 340bp 520bp 315bp,
        clip
    ]{figures/v2-colormap_velocity-3.jpg}%
}

\newcommand{\audiGridContamColorbar}{%
    \includegraphics[
        height=\dimexpr\audiGridCFDHeight*5/10\relax,
        keepaspectratio,
        trim=1045bp 315bp 700bp 320bp,
        clip
    ]{figures/v2-colormap_contamination-1.jpg}%
}

\newcommand{\audiGridCFDPanelWithBar}[3]{%
    \begin{minipage}[c]{0.78\linewidth}
        \centering
        \audiGridCFDPanel[#1]{#2}
    \end{minipage}%
    \hspace{1pt}%
    \begin{minipage}[c]{0.19\linewidth}
        \centering
        #3
    \end{minipage}%
}

\begin{figure*}[t!]
\centering
\footnotesize

\begin{tabular}{@{}ccc@{}}
\begin{subfigure}[b]{0.31\linewidth}
    \centering
    \audiGridTopPanel{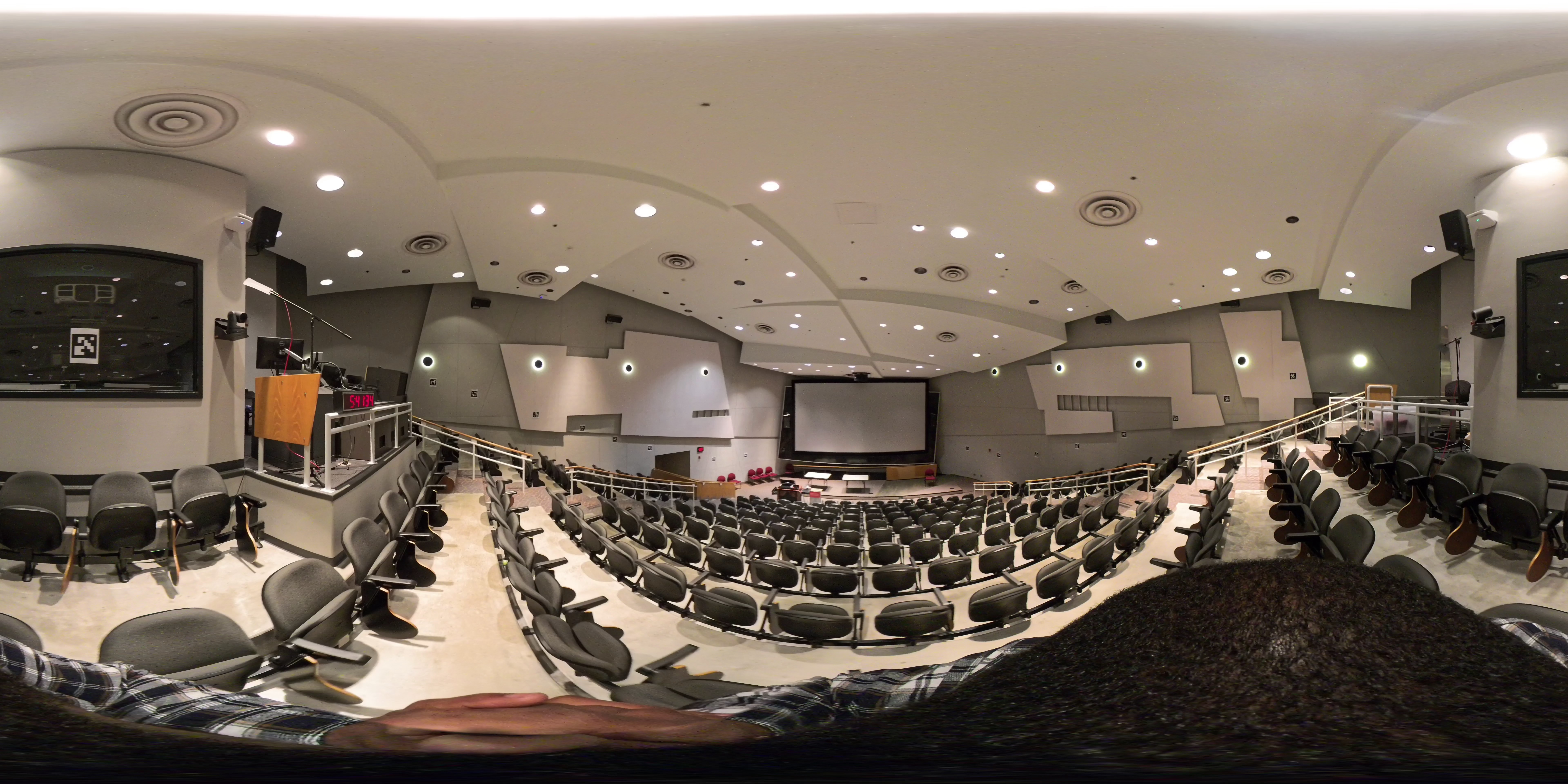}
    \caption{Input 360\textdegree{} frame}
\end{subfigure}
&
\begin{subfigure}[b]{0.31\linewidth}
    \centering
    \audiGridTopPanel{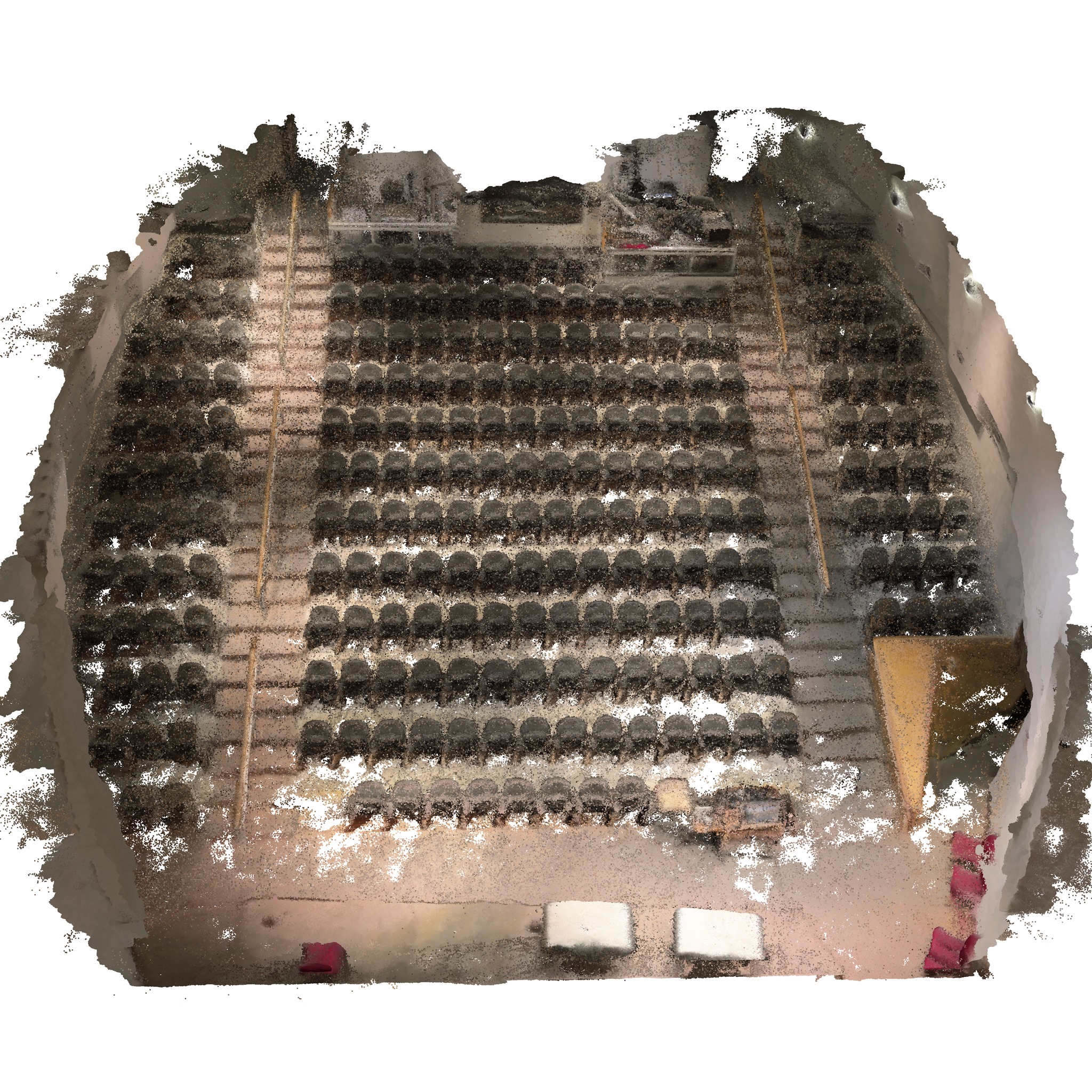}
    \caption{NeRF point cloud}
\end{subfigure}
&
\begin{subfigure}[b]{0.31\linewidth}
    \centering
    \audiGridTopPanel{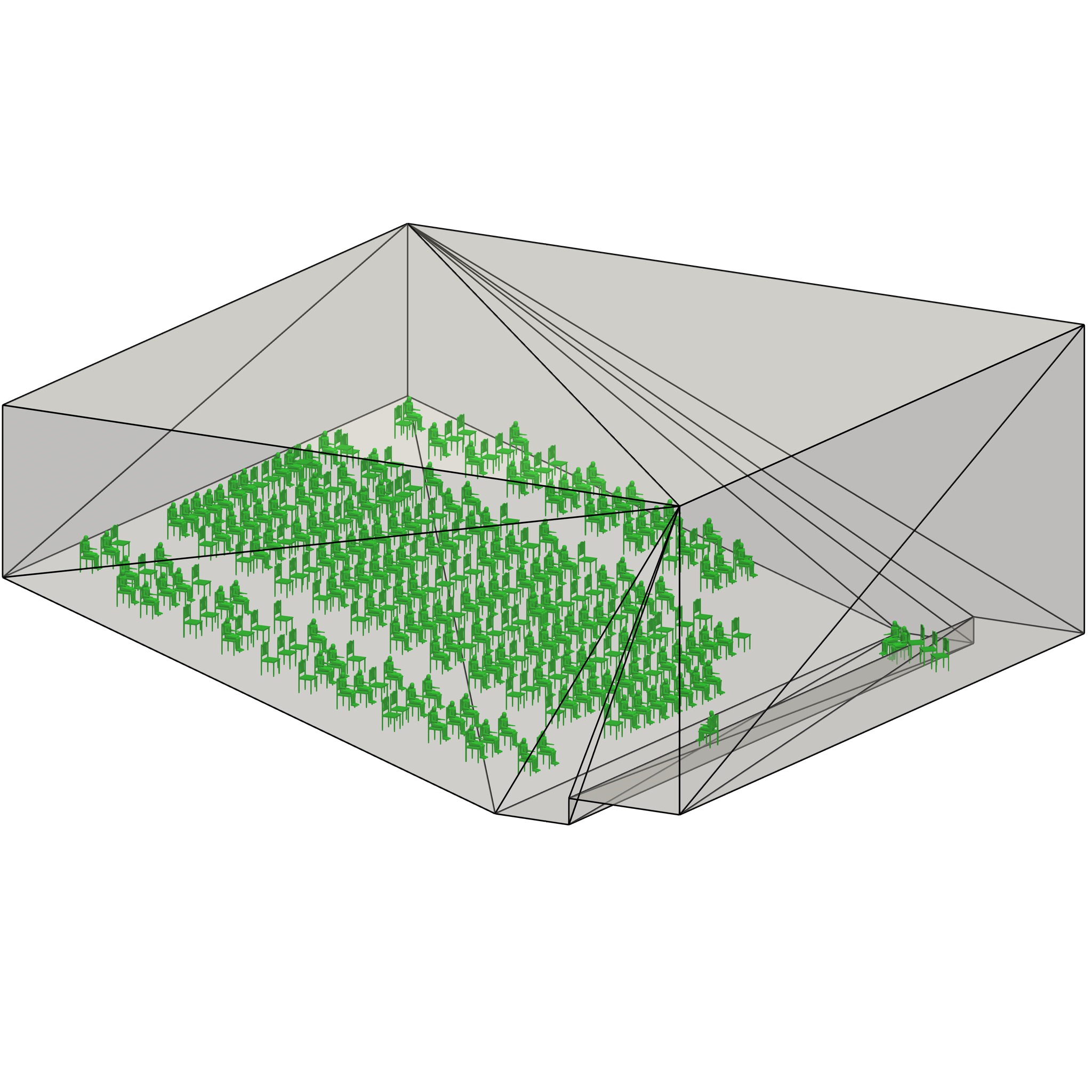}
    \caption{Simulation geometry}
\end{subfigure}
\end{tabular}

\vspace{8pt}

\begin{tabular}{@{}cc@{}}
\begin{subfigure}[b]{0.45\linewidth}
    \centering
    \audiGridCFDPanelWithBar
        {500bp 5bp 500bp 5bp}
        {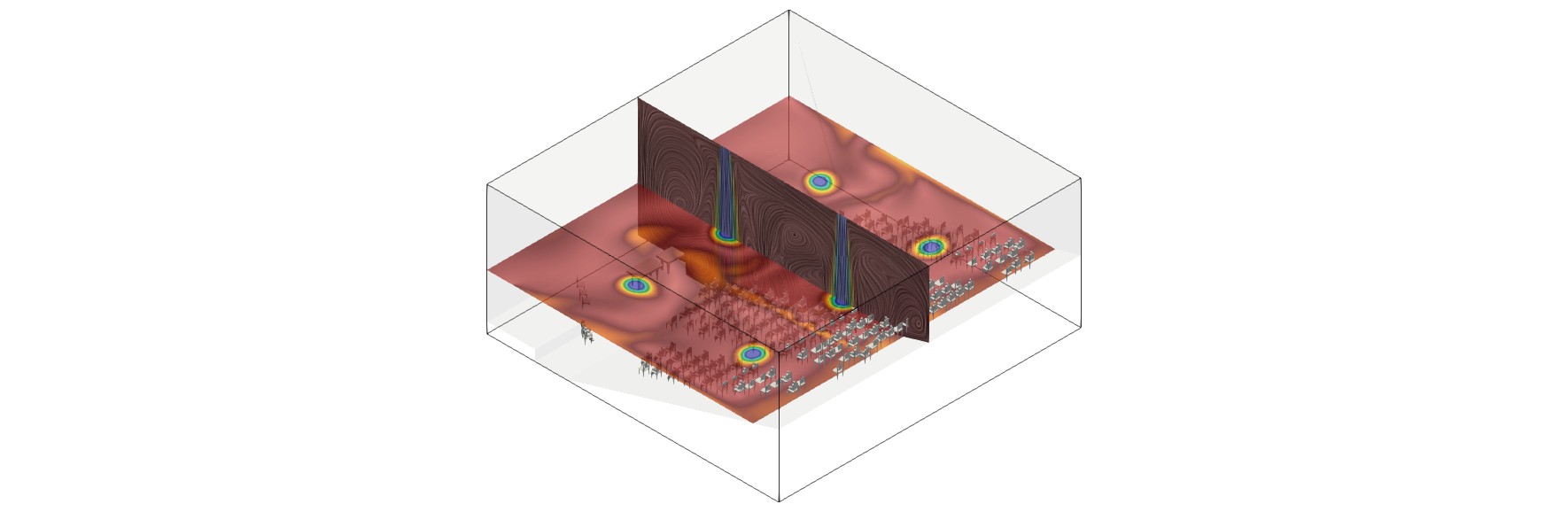}
        {\audiGridVelColorbar}
    \caption{Velocity magnitude, view 1}
\end{subfigure}
&
\begin{subfigure}[b]{0.45\linewidth}
    \centering
    \audiGridCFDPanelWithBar
        {510bp 50bp 510bp 50bp}
        {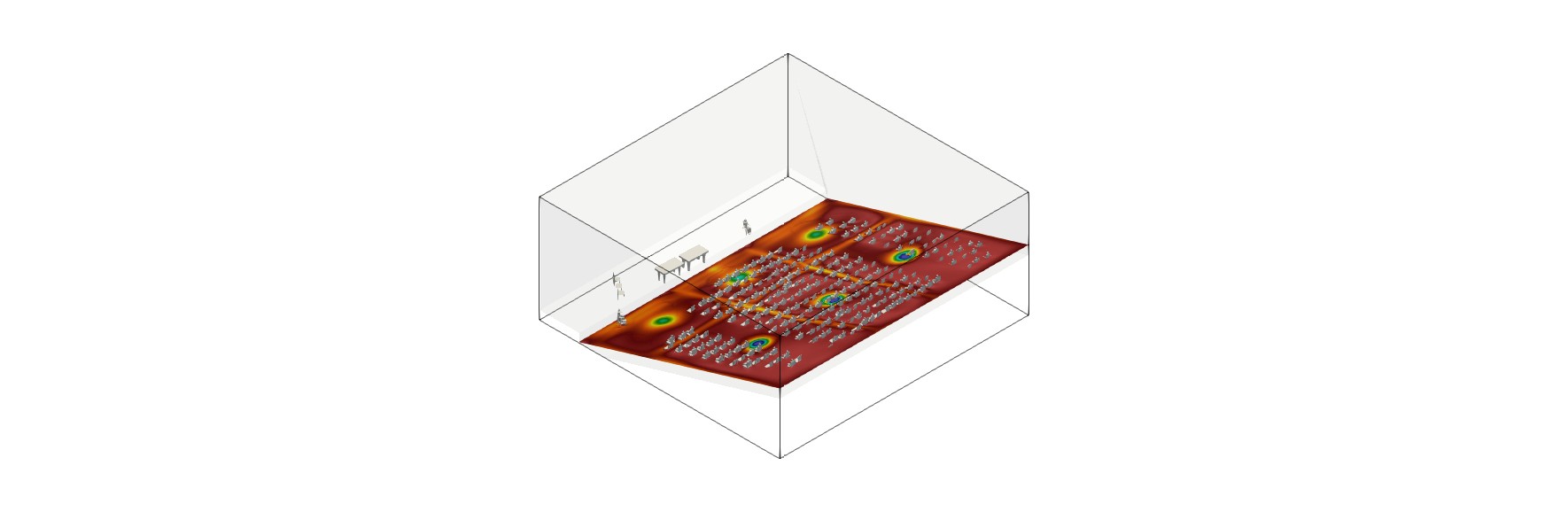}
        {\audiGridVelColorbar}
    \caption{Velocity magnitude, view 2}
\end{subfigure}
\end{tabular}

\vspace{8pt}

\begin{subfigure}[b]{0.49\linewidth}
    \centering
    \begin{minipage}[c][0.24\textheight][c]{\linewidth}
        \centering
        \audiGridCFDPanelWithBar
            {510bp 50bp 510bp 50bp}
            {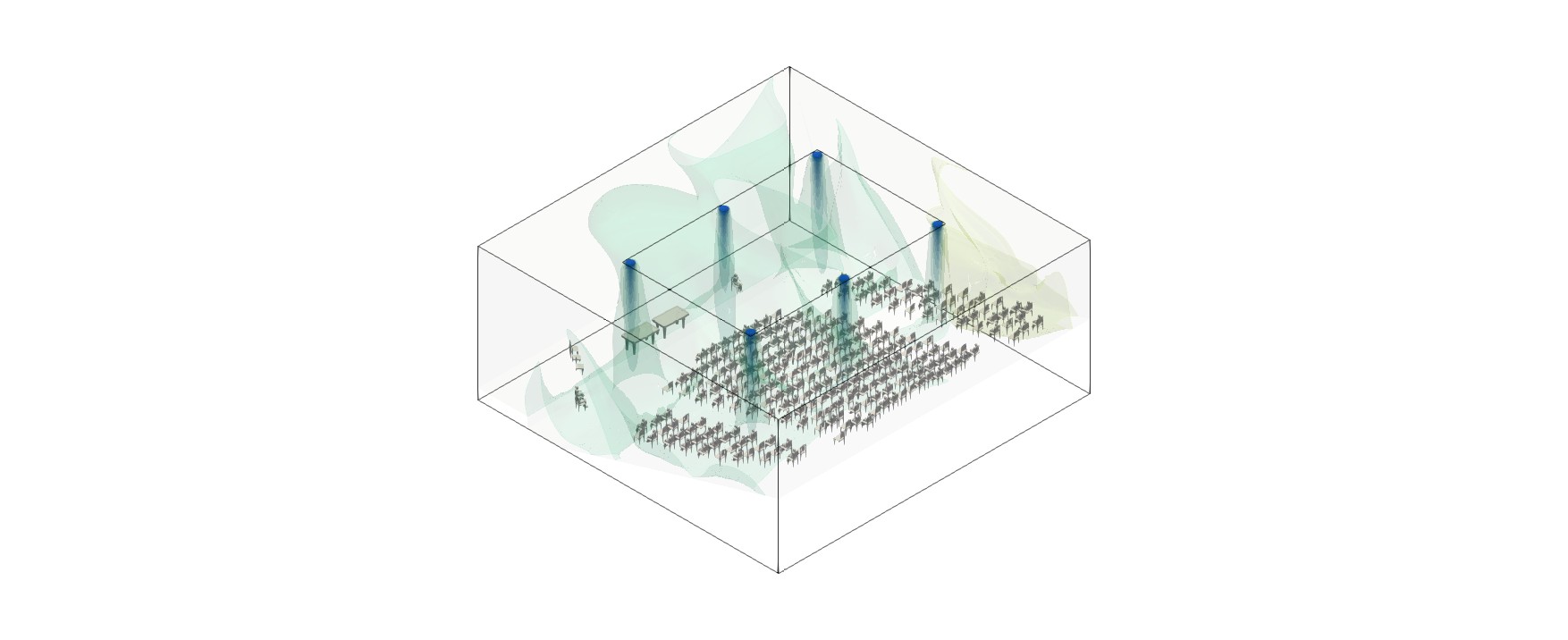}
            {\audiGridContamColorbar}
    \end{minipage}
    \caption{Passive-scalar concentration}
\end{subfigure}
\hfill
\begin{subfigure}[b]{0.49\linewidth}
    \centering
    \begin{minipage}[c][0.24\textheight][c]{\linewidth}
    \centering
    \begin{tikzpicture}
    \begin{axis}[
        width=\linewidth,
        height=0.56\linewidth,
        xlabel={Time (s)},
        ylabel={$C/C_0$},
        xmin=0,
        xmax=4005,
        ymin=0,
        ymax=1.05,
        tick label style={font=\scriptsize},
        label style={font=\scriptsize},
        grid=major,
        grid style={dashed,gray!30},
        axis line style={black},
        tick style={black},
        legend style={
            font=\scriptsize,
            draw=none,
            fill=none,
            at={(0.98,0.98)},
            anchor=north east
        }
    ]

    \addplot[thick] table[
        col sep=comma,
        header=false,
        x index=0,
        y index=1
    ] {figures/Auditorium_Decay.csv};
    \addlegendentry{Front}

    \addplot[thick, dashed] table[
        col sep=comma,
        header=false,
        x index=0,
        y index=2
    ] {figures/Auditorium_Decay.csv};
    \addlegendentry{Middle}

    \addplot[thick, dotted] table[
        col sep=comma,
        header=false,
        x index=0,
        y index=3
    ] {figures/Auditorium_Decay.csv};
    \addlegendentry{Back}

    \addplot[dash dot] coordinates {(0,0.5) (4005,0.5)};
    \addlegendentry{$C/C_0=0.5$}

    \end{axis}
    \end{tikzpicture}
    \end{minipage}
    \caption{Normalized scalar decay at front, middle, and back monitor locations}
\end{subfigure}

\caption{Tiered lecture-hall auditorium reconstruction and CFD-analysis summary. The top row shows (a)~an input 360\textdegree{} frame, (b)~the reconstructed NeRF point cloud, and (c)~the simulation-ready auditorium geometry with reconstructed furniture assets. Panels (d) and (e) show two steady velocity-magnitude views through the room interior, both sharing the velocity-magnitude scale (m/s) indicated beside each panel. Panel~(f) shows the passive-scalar concentration field with its normalized concentration scale, and panel~(g) shows normalized scalar concentration decay histories at front, middle, and back monitor locations, illustrating spatial variation in auditorium clearance.}
\label{fig:auditorium_combined}
\end{figure*}

The auditorium scalar transport was simulated over a \(3999.8\)~s decay window (\tableref{tab:geometry_mesh_summary}). The mean half-clearance time across the three monitor locations is $t_{50}=1489.7$~s, with a range of $1313.8$ to $1581.0$~s across the front, middle, and back monitors. The three monitors do not order monotonically front-to-back: the front row clears fastest ($1313.8$~s), the middle row is the most stagnant ($1581.0$~s), and the back row and the middle row are separated by $6.7$~s ($0.4\%$ of the middle-row value), which is below any resolution the present study establishes, so we treat them as indistinguishable. The order-of-magnitude longer clearance relative to the classrooms is consistent with the substantially larger room volume served by the auditorium supply patches. Per-row $t_{50}$ values are tabulated in \tableref{tab:clearance_metrics}. These values are computed on a geometry that contains 273 reconstructed chairs against a physical count of 242 (\secref{sec:discussion_failures}); they characterize the simulated case rather than the physical auditorium.

\subsection{Summary across Configurations}
\label{sec:results_summary}
\tableref{tab:geometry_mesh_summary} summarizes the geometric configurations and final mesh sizes for all six simulated cases. Mesh sizes range from approximately 4.5 million cells for the empty mixed-furniture classroom
shell to 27.9 million cells for the auditorium, demonstrating successful deployment of the reconstruction-to-CFD workflow across substantially different room and mesh scales.

\begin{table*}[t!]
\centering
\begin{threeparttable}
\caption{Final production CFD cases. Mesh counts were extracted from \texttt{checkMesh} run in parallel on the final OpenFOAM case folders. Airflow endpoint denotes the final iteration of the steady \texttt{simpleFoam} solve used to initialize scalar transport; scalar window denotes elapsed scalar-decay time after initialization from the steady airflow field.}
\label{tab:geometry_mesh_summary}
\setlength{\tabcolsep}{4pt}
\renewcommand{\arraystretch}{1.12}
\begin{tabular}{p{7.5cm}rrrr}
\toprule
\textbf{Geometric content} & \makecell{\textbf{Cells}\\\textbf{(M)}} & \makecell{\textbf{Points}\\\textbf{(M)}} & \makecell{\textbf{Airflow endpoint}\\\textbf{(iterations)}} & \makecell{\textbf{Scalar window}\\\textbf{(s)}} \\
\midrule
Mixed-furniture: room shell & 4.54 & 5.26 & 1000 & 1199.8 \\
Mixed-furniture: +32 chairs & 9.23 & 10.97 & 1000 & 1199.3 \\
Mixed-furniture: +32 chairs + 16 tables & 11.81 & 14.83 & 1000 & 1199.8 \\
Mixed-furniture: +32 chairs + 16 tables + 25 mannequins & 13.94 & 17.42 & 1000 & 1199.4 \\
Chair-dominant: 45 chairs + 2 tables + mannequins & 12.45 & 15.44 & 1000 & 1199.1 \\
Auditorium: 273 chairs + 2 tables & 27.85 & 34.93 & 1000 & 3999.8 \\
\bottomrule
\end{tabular}
\begin{tablenotes}[flushleft]
\footnotesize
\item Scalar-window values are elapsed scalar-decay times, not absolute case times. The steady airflow field is written at \texttt{simpleFoam} iteration 1000 (a pseudo-time index, not seconds), and the transient scalar solve inherits this value as its starting time coordinate; the auditorium scalar solve therefore reached case time \(t=4999.8\), corresponding to \(3999.8\)~s of physical decay.
\end{tablenotes}
\end{threeparttable}
\end{table*}

\begin{table*}[t!]
\centering
\begin{threeparttable}
\caption{Monitor-based scalar-clearance metrics computed from normalized concentration histories. \(t_{50}\), \(t_{80}\), and \(t_{90}\) denote the first elapsed times at which \(C/C_0\) falls below 0.5, 0.2, and 0.1, respectively. AUC is the time integral of \(C/C_0\) over the simulated decay window of duration \(T\), and \(C/C_0(T)\) is the normalized concentration at the end of that window. A dash indicates that the threshold was not reached within the simulated window.}
\label{tab:clearance_metrics}
\setlength{\tabcolsep}{5pt}
\renewcommand{\arraystretch}{1.12}
\begin{tabular}{p{7.5cm}rrrrr}
\toprule
\textbf{Case} & \(\boldsymbol{t_{50}}\) & \(\boldsymbol{t_{80}}\) & \(\boldsymbol{t_{90}}\) & \textbf{AUC} & \(\boldsymbol{C/C_0(T)}\) \\
 & \textbf{(s)} & \textbf{(s)} & \textbf{(s)} & \textbf{(s)} & \\
\midrule
Mixed-furniture: room shell & 130.4 & 480.8 & 791.1 & 271.8 & 0.0357 \\
Mixed-furniture: +32 chairs & 143.3 & 541.5 & 825.6 & 291.4 & 0.0399 \\
Mixed-furniture: +32 chairs + 16 tables & 135.6 & 447.9 & 710.9 & 245.2 & 0.0276 \\
Mixed-furniture: +32 chairs + 16 tables + 25 mannequins & 136.1 & 449.0 & 707.5 & 244.7 & 0.0270 \\
Chair-dominant: 45 chairs + 2 tables + mannequins & 123.1 & 407.8 & 658.5 & 229.4 & 0.0220 \\
Auditorium (front) & 1313.8 & 3094.0 & -- & 1665.6 & 0.1267 \\
Auditorium (middle) & 1581.0 & 3535.8 & -- & 1906.6 & 0.1607 \\
Auditorium (back) & 1574.3 & 3500.8 & -- & 1869.1 & 0.1578 \\
\bottomrule
\end{tabular}
\begin{tablenotes}[flushleft]
\footnotesize
\item Classroom AUC values are integrated over approximately 0--1200~s of scalar-decay time; auditorium AUC values are integrated over approximately 0--4000~s. The three auditorium monitor locations correspond to the front, middle, and back seating regions, respectively.
\end{tablenotes}
\end{threeparttable}
\end{table*}

The end-to-end pipeline produced meshable CFD geometries across three room types and production meshes ranging from approximately 4.5 to 27.9 million cells. In the controlled mixed-furniture classroom sequence, the monitor-based \(t_{50}\) changed from \(130.4\)~s for the room shell to \(143.3\)~s with chairs, \(135.6\)~s with chairs and tables, and \(136.1\)~s with chairs, tables, and mannequins. The largest change in this sequence is the \(12.9\)-s increase between the room-shell and chairs cases; by contrast, the \(0.5\)-s difference between the final two configurations is small and is not interpreted here as evidence of a resolved physical effect. These values characterize the response of the present numerical model at the selected monitor location. The mesh-sensitivity and Annex~20 assessments in \ref{app:verification} provide supporting numerical evidence. Limitations of the present CFD analysis, including the choice of turbulence model, the omission of buoyancy and thermal effects, and the use of passive-obstruction mannequins, are discussed in \secref{sec:cfd_limitations}.

\section{Discussion}
\label{sec:discussion}

The results in \secref{sec:results} demonstrate two complementary capabilities of the proposed pipeline: a controlled what-if study within a single reconstructed classroom (the mixed-furniture classroom), and generalization of the same workflow to a different, chair-dominant classroom and to a substantially larger auditorium. This section interprets these results, examines workflow generalization across the three rooms, discusses pipeline behavior on a notable failure case, states the limitations of the present CFD analysis, and closes with practical implications and future directions.

\subsection{Geometry and Passive-Scalar Clearance}
\label{sec:discussion_classA}

The mixed-furniture classroom sequence provides a controlled numerical test of how progressively modifying the interior geometry changes the computed passive-scalar history at a fixed occupied-zone monitor. With the room envelope, ventilation boundary conditions, turbulence model, scalar model, and monitor location held fixed, adding chairs increased the modeled \(t_{50}\) from \(130.4\) to \(143.3\)~s (\(9.9\%\)). Adding tables subsequently reduced the monitor-based value to \(135.6\)~s. These changes indicate that the computed clearance response varies with the representation of interior geometry. They should not, however, be interpreted as quantified room-wide ventilation penalties or improvements because the classroom metric is based on a single monitor location and the present study does not establish uncertainty bounds on \(t_{50}\).

The three-dimensional velocity fields (\figref{fig:classA_velocity}), the vertical mid-plane slices (\figref{fig:classA_velocity_test2}), and the scalar snapshots (\figref{fig:classA_scalar}) provide qualitative context for the monitor histories. The addition of chairs visibly perturbs the lower-room velocity field and introduces additional wakes around the furniture, while the table surfaces alter the trajectory and lateral redistribution of the ceiling-supplied flow. One plausible interpretation is that the broad horizontal table surfaces redistribute portions of the descending flow and thereby modify the recirculation structure sampled by the monitor. The present diagnostics do not quantify momentum redistribution or recirculation-zone volume, however, so this mechanism remains an interpretation of the flow field and is not independently demonstrated. Likewise, the \(0.5\)-s difference between the chairs+tables and chairs+tables+mannequins cases is too small, relative to the presently quantified numerical evidence, to support a distinct mannequin effect.

The observed model sensitivity to interior geometry is consistent with prior CFD studies showing that furniture and room layout can materially modify indoor airflow and pollutant distributions under otherwise similar ventilation conditions \citep{Zhuang2014FurnitureLayoutIAQ, Askari2022DisplacementVentCFD}. The principal contribution of the present work is therefore the reconstruction and scene-editing workflow itself, which makes controlled geometry-variation studies possible from a single physical capture. We claim no general relationship between furniture and clearance time.

\subsection{Workflow Generalization across Rooms and Scales}
\label{sec:discussion_generalization}
The chair-dominant classroom and the auditorium serve a different function from the mixed-furniture classroom sequence: they probe the generalization and scalability of the reconstruction-to-CFD workflow; they do not support a direct comparative airflow argument. These two cases differ from the mixed-furniture classroom in room envelope, furniture distribution, ventilation layout, and physical scale. The chair-dominant case yields a monitor-based \(t_{50}\) of \(123.1\)~s, while the auditorium yields a mean \(t_{50}\) of \(1489.7\)~s across the monitored locations.

The successful end-to-end deployment on these two cases supports three observations. First, the pipeline structure and its category-level parameterization transferred across three rooms with distinct geometric character; per-environment adjustment was limited to a small set of thresholds, chiefly the consensus count (with the number of registered views), the healing gap, and the auditorium's instance-separation level and chair-placement mode. The framework is therefore reusable across rooms, though not parameter-free. Second, the geometry produced by the pipeline interfaces cleanly with mainstream OpenFOAM meshing across the full range of case sizes (\tableref{tab:geometry_mesh_summary}). Third, the auditorium case tests the workflow at a substantially larger geometric and computational scale: a 280-s 360-degree capture was processed through the same pipeline used for the classrooms, and the resulting geometry was meshed into a 27.9-million-cell CFD domain. The case demonstrates computational scalability of the workflow to a large tiered room, while the chair over-count discussed in \secref{sec:discussion_failures} simultaneously exposes a reconstruction-accuracy limitation that becomes more consequential as object count and geometric repetition increase.

\tableref{tab:clearance_metrics} supports a further cross-room observation. The fully furnished chair-dominant room clears faster than the \emph{empty} mixed-furniture shell on every reported metric ($t_{50}=123.1$ versus $130.4$~s; $t_{80}=407.8$ versus $480.8$~s; AUC $229.4$ versus $271.8$~s), and within the mixed-furniture classroom itself the configurations with tables clear faster than the empty shell on the late-stage metrics ($t_{80}=447.9$ and $449.0$ versus $480.8$~s; $t_{90}=710.9$ and $707.5$ versus $791.1$~s), so the $t_{50}$ increase reported in \secref{sec:results_mixed} is confined to the configurations carrying tables. The chairs-only configuration is slower than the empty shell on all four reported metrics. The two rooms differ in envelope, diffuser layout, and scale, and cross-room absolute values carry the scaling uncertainty discussed in \secref{sec:cfd_limitations}. We therefore report this as an observation only: between-room differences appear to dominate within-room furniture effects. That asymmetry is itself an argument for the workflow. If clearance behavior cannot be extrapolated from a reference room or corrected by a furniture heuristic, then each room must be simulated with its own geometry, and the cost of acquiring that geometry becomes the operative bottleneck that the present pipeline addresses.

In the auditorium, the row-resolved decay curves (\figref{fig:auditorium_combined}) show substantial spatial heterogeneity in clearance, with monitor-based \(t_{50}\) values ranging from \(1313.8\) to \(1581.0\)~s and a mean value of \(1489.7\)~s across the monitored locations. This heterogeneity motivates the use of multi-monitor reporting in larger venues rather than a single representative \(t_{50}\). We caution against over-interpreting these absolute numbers: they reflect the assumed ventilation boundary conditions, including uniform downward velocity at idealized rectangular supply patches, and the RANS \(k\)--\(\varepsilon\) closure with \(\mathit{Sc}_t=0.7\). These assumptions support controlled comparative passive-scalar clearance analysis. The benchmark comparison in \ref{app:verification} provides an independent check of the OpenFOAM momentum-solver setup and meshing approach.
\subsection{Pipeline Robustness and Failure Modes}
\label{sec:discussion_failures}
Reconstruction quality across the three environments is uneven in ways that bear on the workflow's deployment envelope. The clearest reconstruction failure occurs in the auditorium, where 273 chair instances were present in the CFD geometry compared with a manually verified physical count of 242, an over-count of 31 chairs (\(12.8\%\); \tableref{tab:recon_counts}). The most likely source is the tiered seating geometry: a single physical chair viewed from camera positions on adjacent tiers can produce two spatially separated dense regions in the NeRF reconstruction (one corresponding to the seat as observed from above, another to the seat back as observed from the row below), which the octree-based connected-component and placement stages can interpret as two separate instances. Visual inspection of the auditorium reconstruction is consistent with this explanation: the over-count is concentrated in regions where chair-back surfaces in one row are spatially close to chair-seat surfaces in the row above. This is a known weakness of connected-component instance separation when applied to vertically stacked or tiered geometry; mitigations include enforcing minimum inter-instance separation distances or augmenting the instance step with category-specific volumetric priors. The human QA stage caught the discrepancy in the version reported here, but only as an inflated count: the duplicated instances were not localized or removed before meshing. This failure matters because an object-count error propagated beyond reconstruction into the CFD domain, where the additional chair surfaces act as solid obstructions and therefore cannot be assumed a priori to be aerodynamically negligible. The auditorium simulation is consequently retained as a demonstration that the pipeline can construct and execute a large, complex CFD case, but its absolute clearance values are not used as validated predictions of the physical auditorium. The discrepancy also establishes a concrete requirement for future automated QA: repeated-object environments require duplicate-instance detection or comparison against independently known object counts before a geometry is accepted for quantitative CFD interpretation.

The Mixed-Furniture Classroom and the auditorium achieve broadly similar PSNR ($\sim 24.5$~dB, \tableref{tab:nerf_metrics}) despite a substantial scale difference, while the Chair-Dominant Classroom achieves lower PSNR ($23.1$~dB) and notably lower SSIM. The lower scores are consistent with the more uniform wall texture and stronger reflections present in the Chair-Dominant Classroom, both of which are documented failure modes for NeRF reconstruction \citep{Remondino2023CriticalNeRF}. Reconstruction quality at this level was nonetheless sufficient for downstream CFD because the dominant geometry consumed by the mesher is the room enclosure and the placed STL templates, neither of which depends on photometric reconstruction quality below the centimeter scale.

In our experience, the human-in-the-loop editor absorbed three recurring QA tasks per environment: correcting one or two grossly misaligned chairs (cases where multi-start ICP converged on the $90^\circ$-rotated solution, \subfigref{fig:final_alignment}{(a)}), removing a small number of phantom placements that survived the size filter, and adding 25 mannequins for the obstructed mixed-furniture configuration. The total time investment per configuration is small (a few minutes), supporting the framing of the editor as a lightweight quality gate rather than a substantial manual modeling effort. The editor's most consequential capability is scene augmentation: the four mixed-furniture configurations were constructed from a single physical scan with only a few additional minutes of editor work per configuration.

\subsection{Limitations of the CFD Analysis}
\label{sec:cfd_limitations}
Several modeling choices bound what the present CFD results can support. The simulations are isothermal incompressible RANS with a $k$--$\varepsilon$ closure. Buoyancy, thermal stratification, body heat fluxes, breathing jets, and occupant-generated scalar sources are not modeled, and mannequins act only as passive solid obstructions. These simplifications suit a comparative passive-scalar flushing study under fixed mechanical ventilation. Predicting realistic exposure or contaminant dispersion in occupied rooms additionally requires thermal effects, because plumes from occupants and equipment can substantially modify near-body airflow and breathing-zone concentration profiles. Studies that target exposure prediction directly should include thermally driven flows and active scalar sources, typically through coupled buoyant solvers (e.g., \texttt{buoyantPimpleFoam}) and source terms representing breathing and surface heat fluxes; recent indoor LES work \citep{Auvinen2022LESairborne} illustrates the additional fidelity that can be obtained at substantially higher computational cost.

The choice of $k$--$\varepsilon$ as the turbulence closure reflects the intention to keep solver settings within the standard envelope of indoor-CFD practice and to interface cleanly with \texttt{simpleFoam} and the custom turbulent scalar-transport solver configuration documented in \secref{sec:cfd_method}. The closure is known to under-predict separation and recirculation in regions of strong streamline curvature, which are precisely the regions most affected by furniture obstruction. A natural sensitivity check is to repeat one configuration with $k$--$\omega$ SST and compare $t_{50}$; this is identified as future work.

The supply diffusers are represented as idealized rectangular ceiling patches with a uniform downward velocity, rather than as fully-resolved diffuser hardware (e.g., louvered, swirl, or four-way-throw diffusers). The reconstructed geometry preserves the location and footprint of the supply patches as captured from the ceiling, but the prescribed inlet velocity profile is uniform across each patch. The companion work of \citet{Afful2026DiffuserResolvingCFD} demonstrates that fully-resolved diffuser geometry can substantially modify near-field jet behavior compared with uniform-velocity approximations; the present work trades that fidelity for the ability to deploy the workflow across many rooms without diffuser-specific CAD modeling.

A central limitation concerns geometric accuracy. Reconstruction quality is currently evidenced by object counts (\tableref{tab:recon_counts}), photometric novel-view metrics (\tableref{tab:nerf_metrics}), and visual inspection; no direct dimensional-accuracy metric, such as Chamfer distance to a surveyed reference or wall-to-wall dimension error of the room shell, is reported. Because the CFD consumes geometry rather than photometry, quantifying the dimensional error of the reconstructed shell and placed assets against measured references is a priority for follow-up work.

NeRF reconstruction is scale-ambiguous, and each reconstructed room was scaled to physical dimensions through a single height reference. A measurement uncertainty of $1\%$ in the reference height propagates to $\sim 3\%$ in volumetric quantities. Because the supply patches scale with the same factor while the prescribed inlet velocity is held fixed, the supply volume flow rate scales with the square of the scale factor and the nominal residence time scales linearly with it, so the corresponding sensitivity of $t_{50}$ is $\sim 1\%$. The mixed-furniture within-sequence comparison is unaffected because the same scale factor is applied to all four configurations, but cross-room comparisons (e.g., $t_{50}$ from the mixed-furniture to the chair-dominant classroom in absolute units) carry this scaling uncertainty. Multi-reference scaling, or scaling to known fixtures of standardized dimension, would tighten this.

Finally, the present work reports comparative passive-scalar clearance under fixed boundary conditions; validated contaminant-dispersion predictions for the specific rooms simulated remain outside the present scope. Solver fidelity is independently assessed through the benchmark comparison of \ref{app:verification}; per-room validation against tracer-gas decay measurements in the actual rooms used for reconstruction is identified as the priority next step, and is the natural pathway to interpreting the absolute $t_{50}$ values as physical predictions.

\subsection{Practical Implications and Future Directions}
\label{sec:discussion_future}
The most immediate practical implication of the workflow is that CFD-based ventilation assessment can move from a bespoke single-room exercise toward a pipeline that can be deployed with consumer hardware and modest specialist effort. The cost ratio between a 360-degree video capture (a few minutes per room with a sub-\$500 camera) and either manual CAD modeling (days per room) or terrestrial laser scanning (hours of capture and post-processing on \$50k+ hardware), in our estimate, translates directly into the number of rooms that can be assessed within a fixed engineering budget. Whether this matters in practice depends on the application: per-room exposure predictions for high-stakes spaces (operating rooms, infection-control wards) will continue to require careful per-room validation, but portfolio-level ventilation comparison across many rooms (assessment of educational building stock, retrofit prioritization, scenario evaluation under alternative occupancy) is the regime where the geometry-acquisition cost reduction matters most.

Several directions extend the present work. Coupling the reconstructed geometry directly with unfitted finite-element solvers \citep{Yang2025OctreeSBM, Yang2025SBMThermal} would remove the watertight-surface requirement and the corresponding healing and ICP stages from the workflow, potentially simplifying the pipeline at the cost of changing the solver stack. Adding thermally driven flows and active occupant-generated scalar sources would extend the analysis from passive-scalar flushing to exposure prediction. Replacing the rectangular supply patches with diffuser-resolved geometry, by extending the SAM~3 prompt set to include diffuser categories and developing diffuser templates analogous to the chair templates used here, would tighten the inlet-condition fidelity. Finally, incorporating measured tracer-gas validation for the rooms used in this paper would convert the comparative $t_{50}$ values reported here into validated absolute predictions and is the most direct path to broader adoption.

\section{Conclusions}
\label{sec:conclusion}
This paper presented a semi-automated reconstruction-to-CFD workflow that converts a single 360-degree video of an indoor environment into individually editable, simulation-ready surface assets, addressing three persistent bottlenecks in CFD-based ventilation assessment: low-cost geometry acquisition without expensive scanning hardware or manual CAD modeling, semantic decomposition of the captured scene into individually manipulable furniture and occupant objects, and rapid reconfiguration of those objects to evaluate alternative occupancy or layout scenarios without re-capturing the room. The pipeline chains neural reconstruction, text-prompted segmentation, geometric healing, and template/procedural alignment into automated OpenFOAM case generation, with a lightweight browser-based editor for human-in-the-loop reconfiguration. We demonstrated it end-to-end on three university instructional spaces (CFD cases of approximately 4.5--27.9 million cells); a complete capture-to-geometry pass takes two to five hours per room on a consumer workstation.

The demonstrations support one conclusion: geometry acquisition is the operative bottleneck in deploying indoor CFD at scale, and it can be lifted. Within the primary classroom, a single capture supported a four-configuration what-if sequence in which the monitor-based clearance varied non-monotonically with furniture. Across rooms, the fully furnished chair-dominant classroom cleared faster than the empty primary-room shell. The two rooms differ in envelope, diffuser layout, and scale simultaneously, so we report that comparison as an observation consistent with between-room differences dominating within-room furniture effects, and not as a demonstration of it. On the present evidence, clearance behavior in a new room is not reliably obtained by extrapolation from a reference room, and that is the regime in which low-cost per-room geometry acquisition matters most. Solver fidelity and mesh sensitivity were assessed through the mesh-sensitivity study and IEA Annex~20 benchmark comparison of \ref{app:verification}.

Future directions include coupling the reconstructed geometry directly with unfitted finite-element solvers such as the Shifted Boundary Method on adaptive octree meshes \citep{Yang2025OctreeSBM, Yang2025SBMThermal}, which would remove the watertight-surface requirement and the corresponding healing and ICP stages; extending the analysis from passive-scalar flushing to exposure prediction by adding thermally driven flows and active occupant-generated scalar sources; replacing the idealized rectangular supply patches with diffuser-resolved geometry; and validating the absolute clearance metrics against tracer-gas measurements in the rooms used for reconstruction. These extensions would move the workflow from a demonstration of low-cost reconstruction-to-CFD coupling toward a deployable engineering tool for comparative ventilation analysis across the existing building stock.

\section*{Code and Data Availability}
\noindent\textbf{Code.} The reconstruction-to-CFD pipeline (semantic segmentation, multi-view consensus filtering, octree-based instance extraction, graph-based healing, template and procedural STL generation, and the browser-based scene editor) is implemented in Python and released as open-source software at \url{https://github.com/baskargroup/video2cfd}, together with the OpenFOAM case templates (mesh-generation dictionaries, boundary conditions, and solver settings) used in \secref{sec:cfd_method}. The pipeline builds on the publicly available Nerfstudio \citep{Tancik2023Nerfstudio}, SAM~3 \citep{Carion2025SAM3}, Meshroom \citep{Meshroom2021}, COLMAP \citep{Schonberger2016COLMAP}, and OpenFOAM \citep{Weller1998OpenFOAM} packages, which are obtained from their respective sources rather than redistributed here.

\noindent\textbf{Data.} The supporting data (NeRF-exported point clouds, reconstructed STL assets, and representative OpenFOAM cases for the three environments) are available from the corresponding author on reasonable request.

\bibliographystyle{elsarticle-num-names}
\bibliography{bibliography}

\begin{thebibliography}{57}
\expandafter\ifx\csname natexlab\endcsname\relax\def\natexlab#1{#1}\fi
\providecommand{\url}[1]{\texttt{#1}}
\providecommand{\href}[2]{#2}
\providecommand{\path}[1]{#1}
\providecommand{\DOIprefix}{doi:}
\providecommand{\ArXivprefix}{arXiv:}
\providecommand{\URLprefix}{URL: }
\providecommand{\Pubmedprefix}{pmid:}
\providecommand{\doi}[1]{\href{http://dx.doi.org/#1}{\path{#1}}}
\providecommand{\Pubmed}[1]{\href{pmid:#1}{\path{#1}}}
\providecommand{\bibinfo}[2]{#2}
\ifx\xfnm\relax \def\xfnm[#1]{\unskip,\space#1}\fi
\bibitem[{Chen(2009)}]{Chen2009VentilationReviewCFD}
\bibinfo{author}{Q.~Chen},
\newblock \bibinfo{title}{Ventilation performance prediction for buildings: A
  method overview and recent applications},
\newblock \bibinfo{journal}{Building and Environment} \bibinfo{volume}{44}
  (\bibinfo{year}{2009}) \bibinfo{pages}{848--858}.
  \DOIprefix\doi{10.1016/j.buildenv.2008.05.025}.
\bibitem[{Mohamadi and Fazeli(2022)}]{Mohamadi2022ReviewCFDCOVID}
\bibinfo{author}{F.~Mohamadi}, \bibinfo{author}{A.~Fazeli},
\newblock \bibinfo{title}{A review on applications of {CFD} modeling in
  {COVID-19} pandemic},
\newblock \bibinfo{journal}{Archives of Computational Methods in Engineering}
  \bibinfo{volume}{29} (\bibinfo{year}{2022}) \bibinfo{pages}{3567--3586}.
  \DOIprefix\doi{10.1007/s11831-021-09706-3}.
\bibitem[{Zhuang et~al.(2014)Zhuang, Li, and Tu}]{Zhuang2014FurnitureLayoutIAQ}
\bibinfo{author}{R.~Zhuang}, \bibinfo{author}{X.~Li}, \bibinfo{author}{J.~Tu},
\newblock \bibinfo{title}{{CFD} study of the effects of furniture layout on
  indoor air quality under typical office ventilation schemes},
\newblock \bibinfo{journal}{Building Simulation} \bibinfo{volume}{7}
  (\bibinfo{year}{2014}) \bibinfo{pages}{263--275}.
  \DOIprefix\doi{10.1007/s12273-013-0144-5}, \bibinfo{note}{published online
  2013-09-17}.
\bibitem[{Vita et~al.(2023)Vita, Woolf, Avery-Hickmott, and
  Rowsell}]{Vita2023CFDAirborneRisk}
\bibinfo{author}{G.~Vita}, \bibinfo{author}{D.~Woolf},
  \bibinfo{author}{T.~Avery-Hickmott}, \bibinfo{author}{R.~Rowsell},
\newblock \bibinfo{title}{A {CFD}-based framework to assess airborne infection
  risk in buildings},
\newblock \bibinfo{journal}{Building and Environment} \bibinfo{volume}{233}
  (\bibinfo{year}{2023}) \bibinfo{pages}{110099}.
  \DOIprefix\doi{10.1016/j.buildenv.2023.110099}.
\bibitem[{Askari et~al.(2022)Askari, Mahdavinejad, and
  Ansari}]{Askari2022DisplacementVentCFD}
\bibinfo{author}{A.~Askari}, \bibinfo{author}{M.~Mahdavinejad},
  \bibinfo{author}{M.~Ansari},
\newblock \bibinfo{title}{Investigation of displacement ventilation performance
  under various room configurations using computational fluid dynamics
  simulation},
\newblock \bibinfo{journal}{Building Services Engineering Research and
  Technology} \bibinfo{volume}{43} (\bibinfo{year}{2022})
  \bibinfo{pages}{627--643}. \DOIprefix\doi{10.1177/01436244221097312}.
\bibitem[{Morawska et~al.(2020)Morawska, Tang, Bahnfleth, Bluyssen, Boerstra,
  Buonanno, Cao, Dancer, Floto, Franchimon, Haworth, Hogeling, Isaxon, Jimenez,
  Kurnitski, Li, Loomans, Marks, Marr, Mazzarella, Melikov, Miller, Milton,
  Nazaroff, Nielsen, Noakes, Peccia, Querol, Sekhar, Seppanen, Tanabe, Tellier,
  Tham, Wargocki, Wierzbicka, and Yao}]{Morawska2020HowCanAirborne}
\bibinfo{author}{L.~Morawska}, \bibinfo{author}{J.~W. Tang},
  \bibinfo{author}{W.~Bahnfleth}, \bibinfo{author}{P.~M. Bluyssen},
  \bibinfo{author}{A.~Boerstra}, \bibinfo{author}{G.~Buonanno},
  \bibinfo{author}{J.~Cao}, \bibinfo{author}{S.~Dancer},
  \bibinfo{author}{A.~Floto}, \bibinfo{author}{F.~Franchimon},
  \bibinfo{author}{C.~Haworth}, \bibinfo{author}{J.~Hogeling},
  \bibinfo{author}{C.~Isaxon}, \bibinfo{author}{J.~L. Jimenez},
  \bibinfo{author}{J.~Kurnitski}, \bibinfo{author}{Y.~Li},
  \bibinfo{author}{M.~Loomans}, \bibinfo{author}{G.~Marks},
  \bibinfo{author}{L.~C. Marr}, \bibinfo{author}{L.~Mazzarella},
  \bibinfo{author}{A.~K. Melikov}, \bibinfo{author}{S.~Miller},
  \bibinfo{author}{D.~K. Milton}, \bibinfo{author}{W.~Nazaroff},
  \bibinfo{author}{P.~V. Nielsen}, \bibinfo{author}{C.~Noakes},
  \bibinfo{author}{J.~Peccia}, \bibinfo{author}{X.~Querol},
  \bibinfo{author}{C.~Sekhar}, \bibinfo{author}{O.~Seppanen},
  \bibinfo{author}{S.-i. Tanabe}, \bibinfo{author}{R.~Tellier},
  \bibinfo{author}{K.~W. Tham}, \bibinfo{author}{P.~Wargocki},
  \bibinfo{author}{A.~Wierzbicka}, \bibinfo{author}{M.~Yao},
\newblock \bibinfo{title}{How can airborne transmission of {COVID-19} indoors
  be minimised?},
\newblock \bibinfo{journal}{Environment International} \bibinfo{volume}{142}
  (\bibinfo{year}{2020}) \bibinfo{pages}{105832}.
  \DOIprefix\doi{10.1016/j.envint.2020.105832}.
\bibitem[{Bhagat et~al.(2020)Bhagat, Davies~Wykes, Dalziel, and
  Linden}]{Bhagat2020DisplacementVentilation}
\bibinfo{author}{R.~K. Bhagat}, \bibinfo{author}{M.~S. Davies~Wykes},
  \bibinfo{author}{S.~B. Dalziel}, \bibinfo{author}{P.~F. Linden},
\newblock \bibinfo{title}{Effects of ventilation on the indoor spread of
  {COVID-19}},
\newblock \bibinfo{journal}{Journal of Fluid Mechanics} \bibinfo{volume}{903}
  (\bibinfo{year}{2020}) \bibinfo{pages}{F1}.
  \DOIprefix\doi{10.1017/jfm.2020.720}.
\bibitem[{Abreu et~al.(2023)Abreu, Pinto, Matos, and
  Pires}]{Abreu2023ProceduralPointCloudReview}
\bibinfo{author}{N.~Abreu}, \bibinfo{author}{A.~Pinto},
  \bibinfo{author}{A.~Matos}, \bibinfo{author}{M.~Pires},
\newblock \bibinfo{title}{Procedural point cloud modelling in scan-to-{BIM} and
  scan-vs-{BIM} applications: A review},
\newblock \bibinfo{journal}{ISPRS International Journal of Geo-Information}
  \bibinfo{volume}{12} (\bibinfo{year}{2023}) \bibinfo{pages}{260}.
  \DOIprefix\doi{10.3390/ijgi12070260}.
\bibitem[{Khoshelham et~al.(2021)Khoshelham, Tran, Acharya,
  D{\'i}az~Vilari{\~n}o, Kang, and
  Dalyot}]{Khoshelham2021ISPRSBenchmarkIndoorModelling}
\bibinfo{author}{K.~Khoshelham}, \bibinfo{author}{H.~Tran},
  \bibinfo{author}{D.~Acharya}, \bibinfo{author}{L.~D{\'i}az~Vilari{\~n}o},
  \bibinfo{author}{Z.~Kang}, \bibinfo{author}{S.~Dalyot},
\newblock \bibinfo{title}{Results of the isprs benchmark on indoor modelling},
\newblock \bibinfo{journal}{ISPRS Open Journal of Photogrammetry and Remote
  Sensing} \bibinfo{volume}{2} (\bibinfo{year}{2021}) \bibinfo{pages}{100008}.
  \DOIprefix\doi{10.1016/j.ophoto.2021.100008}.
\bibitem[{Skrzypczak et~al.(2022)Skrzypczak, Oleniacz, Le{\'s}niak, Zima,
  Mr{\'o}wczy{\'n}ska, and Kazak}]{Skrzypczak2022ScanToBIMAccuracy}
\bibinfo{author}{I.~Skrzypczak}, \bibinfo{author}{G.~Oleniacz},
  \bibinfo{author}{A.~Le{\'s}niak}, \bibinfo{author}{K.~Zima},
  \bibinfo{author}{M.~Mr{\'o}wczy{\'n}ska}, \bibinfo{author}{J.~K. Kazak},
\newblock \bibinfo{title}{Scan-to-{BIM} method in construction: assessment of
  the {3D} buildings model accuracy in terms inventory measurements},
\newblock \bibinfo{journal}{Building Research \& Information}
  \bibinfo{volume}{50} (\bibinfo{year}{2022}) \bibinfo{pages}{859--880}.
  \DOIprefix\doi{10.1080/09613218.2021.2011703}.
\bibitem[{Mildenhall et~al.(2020)Mildenhall, Srinivasan, Tancik, Barron,
  Ramamoorthi, and Ng}]{Mildenhall2020NeRF}
\bibinfo{author}{B.~Mildenhall}, \bibinfo{author}{P.~P. Srinivasan},
  \bibinfo{author}{M.~Tancik}, \bibinfo{author}{J.~T. Barron},
  \bibinfo{author}{R.~Ramamoorthi}, \bibinfo{author}{R.~Ng},
\newblock \bibinfo{title}{{NeRF}: Representing scenes as neural radiance fields
  for view synthesis},
\newblock in: \bibinfo{booktitle}{European Conference on Computer Vision
  (ECCV)}, \bibinfo{year}{2020}, pp. \bibinfo{pages}{405--421}.
  \DOIprefix\doi{10.1007/978-3-030-58452-8_24}.
\bibitem[{Tancik et~al.(2023)Tancik, Weber, Ng, Li, Yi, Wang, Kristoffersen,
  Austin, Salahi, Ahuja, McAllister, Kerr, and Kanazawa}]{Tancik2023Nerfstudio}
\bibinfo{author}{M.~Tancik}, \bibinfo{author}{E.~Weber},
  \bibinfo{author}{E.~Ng}, \bibinfo{author}{R.~Li}, \bibinfo{author}{B.~Yi},
  \bibinfo{author}{T.~Wang}, \bibinfo{author}{A.~Kristoffersen},
  \bibinfo{author}{J.~Austin}, \bibinfo{author}{K.~Salahi},
  \bibinfo{author}{A.~Ahuja}, \bibinfo{author}{D.~McAllister},
  \bibinfo{author}{J.~Kerr}, \bibinfo{author}{A.~Kanazawa},
\newblock \bibinfo{title}{Nerfstudio: A modular framework for neural radiance
  field development},
\newblock in: \bibinfo{booktitle}{ACM SIGGRAPH 2023 Conference Proceedings},
  \bibinfo{year}{2023}, pp. \bibinfo{pages}{1--12}.
  \DOIprefix\doi{10.1145/3588432.3591516}.
\bibitem[{Kirillov et~al.(2023)Kirillov, Mintun, Ravi, Mao, Rolland, Gustafson,
  Xiao, Whitehead, Berg, Lo, Doll{\'a}r, and Girshick}]{Kirillov2023SAM}
\bibinfo{author}{A.~Kirillov}, \bibinfo{author}{E.~Mintun},
  \bibinfo{author}{N.~Ravi}, \bibinfo{author}{H.~Mao},
  \bibinfo{author}{C.~Rolland}, \bibinfo{author}{L.~Gustafson},
  \bibinfo{author}{T.~Xiao}, \bibinfo{author}{S.~Whitehead},
  \bibinfo{author}{A.~C. Berg}, \bibinfo{author}{W.-Y. Lo},
  \bibinfo{author}{P.~Doll{\'a}r}, \bibinfo{author}{R.~Girshick},
\newblock \bibinfo{title}{Segment anything},
\newblock in: \bibinfo{booktitle}{IEEE/CVF International Conference on Computer
  Vision (ICCV)}, \bibinfo{year}{2023}, pp. \bibinfo{pages}{4015--4026}.
  \DOIprefix\doi{10.1109/ICCV51070.2023.00371}.
\bibitem[{Ravi et~al.(2025)Ravi, Gabeur, Hu, Hu, Ryali, Ma, Khedr, R{\"a}dle,
  Rolland, Gustafson, Mintun, Pan, Alwala, Carion, Wu, Girshick, Doll{\'a}r,
  and Feichtenhofer}]{Ravi2024SAM2}
\bibinfo{author}{N.~Ravi}, \bibinfo{author}{V.~Gabeur}, \bibinfo{author}{Y.-T.
  Hu}, \bibinfo{author}{R.~Hu}, \bibinfo{author}{C.~Ryali},
  \bibinfo{author}{T.~Ma}, \bibinfo{author}{H.~Khedr},
  \bibinfo{author}{R.~R{\"a}dle}, \bibinfo{author}{C.~Rolland},
  \bibinfo{author}{L.~Gustafson}, \bibinfo{author}{E.~Mintun},
  \bibinfo{author}{J.~Pan}, \bibinfo{author}{K.~V. Alwala},
  \bibinfo{author}{N.~Carion}, \bibinfo{author}{C.-Y. Wu},
  \bibinfo{author}{R.~Girshick}, \bibinfo{author}{P.~Doll{\'a}r},
  \bibinfo{author}{C.~Feichtenhofer},
\newblock \bibinfo{title}{{SAM} 2: Segment anything in images and videos},
\newblock in: \bibinfo{booktitle}{International Conference on Learning
  Representations (ICLR)}, \bibinfo{year}{2025}, pp. \bibinfo{pages}{1--42}.
  \bibinfo{note}{ArXiv:2408.00714}.
\bibitem[{Carion et~al.(2025)Carion, Gustafson, Hu, Debnath, Hu, Suris, Ryali,
  Alwala, Khedr, Huang, Lei, Ma, Guo, Kalla, Marks, Greer, Wang, Sun,
  R{\"a}dle, Afouras, Mavroudi, Xu, Wu, Zhou, Momeni, Hazra, Ding, Vaze,
  Porcher, Li, Li, Kamath, Cheng, Doll{\'a}r, Ravi, Saenko, Zhang, and
  Feichtenhofer}]{Carion2025SAM3}
\bibinfo{author}{N.~Carion}, \bibinfo{author}{L.~Gustafson},
  \bibinfo{author}{Y.-T. Hu}, \bibinfo{author}{S.~Debnath},
  \bibinfo{author}{R.~Hu}, \bibinfo{author}{D.~Suris},
  \bibinfo{author}{C.~Ryali}, \bibinfo{author}{K.~V. Alwala},
  \bibinfo{author}{H.~Khedr}, \bibinfo{author}{A.~Huang},
  \bibinfo{author}{J.~Lei}, \bibinfo{author}{T.~Ma}, \bibinfo{author}{B.~Guo},
  \bibinfo{author}{A.~Kalla}, \bibinfo{author}{M.~Marks},
  \bibinfo{author}{J.~Greer}, \bibinfo{author}{M.~Wang},
  \bibinfo{author}{P.~Sun}, \bibinfo{author}{R.~R{\"a}dle},
  \bibinfo{author}{T.~Afouras}, \bibinfo{author}{E.~Mavroudi},
  \bibinfo{author}{K.~Xu}, \bibinfo{author}{T.-H. Wu},
  \bibinfo{author}{Y.~Zhou}, \bibinfo{author}{L.~Momeni},
  \bibinfo{author}{R.~Hazra}, \bibinfo{author}{S.~Ding},
  \bibinfo{author}{S.~Vaze}, \bibinfo{author}{F.~Porcher},
  \bibinfo{author}{F.~Li}, \bibinfo{author}{S.~Li},
  \bibinfo{author}{A.~Kamath}, \bibinfo{author}{H.~K. Cheng},
  \bibinfo{author}{P.~Doll{\'a}r}, \bibinfo{author}{N.~Ravi},
  \bibinfo{author}{K.~Saenko}, \bibinfo{author}{P.~Zhang},
  \bibinfo{author}{C.~Feichtenhofer},
\newblock \bibinfo{title}{{SAM} 3: Segment anything with concepts},
\newblock \bibinfo{journal}{arXiv preprint arXiv:2511.16719}
  (\bibinfo{year}{2025}).
\bibitem[{Remondino et~al.(2023)Remondino, Karami, Yan, Mazzacca, Rigon, and
  Qin}]{Remondino2023CriticalNeRF}
\bibinfo{author}{F.~Remondino}, \bibinfo{author}{A.~Karami},
  \bibinfo{author}{Z.~Yan}, \bibinfo{author}{G.~Mazzacca},
  \bibinfo{author}{S.~Rigon}, \bibinfo{author}{R.~Qin},
\newblock \bibinfo{title}{A critical analysis of {NeRF}-based {3D}
  reconstruction},
\newblock \bibinfo{journal}{Remote Sensing} \bibinfo{volume}{15}
  (\bibinfo{year}{2023}) \bibinfo{pages}{3585}.
  \DOIprefix\doi{10.3390/rs15143585}.
\bibitem[{Fontanini et~al.(2016)Fontanini, Vaidya, Passalacqua, and
  Ganapathysubramanian}]{fontanini2016ventilation}
\bibinfo{author}{A.~D. Fontanini}, \bibinfo{author}{U.~Vaidya},
  \bibinfo{author}{A.~Passalacqua}, \bibinfo{author}{B.~Ganapathysubramanian},
\newblock \bibinfo{title}{Quantifying mechanical ventilation performance: The
  connection between transport equations and {Markov} matrices},
\newblock \bibinfo{journal}{Building and Environment} \bibinfo{volume}{104}
  (\bibinfo{year}{2016}) \bibinfo{pages}{253--262}.
  \DOIprefix\doi{10.1016/j.buildenv.2016.05.019}.
\bibitem[{Liu et~al.(2019)Liu, Zhu, Kim, and Srebric}]{Liu2019ReviewCFDPV}
\bibinfo{author}{J.~Liu}, \bibinfo{author}{S.~Zhu}, \bibinfo{author}{M.~K.
  Kim}, \bibinfo{author}{J.~Srebric},
\newblock \bibinfo{title}{A review of {CFD} analysis methods for personalized
  ventilation ({PV}) in indoor built environments},
\newblock \bibinfo{journal}{Sustainability} \bibinfo{volume}{11}
  (\bibinfo{year}{2019}) \bibinfo{pages}{4166}.
  \DOIprefix\doi{10.3390/su11154166}.
\bibitem[{Tan et~al.(2023)Tan, Gao, Yang, Johnson, Hsu, Passalacqua,
  Krishnamurthy, and Ganapathysubramanian}]{Tan2023ClassroomRiskFramework}
\bibinfo{author}{K.~Tan}, \bibinfo{author}{B.~Gao}, \bibinfo{author}{C.-H.
  Yang}, \bibinfo{author}{E.~L. Johnson}, \bibinfo{author}{M.-C. Hsu},
  \bibinfo{author}{A.~Passalacqua}, \bibinfo{author}{A.~Krishnamurthy},
  \bibinfo{author}{B.~Ganapathysubramanian},
\newblock \bibinfo{title}{A computational framework for transmission risk
  assessment of aerosolized particles in classrooms},
\newblock \bibinfo{journal}{Engineering with Computers} \bibinfo{volume}{40}
  (\bibinfo{year}{2023}) \bibinfo{pages}{235--256}.
  \DOIprefix\doi{10.1007/s00366-022-01773-9}.
\bibitem[{Park et~al.(2022)Park, Lee, Yook, and
  Koo}]{Park2022VerticalAirflowClassroom}
\bibinfo{author}{S.-H. Park}, \bibinfo{author}{K.-R. Lee},
  \bibinfo{author}{S.-J. Yook}, \bibinfo{author}{H.~B. Koo},
\newblock \bibinfo{title}{Enhancement and homogenization of indoor air quality
  in a classroom using a vertical airflow ventilation scheme},
\newblock \bibinfo{journal}{Toxics} \bibinfo{volume}{10} (\bibinfo{year}{2022})
  \bibinfo{pages}{545}. \DOIprefix\doi{10.3390/toxics10090545}.
\bibitem[{H{\r{a}}gbo et~al.(2021)H{\r{a}}gbo, Giljarhus, and
  Hjertager}]{Hagbo2021GeometryAcquisitionPedestrianWind}
\bibinfo{author}{T.-O. H{\r{a}}gbo}, \bibinfo{author}{K.~E.~T. Giljarhus},
  \bibinfo{author}{B.~H. Hjertager},
\newblock \bibinfo{title}{Influence of geometry acquisition method on
  pedestrian wind simulations},
\newblock \bibinfo{journal}{Journal of Wind Engineering and Industrial
  Aerodynamics} \bibinfo{volume}{215} (\bibinfo{year}{2021})
  \bibinfo{pages}{104665}. \DOIprefix\doi{10.1016/j.jweia.2021.104665}.
\bibitem[{Wong et~al.(2025)Wong, Sun, Ying, Yin, Zhou, Brilakis, Kelly, and
  Lam}]{Wong2025ImageScanToBIM}
\bibinfo{author}{M.~O. Wong}, \bibinfo{author}{Y.~Sun},
  \bibinfo{author}{H.~Ying}, \bibinfo{author}{M.~Yin},
  \bibinfo{author}{H.~Zhou}, \bibinfo{author}{I.~Brilakis},
  \bibinfo{author}{T.~Kelly}, \bibinfo{author}{C.~C. Lam},
\newblock \bibinfo{title}{Image-based scan-to-{BIM} for interior building
  component reconstruction},
\newblock \bibinfo{journal}{Automation in Construction} \bibinfo{volume}{173}
  (\bibinfo{year}{2025}) \bibinfo{pages}{106091}.
  \DOIprefix\doi{10.1016/j.autcon.2025.106091}.
\bibitem[{Cui et~al.(2024)Cui, Wang, Hu, Peng, Zhao, Zhang, and
  Wang}]{Cui2024NeRFusionBuildingConstraints}
\bibinfo{author}{D.~Cui}, \bibinfo{author}{W.~Wang}, \bibinfo{author}{W.~Hu},
  \bibinfo{author}{J.~Peng}, \bibinfo{author}{Y.~Zhao},
  \bibinfo{author}{Y.~Zhang}, \bibinfo{author}{J.~Wang},
\newblock \bibinfo{title}{{3D} reconstruction of building structures
  incorporating neural radiation fields and geometric constraints},
\newblock \bibinfo{journal}{Automation in Construction} \bibinfo{volume}{165}
  (\bibinfo{year}{2024}) \bibinfo{pages}{105517}.
  \DOIprefix\doi{10.1016/j.autcon.2024.105517}.
\bibitem[{Gao et~al.(2023)Gao, Cao, and Shan}]{Gao2023SurfelNeRF}
\bibinfo{author}{Y.~Gao}, \bibinfo{author}{Y.-P. Cao},
  \bibinfo{author}{Y.~Shan},
\newblock \bibinfo{title}{{SurfelNeRF}: Neural surfel radiance fields for
  online photorealistic reconstruction of indoor scenes},
\newblock in: \bibinfo{booktitle}{IEEE/CVF Conference on Computer Vision and
  Pattern Recognition (CVPR)}, \bibinfo{year}{2023}, pp.
  \bibinfo{pages}{108--118}.
\bibitem[{Wang et~al.(2023)Wang, Jaiswal, Corpuz, Paudel, Balu, Krishnamurthy,
  Yan, and Hsu}]{Wang2023PhotogrammetryCFD}
\bibinfo{author}{X.~Wang}, \bibinfo{author}{M.~Jaiswal}, \bibinfo{author}{A.~M.
  Corpuz}, \bibinfo{author}{S.~Paudel}, \bibinfo{author}{A.~Balu},
  \bibinfo{author}{A.~Krishnamurthy}, \bibinfo{author}{J.~Yan},
  \bibinfo{author}{M.-C. Hsu},
\newblock \bibinfo{title}{Photogrammetry-based computational fluid dynamics},
\newblock \bibinfo{journal}{Computer Methods in Applied Mechanics and
  Engineering} \bibinfo{volume}{417} (\bibinfo{year}{2023})
  \bibinfo{pages}{116311}. \DOIprefix\doi{10.1016/j.cma.2023.116311}.
\bibitem[{Balu et~al.(2023)Balu, Rajanna, Khristy, Xu, Krishnamurthy, and
  Hsu}]{Balu2023IMGA_PointClouds}
\bibinfo{author}{A.~Balu}, \bibinfo{author}{M.~R. Rajanna},
  \bibinfo{author}{J.~Khristy}, \bibinfo{author}{F.~Xu},
  \bibinfo{author}{A.~Krishnamurthy}, \bibinfo{author}{M.-C. Hsu},
\newblock \bibinfo{title}{Direct immersogeometric fluid flow and heat transfer
  analysis of objects represented by point clouds},
\newblock \bibinfo{journal}{Computer Methods in Applied Mechanics and
  Engineering} \bibinfo{volume}{404} (\bibinfo{year}{2023})
  \bibinfo{pages}{115742}. \DOIprefix\doi{10.1016/j.cma.2022.115742}.
\bibitem[{Ishikawa et~al.(2021)Ishikawa, Kanai, and
  Date}]{Ishikawa2021CartesianGridAsBuiltCFD}
\bibinfo{author}{T.~Ishikawa}, \bibinfo{author}{S.~Kanai},
  \bibinfo{author}{H.~Date},
\newblock \bibinfo{title}{Direct generation of {Cartesian} grid for as-built
  {CFD} analysis from laser-scanned point clouds},
\newblock \bibinfo{journal}{Computer-Aided Design and Applications}
  \bibinfo{volume}{18} (\bibinfo{year}{2021}) \bibinfo{pages}{1341--1358}.
  \DOIprefix\doi{10.14733/cadaps.2021.1341-1358}.
\bibitem[{Shah et~al.(2022)Shah, Ghadai, Gamdha, Schuster, Thomas, Greiner, and
  Krishnamurthy}]{Shah2022GPUCollisionPointCloud}
\bibinfo{author}{H.~Shah}, \bibinfo{author}{S.~Ghadai},
  \bibinfo{author}{D.~Gamdha}, \bibinfo{author}{A.~Schuster},
  \bibinfo{author}{I.~Thomas}, \bibinfo{author}{N.~Greiner},
  \bibinfo{author}{A.~Krishnamurthy},
\newblock \bibinfo{title}{{GPU}-accelerated collision analysis of vehicles in a
  point cloud environment},
\newblock \bibinfo{journal}{IEEE Computer Graphics and Applications}
  \bibinfo{volume}{42} (\bibinfo{year}{2022}) \bibinfo{pages}{37--50}.
  \DOIprefix\doi{10.1109/MCG.2022.3177890}.
\bibitem[{Yang et~al.(2026)Yang, Scovazzi, Krishnamurthy, and
  Ganapathysubramanian}]{Yang2025OctreeSBM}
\bibinfo{author}{C.-H. Yang}, \bibinfo{author}{G.~Scovazzi},
  \bibinfo{author}{A.~Krishnamurthy},
  \bibinfo{author}{B.~Ganapathysubramanian},
\newblock \bibinfo{title}{Simulating incompressible flows over complex
  geometries using the shifted boundary method with incomplete adaptive octree
  meshes},
\newblock \bibinfo{journal}{Journal of Computational Physics}
  \bibinfo{volume}{544} (\bibinfo{year}{2026}) \bibinfo{pages}{114334}.
  \DOIprefix\doi{10.1016/j.jcp.2025.114334}.
\bibitem[{Yang et~al.(2025)Yang, Scovazzi, Krishnamurthy, and
  Ganapathysubramanian}]{Yang2025SBMThermal}
\bibinfo{author}{C.-H. Yang}, \bibinfo{author}{G.~Scovazzi},
  \bibinfo{author}{A.~Krishnamurthy},
  \bibinfo{author}{B.~Ganapathysubramanian},
\newblock \bibinfo{title}{A shifted boundary method for thermal flows},
\newblock \bibinfo{journal}{Journal of Computational Physics}
  (\bibinfo{year}{2025}) \bibinfo{pages}{114333}.
  \DOIprefix\doi{10.1016/j.jcp.2025.114333}.
\bibitem[{Yang et~al.(2019)Yang, Zhou, Su, Zuo, Tang, Liang, Zhu, and
  Li}]{Yang2019IndoorMultiRoomCurvedWalls}
\bibinfo{author}{F.~Yang}, \bibinfo{author}{G.~Zhou}, \bibinfo{author}{F.~Su},
  \bibinfo{author}{X.~Zuo}, \bibinfo{author}{L.~Tang},
  \bibinfo{author}{Y.~Liang}, \bibinfo{author}{H.~Zhu},
  \bibinfo{author}{L.~Li},
\newblock \bibinfo{title}{Automatic indoor reconstruction from point clouds in
  multi-room environments with curved walls},
\newblock \bibinfo{journal}{Sensors} \bibinfo{volume}{19}
  (\bibinfo{year}{2019}) \bibinfo{pages}{3798}.
  \DOIprefix\doi{10.3390/s19173798}.
\bibitem[{Lim and Doh(2021)}]{Lim2021MultiLevelIndoor}
\bibinfo{author}{G.~Lim}, \bibinfo{author}{N.~Doh},
\newblock \bibinfo{title}{Automatic reconstruction of multi-level indoor spaces
  from point cloud and trajectory},
\newblock \bibinfo{journal}{Sensors} \bibinfo{volume}{21}
  (\bibinfo{year}{2021}) \bibinfo{pages}{3493}.
  \DOIprefix\doi{10.3390/s21103493}.
\bibitem[{Liang et~al.(2025)Liang, Liu, Wang, Qin, Chen, Jia, and
  Dai}]{Liang2025GestaltIndoorOcclusion}
\bibinfo{author}{X.~Liang}, \bibinfo{author}{Y.~Liu},
  \bibinfo{author}{J.~Wang}, \bibinfo{author}{Z.~Qin},
  \bibinfo{author}{L.~Chen}, \bibinfo{author}{H.~Jia},
  \bibinfo{author}{C.~Dai},
\newblock \bibinfo{title}{Automatic {3D} reconstruction of indoor building
  point clouds under occlusion based on gestalt rules},
\newblock \bibinfo{journal}{International Journal of Digital Earth}
  \bibinfo{volume}{18} (\bibinfo{year}{2025}) \bibinfo{pages}{2545582}.
  \DOIprefix\doi{10.1080/17538947.2025.2545582}.
\bibitem[{Patil and Kalantari(2025)}]{Patil2025ScanToBIMSegAccuracy}
\bibinfo{author}{J.~Patil}, \bibinfo{author}{M.~Kalantari},
\newblock \bibinfo{title}{Automatic scan-to-{BIM}: The impact of semantic
  segmentation accuracy},
\newblock \bibinfo{journal}{Buildings} \bibinfo{volume}{15}
  (\bibinfo{year}{2025}) \bibinfo{pages}{1126}.
  \DOIprefix\doi{10.3390/buildings15071126}.
\bibitem[{Hu et~al.(2024)Hu, Gan, and Zhai}]{Hu2024DawNetBIMtoScan}
\bibinfo{author}{D.~Hu}, \bibinfo{author}{V.~J.~L. Gan},
  \bibinfo{author}{R.~Zhai},
\newblock \bibinfo{title}{Automated {BIM}-to-scan point cloud semantic
  segmentation using a domain adaptation network with hybrid attention and
  whitening ({DawNet})},
\newblock \bibinfo{journal}{Automation in Construction} \bibinfo{volume}{164}
  (\bibinfo{year}{2024}) \bibinfo{pages}{105473}.
  \DOIprefix\doi{10.1016/j.autcon.2024.105473}.
\bibitem[{Garrido-Jurado et~al.(2014)Garrido-Jurado, Mu{\~n}oz-Salinas,
  Madrid-Cuevas, and Mar{\'i}n-Jim{\'e}nez}]{GarridoJurado2014ArUco}
\bibinfo{author}{S.~Garrido-Jurado}, \bibinfo{author}{R.~Mu{\~n}oz-Salinas},
  \bibinfo{author}{F.~J. Madrid-Cuevas}, \bibinfo{author}{M.~J.
  Mar{\'i}n-Jim{\'e}nez},
\newblock \bibinfo{title}{Automatic generation and detection of highly reliable
  fiducial markers under occlusion},
\newblock \bibinfo{journal}{Pattern Recognition} \bibinfo{volume}{47}
  (\bibinfo{year}{2014}) \bibinfo{pages}{2280--2292}.
  \DOIprefix\doi{10.1016/j.patcog.2014.01.005}.
\bibitem[{Griwodz et~al.(2021)Griwodz, Gasparini, Calvet, Gurdjos, Castan,
  Maujean, De~Lillo, and Lanthony}]{Meshroom2021}
\bibinfo{author}{C.~Griwodz}, \bibinfo{author}{S.~Gasparini},
  \bibinfo{author}{L.~Calvet}, \bibinfo{author}{P.~Gurdjos},
  \bibinfo{author}{F.~Castan}, \bibinfo{author}{B.~Maujean},
  \bibinfo{author}{G.~De~Lillo}, \bibinfo{author}{Y.~Lanthony},
\newblock \bibinfo{title}{{AliceVision Meshroom}: An open-source {3D}
  reconstruction pipeline},
\newblock in: \bibinfo{booktitle}{Proceedings of the 12th ACM Multimedia
  Systems Conference (MMSys '21)}, \bibinfo{year}{2021}, pp.
  \bibinfo{pages}{241--247}. \DOIprefix\doi{10.1145/3458305.3478443}.
\bibitem[{Lowe(2004)}]{Lowe2004SIFT}
\bibinfo{author}{D.~G. Lowe},
\newblock \bibinfo{title}{Distinctive image features from scale-invariant
  keypoints},
\newblock \bibinfo{journal}{International Journal of Computer Vision}
  \bibinfo{volume}{60} (\bibinfo{year}{2004}) \bibinfo{pages}{91--110}.
  \DOIprefix\doi{10.1023/B:VISI.0000029664.99615.94}.
\bibitem[{Sch{\"o}nberger and Frahm(2016)}]{Schonberger2016COLMAP}
\bibinfo{author}{J.~L. Sch{\"o}nberger}, \bibinfo{author}{J.-M. Frahm},
\newblock \bibinfo{title}{Structure-from-motion revisited},
\newblock in: \bibinfo{booktitle}{IEEE Conference on Computer Vision and
  Pattern Recognition (CVPR)}, \bibinfo{year}{2016}, pp.
  \bibinfo{pages}{4104--4113}. \DOIprefix\doi{10.1109/CVPR.2016.445}.
\bibitem[{Besl and McKay(1992)}]{Besl1992ICP}
\bibinfo{author}{P.~J. Besl}, \bibinfo{author}{N.~D. McKay},
\newblock \bibinfo{title}{A method for registration of 3-{D} shapes},
\newblock \bibinfo{journal}{IEEE Transactions on Pattern Analysis and Machine
  Intelligence} \bibinfo{volume}{14} (\bibinfo{year}{1992})
  \bibinfo{pages}{239--256}. \DOIprefix\doi{10.1109/34.121791}.
\bibitem[{Chen and Medioni(1992)}]{Chen1992ICPRange}
\bibinfo{author}{Y.~Chen}, \bibinfo{author}{G.~Medioni},
\newblock \bibinfo{title}{Object modelling by registration of multiple range
  images},
\newblock \bibinfo{journal}{Image and Vision Computing} \bibinfo{volume}{10}
  (\bibinfo{year}{1992}) \bibinfo{pages}{145--155}.
  \DOIprefix\doi{10.1016/0262-8856(92)90066-C}.
\bibitem[{Jolliffe and Cadima(2016)}]{Jolliffe2016PCA}
\bibinfo{author}{I.~T. Jolliffe}, \bibinfo{author}{J.~Cadima},
\newblock \bibinfo{title}{Principal component analysis: a review and recent
  developments},
\newblock \bibinfo{journal}{Philosophical Transactions of the Royal Society A}
  \bibinfo{volume}{374} (\bibinfo{year}{2016}) \bibinfo{pages}{20150202}.
  \DOIprefix\doi{10.1098/rsta.2015.0202}.
\bibitem[{Weller et~al.(1998)Weller, Tabor, Jasak, and
  Fureby}]{Weller1998OpenFOAM}
\bibinfo{author}{H.~G. Weller}, \bibinfo{author}{G.~Tabor},
  \bibinfo{author}{H.~Jasak}, \bibinfo{author}{C.~Fureby},
\newblock \bibinfo{title}{A tensorial approach to computational continuum
  mechanics using object-oriented techniques},
\newblock \bibinfo{journal}{Computers in Physics} \bibinfo{volume}{12}
  (\bibinfo{year}{1998}) \bibinfo{pages}{620--631}.
  \DOIprefix\doi{10.1063/1.168744}.
\bibitem[{{OpenCFD Ltd.}(2024)}]{OpenCFD2024UserGuide}
\bibinfo{author}{{OpenCFD Ltd.}}, \bibinfo{title}{{OpenFOAM} v2412 user guide},
  \bibinfo{howpublished}{\url{https://www.openfoam.com/documentation/user-guide}},
  \bibinfo{year}{2024}.
\bibitem[{Launder and Spalding(1974)}]{Launder1974KEpsilon}
\bibinfo{author}{B.~E. Launder}, \bibinfo{author}{D.~B. Spalding},
\newblock \bibinfo{title}{The numerical computation of turbulent flows},
\newblock \bibinfo{journal}{Computer Methods in Applied Mechanics and
  Engineering} \bibinfo{volume}{3} (\bibinfo{year}{1974})
  \bibinfo{pages}{269--289}. \DOIprefix\doi{10.1016/0045-7825(74)90029-2}.
\bibitem[{Patankar and Spalding(1972)}]{Patankar1972SIMPLE}
\bibinfo{author}{S.~V. Patankar}, \bibinfo{author}{D.~B. Spalding},
\newblock \bibinfo{title}{A calculation procedure for heat, mass and momentum
  transfer in three-dimensional parabolic flows},
\newblock \bibinfo{journal}{International Journal of Heat and Mass Transfer}
  \bibinfo{volume}{15} (\bibinfo{year}{1972}) \bibinfo{pages}{1787--1806}.
  \DOIprefix\doi{10.1016/0017-9310(72)90054-3}.
\bibitem[{Gualtieri et~al.(2017)Gualtieri, Angeloudis, Bombardelli, Jha, and
  Stoesser}]{Gualtieri2017TurbulentSchmidt}
\bibinfo{author}{C.~Gualtieri}, \bibinfo{author}{A.~Angeloudis},
  \bibinfo{author}{F.~A. Bombardelli}, \bibinfo{author}{S.~K. Jha},
  \bibinfo{author}{T.~Stoesser},
\newblock \bibinfo{title}{On the values for the turbulent schmidt number in
  environmental flows},
\newblock \bibinfo{journal}{Fluids} \bibinfo{volume}{2} (\bibinfo{year}{2017})
  \bibinfo{pages}{17}. \DOIprefix\doi{10.3390/fluids2020017}.
\bibitem[{Li et~al.(2018)Li, Liu, Ren, and Cao}]{Li2018ModifiedSchmidt}
\bibinfo{author}{F.~Li}, \bibinfo{author}{J.~Liu}, \bibinfo{author}{J.~Ren},
  \bibinfo{author}{X.~Cao},
\newblock \bibinfo{title}{Predicting contaminant dispersion using modified
  turbulent schmidt numbers from different vortex structures},
\newblock \bibinfo{journal}{Building and Environment} \bibinfo{volume}{130}
  (\bibinfo{year}{2018}) \bibinfo{pages}{120--127}.
  \DOIprefix\doi{10.1016/j.buildenv.2017.12.023}.
\bibitem[{Zhou et~al.(2018)Zhou, Park, and Koltun}]{Zhou2018Open3D}
\bibinfo{author}{Q.-Y. Zhou}, \bibinfo{author}{J.~Park},
  \bibinfo{author}{V.~Koltun},
\newblock \bibinfo{title}{{Open3D}: A modern library for {3D} data processing},
\newblock \bibinfo{journal}{arXiv preprint arXiv:1801.09847}
  (\bibinfo{year}{2018}).
\bibitem[{Dawson-Haggerty et~al.(2019)}]{DawsonHaggerty2019Trimesh}
\bibinfo{author}{M.~Dawson-Haggerty}, et~al., \bibinfo{title}{trimesh: Import,
  export, process, analyze and view triangular meshes},
  \bibinfo{howpublished}{\url{https://trimesh.org/}}, \bibinfo{year}{2019}.
\bibitem[{Stanzione et~al.(2020)Stanzione, West, Evans, Minyard, Ghattas, and
  Panda}]{Stanzione2020Frontera}
\bibinfo{author}{D.~Stanzione}, \bibinfo{author}{J.~West},
  \bibinfo{author}{R.~T. Evans}, \bibinfo{author}{T.~Minyard},
  \bibinfo{author}{O.~Ghattas}, \bibinfo{author}{D.~K. Panda},
\newblock \bibinfo{title}{Frontera: The evolution of leadership computing at
  the {Texas Advanced Computing Center}},
\newblock in: \bibinfo{booktitle}{Practice and Experience in Advanced Research
  Computing (PEARC '20)}, \bibinfo{year}{2020}, pp. \bibinfo{pages}{106--111}.
  \DOIprefix\doi{10.1145/3311790.3396656}.
\bibitem[{Pellegrini and Roman(1996)}]{Pellegrini1996Scotch}
\bibinfo{author}{F.~Pellegrini}, \bibinfo{author}{J.~Roman},
\newblock \bibinfo{title}{{SCOTCH}: A software package for static mapping by
  dual recursive bipartitioning of process and architecture graphs},
\newblock in: \bibinfo{booktitle}{High-Performance Computing and Networking
  (HPCN Europe)}, \bibinfo{year}{1996}, pp. \bibinfo{pages}{493--498}.
  \DOIprefix\doi{10.1007/3-540-61142-8_588}.
\bibitem[{Ahrens et~al.(2005)Ahrens, Geveci, and Law}]{Ahrens2005ParaView}
\bibinfo{author}{J.~Ahrens}, \bibinfo{author}{B.~Geveci},
  \bibinfo{author}{C.~Law},
\newblock \bibinfo{title}{{ParaView}: An end-user tool for large-data
  visualization},
\newblock in: \bibinfo{booktitle}{Visualization Handbook},
  \bibinfo{publisher}{Elsevier}, \bibinfo{year}{2005}, pp.
  \bibinfo{pages}{717--731}. \DOIprefix\doi{10.1016/B978-012387582-2/50038-1}.
\bibitem[{Auvinen et~al.(2022)Auvinen, Kuula, Gr{\"o}nholm, S{\"u}hring, and
  Hellsten}]{Auvinen2022LESairborne}
\bibinfo{author}{M.~Auvinen}, \bibinfo{author}{J.~Kuula},
  \bibinfo{author}{T.~Gr{\"o}nholm}, \bibinfo{author}{M.~S{\"u}hring},
  \bibinfo{author}{A.~Hellsten},
\newblock \bibinfo{title}{High-resolution large-eddy simulation of indoor
  turbulence and its effect on airborne transmission of respiratory
  pathogens---model validation and infection probability analysis},
\newblock \bibinfo{journal}{Physics of Fluids} \bibinfo{volume}{34}
  (\bibinfo{year}{2022}) \bibinfo{pages}{015124}.
  \DOIprefix\doi{10.1063/5.0076495}.
\bibitem[{Afful et~al.(2026)Afful, Passe, and
  Ganapathysubramanian}]{Afful2026DiffuserResolvingCFD}
\bibinfo{author}{J.~Afful}, \bibinfo{author}{U.~Passe},
  \bibinfo{author}{B.~Ganapathysubramanian}, \bibinfo{title}{Diffuser-resolving
  {CFD} of {K}--12 classrooms under {ASHRAE} 62.1 and 241-aligned ventilation
  scenarios}, \bibinfo{year}{2026}. \bibinfo{note}{Manuscript in preparation}.
\bibitem[{Chen and Srebric(2002)}]{ChenSrebric2002CFDVerification}
\bibinfo{author}{Q.~Chen}, \bibinfo{author}{J.~Srebric},
\newblock \bibinfo{title}{A procedure for verification, validation, and
  reporting of indoor environment {CFD} analyses},
\newblock \bibinfo{journal}{HVAC\&R Research} \bibinfo{volume}{8}
  (\bibinfo{year}{2002}) \bibinfo{pages}{201--216}.
  \DOIprefix\doi{10.1080/10789669.2002.10391437}.
\bibitem[{Nielsen(1990)}]{nielsen1990specification}
\bibinfo{author}{P.~V. Nielsen}, \bibinfo{title}{Specification of a
  Two-Dimensional Test Case}, \bibinfo{type}{IEA Annex 20 Research Report},
  Aalborg University, Department of Building Technology and Structural
  Engineering, \bibinfo{address}{Aalborg, Denmark}, \bibinfo{year}{1990}.

\end{thebibliography}

\newpage
\setcounter{figure}{0}
\setcounter{table}{0}

\appendix
\section{CFD Framework Verification and Validation}
\label{app:verification}
Verification and validation are distinct requirements in indoor-environment CFD. Following the framework articulated by \citet{ChenSrebric2002CFDVerification}, numerical verification concerns the sensitivity and consistency of the computational formulation, whereas validation requires comparison against suitable physical observations for the problem of interest. This appendix assesses the fidelity of the OpenFOAM solver configuration and the sensitivity of the production meshes used throughout \secref{sec:cfd_method}, independent of the reconstruction-to-CFD demonstrations of \secref{sec:results}. \ref{app:mesh_convergence} reports a four-level mesh-sensitivity assessment performed on the Mixed-Furniture Classroom configuration with mannequins. \ref{app:room_scale_patch_checks} documents the metric dimensions of the final room geometries and the height references used to assign their physical scale. \ref{app:annex20_validation} reports an IEA Annex~20 benchmark comparison, examining key flow indicators against an established reference case to test whether the present solver settings reproduce expected behavior.

\subsection{Mesh-Sensitivity Assessment}
\label{app:mesh_convergence}
Mesh sensitivity was assessed using the normalized passive-scalar concentration \(C/C_0\), which is the transported quantity directly associated with the contaminant-clearance analysis in this study. Vertical scalar profiles were sampled along a representative occupied-zone line at \(t=2000\)~s OpenFOAM time, corresponding to \(1000\)~s of elapsed scalar-transport time after initialization from the steady airflow field. The extra-fine mesh produced valid samples over a shorter portion of the requested vertical line than the other meshes. Therefore, the four-mesh comparison was performed over the common valid sampling interval shared by all meshes, approximately \(z=0.05\)--\(2.00\)~m above the floor. The scalar line integral \(\int C\,ds\) was computed over this same interval as a summary metric for comparison across mesh densities.

\begin{table}[h!]
\centering
\caption{Mesh-sensitivity summary for the fully obstructed mixed-furniture classroom configuration
using \(\mathit{Sc}_t=0.7\). The scalar integral was computed over the common
valid portion of the vertical occupied-zone sampling line, from approximately
\(z=0.05\)~m to \(z=2.00\)~m above the floor.}
\label{tab:mesh_convergence_summary}

\setlength{\tabcolsep}{8pt}
\begin{tabular}{lrrr}
\toprule
\textbf{Mesh level} & \textbf{Cells} & \textbf{Points} & \(\boldsymbol{\int C\,ds}\) \textbf{(m)} \\
\midrule
Coarse     & 2.15M  & 2.65M  & 0.03322 \\
Medium     & 5.99M  & 7.36M  & 0.03473 \\
Fine       & 15.91M & 18.82M & 0.04648 \\
Extra-fine & 55.37M & 62.85M & 0.04654 \\
\bottomrule
\end{tabular}
\end{table}

\begin{figure*}[t!]
\centering

\begin{subfigure}[b]{0.48\linewidth}
\centering
\begin{tikzpicture}
\begin{axis}[
    meshplot,
    xlabel={Height above floor, $z$ (m)},
    ylabel={Normalized concentration, $C/C_0$},
    xmin=0, xmax=2.05,
    ymin=-0.05, ymax=1.01,
    ymajorgrids=true,
    xmajorgrids=true,
]
\addplot[meshcoarse, thick]
    table[x=z_above_floor_m, y=T, col sep=comma]
    {figures/coarse_vertical_profile_common_2000.csv};
\addplot[meshmedium, thick]
    table[x=z_above_floor_m, y=T, col sep=comma]
    {figures/medium_vertical_profile_common_2000.csv};
\addplot[meshfine, thick]
    table[x=z_above_floor_m, y=T, col sep=comma]
    {figures/fine_vertical_profile_common_2000.csv};
\addplot[meshextrafine, thick]
    table[x=z_above_floor_m, y=T, col sep=comma]
    {figures/extrafine_vertical_profile_common_2000.csv};
\legend{
    Coarse ($2.15$M cells),
    Medium ($5.99$M cells),
    Fine ($15.91$M cells),
    Extra-fine ($55.37$M cells)
}
\end{axis}
\end{tikzpicture}
\caption{Vertical scalar concentration profile over the common valid sampling interval at $t=2000$~s OpenFOAM time.}
\label{fig:mesh_contamination_profile}
\end{subfigure}\hfill
\begin{subfigure}[b]{0.48\linewidth}
\centering
\begin{tikzpicture}
\begin{axis}[
    meshplot,
    xlabel={Mesh size (million cells)},
    ylabel={Integrated concentration, $\int C\,ds$ (m)},
    xmin=0,
    ymin=0.030,
    ymax=0.050,
    ymajorgrids=true,
    xmajorgrids=true,
]
\addplot[black, thick, mark=*, mark size=2.2pt]
    table[x=cells, y=IT, col sep=comma]
    {figures/mesh_integrals_common_range.csv};
\end{axis}
\end{tikzpicture}
\caption{Integrated scalar concentration over the common valid sampling interval.}
\label{fig:mesh_integrated_contamination}
\end{subfigure}

\caption{Mesh-sensitivity assessment for the mixed-furniture classroom (fully obstructed) using the turbulent scalar-transport framework with $\mathit{Sc}_t=0.7$. (a) Vertical profile of normalized scalar concentration, $C/C_0$, along the occupied-zone sampling line for the coarse, medium, fine, and extra-fine meshes. Because the extra-fine mesh produced valid samples over a shorter portion of the requested vertical line, all four profiles and line integrals are compared over the common valid interval, approximately $z=0.05$--$2.00$~m above the floor. (b) Integrated scalar concentration, $\int C\,ds$, computed over the same interval and plotted against mesh size. The meshes contain approximately $2.15$M, $5.99$M, $15.91$M, and $55.37$M cells, respectively.}
\label{fig:mesh_convergence}
\end{figure*}
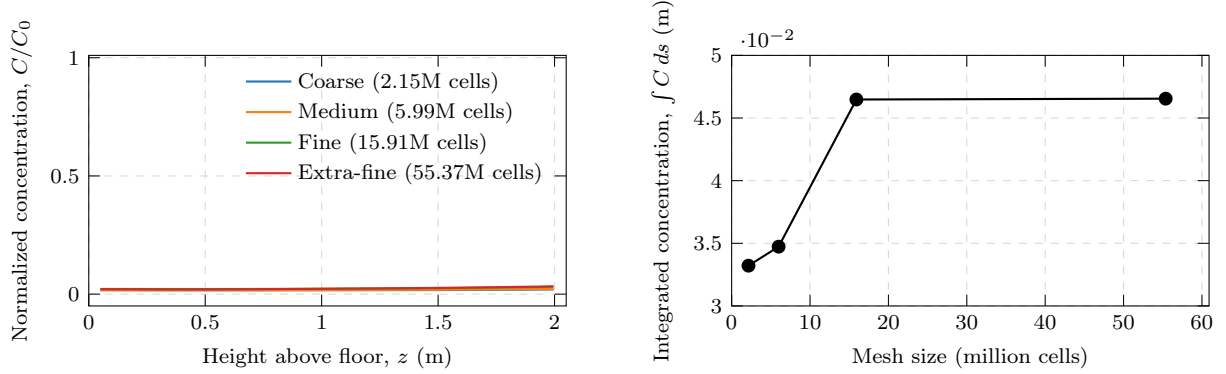

\subfigref{fig:mesh_convergence}{(a)} shows the scalar profiles and
\subfigref{fig:mesh_convergence}{(b)} the integrated scalar metric for the
four mesh levels.

Because refinement was introduced through local \texttt{snappyHexMesh}
settings rather than uniform global refinement, adjacent mesh levels do not
have a constant spatial refinement ratio. The assessment is therefore reported
as a practical mesh-sensitivity study rather than a formal Grid Convergence
Index (GCI) analysis.

Over the common valid sampling interval, the integrated scalar concentration
is \(0.03322\), \(0.03473\), \(0.04648\), and \(0.04654\) for the coarse,
medium, fine, and extra-fine meshes, respectively
(\tableref{tab:mesh_convergence_summary}). The fine and extra-fine
values differ by approximately \(0.14\%\), indicating that this scalar-profile
metric is practically insensitive to further refinement beyond the fine mesh
over the sampled interval. Because passive-scalar transport and contaminant
clearance are the primary quantities of interest in the present comparative
analysis, the fine mesh was retained as the production-mesh target. The mesh-sensitivity assessment was performed using scalar profiles and integrated line metrics rather than full repeated \(t_{50}\) calculations for every production case. Also, the scalar-transport simulations use first-order upwind convection for robustness on complex hex-dominant meshes.

\subsection{Room dimensions and metric scale assignment}
\label{app:room_scale_patch_checks}

\begin{figure}[t!]
\centering
\begin{subfigure}[b]{0.3\linewidth}
    \centering
    \includegraphics[width=0.9\linewidth]{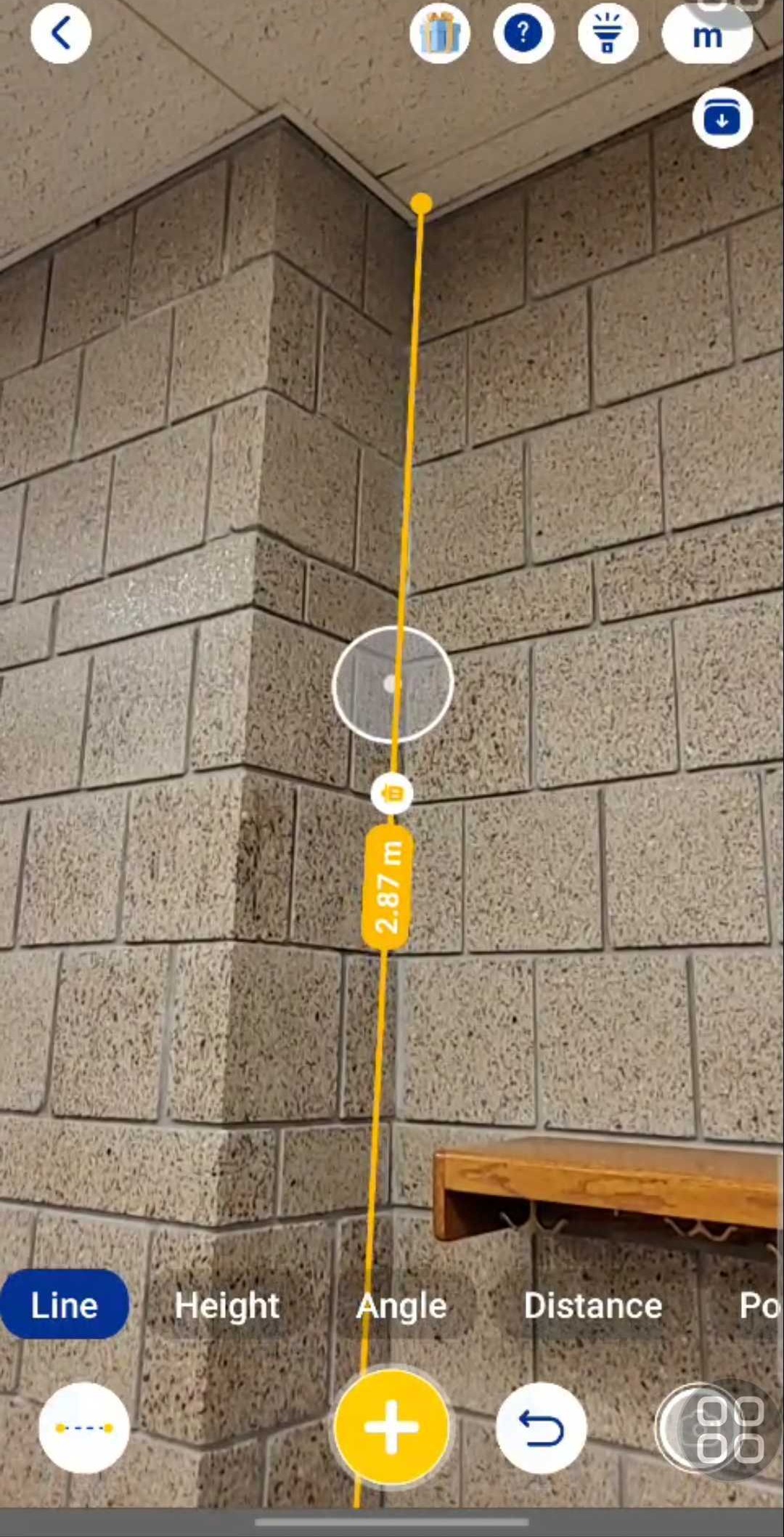}
    \caption{Classroom height measurement}
    \label{fig:ar_ruler_classroom}
\end{subfigure}
\begin{subfigure}[b]{0.3\linewidth}
    \centering
    \includegraphics[width=0.9\linewidth]{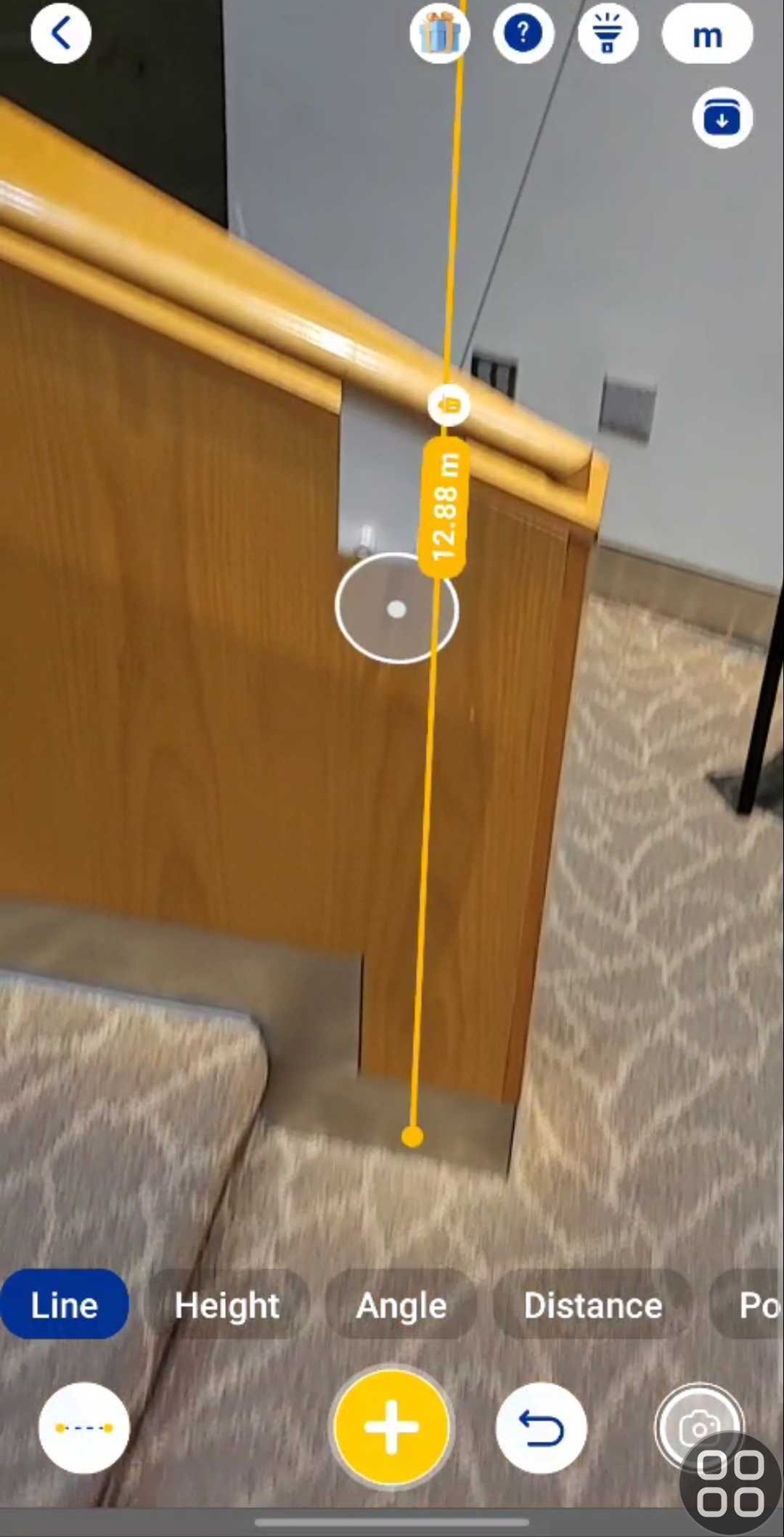}
    \caption{Auditorium height measurement}
    \label{fig:ar_ruler_auditorium}
\end{subfigure}
\caption{Room-height reference measurements using the AR Ruler mobile-phone application. These heights establish the metric scale of the reconstructed OpenFOAM domains and are therefore not an independent check of reconstruction accuracy; they are not survey-grade measurements.}
\label{fig:ar_ruler_measurements}
\end{figure}

To document the geometric scale used in the CFD cases, \tableref{tab:room_scale_checks} compares the modeled room heights with the physical room-height references collected using the AR Ruler mobile-phone application (\figref{fig:ar_ruler_measurements}). These references establish the metric scale of the reconstructed geometry and are therefore not an independent check of reconstruction accuracy.

\begin{table}[htbp]
\centering
\caption{Metric dimensions of the final OpenFOAM room geometries and physical room-height references used for global scale assignment. The in-plane dimensions \(L_x\) and \(L_y\) are extracted from the scaled computational geometry. The AR Ruler height measurements establish the metric scale and are therefore not treated as independent validation of reconstruction accuracy.}
\label{tab:room_scale_checks}
\begin{tabular}{lccccc}
\toprule
Case & \(L_x\) (m) & \(L_y\) (m) & \(H\) model (m) & \(H\) reference (m) & Difference \\
\midrule
Mixed-furniture: empty   & 5.71  & 7.22  & 2.90  & 2.87  & \(+0.03\) m \;(\(+1.0\%\)) \\
Mixed-furniture: +chairs & 5.71  & 7.22  & 2.90  & 2.87  & \(+0.03\) m \;(\(+1.0\%\)) \\
Mixed-furniture: +tables & 5.71  & 7.22  & 2.90  & 2.87  & \(+0.03\) m \;(\(+1.0\%\)) \\
Mixed-furniture: full    & 5.71  & 7.22  & 2.90  & 2.87  & \(+0.03\) m \;(\(+1.0\%\)) \\
Chair-dominant: complex  & 6.04  & 6.29  & 2.90  & 2.87  & \(+0.03\) m \;(\(+1.0\%\)) \\
Auditorium        & 30.01 & 28.98 & 12.87 & 12.88 & \(-0.01\) m \;(\(-0.08\%\)) \\
\bottomrule
\end{tabular}
\end{table}

\subsection{Annex~20 Benchmark Comparison}
\label{app:annex20_validation}

We compared the momentum and turbulence modeling to the canonical IEA Annex~20 two-dimensional benchmark \citep{nielsen1990specification} before deploying the framework to the full-scale classroom models. The benchmark tests the solver's representation of a wall-attached supply jet and the resulting room-scale recirculation, including the Coanda-type attachment of the inlet jet to the ceiling. These features are relevant to ceiling-supplied indoor airflow, although the benchmark does not reproduce the geometry or diffuser conditions of the reconstructed classrooms. We discretized the domain with purely hexahedral elements, applying 5:1 spatial grading near the ceiling to capture the steep boundary-layer velocity gradients, and performed a grid-independence study to confirm that spatial discretization does not artificially diffuse the supply jet. Simulations used the steady-state incompressible Reynolds-averaged Navier--Stokes (RANS) solver in OpenFOAM v2412 with the standard $k$--$\varepsilon$ turbulence model and wall functions.

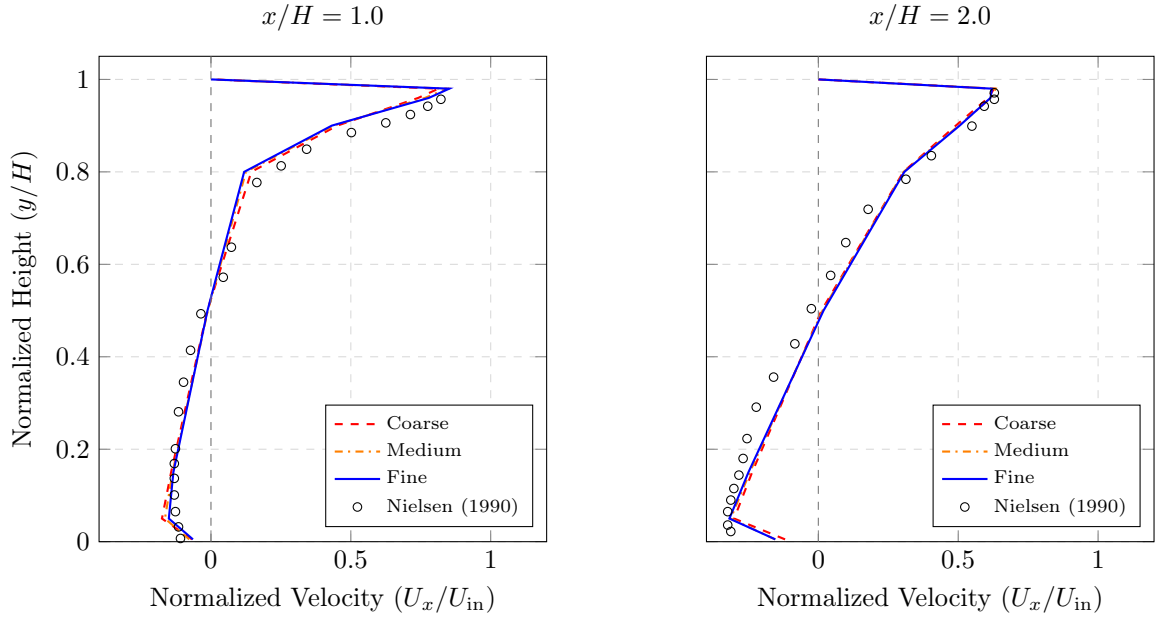
\begin{figure}[t!]
    \centering

    \begin{minipage}{0.48\textwidth}
        \centering
        \begin{tikzpicture}
            \begin{axis}[
                width=0.95\linewidth,
                height=8cm,
                xlabel={Normalized Velocity ($U_x / U_{\mathrm{in}}$)},
                ylabel={Normalized Height ($y / H$)},
                xmin=-0.4, xmax=1.2,
                ymin=0, ymax=1.05,
                grid=both,
                grid style={dashed, gray!30},
                unbounded coords=discard,
                legend cell align={left},
                legend pos=south east,
                legend style={font=\scriptsize},
                title={$x/H = 1.0$}
            ]

            \addplot[color=red, thick, dashed]
                table[x=U_Coarse, y=y_H, col sep=comma]
                {figures/Annex20_Overleaf_Line1.csv};
            \addlegendentry{Coarse}

            \addplot[color=orange, thick, dashdotted]
                table[x=U_Medium, y=y_H, col sep=comma]
                {figures/Annex20_Overleaf_Line1.csv};
            \addlegendentry{Medium}

            \addplot[color=blue, thick]
                table[x=U_Fine, y=y_H, col sep=comma]
                {figures/Annex20_Overleaf_Line1.csv};
            \addlegendentry{Fine}

            \addplot[
                only marks,
                mark=o,
                mark options={fill=white, scale=0.8},
                color=black
            ] coordinates {
                (0.822, 0.957)
                (0.775, 0.942)
                (0.713, 0.924)
                (0.625, 0.906)
                (0.502, 0.885)
                (0.342, 0.849)
                (0.251, 0.813)
                (0.164, 0.777)
                (0.073, 0.637)
                (0.044, 0.572)
                (-0.036, 0.493)
                (-0.073, 0.414)
                (-0.098, 0.345)
                (-0.116, 0.281)
                (-0.127, 0.201)
                (-0.131, 0.169)
                (-0.131, 0.137)
                (-0.131, 0.101)
                (-0.127, 0.065)
                (-0.116, 0.032)
                (-0.109, 0.007)
            };
            \addlegendentry{Nielsen (1990)}

            \draw[gray, thin, dashed] (axis cs:0,0) -- (axis cs:0,1.05);

            \end{axis}
        \end{tikzpicture}
    \end{minipage}\hfill
    \begin{minipage}{0.48\textwidth}
        \centering
        \begin{tikzpicture}
            \begin{axis}[
                width=0.95\linewidth,
                height=8cm,
                xlabel={Normalized Velocity ($U_x / U_{\mathrm{in}}$)},
                yticklabel=\empty,
                xmin=-0.4, xmax=1.2,
                ymin=0, ymax=1.05,
                grid=both,
                grid style={dashed, gray!30},
                unbounded coords=discard,
                legend cell align={left},
                legend pos=south east,
                legend style={font=\scriptsize},
                title={$x/H = 2.0$}
            ]

            \addplot[color=red, thick, dashed]
                table[x=U_Coarse, y=y_H, col sep=comma]
                {figures/Annex20_Overleaf_Line2.csv};
            \addlegendentry{Coarse}

            \addplot[color=orange, thick, dashdotted]
                table[x=U_Medium, y=y_H, col sep=comma]
                {figures/Annex20_Overleaf_Line2.csv};
            \addlegendentry{Medium}

            \addplot[color=blue, thick]
                table[x=U_Fine, y=y_H, col sep=comma]
                {figures/Annex20_Overleaf_Line2.csv};
            \addlegendentry{Fine}

            \addplot[
                only marks,
                mark=o,
                mark options={fill=white, scale=0.8},
                color=black
            ] coordinates {
                (0.629, 0.971)
                (0.629, 0.957)
                (0.593, 0.942)
                (0.549, 0.899)
                (0.404, 0.835)
                (0.313, 0.784)
                (0.178, 0.719)
                (0.098, 0.647)
                (0.044, 0.576)
                (-0.025, 0.504)
                (-0.084, 0.428)
                (-0.160, 0.356)
                (-0.222, 0.291)
                (-0.255, 0.223)
                (-0.269, 0.180)
                (-0.284, 0.144)
                (-0.302, 0.115)
                (-0.313, 0.090)
                (-0.324, 0.065)
                (-0.324, 0.036)
                (-0.313, 0.022)
            };
            \addlegendentry{Nielsen (1990)}

            \draw[gray, thin, dashed] (axis cs:0,0) -- (axis cs:0,1.05);

            \end{axis}
        \end{tikzpicture}
    \end{minipage}

    \caption{Comparison of normalized streamwise velocity profiles predicted using coarse, medium, and fine meshes with digitized measurement data from \citet{nielsen1990specification} for the IEA Annex~20 benchmark room at $x/H=1.0$ (left) and $x/H=2.0$ (right). The experimental points were digitized from the published figure and therefore represent approximate measurement locations rather than original tabulated data.}
    \label{fig:annex20_validation}
\end{figure}

\figref{fig:annex20_validation} compares the normalized streamwise velocity profiles ($U_x/U_{\mathrm{in}}$) at two streamwise stations ($x/H = 1.0$ and $x/H = 2.0$) across the coarse, medium, and fine meshes. The predictions converge toward the digitized experimental measurements under mesh refinement, and the selected mesh and turbulence closure capture both the near-wall jet and the velocity of the primary recirculation zone.

\end{document}